 \documentclass[10pt,twocolumn,singlespace]{IEEEtran}
 \newcommand{\DRAFT}[1]{}
 \newcommand{\FINAL}[1]{#1}
  \usepackage{graphicx,color}
  \usepackage{amsthm,amssymb}
  \usepackage{setspace}
  \usepackage{algorithm2e}
\newsavebox{\fminibox}
\newlength{\fminilength}

  \usepackage{cite}
 \usepackage[usenames,dvipsnames]{pstricks}
 \usepackage{epsfig}
 \usepackage{pst-grad} % For gradients
 \usepackage{pst-plot} % For axes

\usepackage[cmex10]{amsmath}
\graphicspath{{./figures/},{./figures_simu/},{./raman/}} 

\makeatletter
\newcommand{\removelatexerror}{\let\@latex@error\@gobble}
\makeatother
\begin{document}

\title{Dynamical spectral unmixing of multitemporal hyperspectral images}

\author{Simon~Henrot, Jocelyn~Chanussot,\textcolor{black}{~\emph{Fellow, IEEE}}, and Christian~Jutten,~\textcolor{black}{\emph{Fellow, IEEE}}
%\thanks{Copyright (c) 2013 IEEE. Personal use of this material is permitted. However, permission to use this material for any other purposes must be obtained from the IEEE by sending a request to pubs-permissions@ieee.org.}
\thanks{This work was supported by the European Research Council project CHESS, under Grant 2012-ERC-AdG-320684. The authors are with GIPSA-lab/DIS, F-38402 Saint Martin d'H\`eres Cedex- France (e-mail: firstname.lastname@gipsa-lab.grenoble-inp.fr). \textcolor{black}{Jocelyn Chanussot and Christian Jutten are also members of Institut Universitaire de France (IUF).}}}

% The paper headers
%\markboth{ IEEE Transactions on Image Processing}%
%{Shell \MakeLowercase{\textit{et al.}}: Bare Demo of IEEEtran.cls for Journals}

\maketitle

\begin{abstract}
\textcolor{black}{In this paper, we consider the problem of unmixing a time series of hyperspectral images. We propose a dynamical model based on linear mixing processes at each time instant. The spectral signatures and fractional abundances of the pure materials in the scene are seen as latent variables, and assumed to follow a general dynamical structure. Based on a simplified version of this model, we derive an efficient spectral unmixing algorithm to estimate the latent variables by performing alternating minimizations. The performance of the proposed approach is demonstrated on synthetic and real multitemporal hyperspectral images.}
\end{abstract}

\begin{IEEEkeywords}
Hyperspectral imaging, Remote sensing, Source separation, Tensor decomposition
\end{IEEEkeywords}

\section{Introduction}
	\label{sec:intro}
	
	Hyperspectral imaging consists in acquiring a set of images capturing a spatial scene at a few hundreds of wavelengths across the visible and near-infrared regions of the electromagnetic spectrum. The resulting data cube may equivalently be viewed as a set of two-dimensional (2D) gray-scale images, each corresponding to a particular spectral band, or as a collection of spectra, one per pixel of the image. In most cases, the image comprises a small number of pure materials, termed \emph{endmembers} or \emph{sources}, whose spectral signatures are mixed in each pixel. Spectral unmixing (SU) refers to the process of extracting the endmembers and estimating their corresponding mixing coefficients, or \emph{abundances}, for each pixel in the image~\cite{KES02,BIO12}.

In the \emph{linear mixing model} (LMM), a given pixel spectrum $\mathbf{x}^n$ in the image $\mathbf{x}$ can be expressed as a linear combination of the pure spectra $\{ \mathbf{s}^1, \dots, \mathbf{s}^P\}$, weighted by \emph{abundance coefficients} $\mathbf{a}^n$ representing the contribution of each source to $\mathbf{x}^n$:
\begin{equation*}
	\mathbf{x}^n = \mathbf{S} \mathbf{a}^n
\end{equation*}
where each column of matrix $\mathbf{S}$ is a source spectrum, and $n$ denotes the pixel index. In some cases, the mixing process is instead known to be nonlinear~\cite{CHE13}. When incident light interacts with several endmembers before reaching the sensor (\emph{e.g.} in \emph{multilayered} configurations), the mixing process can be approximated as bilinear. When the mixing process occurs at a microscopic scale, the model exhibits a different type of nonlinearity~\cite{DOB13}. In recent years, a number of papers have turned to a 'data-driven' approach to represent hyperspectral images in a high-dimensional manifold~\cite{NGU12,LUN14}, which has the advantage of being robust over a wide range of mixing scenarios. The extent of nonlinear effects varies from a hyperspectral image to another, ranging from only a few pixels to the whole spatial scene. However, nonlinear SU methods typically exhibit a much higher computational complexity than their linear counterparts. Given the high dimensionality of hyperspectral data, the LMM remains widely used as the basis of many SU algorithms.

In many applications, sensors acquire the data at multiple time frames, yielding so-called \emph{multitemporal} hyperspectral images. Multitemporal imaging allows to capture the dynamics of the underlying processes in the scene. The price to pay for this additional diversity is an even greater computational load, which gives an additional incentive to model the mixing process as linear for each time frame of the data. However, independently performing linear SU on each time frame is not sufficient. A first problem of this approach is the standard \emph{permutation problem} which arises in blind source separation: the index of each extracted endmember changes from a time frame to another. More importantly, unmixing each time frame in a separate manner fails to exploit the temporal information in the data, and thus does not reach the true potential of multitemporal image processing.

Multitemporal hyperspectral imaging has garnered increasing interest in recent years, mainly focused on classification problems~\cite{MER14,HU15}. For instance, working in a manifold-learning based framework allows to jointly process two time frames of the data by \emph{aligning} the manifolds obtained at each time frame~\cite{TUI13,YAN11}, enabling the use of labels from a time frame to be transferred to another~\cite{MA10,YAN13}. Unfortunately, the resulting computational complexity can be a significant hurdle when processing many time frames. 

Dedicated methods in the field of multitemporal spectral unmixing have only begun to emerge~\cite{IOR14}. In this paper, we aim at providing a framework for modeling and efficient unmixing of a time series of hyperspectral images. 
\textcolor{black}{Specifically, the main contribution of this paper is a model for multitemporal hyperspectral images, based on the LMM to retain the low complexity of linear SU methods. The temporal information in the data is accounted for by making assumptions on the dynamics of the underlying source spectra and abundance maps, seen as latent variables. Based on this model, we also propose an efficient SU algorithm which jointly processes all time frames in the image, and recovers source spectra and abundance maps fitting the proposed model.}
We demonstrate the performance of the proposed approach on real multitemporal hyperspectral images.

The remainder of the paper is organized as follows. In section \ref{sec:model}, we introduce and discuss our model for mutlitemporal hyperspectral images. We derive a joint spectral unmixing algorithm in section \ref{sec:optim}. The performances of the proposed approach are evaluated on synthetic and real multitemporal hyperspectral images in section \ref{sec:exp}. Finally, we conclude in section \ref{sec:ccl}.

\section{Proposed model for multitemporal hyperspectral images}
	\label{sec:model}

Let $\underline{\mathbf{X}} = \{ \mathbf{X}_k, k \in [\![1, K]\!] \}$ denote a time sequence of $K$ hyperspectral images, where $k$ is the discrete time index. Each image $\mathbf{X}_k$ is encoded in a $L \times N$ matrix where $L$ and $N$ respectively (\emph{resp.}) denote the number of wavelengths (\emph{channels}) and the number of pixels in the image. In the following, $\mathbf{x}_k^n$ denotes the $n$-th column of $\mathbf{X}_k$, \emph{i.e.} the observed spectrum of the $n$-th pixel of image $k$, and similar notations are used for other matrices. Equations involving indices $k$, $n$, $l$, $p$ will resp. hold for $k \in [\![1, K]\!], n \in [\![1, N]\!], l \in [\![1, L]\!], p \in [\![1, P]\!]$ unless stated otherwise.

At each time $k$, we suppose that the standard linear mixing model holds, \emph{i.e.} each pixel spectrum is obtained as a linear combination of $P$ pure spectra or \emph{sources}~\cite{BIO12}:
\begin{equation}
	\label{model}
	\mathbf{X}_k = \mathbf{S}_k \mathbf{A}_k + \mathbf{E}_k
\end{equation}
where $L \times P$ matrix $\mathbf{S}_k$ gathers the spectral signatures of the $P$ sources, and $P \times N$ abundance matrix $\mathbf{A}_k$ contains the mixing coefficients. Matrix $\mathbf{E}_k$ is an additive noise term which accounts for both measurements and model errors. The entries in $\underline{\mathbf{S}} = \{ \mathbf{S}_k \}$ and $\underline{\mathbf{A}} = \{ \mathbf{A}_k \}$ are known to be positive. \textcolor{black}{The number of sources $P$ is assumed to be known, \emph{e.g.} using one of the methods presented in~\cite{CHA04,BIO08,LUO13}. Since these methods are designed to be applied imagewise, their application will result in a vector of $K$ values of $P$. The number of sources will be fixed as the maximum value of this vector. We will see later that the proposed model contains a parameter which allows to deal with the appearance or disappearance of a particular endmember.}

The set of sources $\underline{\mathbf{S}}$ and their corresponding abundance coefficients $\underline{\mathbf{A}}$ may readily be obtained in a separate manner, that is, by running standard spectral unmixing algorithms at each time $k$. However, this approach may be viewed as flawed because it does not account for the dynamics of sources and abundance coefficients. Indeed, while the spectral signatures of sources extracted from the same spatial scenes at different times may not be strictly identical, we can expect them to bear some resemblance to each other. This similarity can be captured by modeling the dynamic \emph{spectral variability} of the sources. Likewise, the abundance maps of sources extracted at neighboring time frames should be highly similar in most cases. \textcolor{black}{In mathematical terms, we propose the following dynamic system:}
\textcolor{black}{
\begin{align}
	\label{general_dynamic_model}
	& \left\{ \begin{array}{ll} \mathbf{X}_{k} &= \mathbf{S}_{k} \mathbf{A}_{k} + \mathbf{E}_{k}\\
	\mathbf{S}_{k} &= f_S (\mathbf{S}_{k-1}) + \mathbf{V}_{k}\\
	\mathbf{A}_{k} &= f_A (\mathbf{A}_{k-1}) + \mathbf{D}_{k} \end{array} \right.
\end{align}}
\textcolor{black}{where functions $f_S$ and $f_A$ and noise terms $\underline{\mathbf{E}}$, $\underline{\mathbf{V}}$ and $\underline{\mathbf{D}}$ must be tailored to the problem at hand. Since assumptions are only made on the dynamics of the sources and abundance maps, rather than on the data themselves, our approach may be thought of as 'data-driven'.}

\textcolor{black}{We now propose a simplified version of \eqref{general_dynamic_model}} that is both general enough to encompass a wide range of situations and imaging scenarios, and tractable to accommodate the very large number of variables involved in multitemporal hyperspectral imaging. 

The spectral shape of various instances of a source spectral signature is known to be mainly invariant~\cite{SHA03}. Based on this rationale, \cite{NAS05} models the \emph{spectral variability} of sources in the spatial dimension (from one pixel to another) by a scale change and an additive noise term. We propose here to use a similar model to characterize the spectral variability of sources in the temporal dimension:
\begin{equation}
	\label{spectrum}
	\mathbf{s}_k^p = \psi_k^p \mathbf{s}_0^p + \mathbf{v}_k^p
\end{equation}
where $\psi_k^p$ is a (scalar) nonnegative scale factor, $\mathbf{s}_0^p$ is the \emph{reference} spectral signature for the $p$-th source and $\mathbf{v}_k^p$ is a zero-mean additive noise term accounting for a nonlinear distortion. For each source, any of the following methods can be employed to obtain $\mathbf{s}_0^p$:
\begin{itemize}
	\item selection in a dictionary when ground truth is available, or when the pure components in the image are known beforehand. Pruning methods may be applied to select the atoms of the dictionary~\cite{IOR14};
	\item polynomial regression from a bundle of extracted spectra, if one wishes to compute average reference spectra;
	\item or simply setting the reference spectrum to the one extracted independently from the first image, \emph{i.e.} $\mathbf{s}_0^p = \mathbf{s}_1^p$.
\end{itemize}

\textcolor{black}{We make the important assumption that all images in the temporal sequence have been either acquired with a fixed sensor or coregistered beforehand, \emph{e.g.} using methods based on cross-correlation~\cite{SCA92}.} Since each image is acquired over the same spatial scene, we assume that the abundance maps are given by
\begin{equation}
	\label{abund}
	\mathbf{a}_k^n = \mathbf{a}_{k-1}^n + \mathbf{d}_k^n
\end{equation}
where the noise term $\mathbf{d}_k^n$ models a potential \emph{change} in the spatial distribution (\emph{e.g.} the replacement of a crop by another in agricultural remote sensing, or the displacement of a molecule within the medium in biological spectroscopy). Each entry of $\mathbf{d}_k^n$ is non-zero only if the abundance on the corresponding endmember changes in the pixel; hence, we assume that $\mathbf{D}_k$ is sparse, and consequently $\underline{\mathbf{D}}$ too (see~\cite{CHE14} for a similar assumption in the spatial dimension of the image). \footnote{\textcolor{black}{Since the normalization step would negate the sparse structure of the difference matrices $\underline{\mathbf{D}}$, we do not apply the sum-to-one constraint on the abundance maps.}}

Gathering \eqref{model}, \eqref{spectrum} and \eqref{abund} in matrix from, we obtain the \textcolor{black}{simplified} dynamic system
\begin{align}
	\label{dynamic_model}
	& \left\{ \begin{array}{ll} \mathbf{X}_{k} &= \mathbf{S}_{k} \mathbf{A}_{k} + \mathbf{E}_{k}\\
	\mathbf{S}_{k} &= \mathbf{S}_0 \boldsymbol{\psi}_{k} + \mathbf{V}_{k}\\
	\mathbf{A}_{k} &= \mathbf{A}_{k-1} + \mathbf{D}_{k} \end{array} \right.
\end{align}
where $\boldsymbol{\psi}_k$ is a diagonal matrix with the main diagonal gathering the values $(\psi_k^1, \dots, \psi_k^P)$.\\

\textcolor{black}{In the next three subsections, we briefly consider the proposed model from different perspectives, in terms of potential links with the tensor decompositions and the standard state-space models. We then discuss the applicability of the approximation \eqref{dynamic_model} to various real life datasets, in terms of acceptable limits on spectral invariability.}

\subsection{Similarity to tensor decompositions}

Since $\mathbf{\underline{X}}$ is a tensor, model \eqref{dynamic_model} can be viewed as a tensor decomposition. Specifically, if one cancels the additive noise term in \eqref{spectrum}, equations \eqref{dynamic_model} and \eqref{spectrum} combine to write
\begin{equation}
	\label{tensor}
	\mathbf{X}_k = \mathbf{S}_0 \boldsymbol{\psi}_{k} \mathbf{A}_k + \mathbf{E}_k
\end{equation}
which is known as the \emph{nonnegative tensor factorization 1} (NTF1) under nonnegativity constraints on $\mathbf{S}_0$ and $\mathbf{\underline{A}}$~\cite{CIC09}. Here, spectral variability is only accounted for by a varying scale factor at each time frame. Model $\eqref{dynamic_model}$ thus corresponds to a modified version of NTF1, which relaxes the factorization structure and imposes additional constraints on the dynamics on the abundance maps given by \eqref{abund}.

\subsection{Similarity to state-space models}

Another interpretation of model \eqref{dynamic_model} lies in the framework of state-space representations. Here, observations $\mathbf{\underline{X}}$ are viewed as the output of a discrete time system with unknown internal states $\{ \mathbf{\underline{S}}, \mathbf{\underline{A}}, \boldsymbol{\underline{\psi}} \}$. In control theory terms, the first equation of \eqref{dynamic_model} would be referred to as the \emph{measurement model}, and the second and third equations rewrite as \emph{process models}:
\begin{align*}
	& \left\{ \begin{array}{ll}
	\mathbf{S}_k = \mathbf{S}_{k-1} \boldsymbol{\hat{\psi}}_k + \mathbf{\hat{V}}_k\\
	\mathbf{A}_k = \mathbf{A}_{k-1} + \mathbf{D}_k \end{array} \right.
\end{align*}
using straightforward recursive computations. Because the measurements $\mathbf{\underline{X}}$ and the update of $\mathbf{\underline{S}}$ are both nonlinear functions of the state variables, standard Kalman filtering does not apply to the model. However, we will similarly aim at recovering the unknown variables $\{ \mathbf{\underline{S}}, \mathbf{\underline{A}}, \boldsymbol{\underline{\psi}} \}$ based on the whole time series of images, rather than by considering single images~\cite{KAL60}.\\

\subsection{Validity of the simplified dynamic system}

\textcolor{black}{A key assumption of the simplified model \eqref{dynamic_model} is that endmembers have mainly invariant spectral shape. This is a strong assumption, which works well \emph{e.g.} when the data sequence is acquired with a high temporal resolution, as demonstrated in Section \ref{sec:real} on images capturing the release of a gas plume. Other typical applications with high temporal resolution include sequences of multispectral images acquired in fluorescence microscopy / spectroscopy. In other practical applications such as seasonally varying vegetation over the course of weeks or months, the proposed model may not be valid and adjustments must be made to fit the underlying physical models. When crops are replaced with bare soil, for instance, one simple adjustment may consist in introducing an additional endmember in the model. If information on the seasonal $\mathbf{S}_k$'s is available via physical priors, the second equation of the model can be modified in a straightforward manner to allow variations around the nominal values instead of using a fixed matrix $\mathbf{S}_0$. Additionally, other priors may be assumed on the difference matrices $\underline{\mathbf{D}}$ depending on the spatial layout of the crops.}

\textcolor{black}{As for spectral variations due to variable illumination and environmental, atmospheric and temporal conditions, more complex changes may be needed. In~\cite{ZAR14}, the authors characterize methods accounting for spectral variability by either treating endmembers as sets or \emph{bundles} of spectra, or as statistical distributions. If spectral variations of the endmembers belong to a known spectral library encoded in matrix $\mathbf{S}_0$, one could extract the relevant spectral signatures at different time frames using a variation of the MESMA algorithm~\cite{ROB98}. Another approach, belonging to the second class of methods, would consist in modeling each pixel as a linear combination of random endmembers following a Gaussian distribution as in~\cite{HAL15}. Since deriving closed-from minimizers is difficult in this context, the optimization procedure would then require using a \emph{Markov Chain Monte Carlo} approach which generates samples asymptotically distributed according to the joint posterior distribution of the unknown parameters. In both cases, one of the biggest challenges would consist in implementing efficient algorithms to deal with the high dimensionality of multitemporal hyperspectral images. We will see in the following section how it can be done in the context of the simplified model \eqref{dynamic_model}.}

\section{Joint spectral unmixing}
	\label{sec:optim}
	
We propose to estimate the unknown variables as a minimizer of the following objective function:
\DRAFT{
\begin{equation}
	\label{objective}
	{\cal J} (\mathbf{\underline{S}}, \mathbf{\underline{A}}, \boldsymbol{\underline{\psi}} | \mathbf{\underline{X}}) = \frac{1}{2} \sum_{k=1}^{K} \| \mathbf{X} _k - \mathbf{S} _k \mathbf{A} _k \|_F^2 + \frac{\lambda_S}{2} \sum_{k=1}^{K} \| \mathbf{S}_k - \mathbf{S}_0 \boldsymbol{\psi}_k \|_F^2 + \lambda_A \sum_{k=2}^{K} \| \mathbf{A} _k - \mathbf{A}_{k-1} \|_{\ell_1}
\end{equation}}
\FINAL{
\begin{align}
	\label{objective}
	&{\cal J} (\mathbf{\underline{S}}, \mathbf{\underline{A}}, \boldsymbol{\underline{\psi}} | \mathbf{\underline{X}}) = \frac{1}{2} \sum_{k=1}^{K} \| \mathbf{X} _k - \mathbf{S} _k \mathbf{A} _k \|_F^2 \nonumber \\ 
	& + \frac{\lambda_S}{2} \sum_{k=1}^{K} \| \mathbf{S}_k - \mathbf{S}_0 \boldsymbol{\psi}_k \|_F^2 + \lambda_A \sum_{k=2}^{K} \| \mathbf{A} _k - \mathbf{A}_{k-1} \|_{\ell_1}
\end{align}}
where $\|.\|_F$ and $\|.\|_{{\ell_1}}$ resp. denote the Frobenius and $\ell_1$ norm, and the notation ${\cal J} (\mathbf{\underline{S}}, \mathbf{\underline{A}}, \boldsymbol{\underline{\psi}} | \mathbf{\underline{X}})$ states that the objective function is minimized w.r.t. the unknown variables $\{ \mathbf{\underline{S}}, \mathbf{\underline{A}}, \boldsymbol{\underline{\psi}} \}$ given the observations $\mathbf{\underline{X}}$. $\lambda_S$ and $\lambda_A$ are scalar regularization hyperparameters which control the weight of the different terms in \eqref{objective}. The first term in \eqref{objective} measures the fitness of the data to the model, and the second and third term allow to enforce the prior information of \eqref{dynamic_model}.

\textcolor{black}{Before detailing our optimization strategy, we propose a way to set hyperparameters $\lambda_S$ and $\lambda_A$ in criterion \eqref{objective} from a probabilistic standpoint.}

\subsection{Hyperparameter tuning}
	\label{sec:proba}
	
	\textcolor{black}{In a Bayesian framework, criterion \eqref{objective} may be viewed as the \emph{maximum a posteriori} (MAP) estimator of the unknown variables $\{ \mathbf{\underline{S}}, \mathbf{\underline{A}}, \boldsymbol{\underline{\psi}} \}$ given the observations $\mathbf{\underline{X}}$. Hence, by making the following assumptions
\begin{enumerate}
\item all entries of tensor $\mathbf{\underline{E}}$ are i.i.d. centered Gaussian random variables of variance $\sigma_e^2$;
\item all entries of tensor $\mathbf{\underline{V}}$ are i.i.d. centered Gaussian random variables of variance $\sigma_v^2$;
\item all entries of tensor $\mathbf{\underline{W}}$ are i.i.d. centered Laplacian random variables of scale $b$,
\end{enumerate}
standard computations yield the following values:
\begin{align}
& \left\{ \begin{array}{ll} 
\textcolor{black}{\lambda_S} &\textcolor{black}{= \sigma_e^2 / \sigma_v^2}\\
\textcolor{black}{\lambda_A} &\textcolor{black}{= \sigma_e^2 / b.}
\end{array} \right.
\label{eq:tuning}
\end{align}
Equation \eqref{eq:tuning} allows to set parameters $\lambda_S$ and $\lambda_A$ providing $\sigma_e$, $\sigma_v$ and $b$ are known or can be estimated from the data set at hand. When this is not the case, $\lambda_S$ and $\lambda_A$ can be estimated using the approach recently proposed in~\cite{SON15}, or by a sub-optimal strategy of trial and error.}

In the next section, we detail an optimization procedure to minimize this criterion. 

\subsection{Optimization}

Since the number of variables involved is very large, we propose to minimize \eqref{objective} using an \emph{projected alternated least squares} strategy or \emph{alternating nonnegative least squares} (ANLS)~\cite{BER07} similarly to the approach of~\cite{CIC07}. The method consists in alternatively solving the following problems
\begin{align}
\left\{ \begin{array}{ll} & \min_{\mathbf{\underline{S}}} {\cal J} (\mathbf{\underline{S}}, \mathbf{\underline{A}}, \boldsymbol{\underline{\psi}} ; \lambda_A, \lambda_S) \emph{ s.t. } \mathbf{\underline{S}} \geqslant 0\\
& \min_{\mathbf{\underline{A}}} {\cal J} (\mathbf{\underline{S}}, \mathbf{\underline{A}}, \boldsymbol{\underline{\psi}} ; \lambda_A, \lambda_S) \emph{ s.t. } \mathbf{\underline{A}} \geqslant 0\\
& \min_{\boldsymbol{\underline{\psi}}} {\cal J} (\mathbf{\underline{S}}, \mathbf{\underline{A}}, \boldsymbol{\underline{\psi}} ; \lambda_A, \lambda_S)
\end{array} \right.
\end{align}
until some stopping criterion is satisfied. \textcolor{black}{We now detail the resolution of each subproblem.}\\

\subsubsection{Minimization w.r.t. $\underline{\mathbf{S}}$}

\textcolor{black}{We use the \emph{Alternated Direction Method of Multipliers} (ADMM)~\cite{BOY11} to tackle the nonnegativity contraints, in which the minimization of \eqref{objective} w.r.t. $\{ \mathbf{S}_k, k = 1, \dots, K \}$ is rewritten using a set $\underline{\mathbf{M}}$ of auxiliary matrices $\{ \mathbf{M}_k, k = 1, \dots, K \}$ of size $L \times P$:
\begin{align}
	\label{problem_admm_S}
	\min_{\underline{\mathbf{S}},\underline{\mathbf{M}}} & \sum_{k=1}^{K} \frac{1}{2} \bigg ( \| \mathbf{X} _k - \mathbf{S} _k \mathbf{A} _k \|_F^2   + \frac{\lambda_S}{2} \sum_{k=1}^{K} \| \mathbf{S}_k - \mathbf{S}_0 \boldsymbol{\psi}_k \|_F^2 \nonumber \\
	& + I_{(\mathbb{R}^{+})^{L \times P}} (\mathbf{M}_k) \bigg )\nonumber \\
	& \emph{ s.t. } \mathbf{S} _k - \mathbf{M}_{k} = \mathbf{0}, k \in \{ 1, \dots, K \}
\end{align}
where $ I_{(\mathbb{R}^{+})^{L \times P}}$ denotes the indicator function of the positive orthant of $\mathbb{R}^{L \times P}$. The augmented Lagrangian of the problem is given by
\DRAFT{
\begin{equation}
	\label{lagrangian}
	{\cal L}_{\rho} = \sum_{k=1}^{K} \frac{1}{2} \bigg ( \| \mathbf{X} _k - \mathbf{S} _k \mathbf{A} _k \|_F^2  +  \frac{\lambda_S}{2} \sum_{k=1}^{K} \| \mathbf{S}_k - \mathbf{S}_0 \boldsymbol{\psi}_k \|_F^2 + I_{(\mathbb{R}^{+})^{L \times P}} (\mathbf{M}_k) +  \frac{\rho}{2} \| \mathbf{S} _k - \mathbf{M}_{k} + \mathbf{U}_k \|_2^{2} - \frac{\rho}{2} \| \mathbf{U}_k \|_F^2 \bigg )
\end{equation}}
\FINAL{
\begin{align}
	\label{lagrangian}
& {\cal L}_{\rho} =  \sum_{k=1}^{K} \frac{1}{2} \bigg ( \| \mathbf{X} _k - \mathbf{S} _k \mathbf{A} _k \|_F^2  +  \frac{\lambda_S}{2} \sum_{k=1}^{K} \| \mathbf{S}_k - \mathbf{S}_0 \boldsymbol{\psi}_k \|_F^2 \nonumber \\
& + I_{(\mathbb{R}^{+})^{L \times P}} (\mathbf{M}_k) + \frac{\rho}{2} \| \mathbf{S} _k - \mathbf{M}_{k} + \mathbf{U}_k \|_2^{2} - \frac{\rho}{2} \| \mathbf{U}_k \|_F^2 \bigg )
\end{align}}
where $\rho$ is called the \emph{barrier parameter} and $\underline{\mathbf{U}} = \{ \mathbf{U}_k, k = 1 \dots K \}$ are the so-called \emph{normalized Lagrange multipliers}, which are scaled by a factor of $\rho$ so that the augmented Lagrangian only incorporates quadratic terms. The procedure then consists in alternatively minimizing ${\cal L}_{\rho}$ w.r.t. $\underline{\mathbf{S}}$ and $\underline{\mathbf{M}}$ and updating the normalized Lagrange multipliers $\underline{\mathbf{U}}$ (\emph{dual update}), until the stopping criteria based on the primal and dual residuals are satisfied~\cite{BOY11}. The minimization w.r.t. $\underline{\mathbf{S}}$ is performed by cancelling the gradients of ${\cal L}_{\rho}$, while the minimization w.r.t. $\underline{\mathbf{M}}$ uses the frameworks of \emph{proximal operators}\footnote{\textcolor{black}{The \emph{proximal operator} of a function $h$ with penalty $\rho$ is defined as $$\text{prox}_{h,\rho} (t) \overset{\Delta}{=} \underset{z}{\text{arg min}} \{ h(z) + (\rho/2) (z - t)^2\}.$$}} and the dual update consists in maximizing the dual problem w.r.t. $\underline{\mathbf{U}}$:
\DRAFT{
\begin{align}
\label{S_update}
\left\{ \begin{array}{ll} 
\mathbf{S}_k &\leftarrow
(\mathbf{X}_k \mathbf{A}_k^T + \lambda_S \mathbf{S}_0 \boldsymbol{\psi}_k + \rho (\mathbf{M}_k - \mathbf{U}_k) ) (\mathbf{A}_k \mathbf{A}_k^T + (\lambda_S + \rho) \mathbf{I}_{P})^{-1} \\
\mathbf{M}_k \leftarrow
\PI_{(\mathbb{R}^{+})^{L \times P}} ( \mathbf{S}_k + \mathbf{U}_k) \\
\mathbf{U}_K &\leftarrow
\mathbf{U}_k + \mathbf{S}_k - \mathbf{M}_k.
 \end{array} \right.
\end{align}}
\FINAL{
\begin{align}
\label{S_update}
\left\{ \begin{array}{ll} 
\mathbf{S}_k &\leftarrow
(\mathbf{X}_k \mathbf{A}_k^T + \lambda_S \mathbf{S}_0 \boldsymbol{\psi}_k + \rho (\mathbf{M}_k - \mathbf{U}_k) )\\ & (\mathbf{A}_k \mathbf{A}_k^T + (\lambda_S + \rho) \mathbf{I}_{P})^{-1} \\
\mathbf{M}_k &\leftarrow
\Pi_{(\mathbb{R}^{+})^{L \times P}} ( \mathbf{S}_k + \mathbf{U}_k) \\
\mathbf{U}_K &\leftarrow
\mathbf{U}_k + \mathbf{S}_k - \mathbf{M}_k
 \end{array} \right.
\end{align}}
where $\mathbf{I}_P$ is the $P \times P$ identity matrix and the projector onto $(\mathbb{R}^{+})^{L \times P}$, denoted by $\Pi_{(\mathbb{R}^{+})^{L \times P}}$, is the proximal operator of $I_{(\mathbb{R}^{+})^{L \times P}}$. Note that all updates have a closed-form expression and the computational complexity is dominated by cheap inversions of $P \times P$ matrices.}

\subsubsection{Minimization w.r.t. $\underline{\mathbf{A}}$}

\textcolor{black}{Likewise, the $\ell_1$ penalty term suggests the use of the ADMM. We introduce two sets of auxiliary matrices: $\{ \mathbf{Q}_k, k = 1, \dots, K \}$ of size $P \times N$ and $\{ \mathbf{D}_k, k = 2, \dots, K \}$ of size $P \times N$ and rewrite the problem as
\begin{align}
	\label{problem_admm}
	\min_{\underline{\mathbf{A}}, \underline{\mathbf{D}}, \underline{\mathbf{Q}}} & \sum_{k=1}^{K} \frac{1}{2} \bigg ( \| \mathbf{X} _k - \mathbf{S} _k \mathbf{A} _k \|_F^2 + I_{(\mathbb{R}^{+})^{P \times N}} (\mathbf{Q}_k) \bigg )   \nonumber \\
	&  + \lambda_A \sum_{k=2}^{K} \| \mathbf{D} _k \|_{\ell_1}  \nonumber \\
	& \emph{ s.t. } \mathbf{A} _k - \mathbf{A}_{k-1} - \mathbf{D}_k = \mathbf{0}, k \in \{ 2, \dots, K \}\nonumber \\
	& \emph{ s.t. } \mathbf{A} _k - \mathbf{Q}_{k} = \mathbf{0}, k \in \{ 1, \dots, K \}.
\end{align}
The augmented Lagrangian of the problem is
\DRAFT{
\begin{equation}
	\label{lagrangian}
	{\cal L}_{\rho} = \sum_{k=1}^{K} \frac{1}{2} \bigg ( \| \mathbf{X} _k - \mathbf{S} _k \mathbf{A} _k \|_F^2 + I_{(\mathbb{R}^{+})^{P \times N}}(\mathbf{Q}_k) \bigg + \frac{\rho}{2} \| \mathbf{A} _k - \mathbf{Q}_{k} + \mathbf{W}_k \|_2^{2} - \frac{\rho}{2} \| \mathbf{W}_k \|_F^2) + \sum_{k=2}^{K} \bigg ( \lambda_A \| \mathbf{D} _k \|_{\ell_1} + \frac{\rho}{2} \| \mathbf{A} _k - \mathbf{A}_{k-1} - \mathbf{D}_k + \mathbf{Z}_k \|_2^{2} - \frac{\rho}{2} \| \mathbf{Z}_k \|_F^2 \bigg )
\end{equation}}
\FINAL{
\begin{align}
	\label{lagrangian}
	&{\cal L}_{\rho} = \sum_{k=1}^{K} \frac{1}{2} \bigg ( \| \mathbf{X} _k - \mathbf{S} _k \mathbf{A} _k \|_F^2 + I_{(\mathbb{R}^{+})^{P \times N}}(\mathbf{Q}_k) \nonumber\\
	&  + \frac{\rho}{2} \| \mathbf{A} _k - \mathbf{Q}_{k} + \mathbf{W}_k \|_2^{2} - \frac{\rho}{2} \| \mathbf{W}_k \|_F^2 \bigg ) \nonumber\\
	&+ \sum_{k=2}^{K} \bigg ( \lambda_A \| \mathbf{D} _k \|_{\ell_1} + \frac{\rho}{2} \| \mathbf{A} _k - \mathbf{A}_{k-1} - \mathbf{D}_k + \mathbf{Z}_k \|_2^{2} - \frac{\rho}{2} \| \mathbf{Z}_k \|_F^2 \bigg )
\end{align}}
where $\underline{\mathbf{W}}$ and $\underline{\mathbf{Z}}$ are the normalized Lagrange multipliers associated to both sets of constraints. The procedure is similar to the one exposed in the previous section and yields the following updates w.r.t. $\underline{\mathbf{A}}$:
\begin{align}
\label{A_update}
\left\{ \begin{array}{ll} 
\mathbf{A}_1 &\leftarrow
(\mathbf{S}_1^t \mathbf{S}_1 + 2 \rho \mathbf{I}_P)^{-1} \\
& (\mathbf{S}_1^t \mathbf{X}_1 + \rho (\mathbf{A}_2 - \mathbf{D}_2 + \mathbf{Z}_2 + \mathbf{Q}_1 - \mathbf{W}_1)) \\
\mathbf{A}_{k} &\leftarrow
(\mathbf{S}_k^t \mathbf{S}_k + 3 \rho \mathbf{I}_P)^{-1} \\
& (\mathbf{S}_k^t \mathbf{X}_k + \rho (\mathbf{A}_{k+1} + \mathbf{A}_{k-1} - \mathbf{D}_{k+1} + \mathbf{D}_{k}\\
 & + \mathbf{Z}_{k+1} - \mathbf{Z}_k + \mathbf{Q}_k - \mathbf{W}_k), k \in \{ 2 \dots K-1 \}\\
\mathbf{A}_K &\leftarrow
(\mathbf{S}_K^t \mathbf{S}_K + 2 \rho \mathbf{I}_P)^{-1} \\
& (\mathbf{S}_K^t \mathbf{X}_K + \rho (\mathbf{A}_{K-1} + \mathbf{D}_K - \mathbf{Z}_K + \mathbf{Q}_K - \mathbf{W}_K).
 \end{array} \right.
\end{align}
Recalling that the proximal operator of the absolute value is the \emph{soft-thresholding} operator, the updates w.r.t. $\underline{\mathbf{D}}$ and $\underline{\mathbf{Q}}$ are:
\DRAFT{
\begin{align}
\label{DQ_update}
	\mathbf{D}_k &\leftarrow \textcolor{black}{\text{max} (\mathbf{A}_k - \mathbf{A}_{k-1} + \mathbf{U}_k - \lambda_A / \rho, 0) - \text{max} (- \mathbf{A}_k + \mathbf{A}_{k-1} - \mathbf{U}_k - \lambda_A / \rho, 0)}, \ k \in \{2, \dots, K\}.
\end{align}}
\FINAL{
\begin{align}
\label{DQ_update}
\left\{ \begin{array}{ll} 
	\mathbf{Q}_k &\leftarrow \Pi_{(\mathbb{R}^{+})^{P \times N}} ( \mathbf{A}_k + \mathbf{W}_k), \ k \in \{1, \dots, K\} \\
	\mathbf{D}_k &\leftarrow \text{max} (\mathbf{A}_k - \mathbf{A}_{k-1} + \mathbf{U}_k - \lambda_A / \rho, 0)  \\ 
	& - \text{max} (- \mathbf{A}_k + \mathbf{A}_{k-1} - \mathbf{U}_k - \lambda_A / \rho, 0), \ k \in \{2, \dots, K\}. \\
\end{array} \right.
\end{align}}
Finally, the dual updates are given by
\begin{align}
\label{WZ_update}
\left\{ \begin{array}{ll} 
\mathbf{W}_k & \leftarrow \mathbf{W}_k + \mathbf{A}_k - \mathbf{Q}_k, \ k \in \{1, \dots, K\}\\\
\mathbf{Z}_k & \leftarrow \mathbf{Z}_k + \mathbf{A} _k - \mathbf{A}_{k-1} - \mathbf{D}_k, \ k \in \{2, \dots, K\}.\\
 \end{array} \right.
\end{align}
Again, the computational complexity of all involved operations is very low.
}

\subsubsection{Minimization w.r.t. $\boldsymbol{\underline{\psi}}$}

Since $\boldsymbol{\psi}_k$ is a diagonal matrix, we only need to update the diagonal coefficients. The gradient of criterion \eqref{objective} w.r.t. coefficient $\psi_k^p$ is given by
\begin{align}
\label{gradientpsi} \frac{\partial {\cal J}}{\partial \psi_k^p} &=
(\mathbf{s}_0^p)' \mathbf{s}_0^p \psi_k^p - (\mathbf{s}_0^p)' \mathbf{s}_k^p
\end{align}
hence, the update rule is
\begin{align}
\label{psi_update} \psi_k^p &\leftarrow \frac{(\mathbf{s}_0^p)' \mathbf{s}_k^p}{(\mathbf{s}_0^p)' \mathbf{s}_0^p}.
\end{align}
Note that this update guarantees the nonnegativity of coefficients $\{ \psi_k, k = 1, \dots, K \}$. \textcolor{black}{It is worth noting at this point that the algorithm only iterates closed-form updates for all variables of interest. The method is summarized in the next section}.

\subsection{Algorithm outline}

Initialization of the unknown matrices $\{ \mathbf{S}_k, \mathbf{A}_k, k = 1, \dots, K \}$ can easily be carried out in a separate fashion, \emph{e.g.} using \emph{Vertex Component Analysis} (VCA)~\cite{NAS05b} or \emph{Minimum Volume Simplex Analysis} (MVSA)~\cite{LI08}, and \emph{Fully Constrained Least Squares} (FCLS)~\cite{HEI01}. The algorithm structure is displayed in table \ref{tab:algo}.

\begin{table}
 \removelatexerror
\begin{center}
\DRAFT{
\begin{tabular}{| p{.5\textwidth} |}}
\FINAL{
\begin{tabular}{| p{.4\textwidth} |}}
\hline
\begin{algorithm}[H]
\KwData{$\{ \mathbf{X}_k, k = 1, \dots, K \}$}
\KwResult{$\{ \mathbf{S}_k, \mathbf{A}_k, \boldsymbol{\psi}_k, k = 1, \dots, K \}$}
Set $\lambda_A$, $\lambda_S$, $\epsilon_S$, $\epsilon_A$, $\underline{\mathbf{S}}^{\text{init}}$, $\underline{\mathbf{A}}^{\text{init}}$, $\underline{\boldsymbol{\psi}}^{\text{init}}$\;
\end{algorithm}
\\
\hline
\begin{algorithm}[H]
\label{algo:HQ}
 \SetAlgoLined
  \Repeat{ALS stopping criterion is satisfied}{
\Repeat{ADMM stopping criterion is satisfied}{
  \textcolor{black}{Compute $\underline{\mathbf{S}}^{\text{new}}$, $\underline{\mathbf{M}}^{\text{new}}$ and $\underline{\mathbf{U}}^{\text{new}}$ using \eqref{S_update}}\;
  }
 \Repeat{ADMM stopping criterion is satisfied}{
  \textcolor{black}{Compute $\underline{\mathbf{A}}^{\text{new}}$ using \eqref{A_update}\;
  Compute $\underline{\mathbf{D}}^{\text{new}}$ and $\underline{\mathbf{Q}}^{\text{new}}$ using \eqref{DQ_update}\;
  Compute $\underline{\mathbf{W}}^{\text{new}}$ and $\underline{\mathbf{Z}}^{\text{new}}$using \eqref{WZ_update}\;}
  }
  Compute $\underline{\boldsymbol{\psi}}^{\text{new}}$ using \eqref{psi_update}\;}
   \caption{Joint spectral unmixing of multi\-temporal images}
   \end{algorithm}\\
   \hline
   \end{tabular}
   \caption{Proposed method for jointly unmixing hyperspectral multitemporal images.}
      \label{tab:algo}
   \end{center}
   \end{table}

\subsection{Remarks on the convergence of the method}

We propose the following stopping criteria:
\begin{align}
\frac{\sum_{k=1}^K \| \mathbf{A}_k^{\text{new}} - \mathbf{A}_k \|_F^2}{\sum_{k=1}^K \| \mathbf{A}_k \|_F^2} &< \epsilon_A\\
\frac{\sum_{k=1}^K \| \mathbf{S}_k^{\text{new}} - \mathbf{S}_k \|_F^2}{\sum_{k=1}^K \| \mathbf{S}_k \|_F^2} &< \epsilon_S,
\end{align}
\emph{i.e.} the optimization stops when the residuals on $\mathbf{\underline{S}}$ and $\mathbf{\underline{A}}$ are sufficiently small.

The optimization problem falls in the general class of \emph{Nonnegative Matrix Factorization} (NMF) problems, which are non convex and typically have local minima and non-uniqueness issues. \textcolor{black}{Moreover, since the ADMM steps only provide approximated minimizers of the sub-problems (if sufficient inner iteration numbers are used), only convergence to stationary points can be 'approximately' guaranteed}. However, it is well known that the incorporation of additional constraints into the NMF model is a way to obtain a more well-posed NMF problem~\cite{GIL12}. For instance, it has been shown that the non-negative problem admits a unique solution under well formulated conditions~\cite{MOU05}. While theoretical guarantees are out of the scope of this paper, we believe that the two regularization terms in \eqref{objective} also help constraining the range of admissible solutions. \textcolor{black}{This belief is confirmed by our experimental results, presented in the following section.}

\section{Experimental results}
	\label{sec:exp}
	
	In this section, we evaluate the performance of the proposed method, thereafter called 'joint unmixing'. We compare it to a 'separate unmixing' approach where each time frame is processed separately by first extracting endmembers using VCA~\cite{NAS05} or MVSA~\cite{LI08}, then estimating abundances using SUNSAL~\cite{BIO10}. \textcolor{black}{The comparison metric is the scaled mean square error (MSE), defined by}
\begin{align}
e_A &= \displaystyle \frac{\sum_{k=1}^K \| \mathbf{A}_k^{\text{est}} - \mathbf{A}_k^{\text{true}} \|_F^2}{\sum_{k=1}^K \| \mathbf{A}_k^{\text{true}} \|_F^2} \nonumber \\
e_S &= \displaystyle \frac{\sum_{k=1}^K \| \mathbf{S}_k^{\text{est}} - \mathbf{S}_k^{\text{true}} \|_F^2}{\sum_{k=1}^K \| \mathbf{S}_k^{\text{true}} \|_F^2}.
\label{eq:scaledMSE}
\end{align}

\subsection{Experiments on synthetic data}

\begin{table*}[ht!]
\center
\begin{tabular}{| c | c | c | c | c | c |}
\hline
& \textcolor{black}{Running time} & \textcolor{black}{$\text{mean}(e_S)$} & \textcolor{black}{$\text{std}(e_S)$} & \textcolor{black}{$\text{mean}(e_A)$} & \textcolor{black}{$\text{std}(e_A)$}\\
\hline
\textcolor{black}{Separate unmixing} & \textcolor{black}{15 sec } & \textcolor{black}{0.95} & \textcolor{black}{0.03} & \textcolor{black}{1.11} & \textcolor{black}{0.03}\\
\hline
\textcolor{black}{Joint unmixing} & \textcolor{black}{18 min} & \textcolor{black}{0.63} & \textcolor{black}{0.02} & \textcolor{black}{0.66} & \textcolor{black}{0.01}\\
\hline
\end{tabular}
\caption{\textcolor{black}{Mean and standard deviations (denoted by 'std') of the scaled mean square errors of the estimated spectra and abundance maps, respectively denoted by $e_S$ and $e_A$, defined by equation \eqref{eq:scaledMSE}.}}
\label{tab:synth_results}
\end{table*}

We consider a synthetic time series of ten hyperspectral images, composed of three endmembers. First, we design an abundance map for the first frame composed of three overlapping circular regions. Three real spectra corresponding to various gases are randomly picked from a dictionary (see next section), sampled on 129 wavelengths. We set spectral scale factors as one period of a sinusoid over the ten time frames. \textcolor{black}{Model \eqref{dynamic_model} is then applied based on the first image to create a time series of hyperspectral images, using sparse perturbation matrices $\underline{\mathbf{D}}$ generated using Laplacian distributions with $\sigma_e = \sigma_v = 5e^{-2}$ and $b = 1e^{-2}$ (these parameters are assumed to be known in these experiments).}

\textcolor{black}{In the case of separate unmixing, we compare the results yielded by VCA and MVSA, by testing values of the tuning parameters (spherization parameter and maximum eigenvalue of the quadratic approximation term) discretized on 15 points on a logarithmic scale from $10^{-10}$ to $10^4$. Since the true abundance maps contain pure pixels, VCA produced the best estimates of the endmembers in the mean square error (MSE) sense. Likewise, the impact of several values of the regularization parameter of SUNSAL was tested with a similar scheme. The resulting sources are sorted out by minimizing the spectral angle distance between the estimated matrix and matrix $\mathbf{S}_0$ over all possible permutations. In the case of joint unmixing, the algorithm is initialized by unity scale factors and constant abundance maps and $\mathbf{S}_0$ is fixed to the true endmember matrix. Note that here, contrarily to the separate unmixing approach, the proposed algorithm does not suffer from the permutation ambiguity: that is, the structure of the problem forces the index for each physical source to be the same at each time frame. Hyperparameters $\lambda_S$ and $\lambda_A$ are set using the probabilistic interpretation of section \ref{sec:proba}.}

\begin{figure*}[ht!]
\center
	\begin{tabular}{ccc}
	\multicolumn{3}{c}{Time 10}\\
	Source 1 & Source 2 & Source 3\\
	\multicolumn{3}{c}{Separate unmixing}\\
	\includegraphics[width=.23\textwidth,height=.1\textheight]{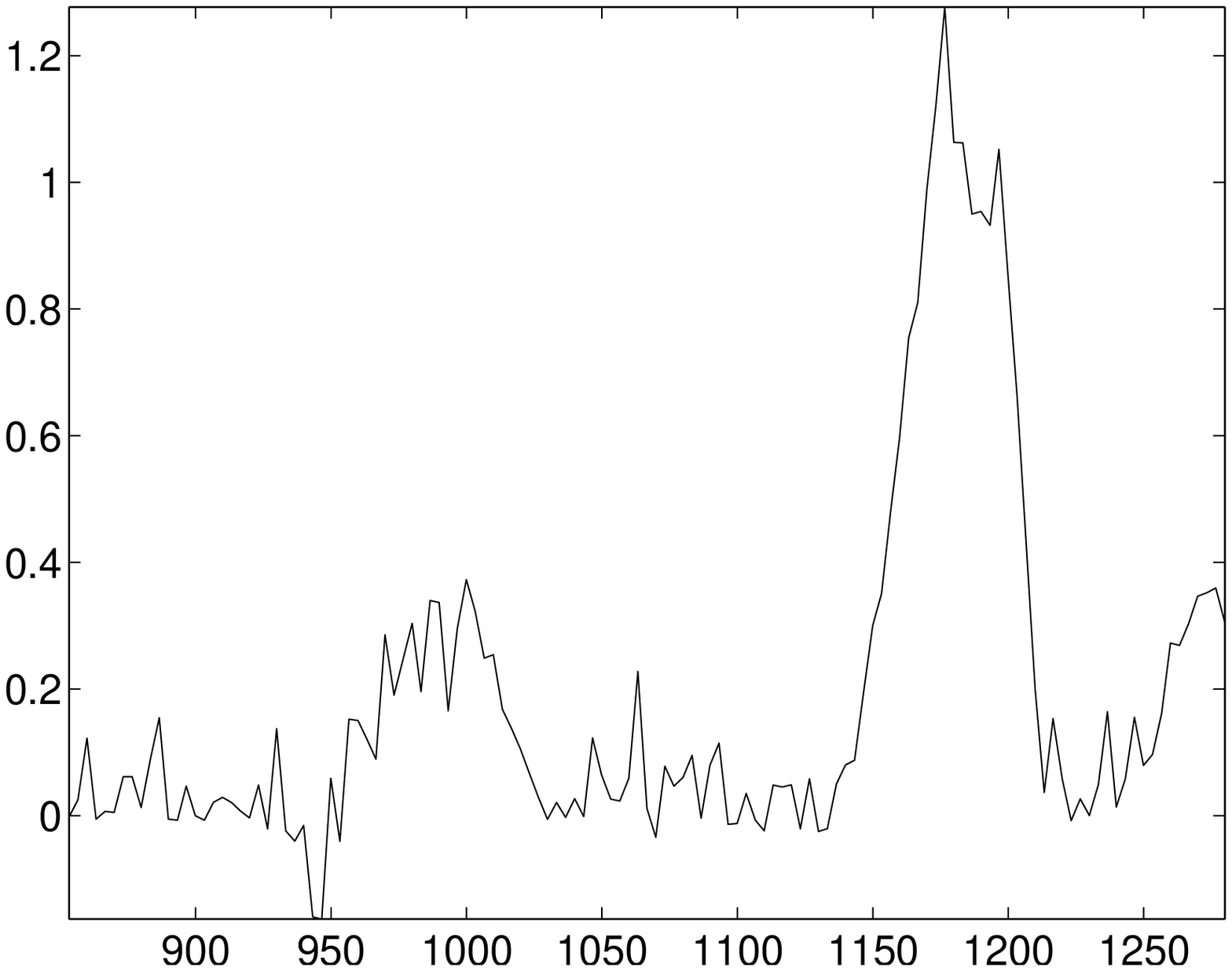}&
	\includegraphics[width=.23\textwidth,height=.1\textheight]{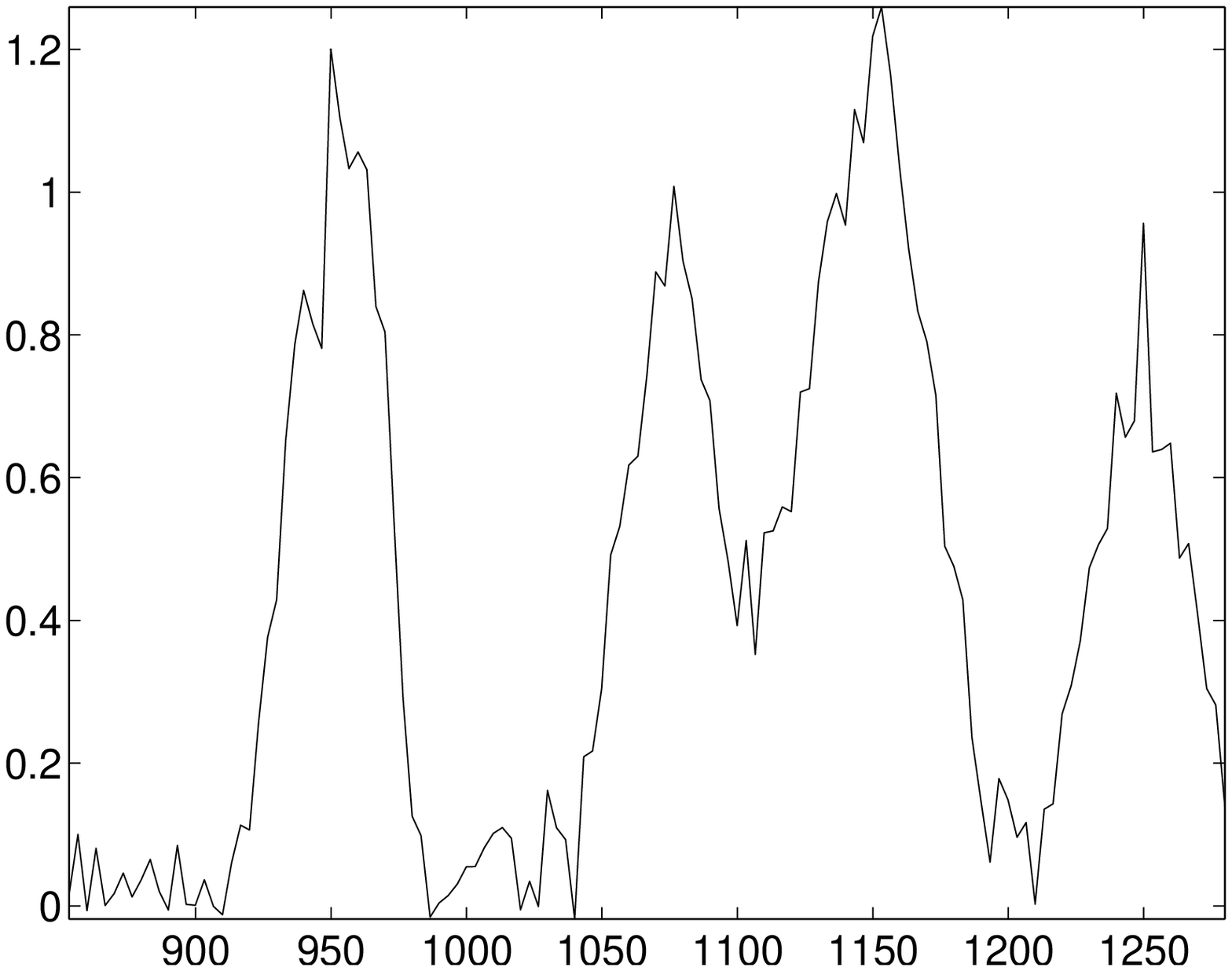}&
	\includegraphics[width=.23\textwidth,height=.1\textheight]{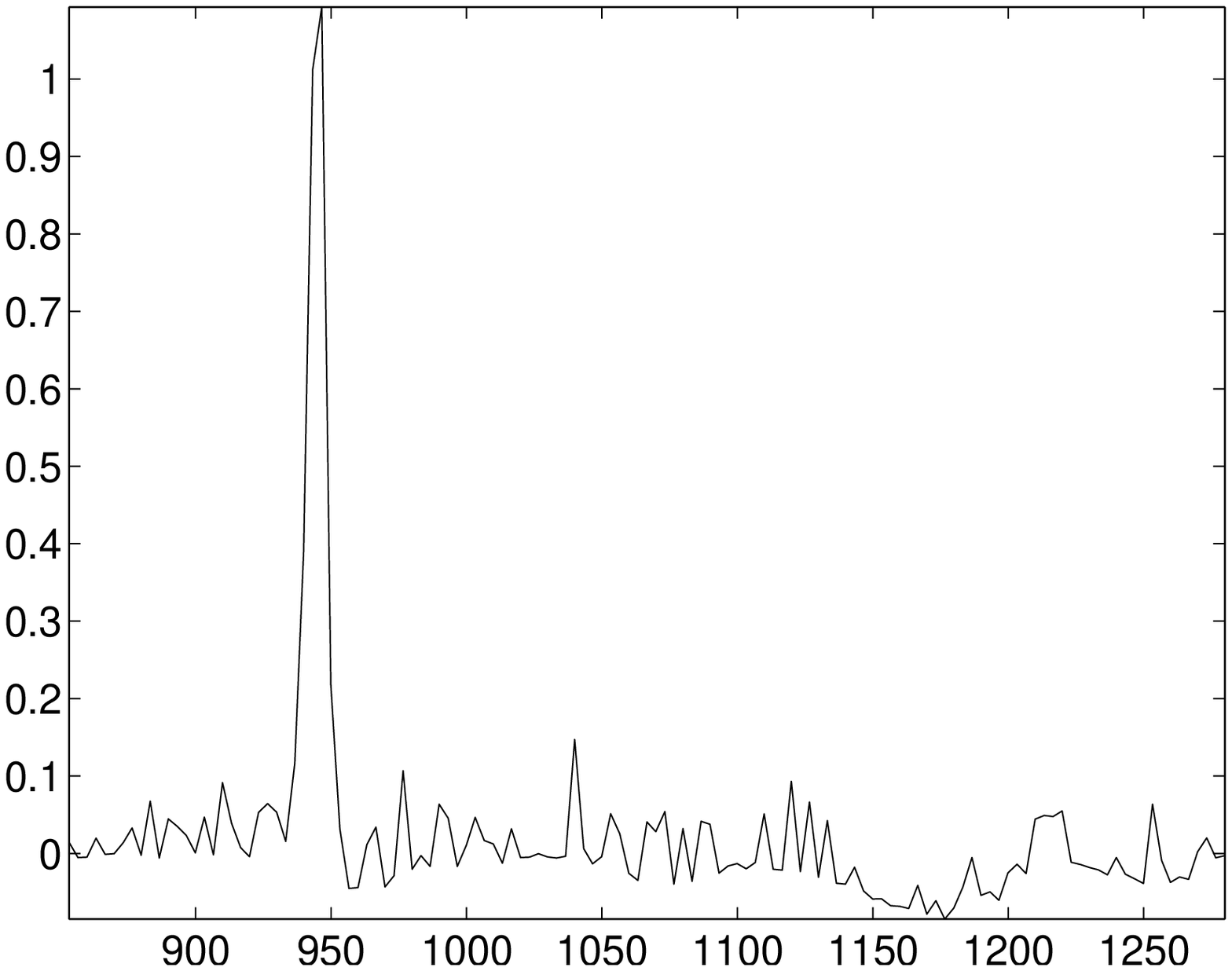}\\
	\includegraphics[width=.23\textwidth]{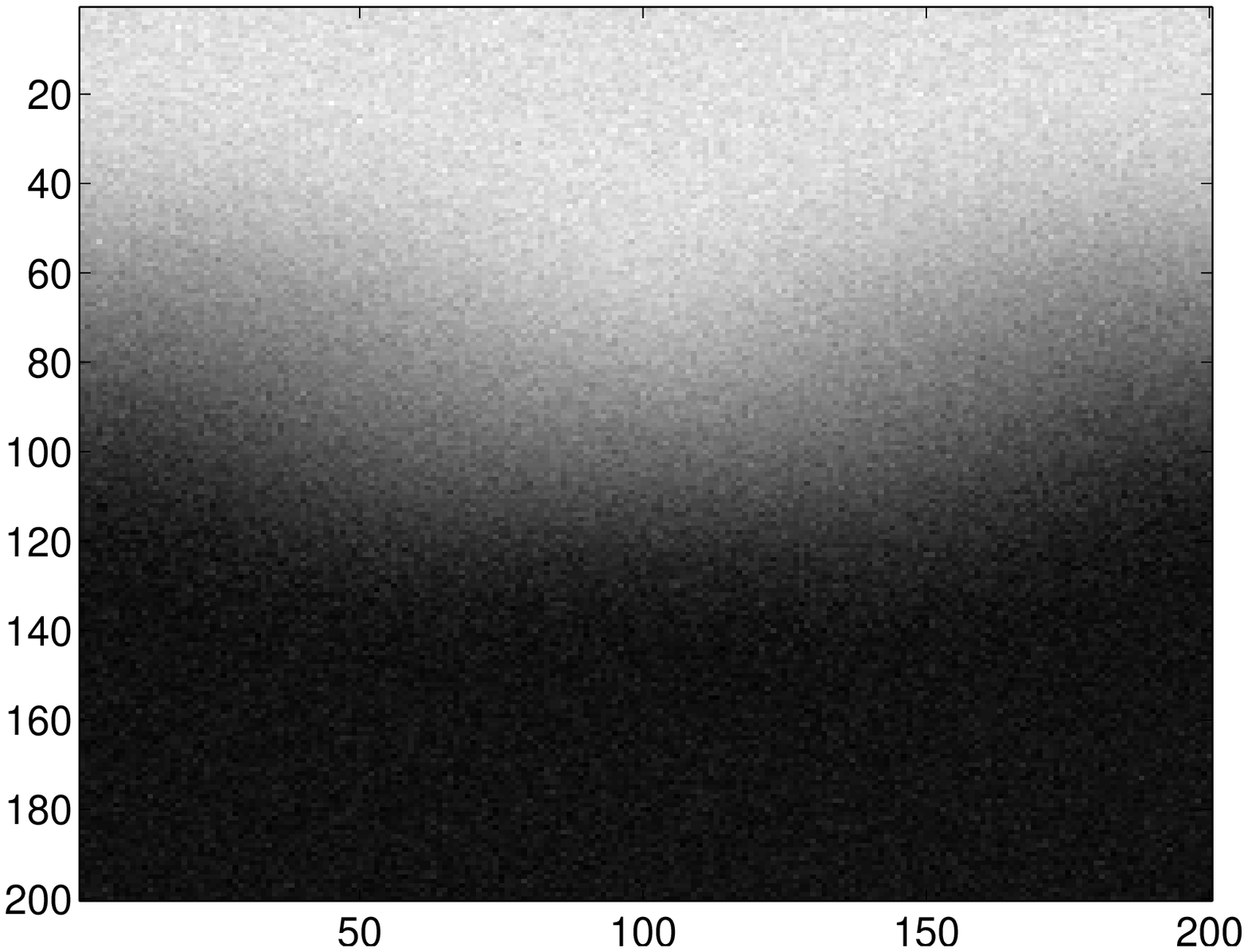}&
	\includegraphics[width=.23\textwidth]{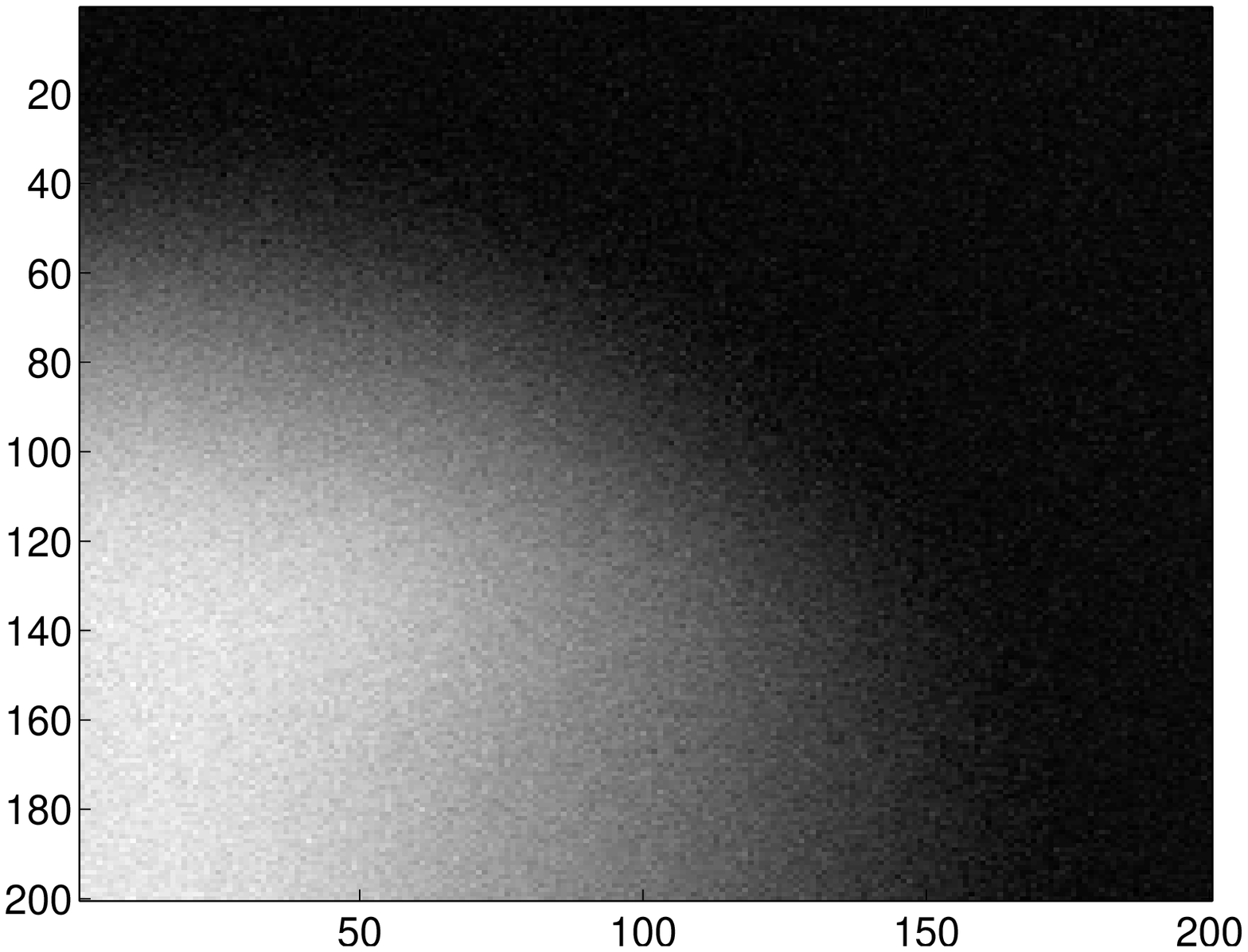}&
	\includegraphics[width=.23\textwidth]{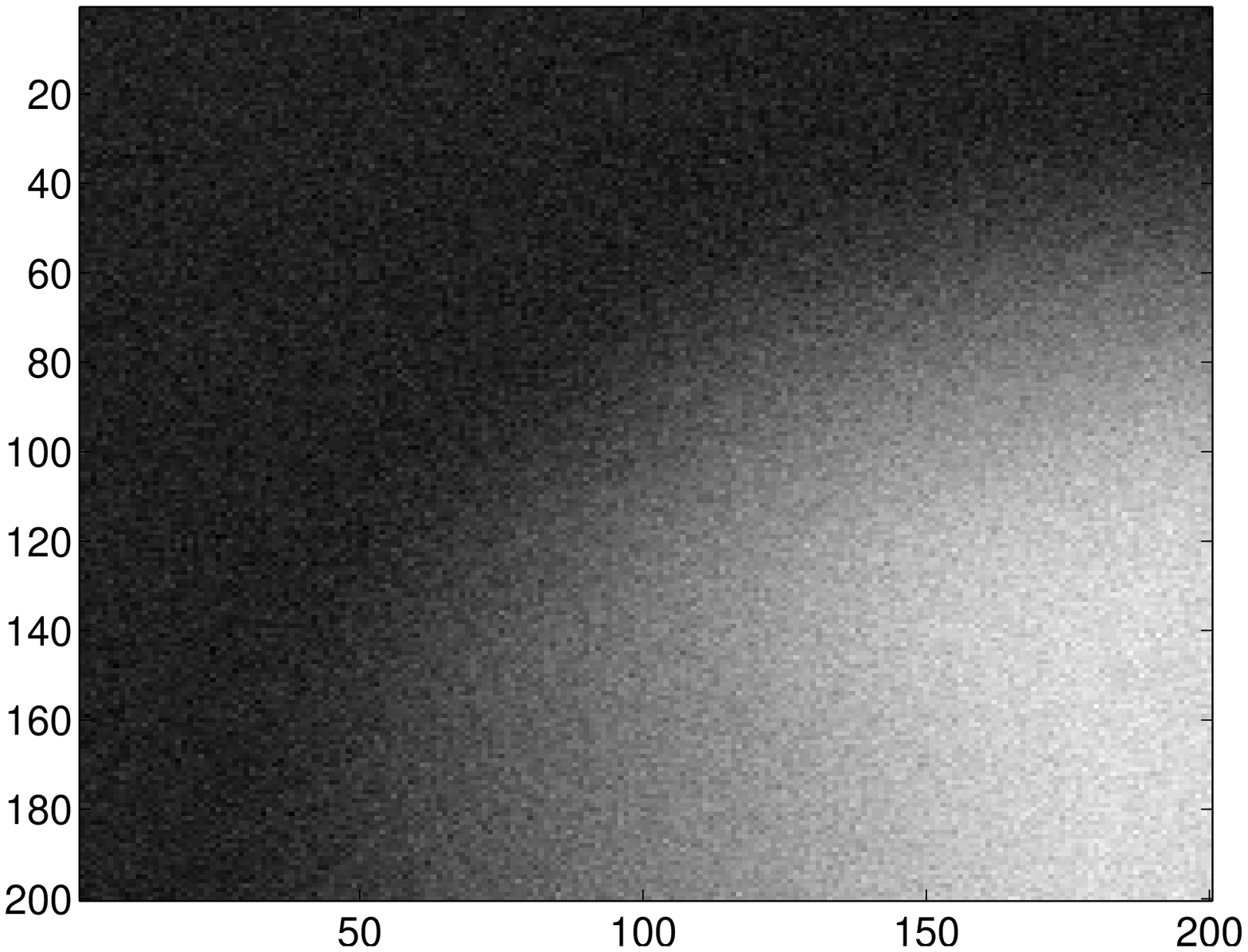}\\
	\multicolumn{3}{c}{Joint unmixing}\\
	\includegraphics[width=.23\textwidth,height=.1\textheight]{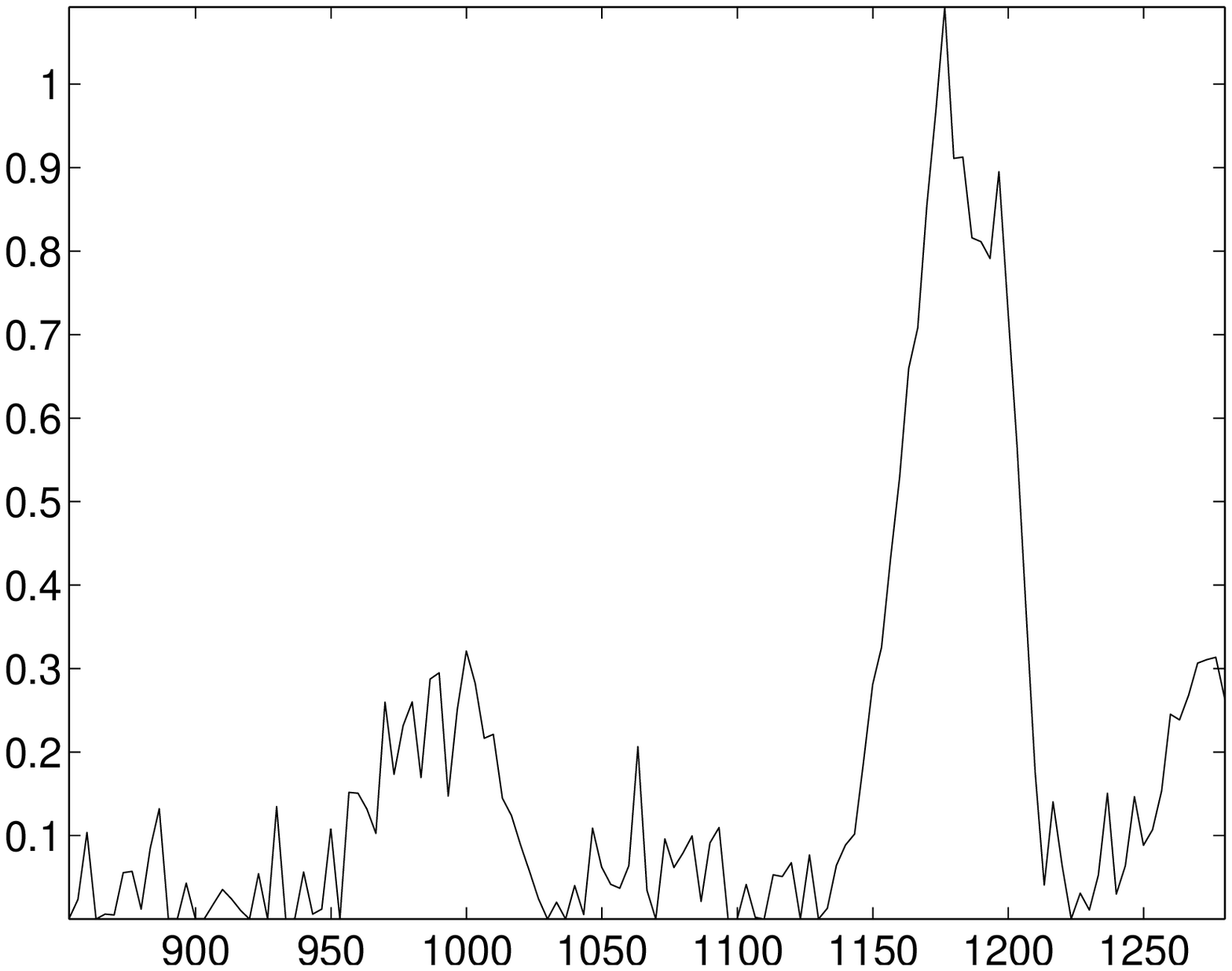}&
	\includegraphics[width=.23\textwidth,height=.1\textheight]{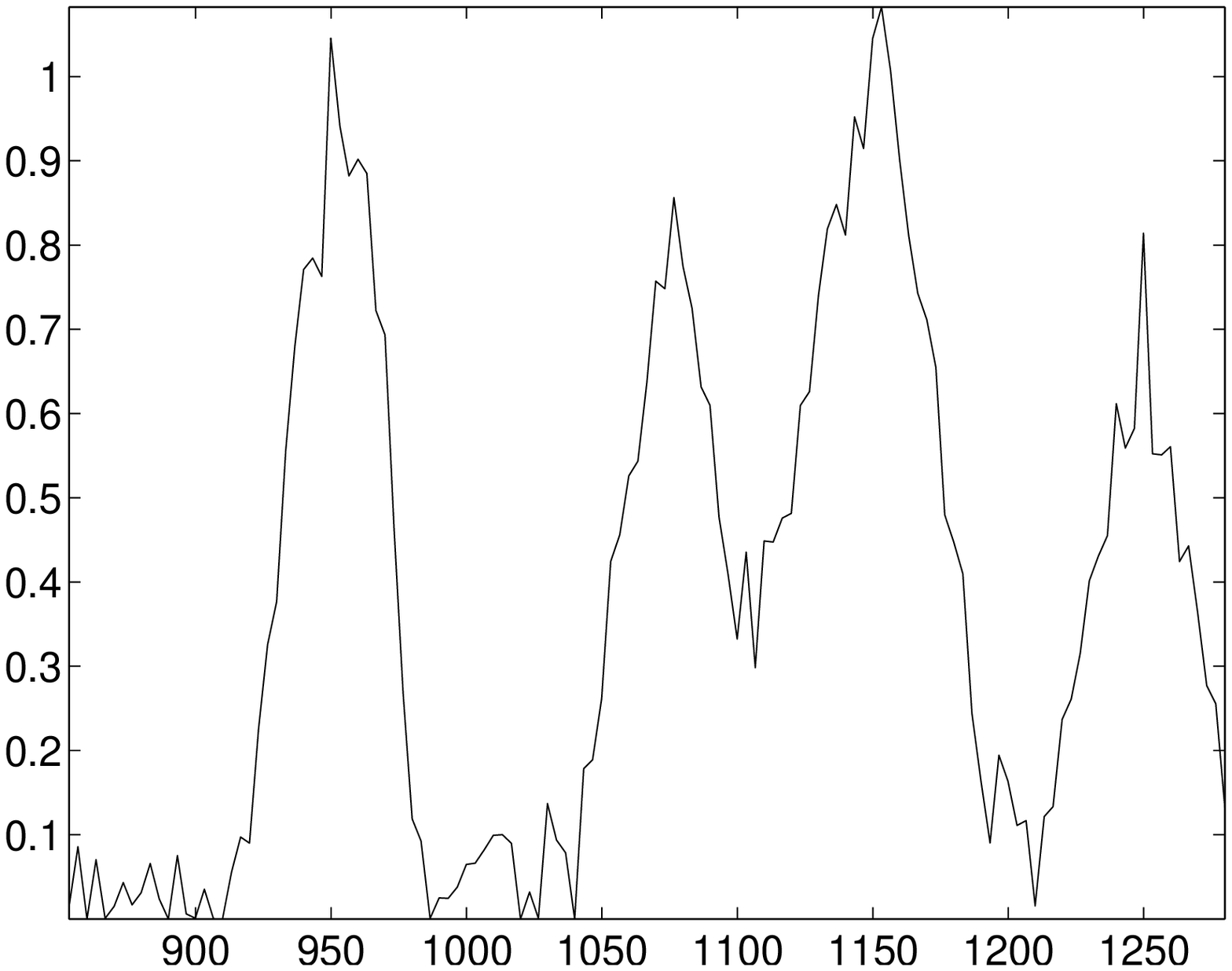}&
	\includegraphics[width=.23\textwidth,height=.1\textheight]{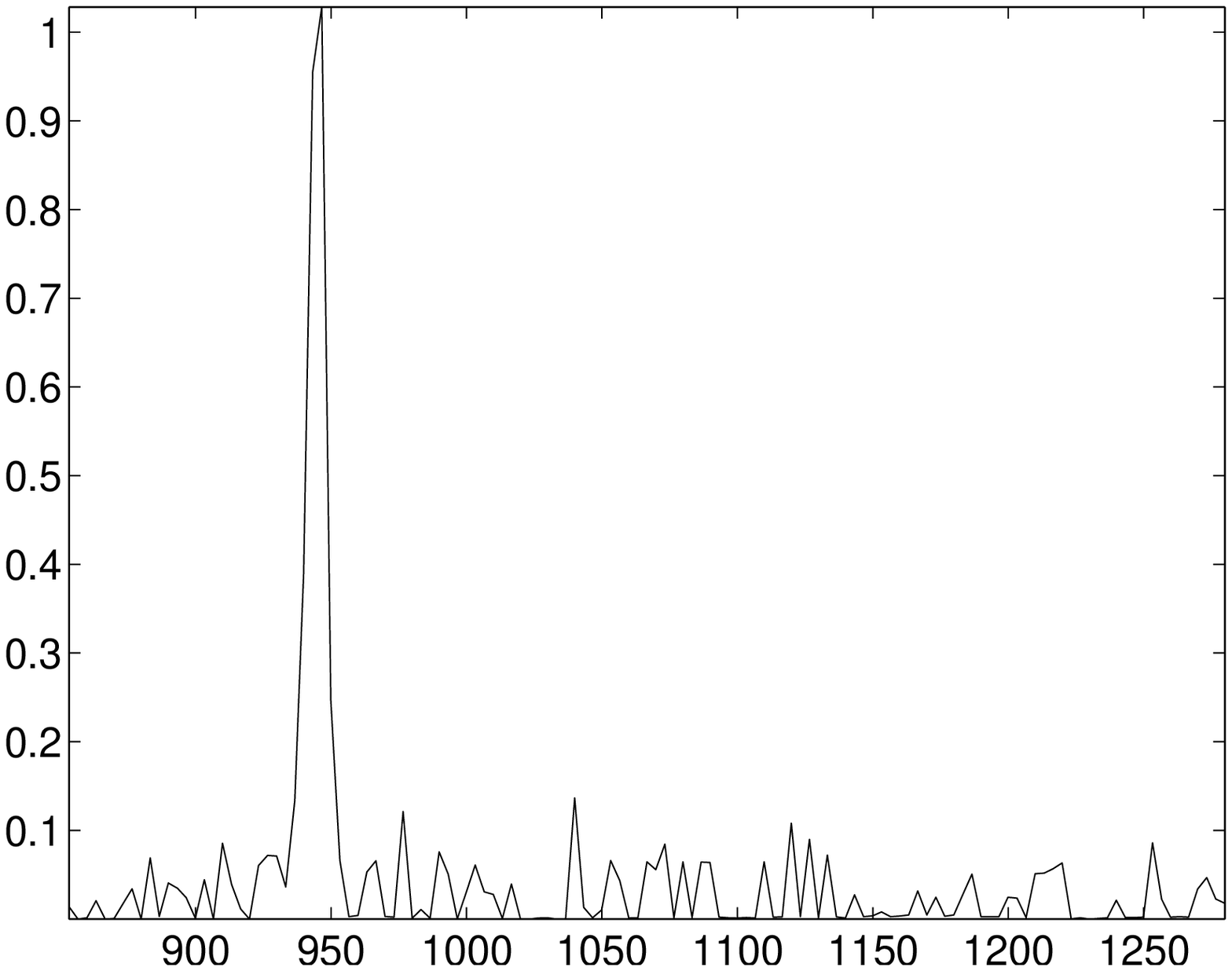}\\
	\includegraphics[width=.23\textwidth]{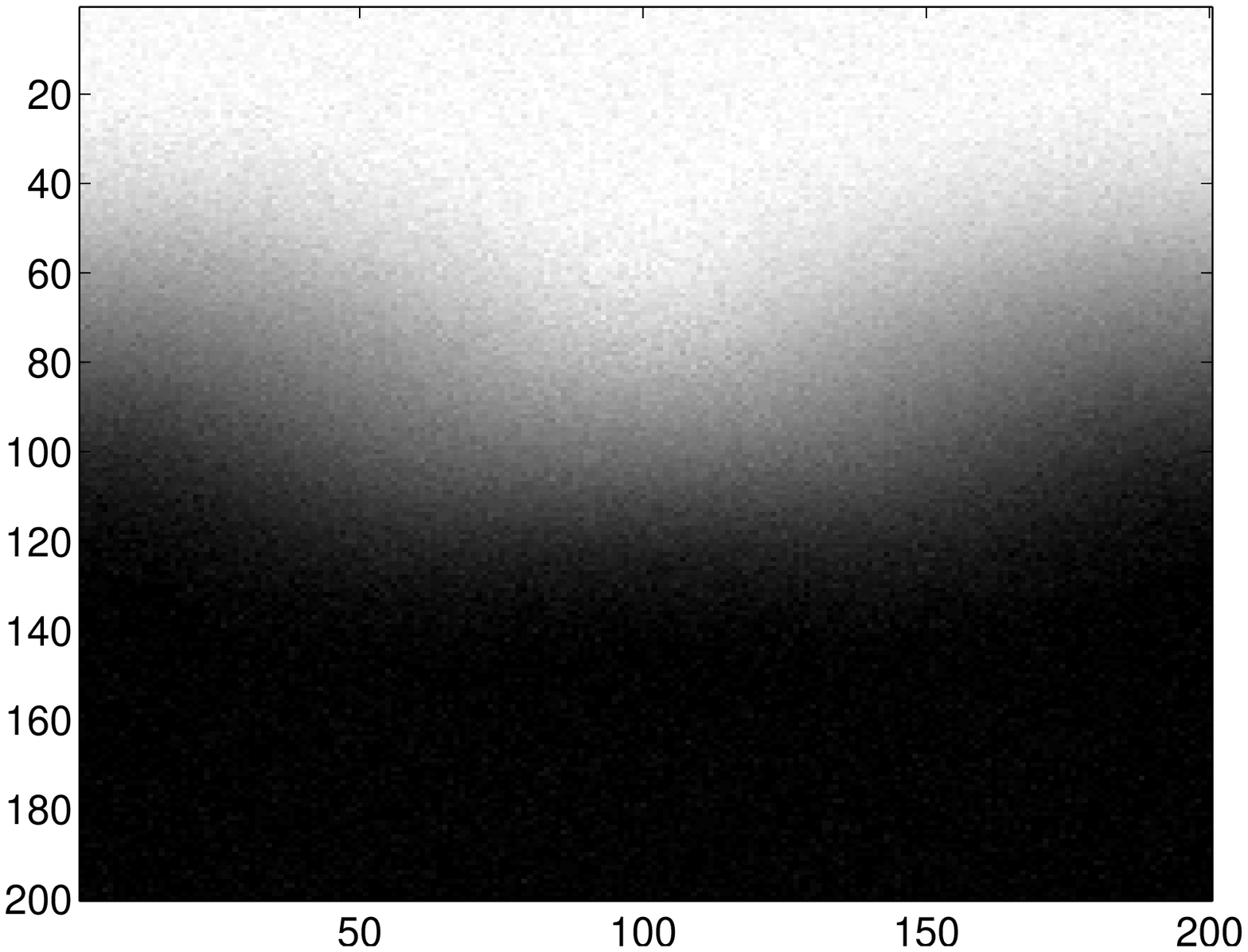}&
	\includegraphics[width=.23\textwidth]{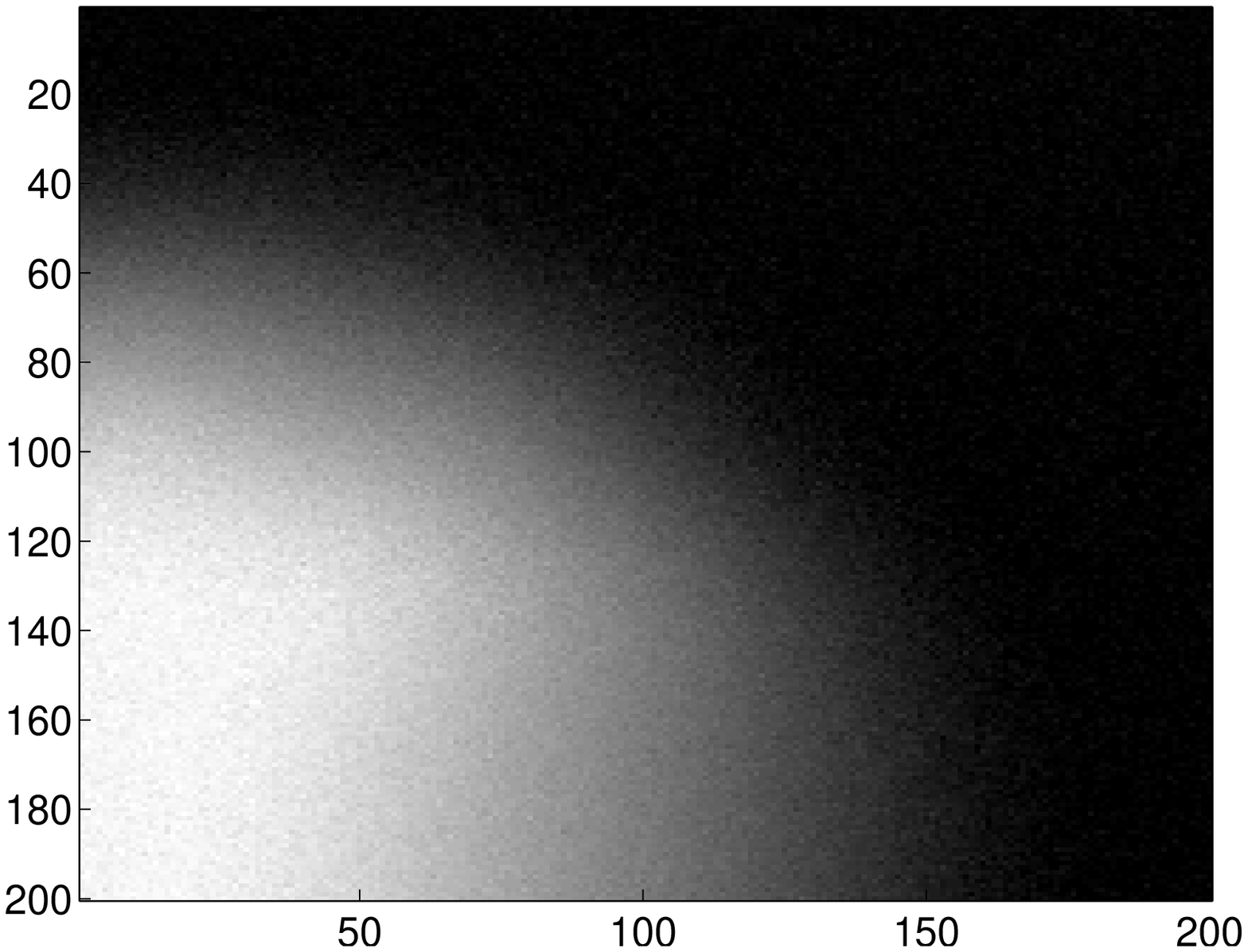}&
	\includegraphics[width=.23\textwidth]{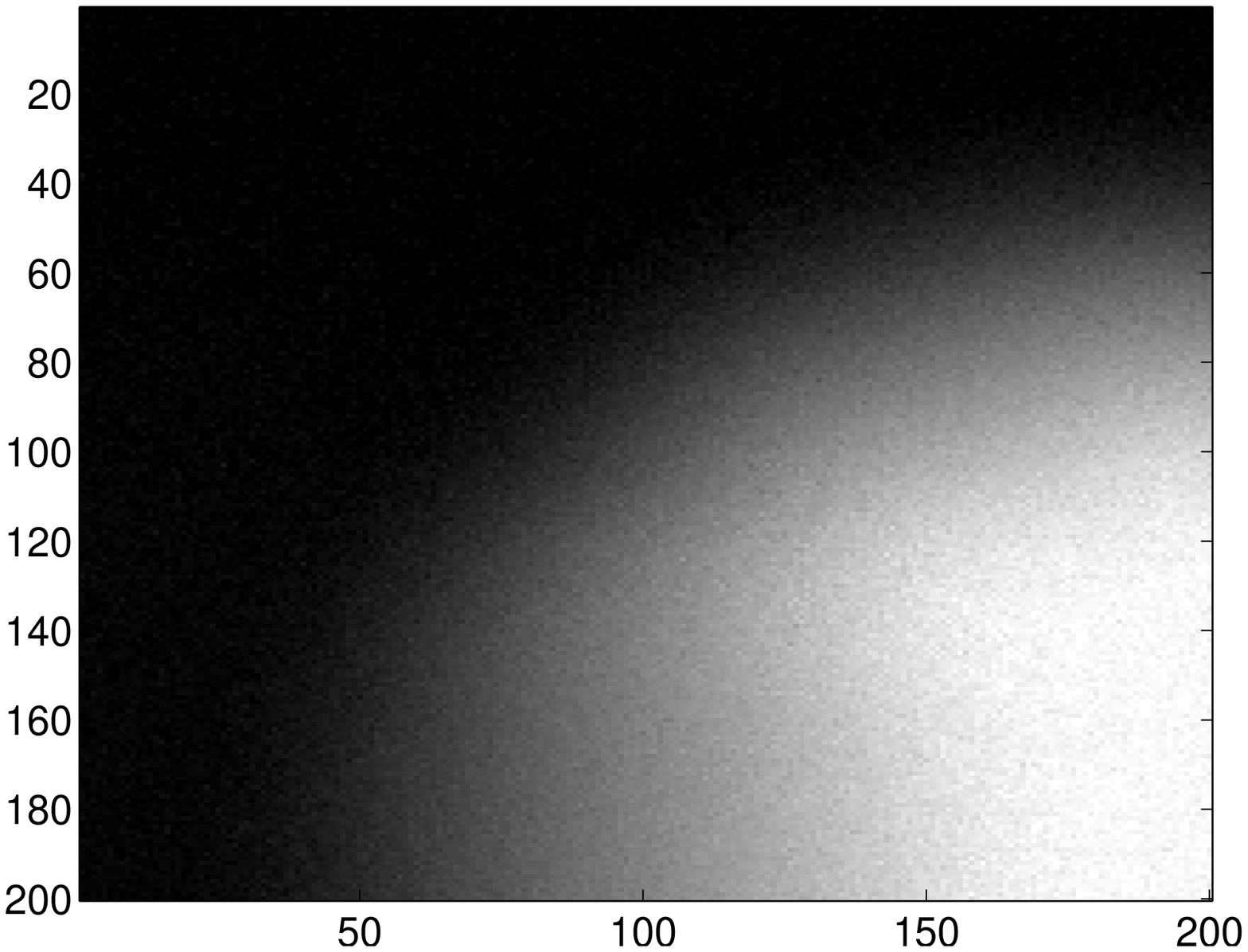}\\
	\end{tabular}
	\caption{\textcolor{black}{Unmixing of the tenth time frame. Each column corresponds to a particular endmember. The first and second row resp. display the spectra and abundance maps estimated by the separate approach, while the third and fourth row present the results of the joint approach. Both methods yield visually similar performances.}}
	\label{fig:synth_results}
\end{figure*}

We then compare the separate and joint unmixing approach on this data set; results are displayed for the tenth time frame in figure \ref{fig:synth_results}. Visually, both methods seem to give similar performances; thus, we turn to a quantitative measure to compare the two algorithms. \textcolor{black}{We run the comparison for ten different noise trials and the mean scaled MSEs of the estimated spectra and abundance maps reveal that joint unmixing outperforms separate unmixing, as shown in table \ref{tab:synth_results}.}

\begin{figure*}[ht]
	\center
	\begin{tabular}{ccc}
	Source 1 & Source 2 & Source 3\\
	\includegraphics[width=.31\textwidth,height=.15\textheight]{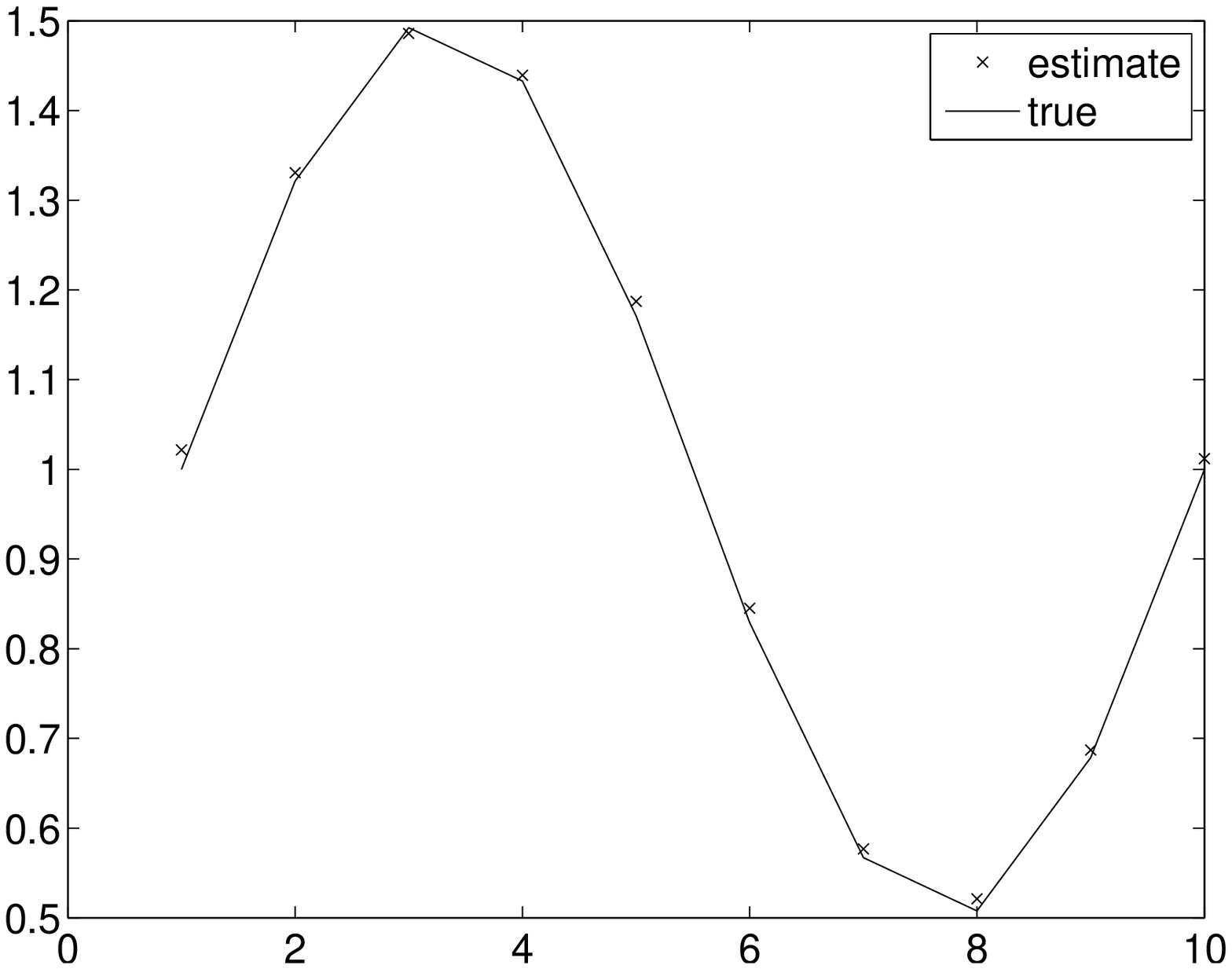}&
		\includegraphics[width=.31\textwidth,height=.15\textheight]{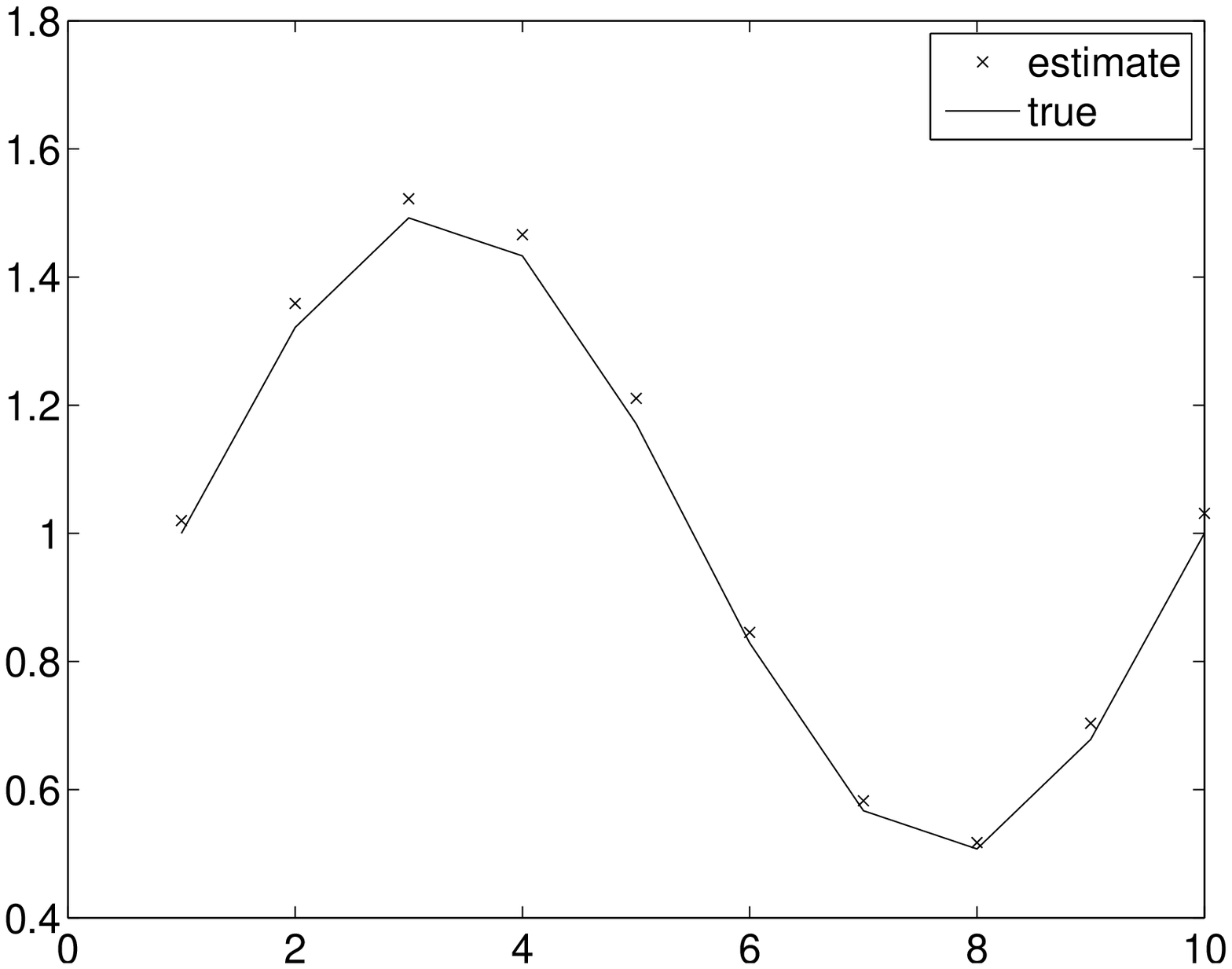}&
			\includegraphics[width=.31\textwidth,height=.15\textheight]{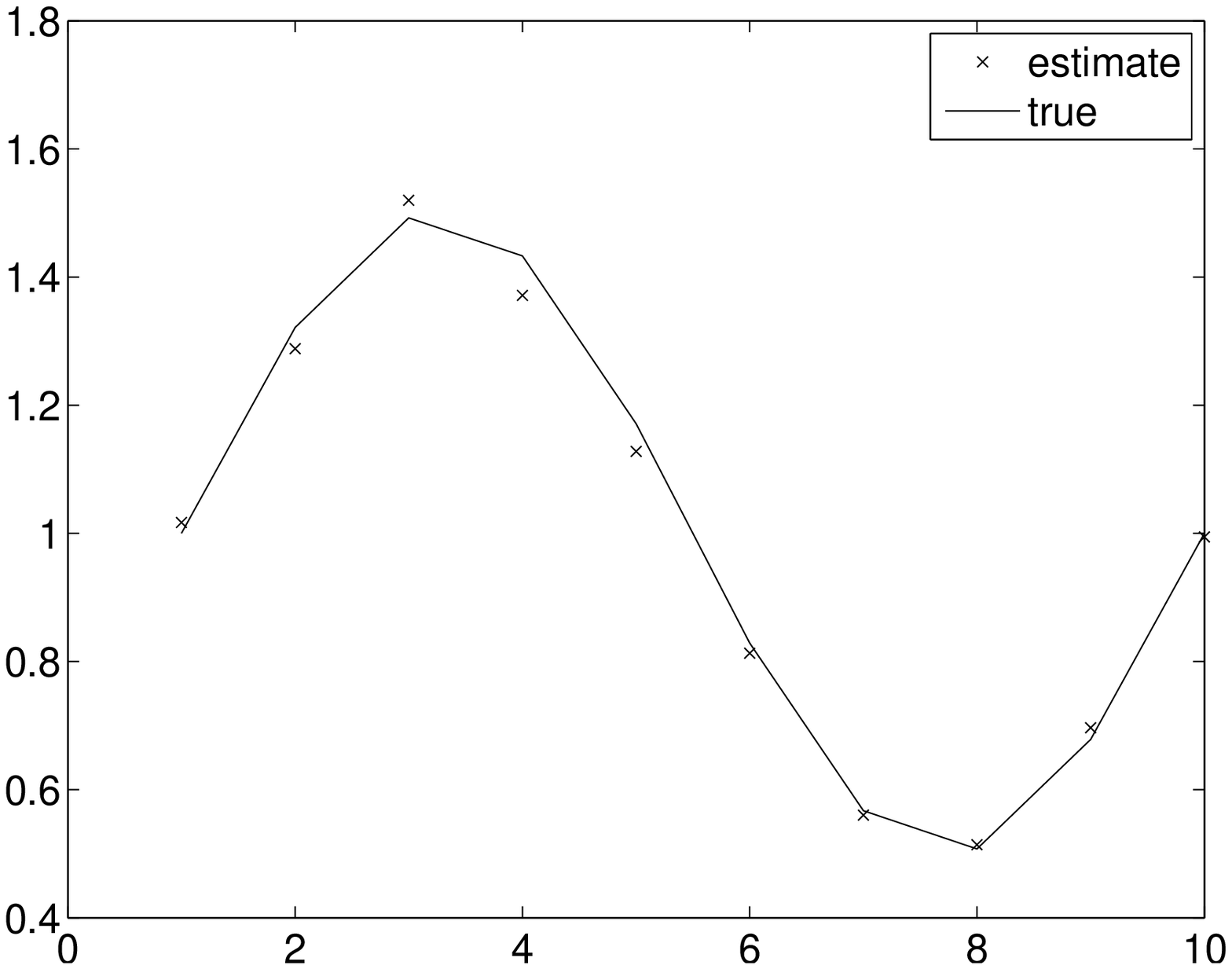}
	\end{tabular}
	\caption{\textcolor{black}{Evolution of scale factors over time. Straight lines correspond to the true scale factors; estimates are displayed using crosses. The joint method provides a good estimate of the true scale factors without prior knowledge on their values.}}
	\label{fig:scales_synth}
\end{figure*}

Figure \ref{fig:scales_synth} displays the estimated scale factors (represented as crosses; true values are indicated as straight lines). The scaled MSE value of the scale factors estimate is very low, at 0.02 and the good estimation performance is confirmed by the visual plots. The simulation thus shows that the proposed method can produce good estimates of the scale factors, even with no prior knowledge on their values. \textcolor{black}{The next step then consists in applying the algorithm to a real data set.}

\subsection{Experiments on real data}
\label{sec:real}

\begin{figure*}
	\center
	\begin{tabular}{cccc}
	        Time 1 & Time 2 & Time 3 & Time 4\\
		\includegraphics[width=.2\textwidth,height=.1\textheight]{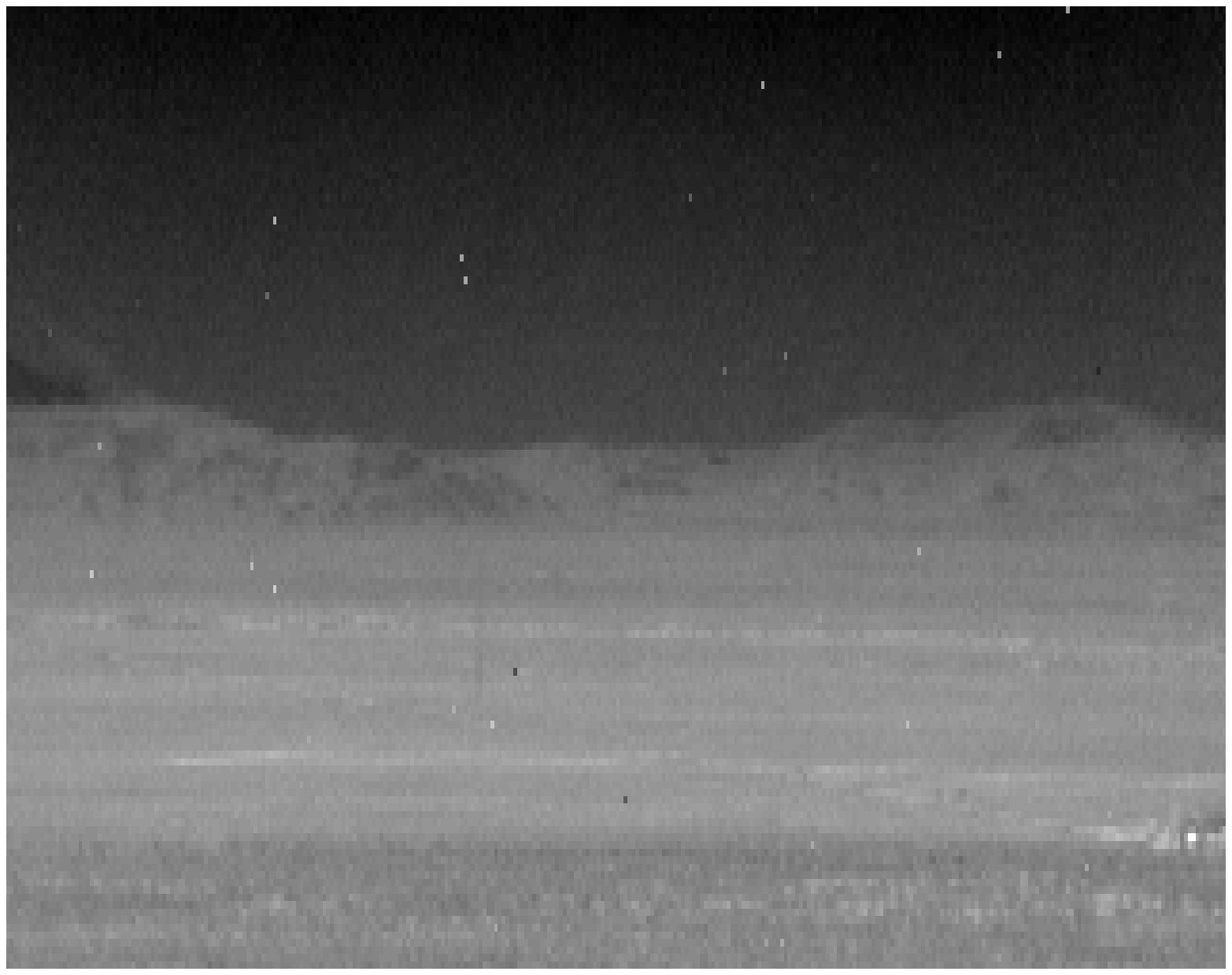}&
		\includegraphics[width=.2\textwidth,height=.1\textheight]{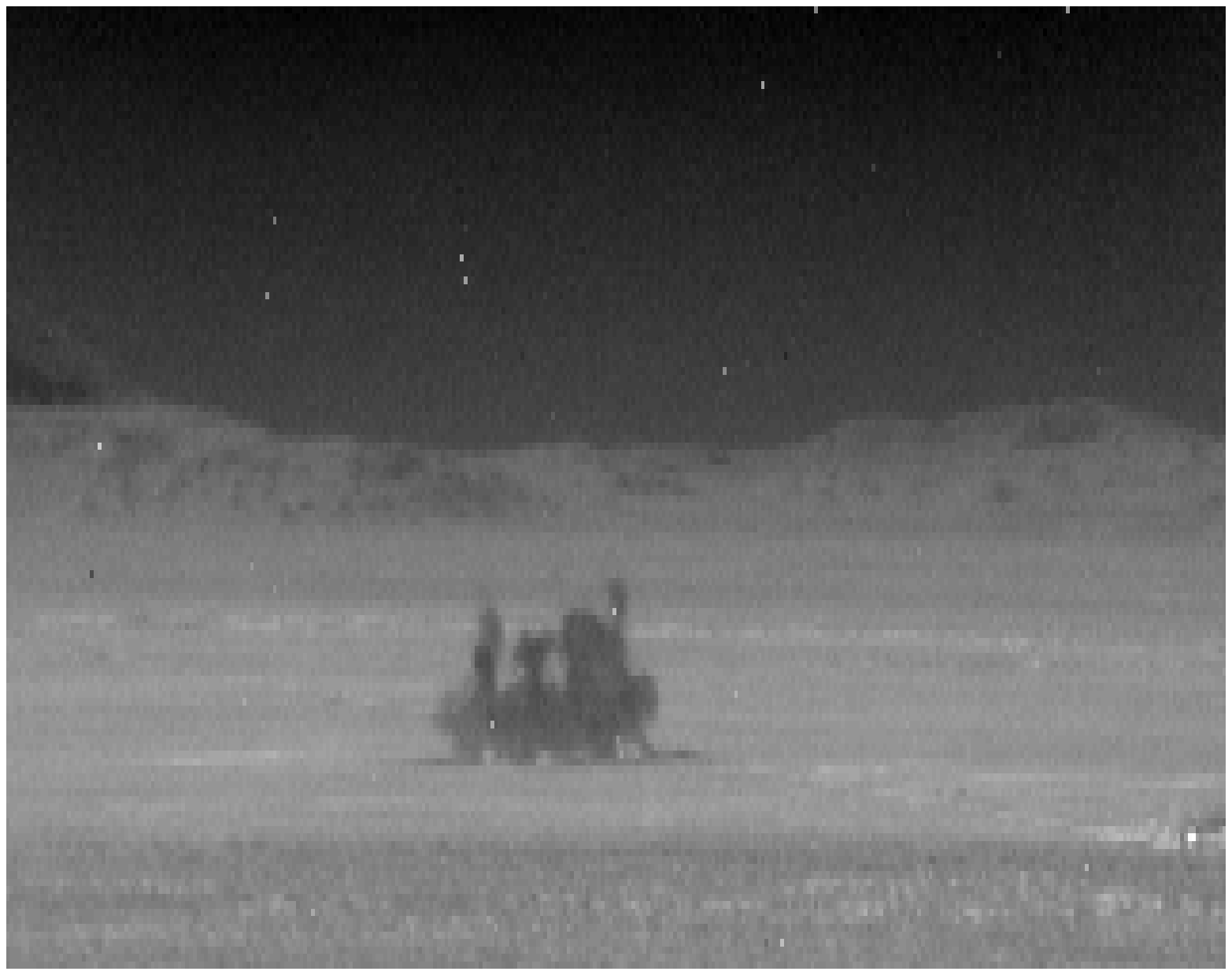}&
		\includegraphics[width=.2\textwidth,height=.1\textheight]{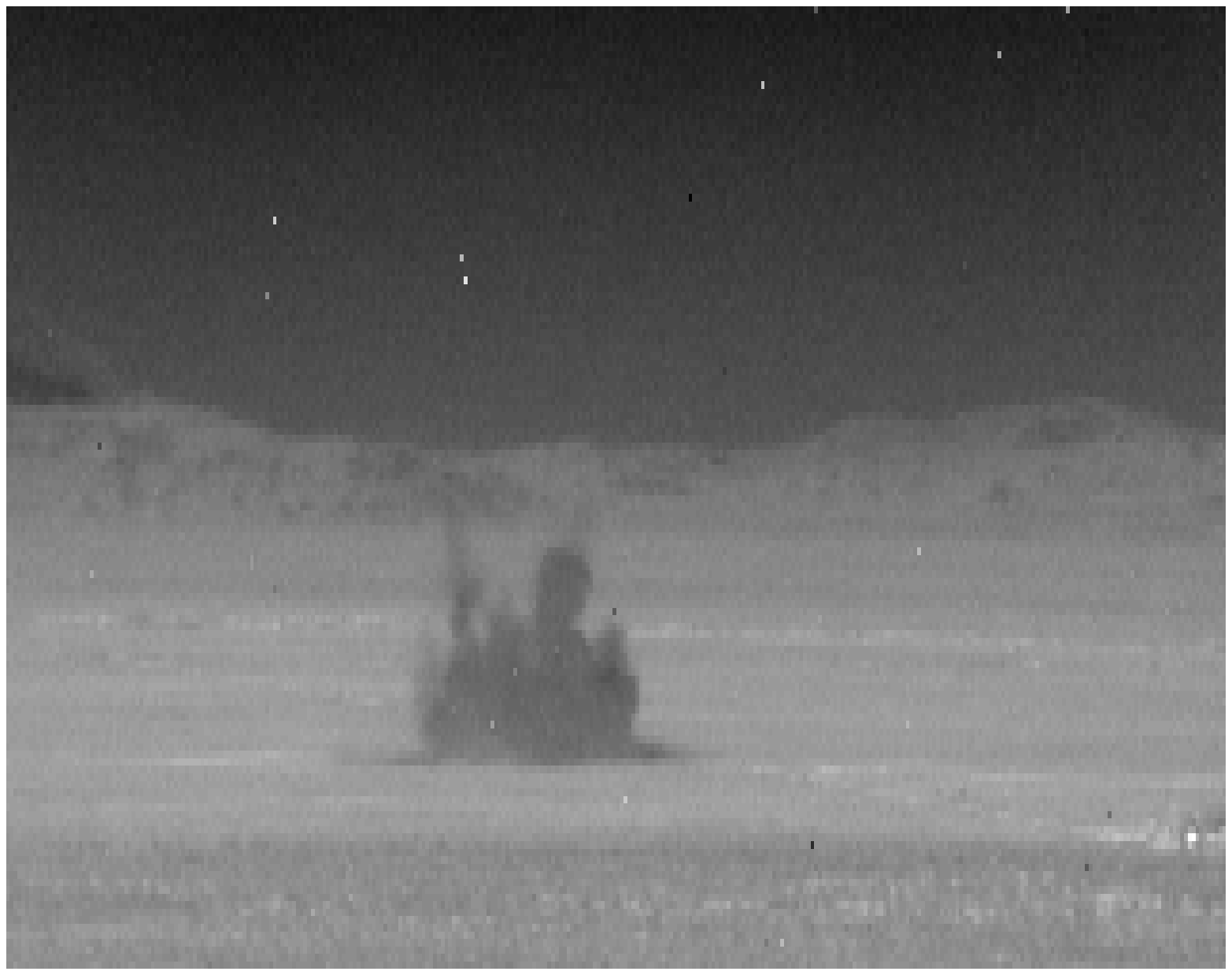}&
		\includegraphics[width=.2\textwidth,height=.1\textheight]{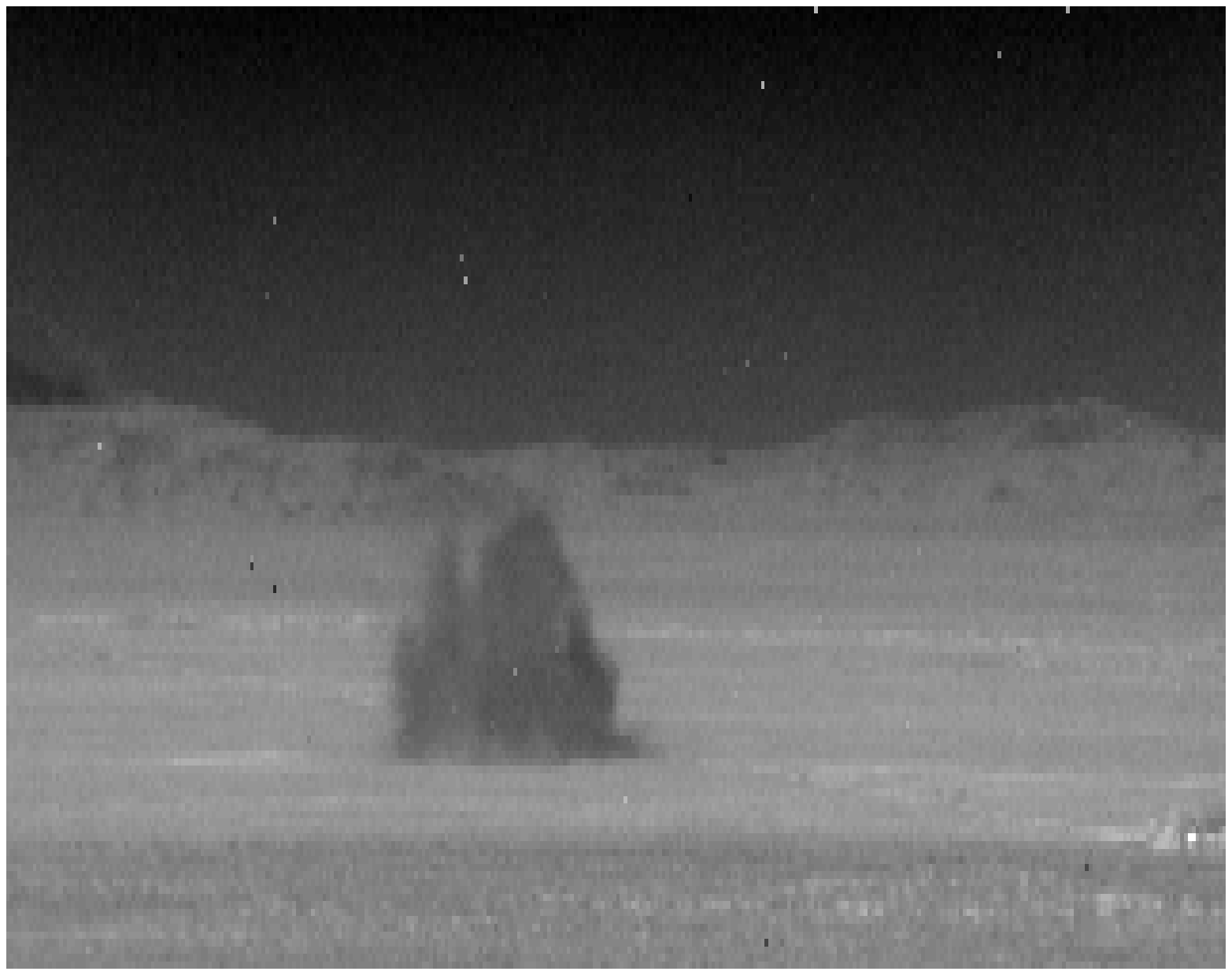}\\
		Time 5 & Time 6 & Time 7 & Time 8\\
		\includegraphics[width=.2\textwidth,height=.1\textheight]{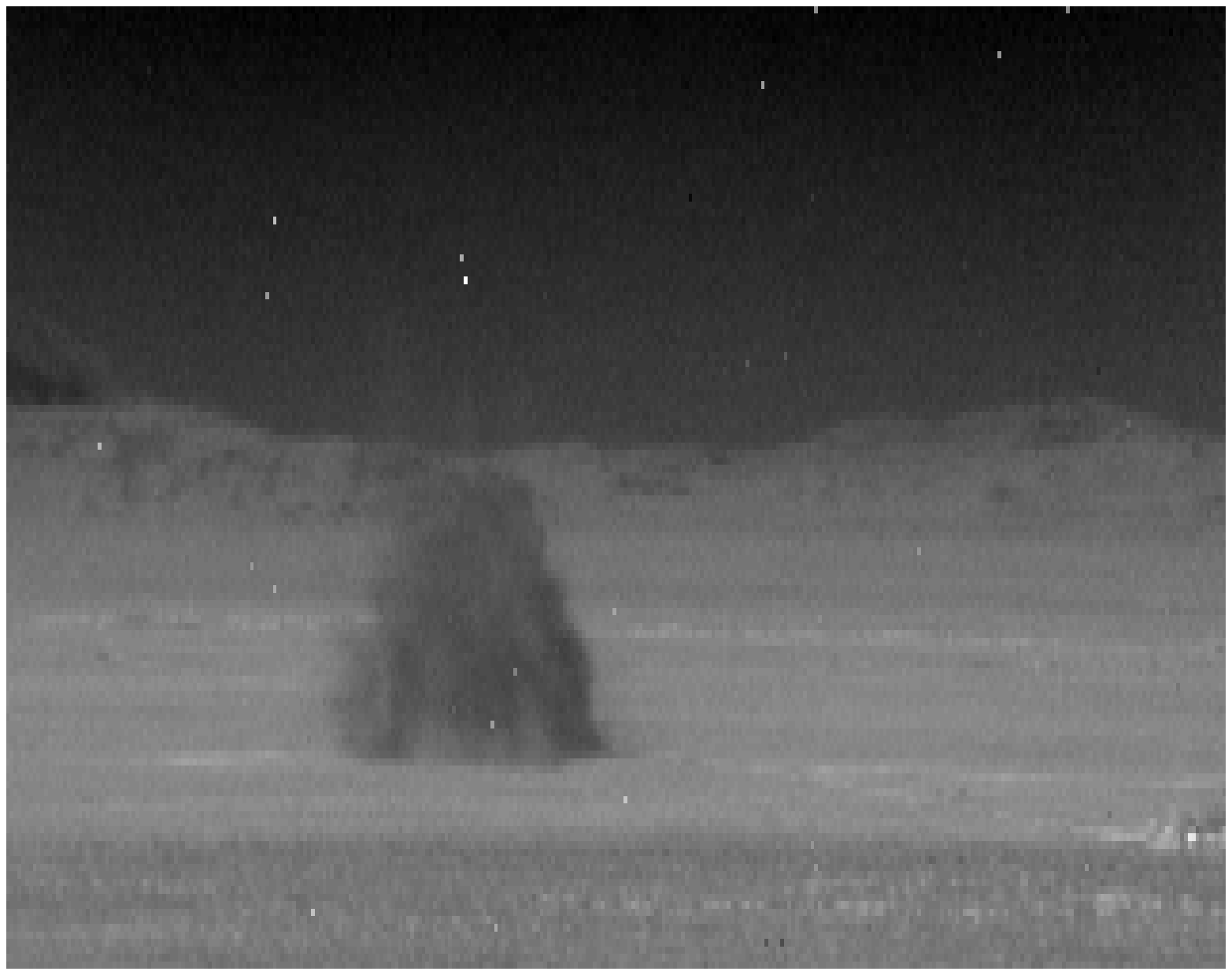}&
		\includegraphics[width=.2\textwidth,height=.1\textheight]{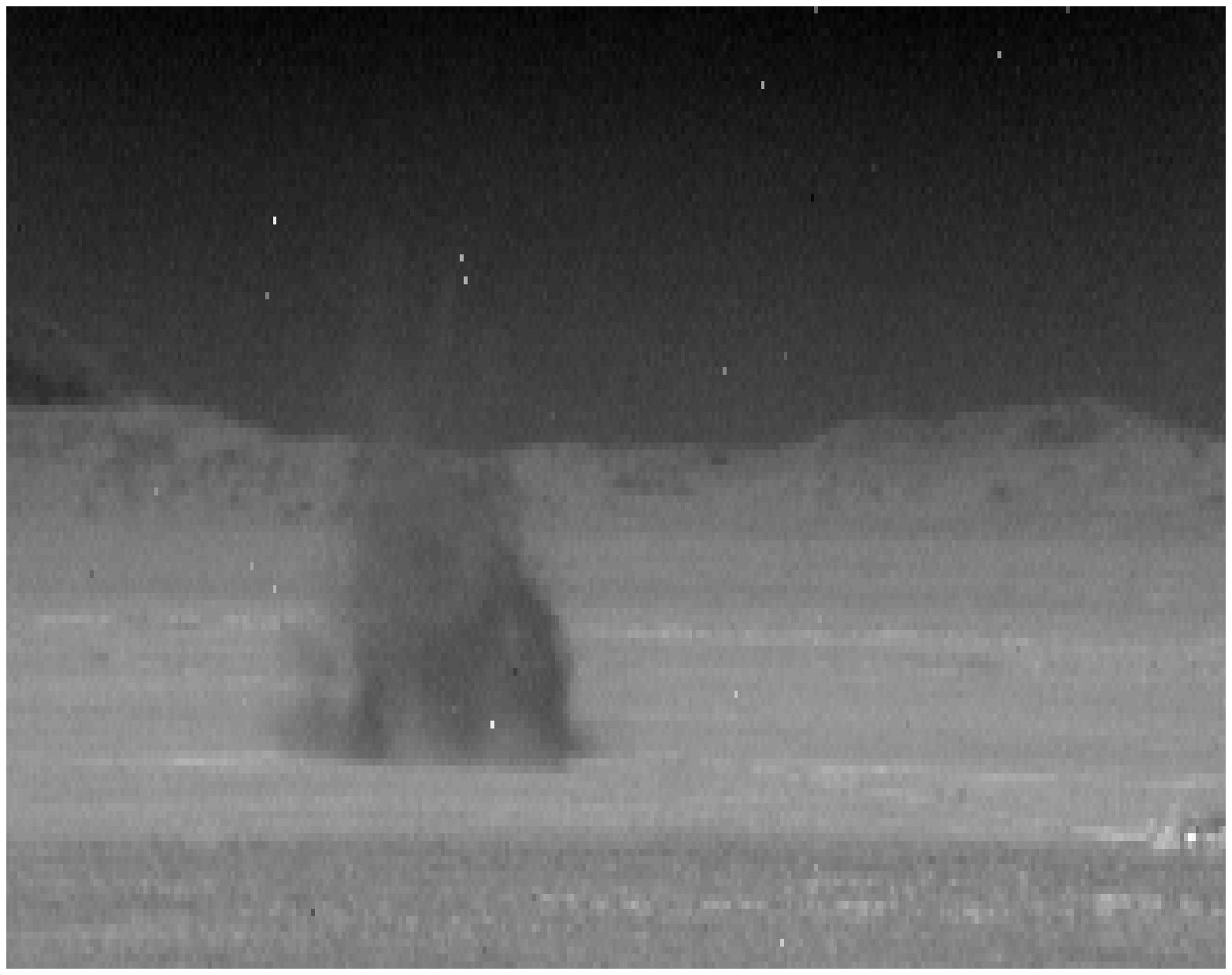}&
		\includegraphics[width=.2\textwidth,height=.1\textheight]{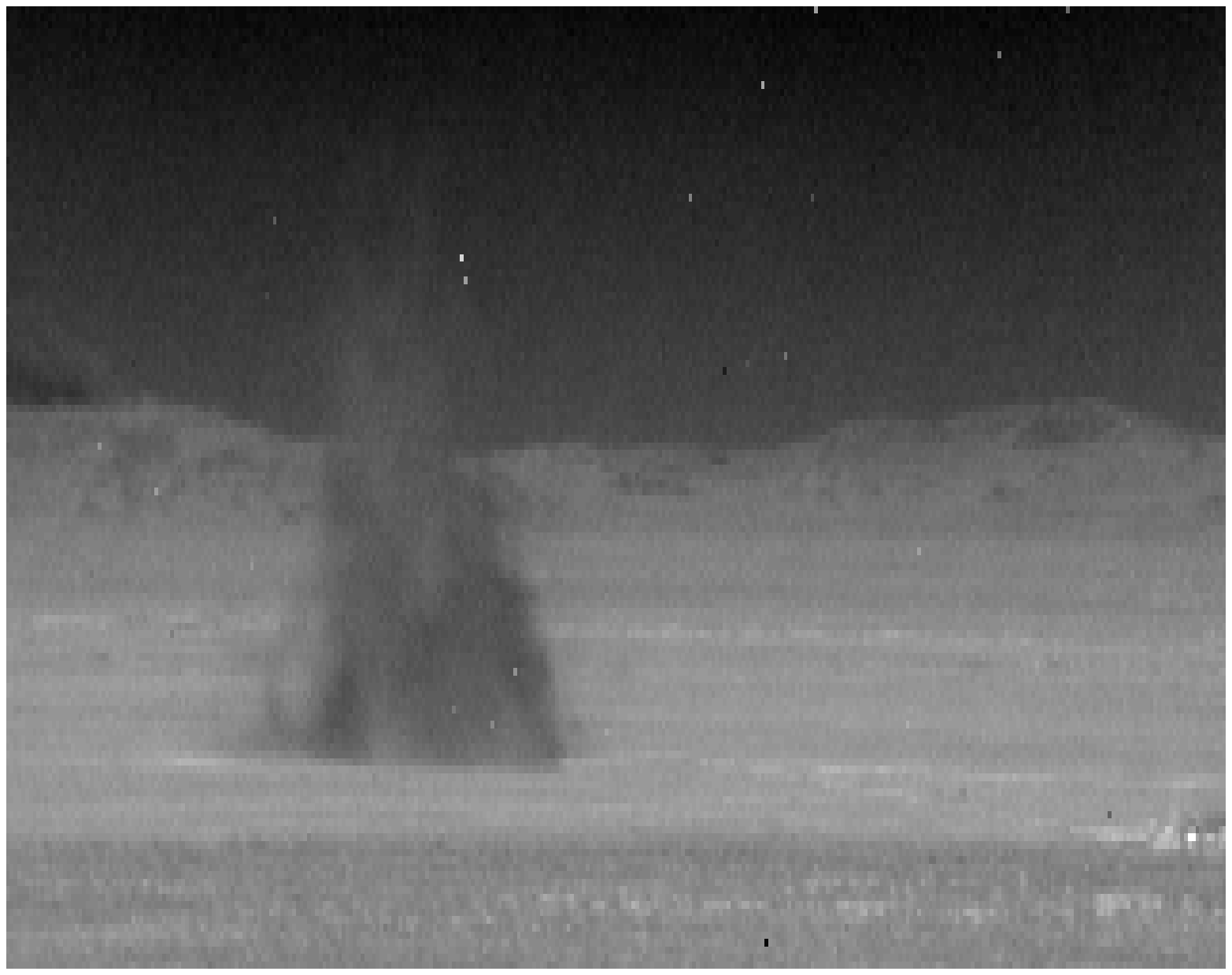}&
		\includegraphics[width=.2\textwidth,height=.1\textheight]{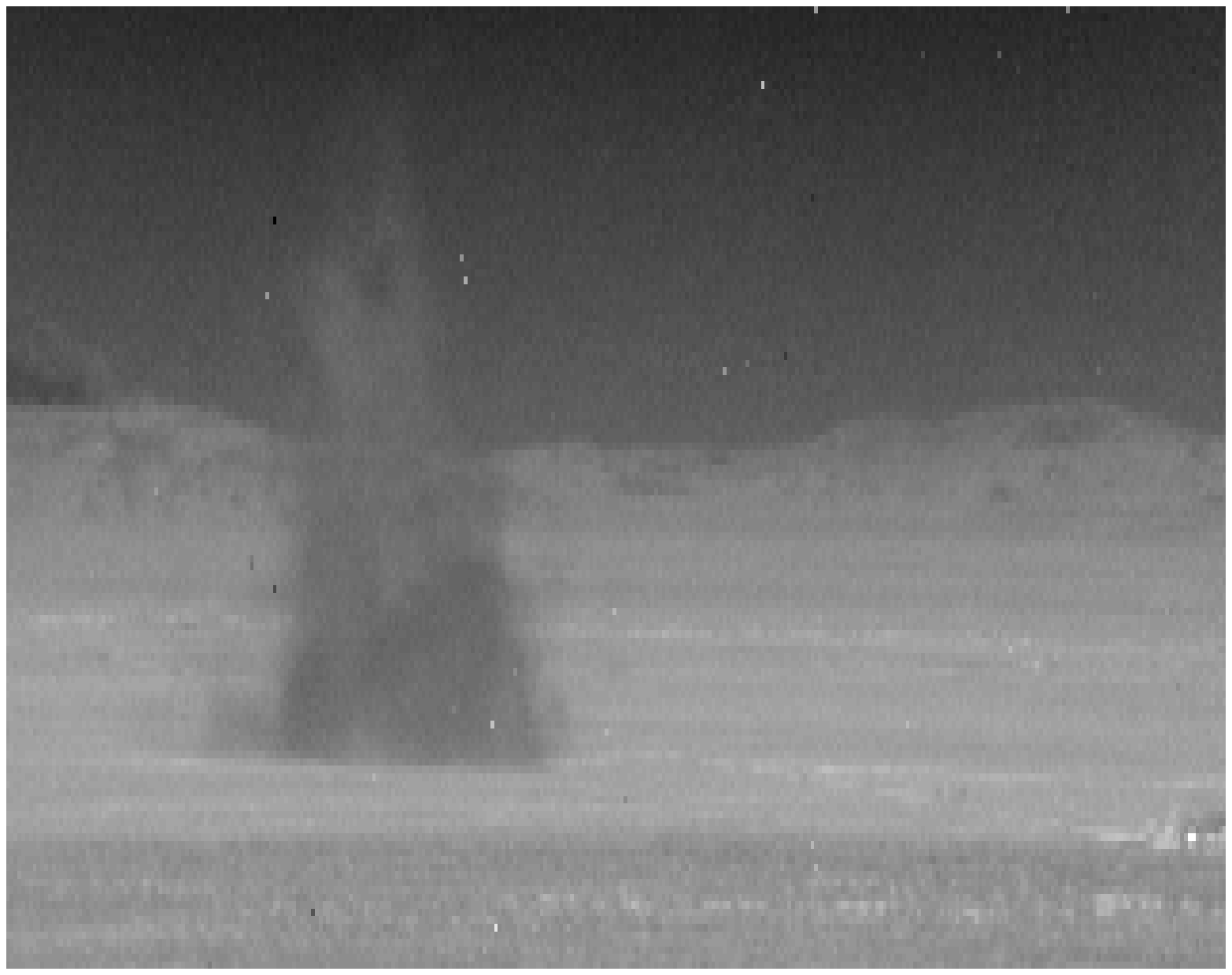}\\
		Time 9 & Time 10 & Time 11 & Time 12\\
		\includegraphics[width=.2\textwidth,height=.1\textheight]{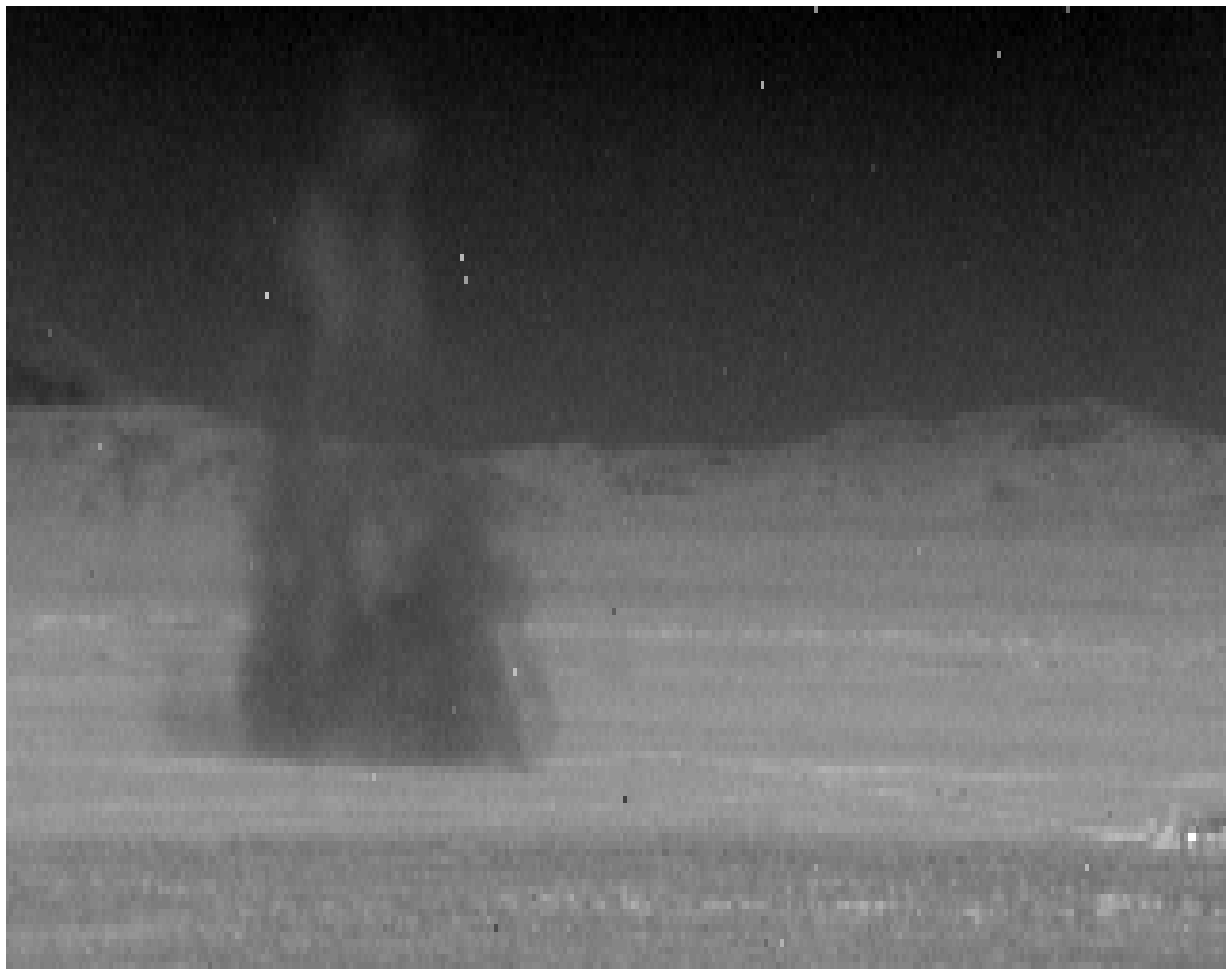}&
		\includegraphics[width=.2\textwidth,height=.1\textheight]{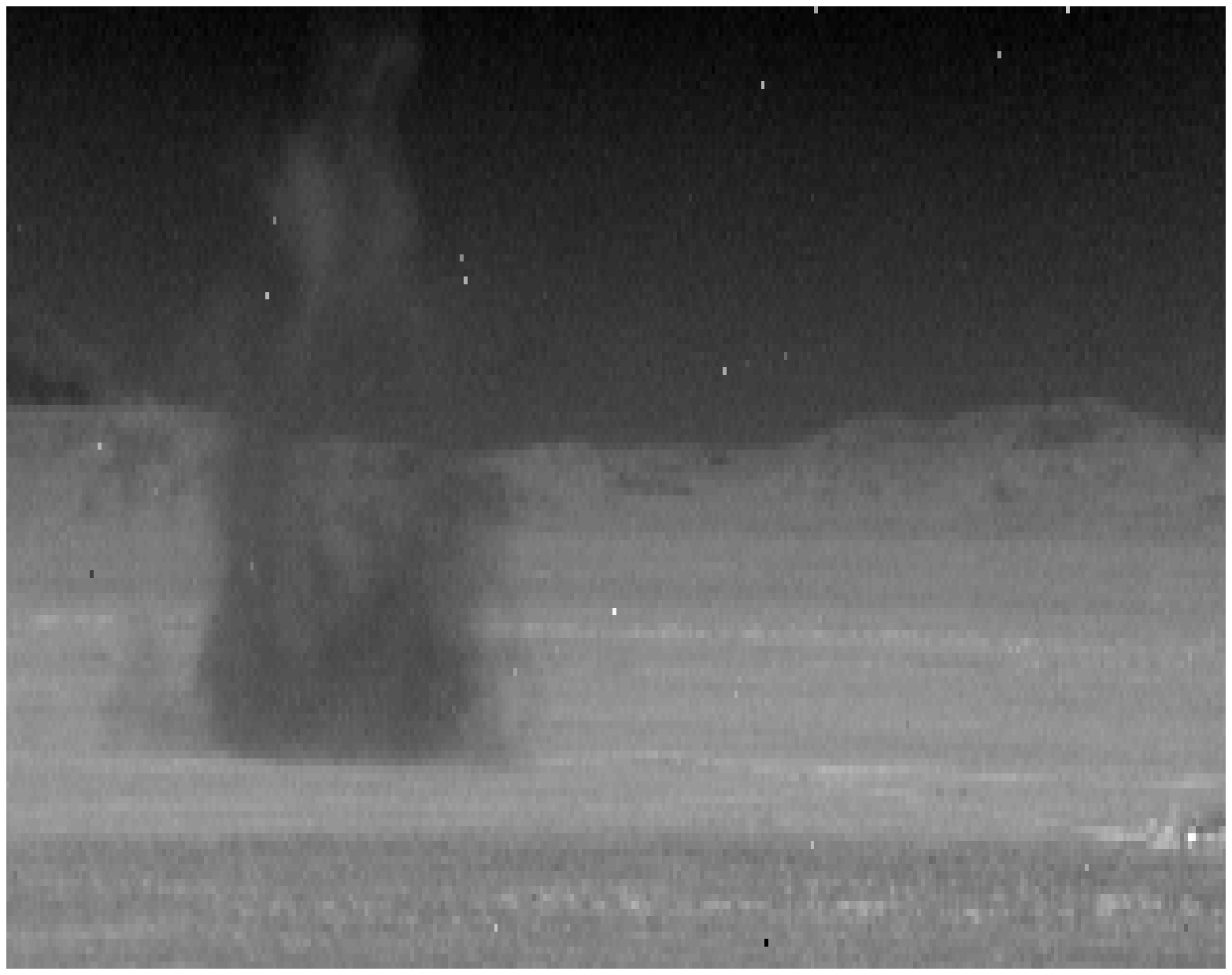}&
		\includegraphics[width=.2\textwidth,height=.1\textheight]{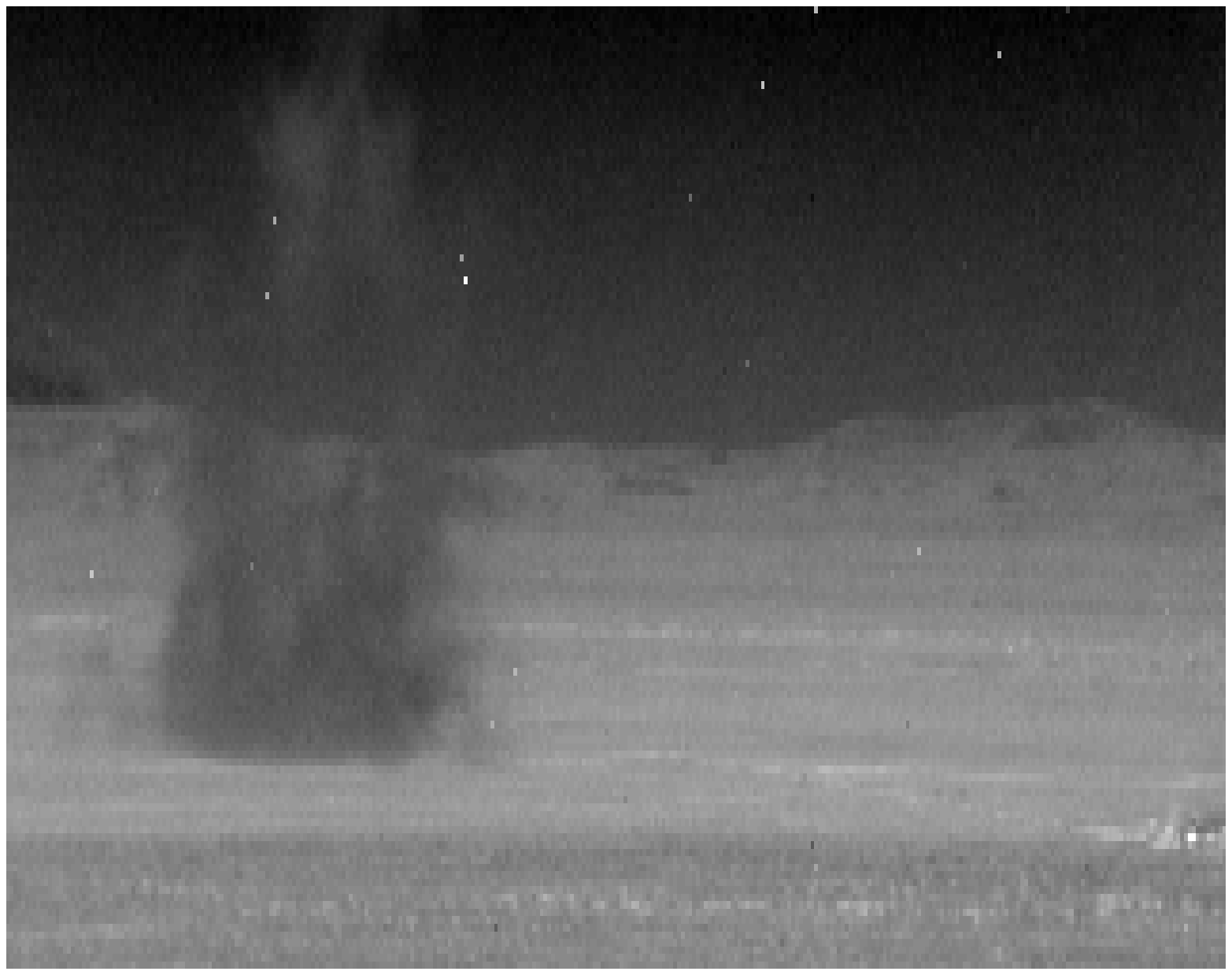}&
		\includegraphics[width=.2\textwidth,height=.1\textheight]{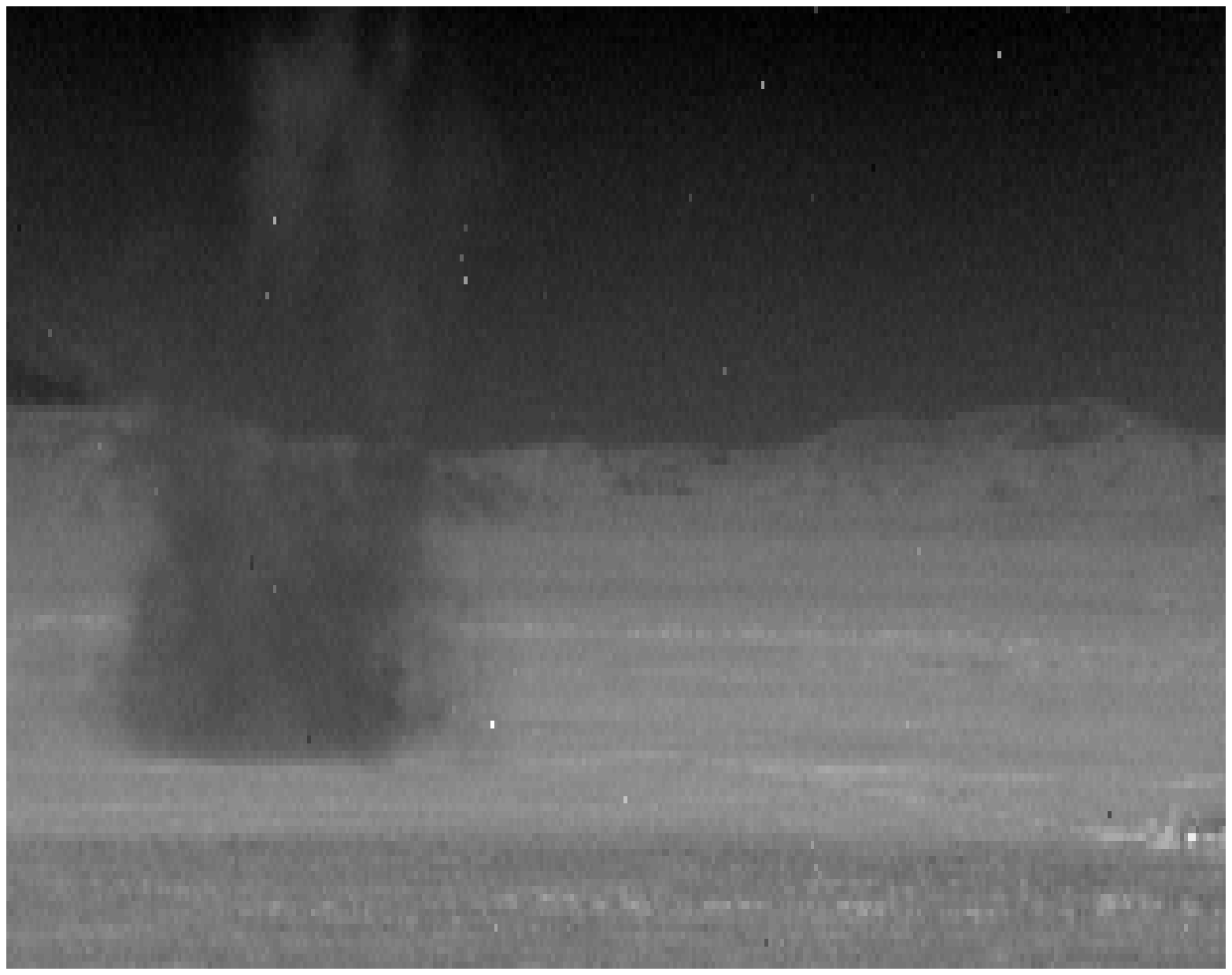}\\
	\end{tabular}
	\caption{First principal component of each time frame: the sequence depicts the evolution of the gas plume in the scene over time.}
	\label{fig:time_sequence}
\end{figure*}

In this section, we evaluate the performance of the proposed approach on \emph{longwave infrared} (LWIR) spectroscopy data. LWIR imaging allows to capture a scene based on thermal emissions only, requiring no illumination. The scene of interest was acquired in a desert in 2006~\cite{BRO11}, and captures a chemical gas plume emitted from a specific location. Specifically, we focus on a time series of hyperspectral images observed by a Fourier-Transform (FTIR) sensor located 2.82km from the release ground. We consider twelve time frames, which account for the dynamics of the release of the gas plume. The first time instant is taken just before the release, and the subsequent eleven frames cover the early gas emission. Each hyperspectral data cube comprises $128 \times 320$ spatial pixels and 129 wavelengths taken with a $4 \text{ cm}^{-1}$ frequency spacing. The first principal component of each time frame is displayed in figure \ref{fig:time_sequence}.

\begin{figure}[h!!]
	\center 
	\includegraphics[width=.5\textwidth]{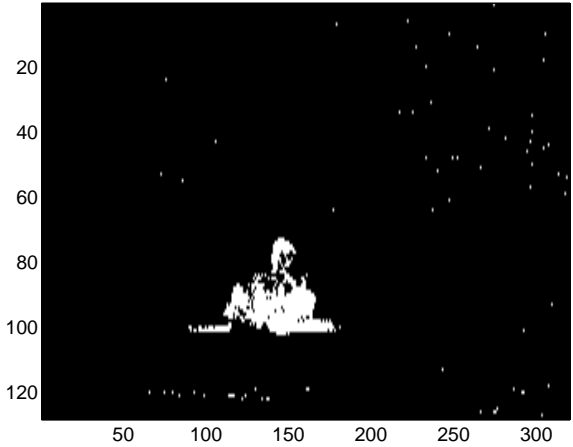}
	\caption{\textcolor{black}{Spatial display of the binary detection test result for each pixel of the third time frame, using a general likelihood ratio on the value of the estimated nonlinearity parameter. A value of one (white) indicates a nonlinear mixing process, a value of 0 (black) corresponds to the LMM. As can be seen, the LMM mainly fails to hold within the spatial support of the gas plume, isolated pixels corresponding to remaining outliers in the scene. Anywhere else, the LMM can be assumed to be a good approximation of the actual physical mixing process.}}
	\label{nonlinearity}
\end{figure}

In this scene, we expect multiple scattering effects to take place in the spatial support of the gas plume. To verify this, we perform nonlinear detection on the pixels of the third time frame, using the method presented in~\cite{ALT13}. The first step consists in carrying out nonlinear unmixing on the image. In this approach, each pixel in the image is given by the \emph{polynomial post-nonlinear mixing model} (PPNMM)~\cite{ALT13}:
\begin{equation}
\mathbf{x}_k^n = \mathbf{S}_k \mathbf{a}_k^n + b_k^n (\mathbf{S}_k \mathbf{a}_k^n) \odot (\mathbf{S}_k \mathbf{a}_k^n) + \mathbf{e}_k^n
\end{equation} 
where $\odot$ stands for the Hadamard (entry-wise) product, and the nonlinear term is scaled by the scalar parameter $b_k^n$. In other words, $b_k^n$ quantifies the nonlinearity of the mixing process for each pixel in the image, and the PPNMM reduces to the LMM for $\mathbf{\underline{B}} = \mathbf{\underline{0}}$. The nonlinearity parameter is estimated for each pixel in the image and a general likelihood ratio test is then used on $\mathbf{\underline{B}}$ to decide whether each pixel results from the LMM or the PPNMM. Figure \ref{nonlinearity} displays the decision for each pixel of the third time frame, with the false alarm probability set as $0.05$. As expected, the mixing process is mostly nonlinear within the spatial support of the gas plume only (isolated pixels appearing on the map correspond to remaining outliers). Hence, we expect traditional unmixing methods to fail in this specific area.

We preprocess the data to filter out outliers caused by defaults in the sensor. First, their location is determined by detecting pixels whose value differ from the mean of the data by more than 5 times the standard deviation. These pixels are then replaced by a surrogate pixel whose value is computed using a median filter.

\begin{figure*}[ht]
	\begin{tabular}{cccc}
	Source 1 & Source 2 & Source 3 & Source 4\\
		\multicolumn{4}{c}{Time frame 1}\\
	\includegraphics[width=.23\textwidth]{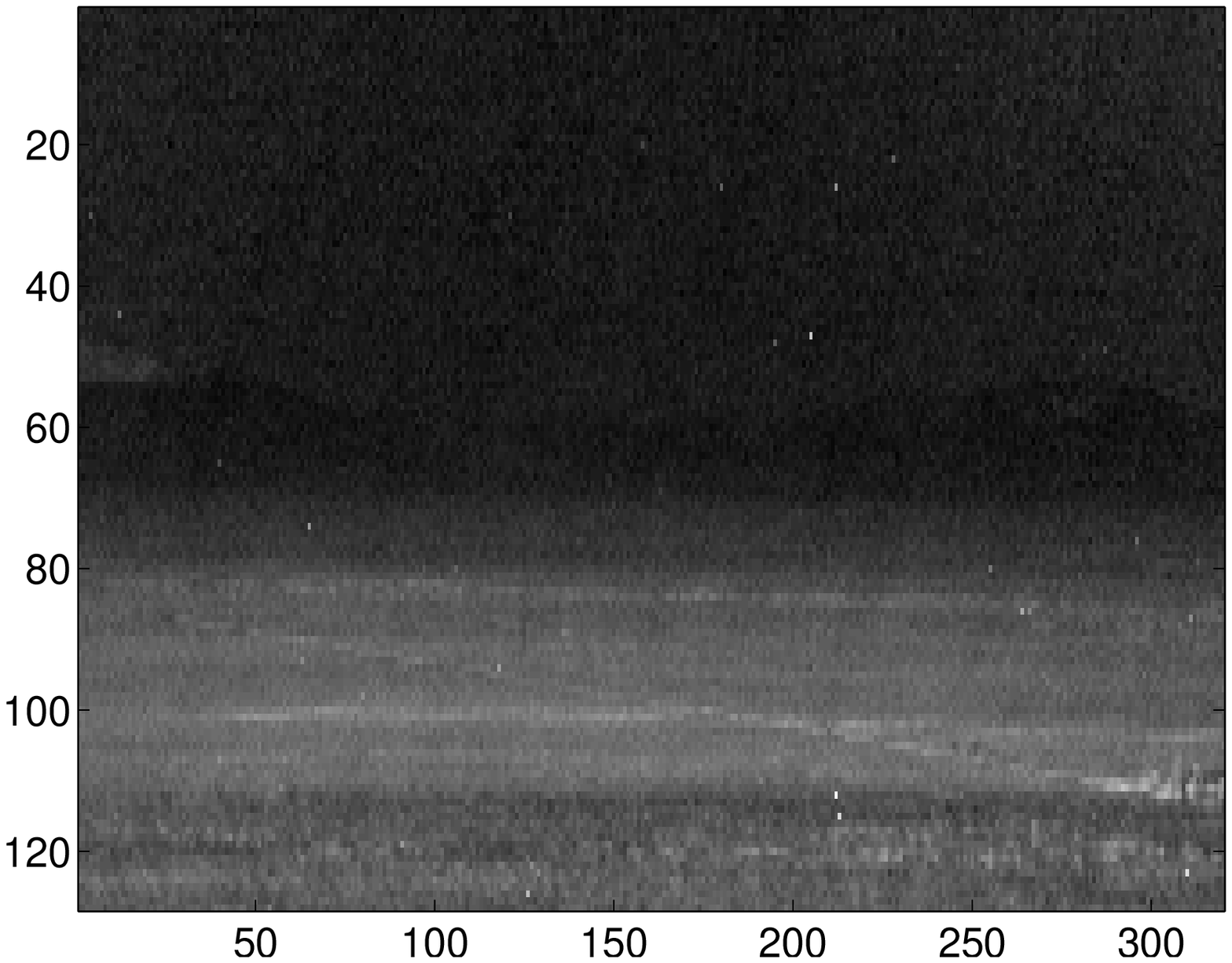}&
	\includegraphics[width=.23\textwidth]{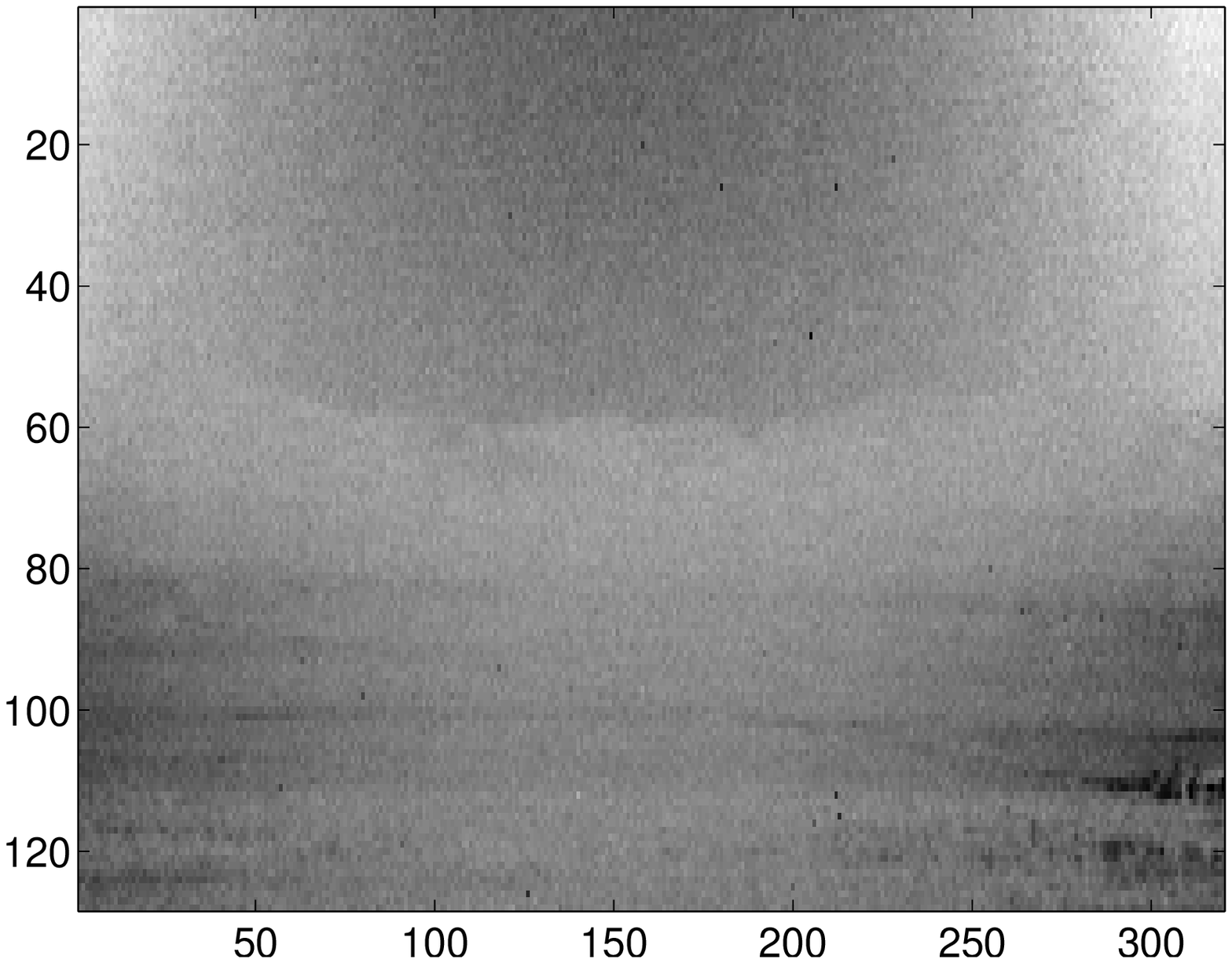}&
	\includegraphics[width=.23\textwidth]{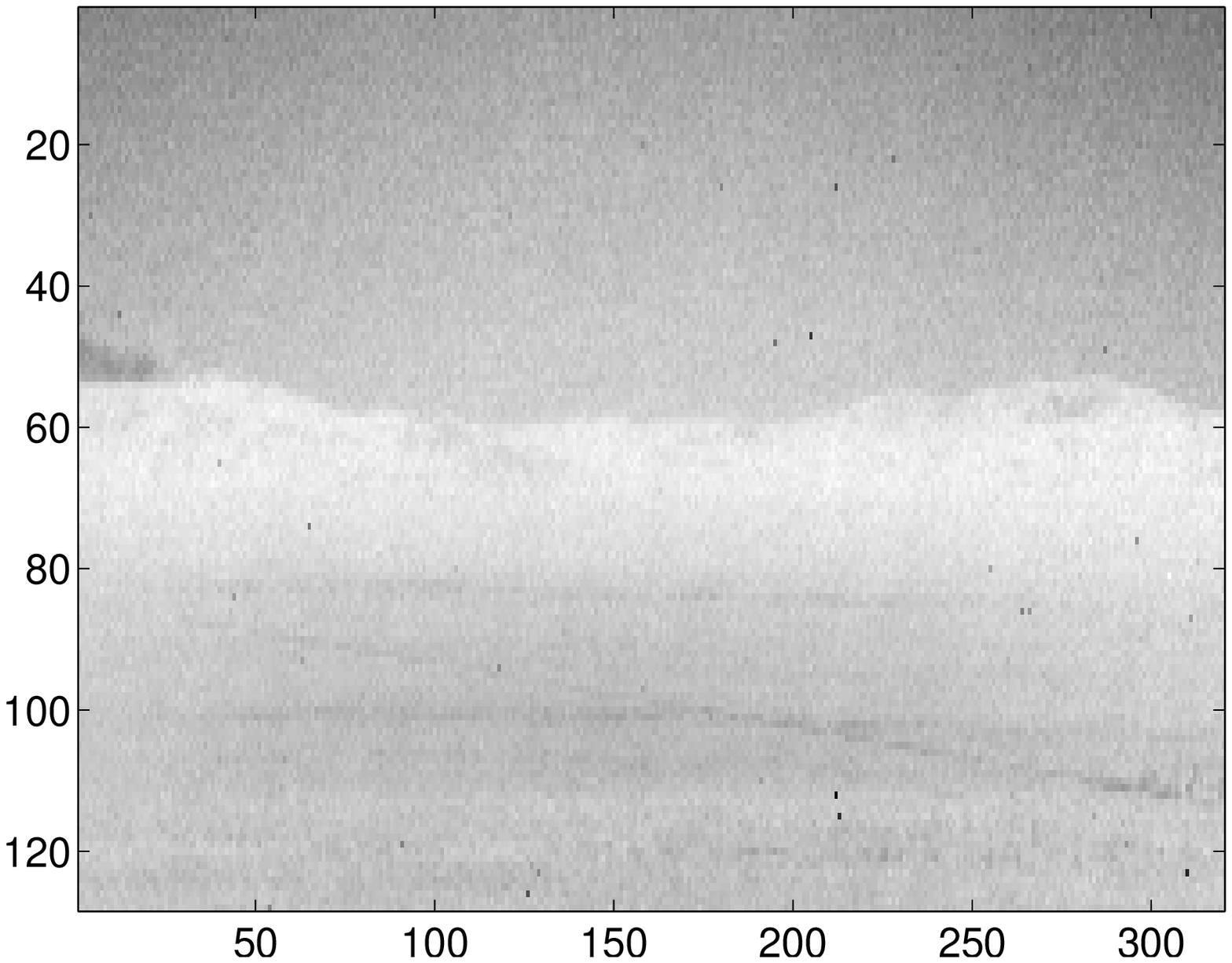}&
	\includegraphics[width=.23\textwidth]{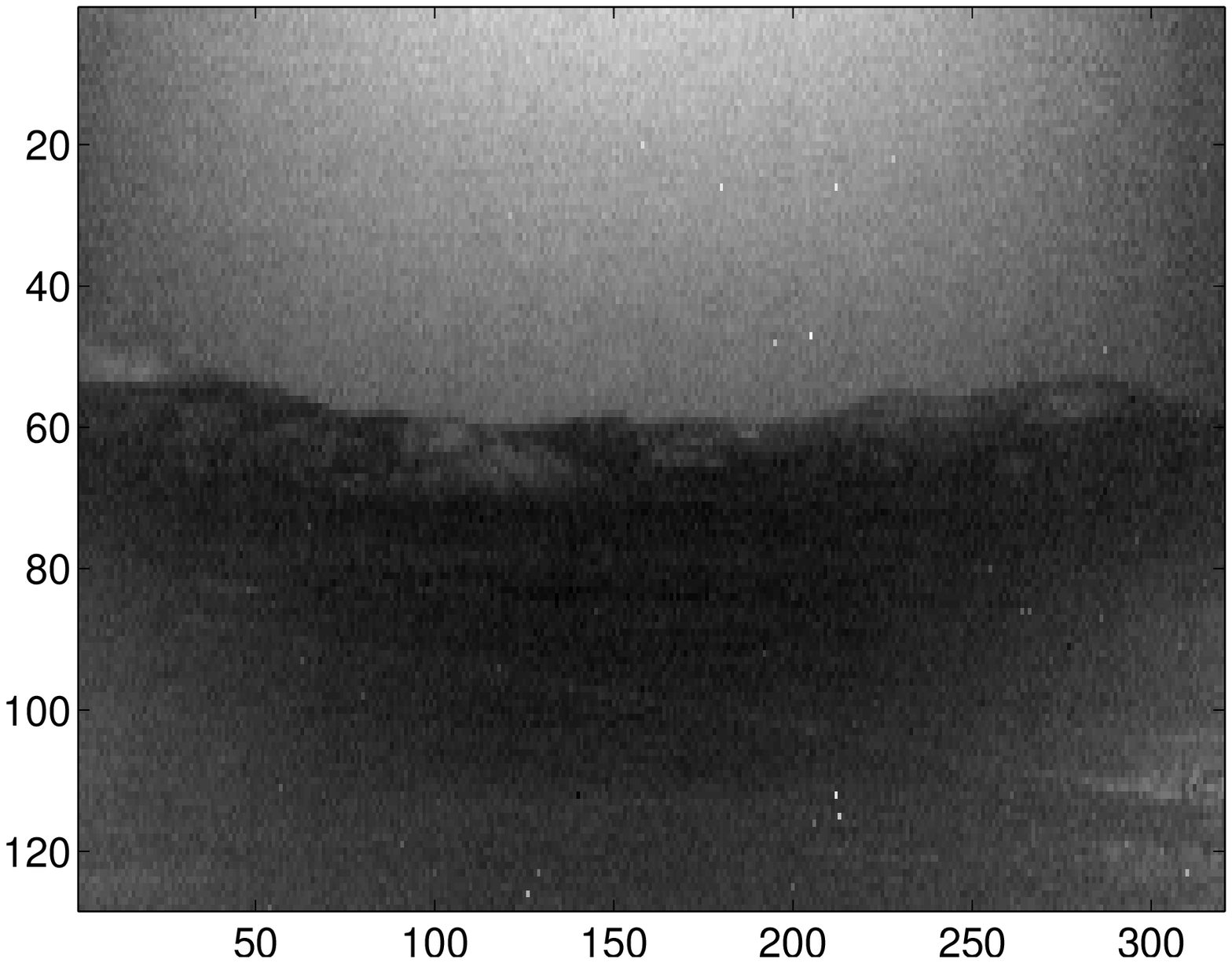}\\
		\multicolumn{4}{c}{Time frame 3}\\
	\includegraphics[width=.23\textwidth]{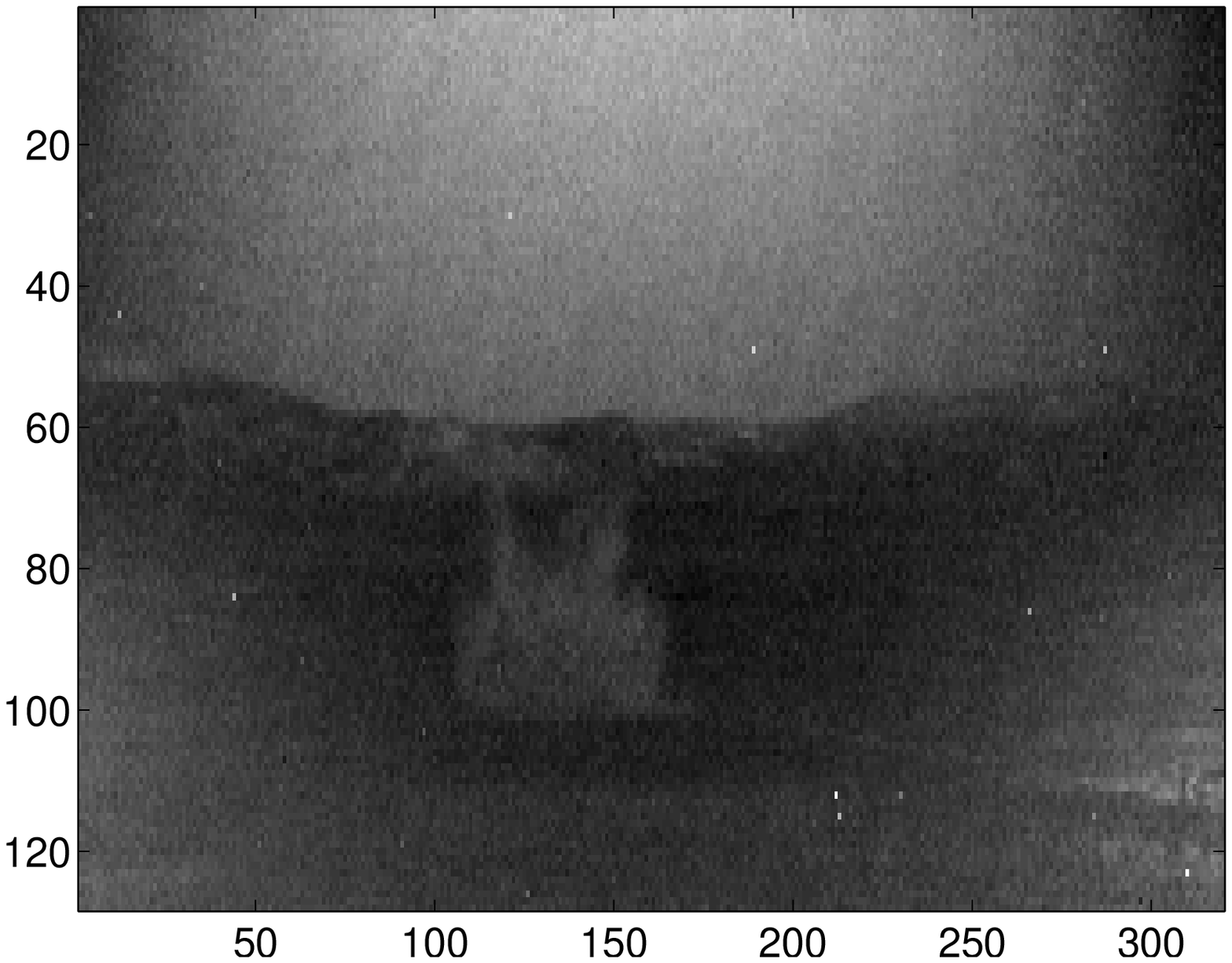}&
	\includegraphics[width=.23\textwidth]{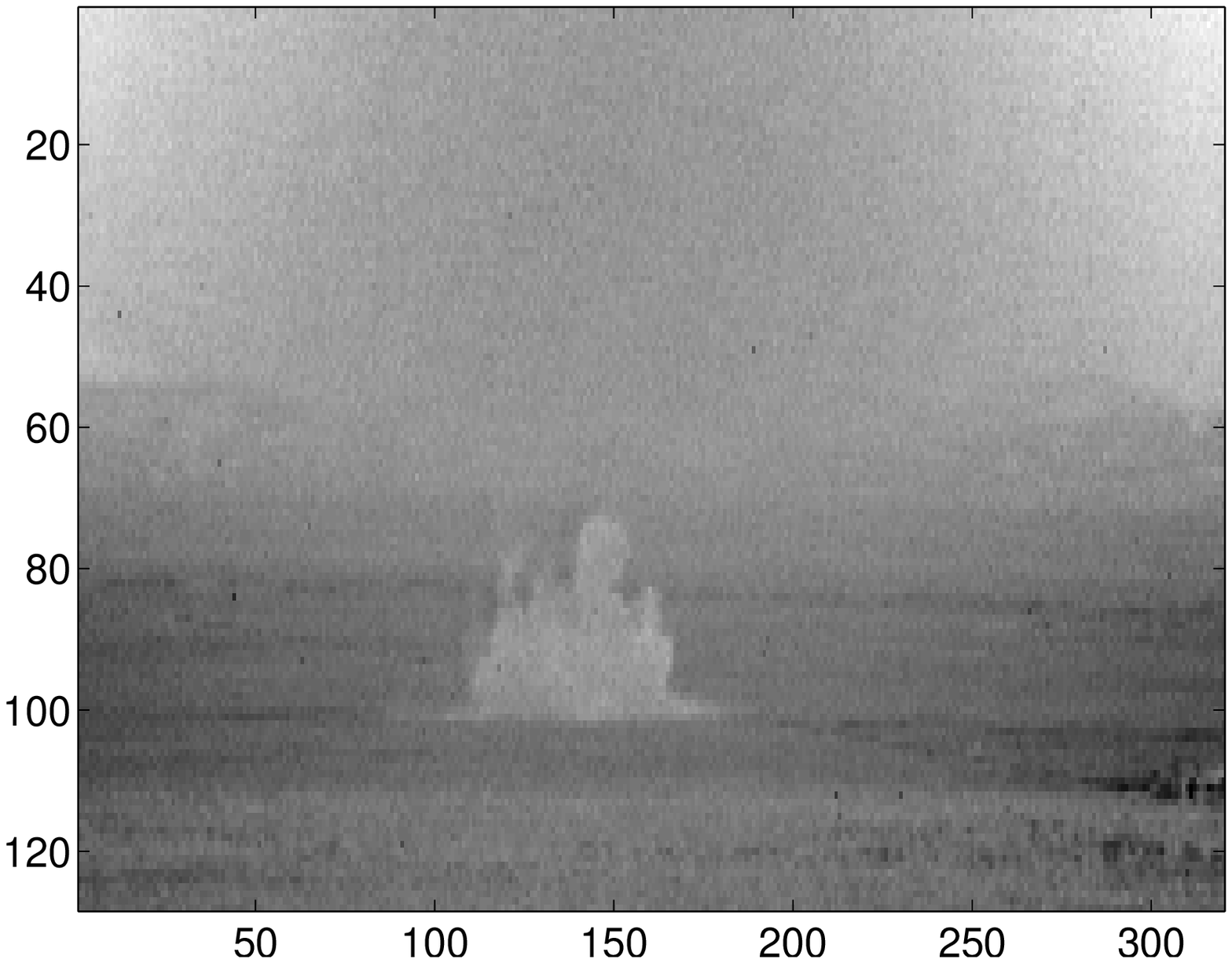}&
	\includegraphics[width=.23\textwidth]{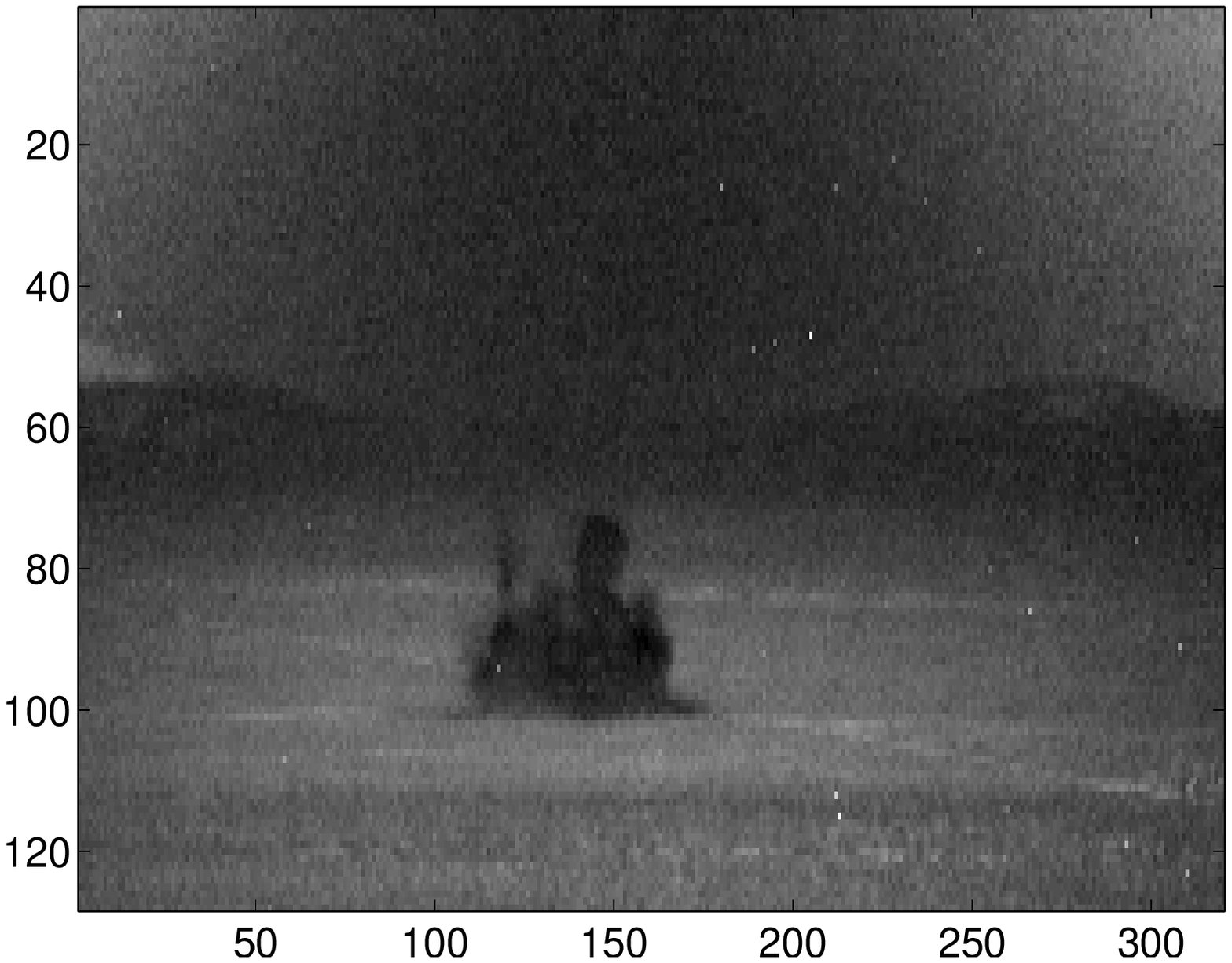}&
	\includegraphics[width=.23\textwidth]{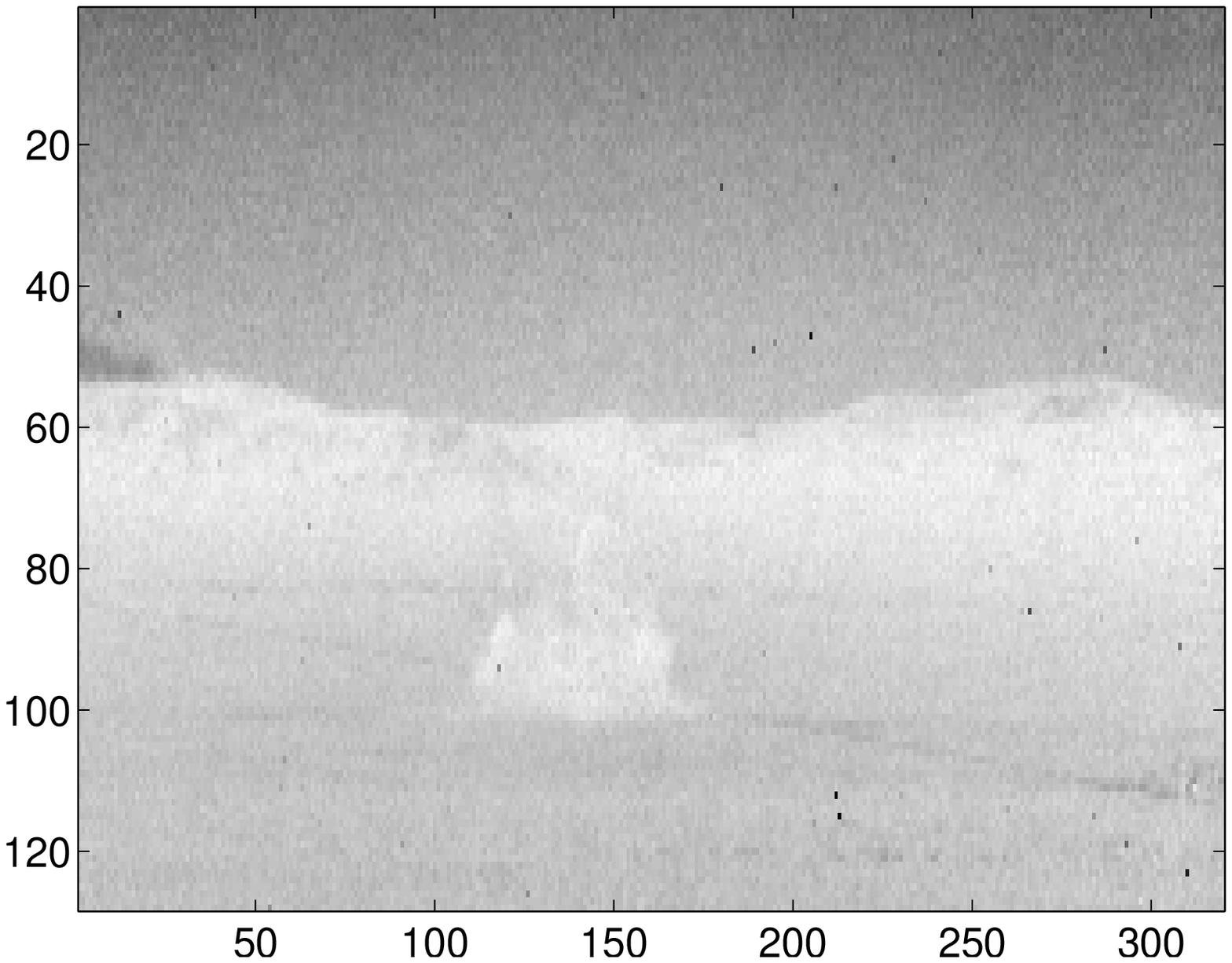}\\
		\multicolumn{4}{c}{Time frame 7}\\
	\includegraphics[width=.23\textwidth]{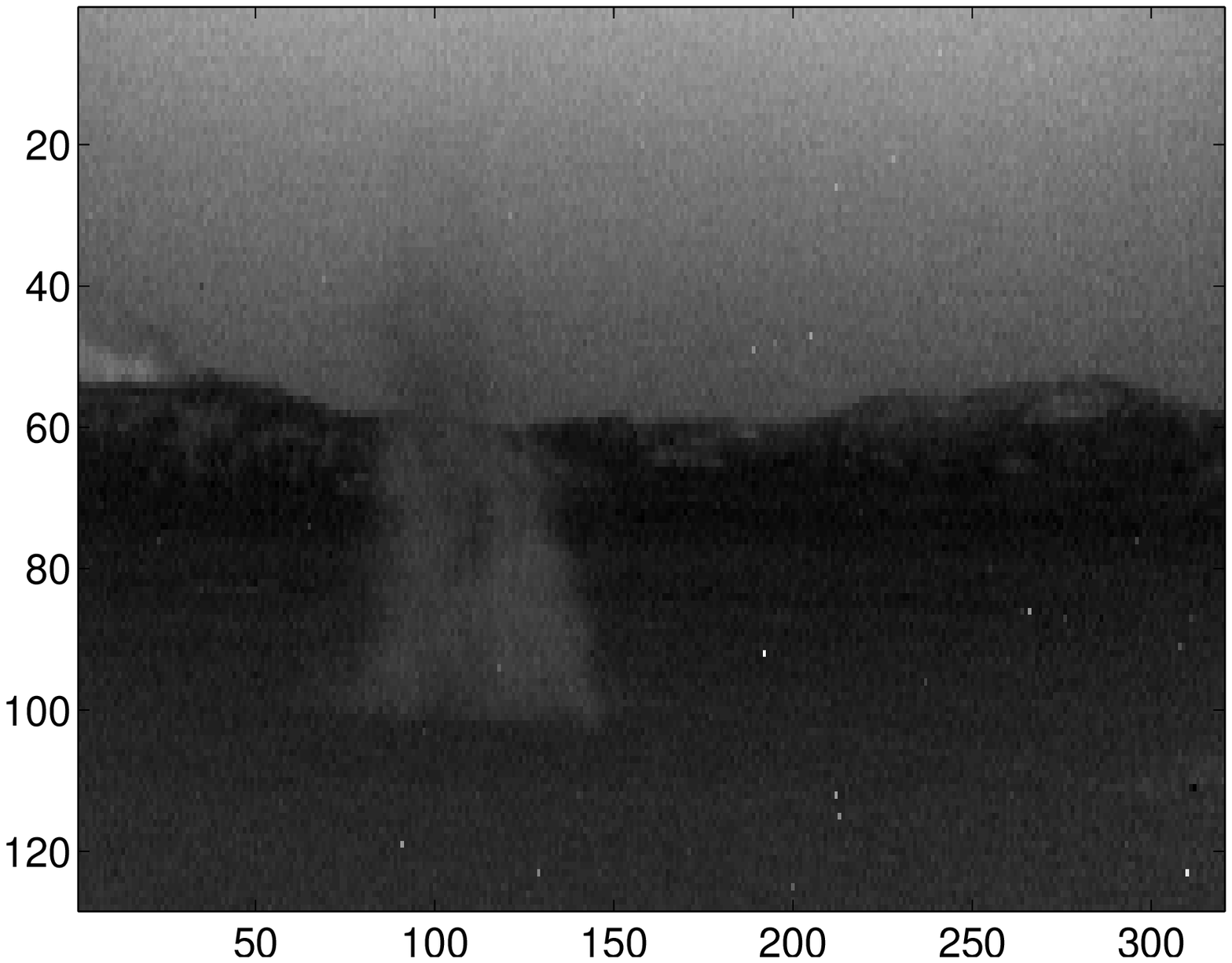}&
	\includegraphics[width=.23\textwidth]{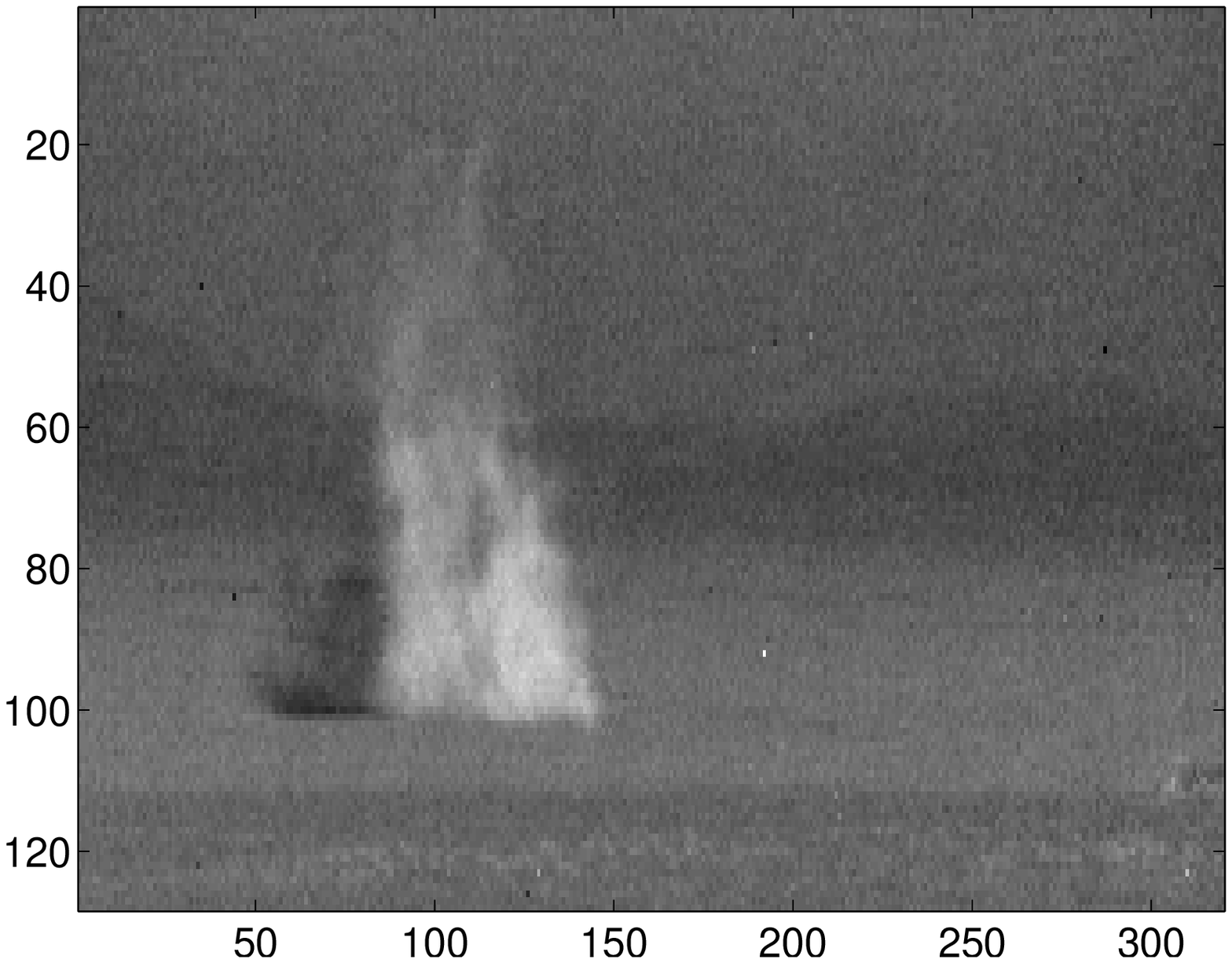}&
	\includegraphics[width=.23\textwidth]{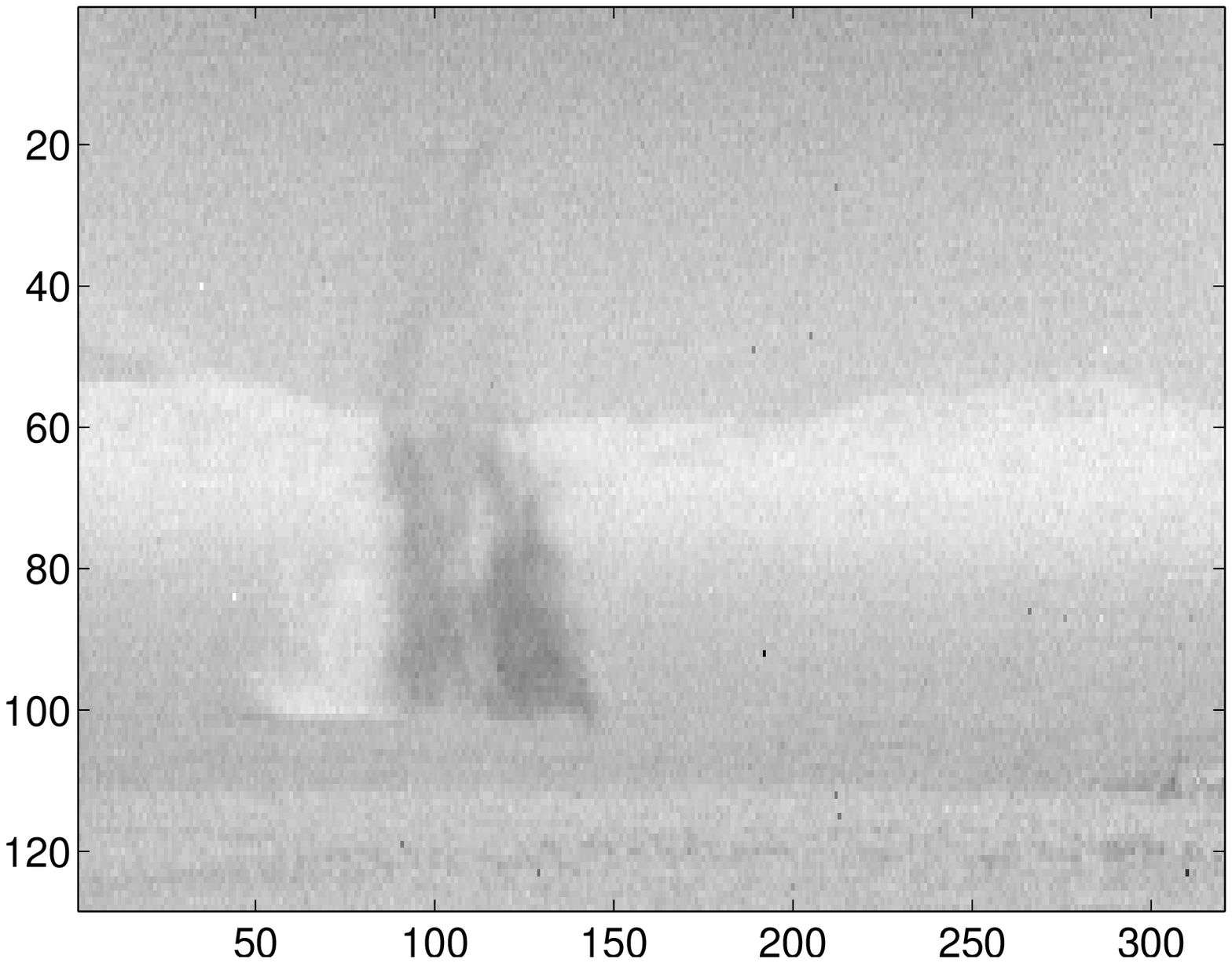}&
	\includegraphics[width=.23\textwidth]{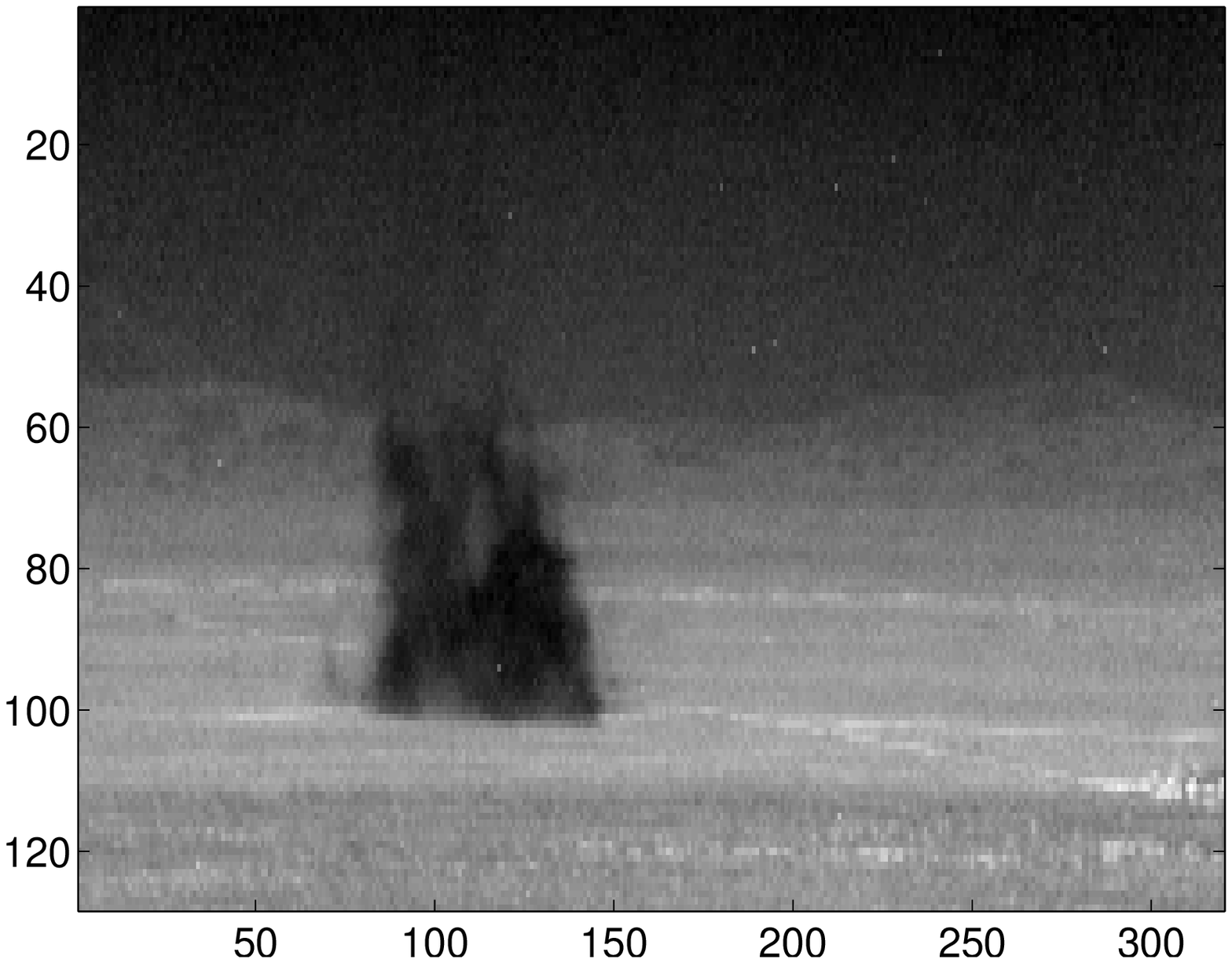}\\
		\multicolumn{4}{c}{Time frame 10}\\
	\includegraphics[width=.23\textwidth]{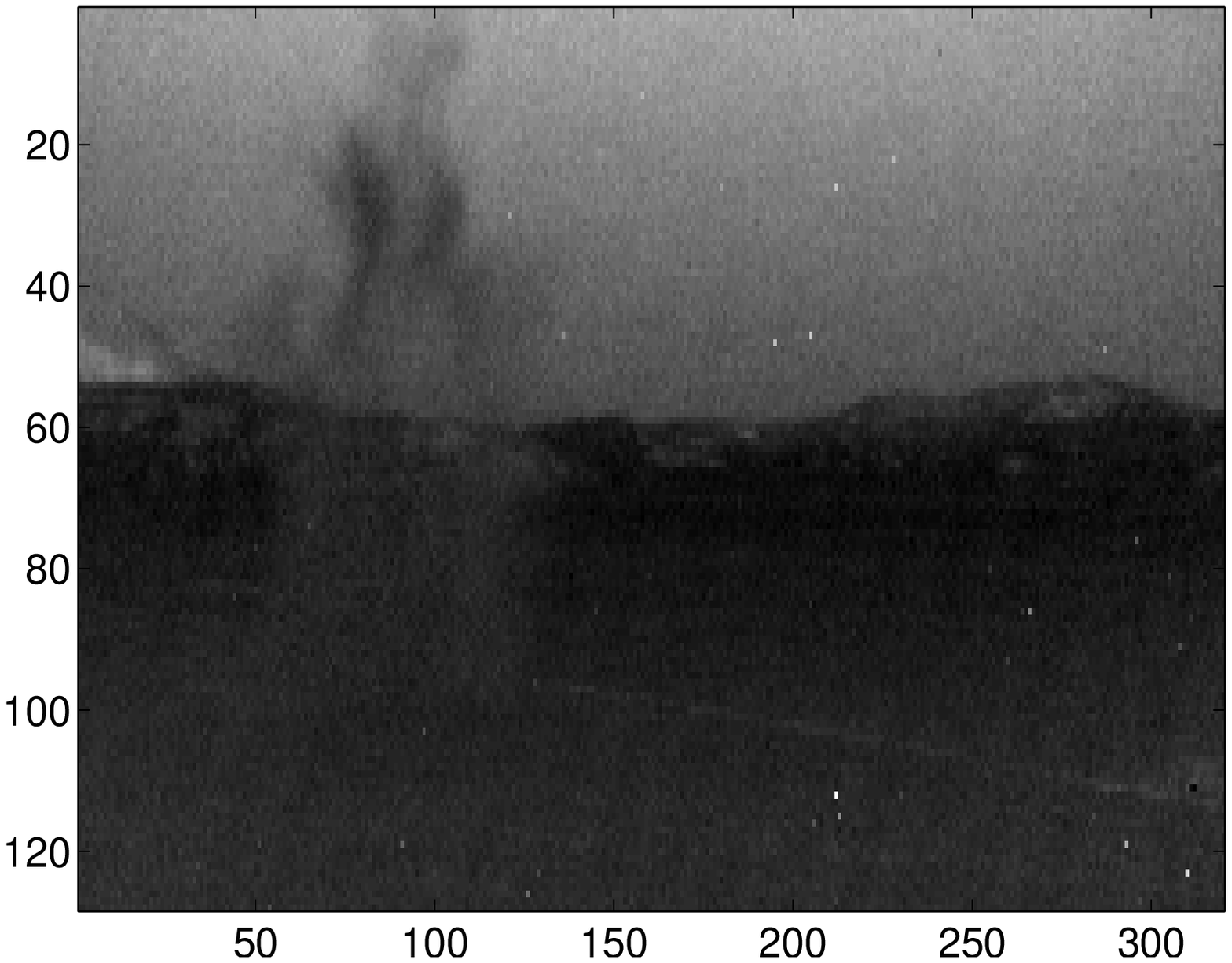}&
	\includegraphics[width=.23\textwidth]{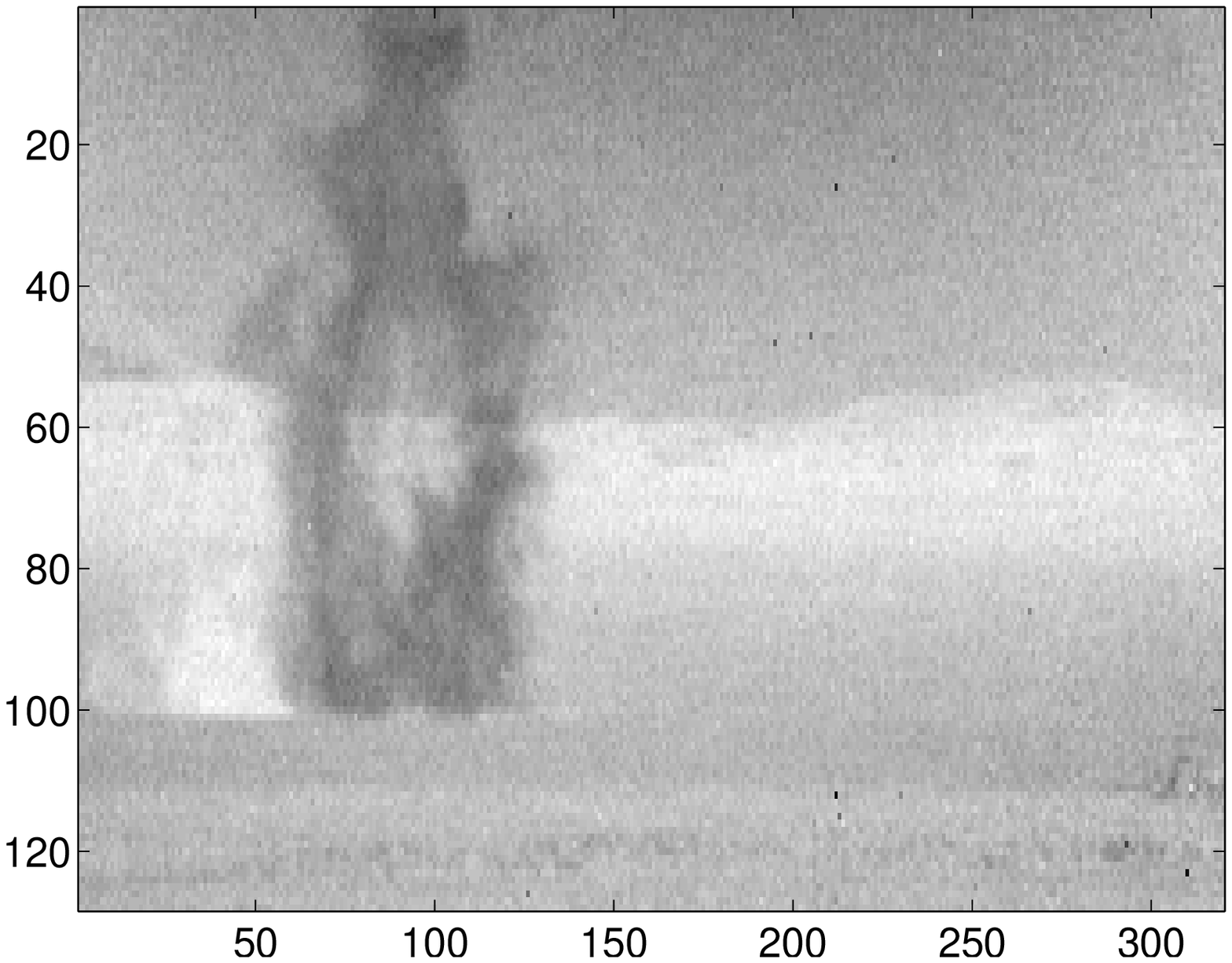}&
	\includegraphics[width=.23\textwidth]{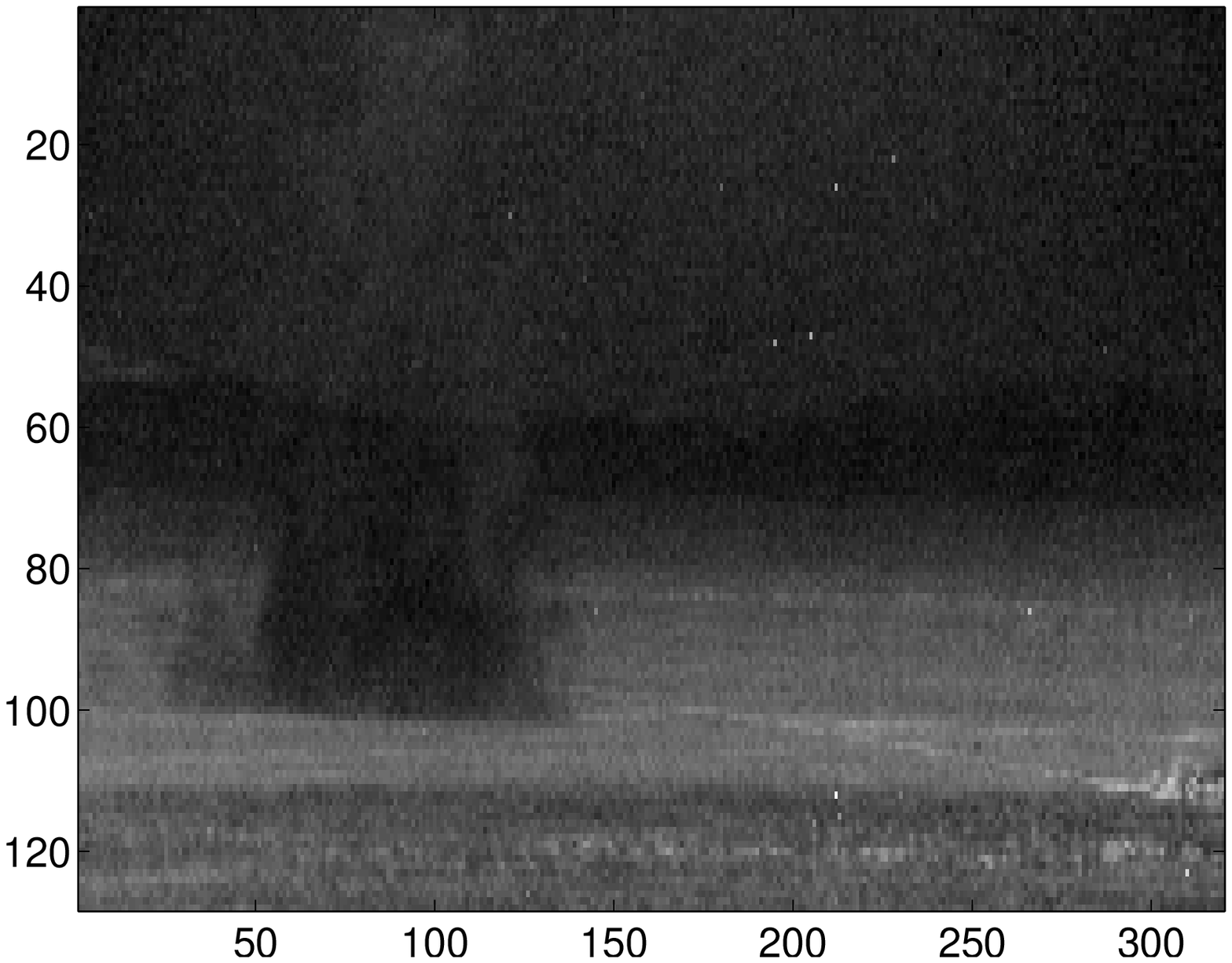}&
	\includegraphics[width=.23\textwidth]{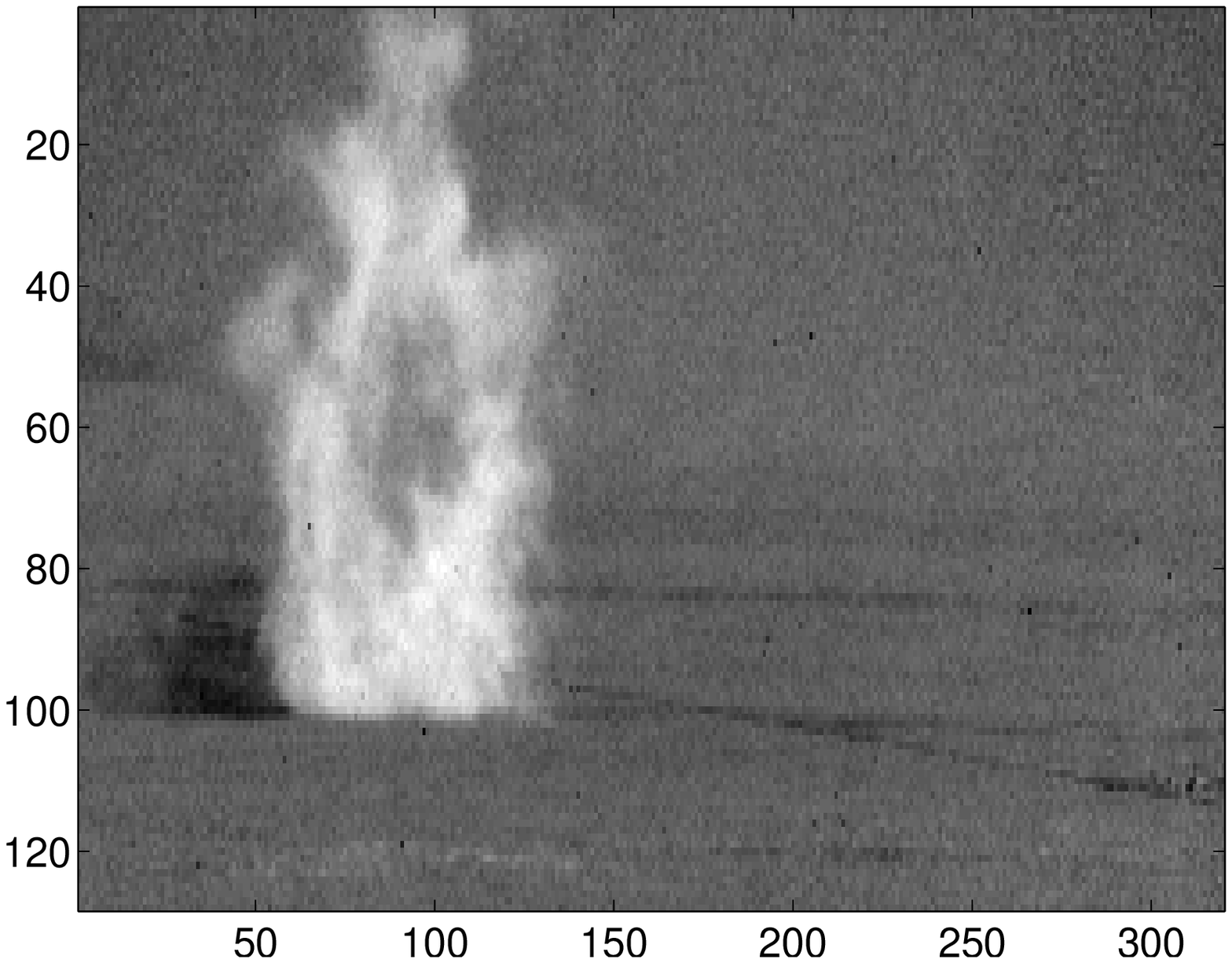}\\
	\end{tabular}
	\caption{\textcolor{black}{Abundance maps obtained by the separate unmixing method : estimated abundance maps of the first, third, seventh and tenth time frame. Each row corresponds to a time frame and each column to a particular source. After occurrence of the gas plume, 'ghosts' corrupt all extracted abundance maps: the gas plume cannot be uniquely attributed to any endmember, preventing a clear physical interpretation of the maps.}}
		\label{fig:abund_sepa}
\end{figure*}

\begin{figure*}[ht]
	\begin{tabular}{cccc}
	Source 1 & Source 2 & Source 3 & Source 4\\
		\multicolumn{4}{c}{Time frame 1}\\
	\includegraphics[width=.23\textwidth]{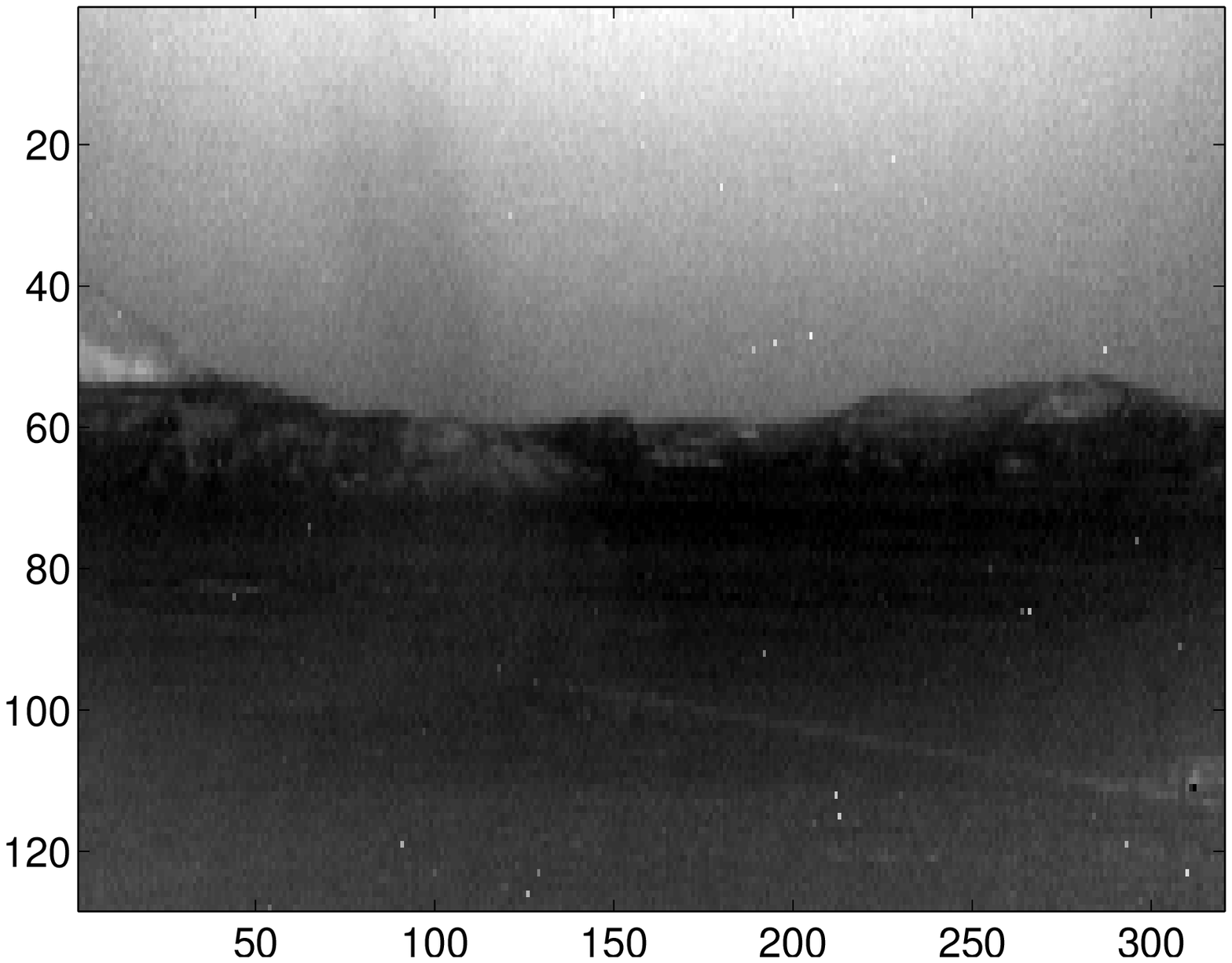}&
	\includegraphics[width=.23\textwidth]{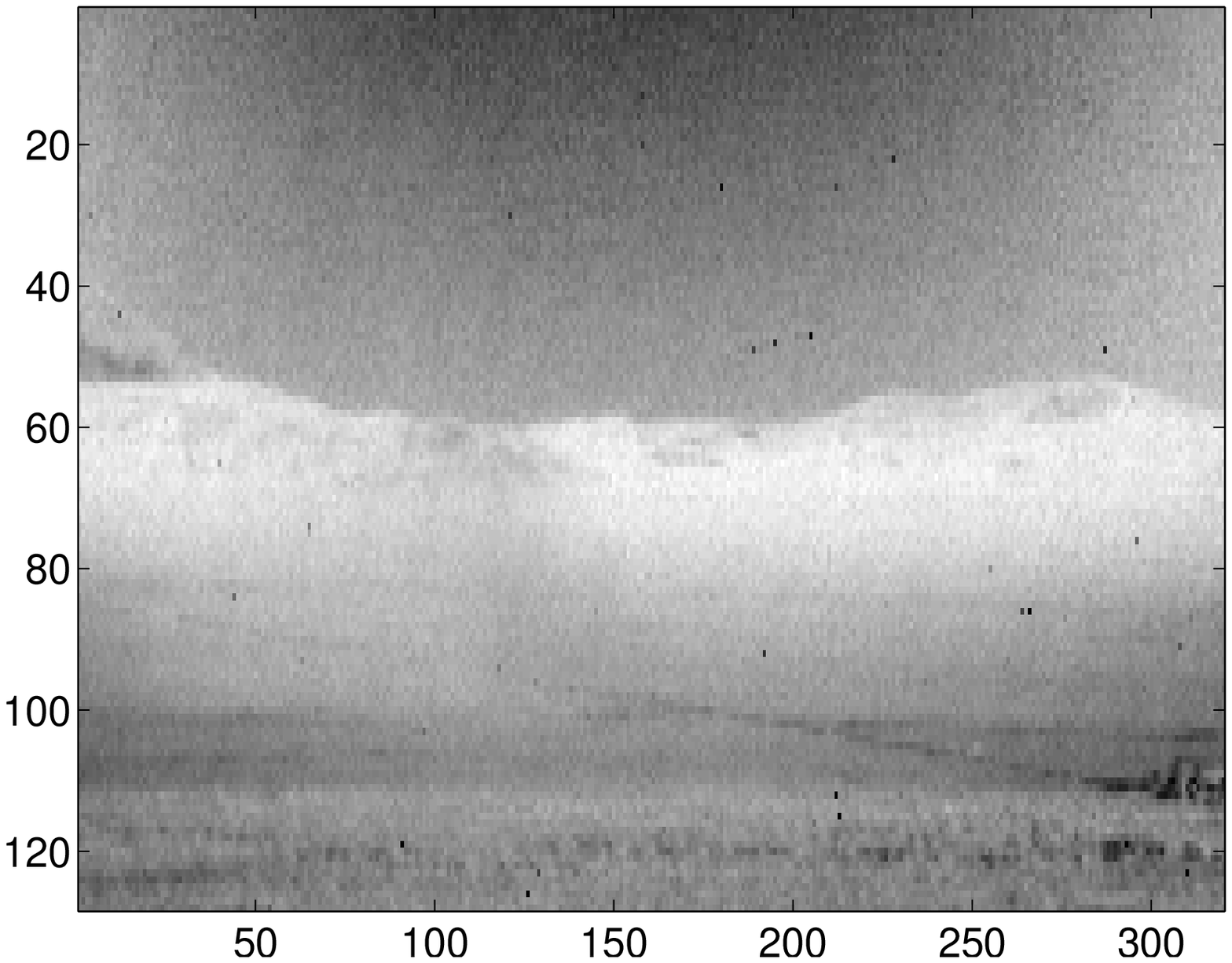}&
	\includegraphics[width=.23\textwidth]{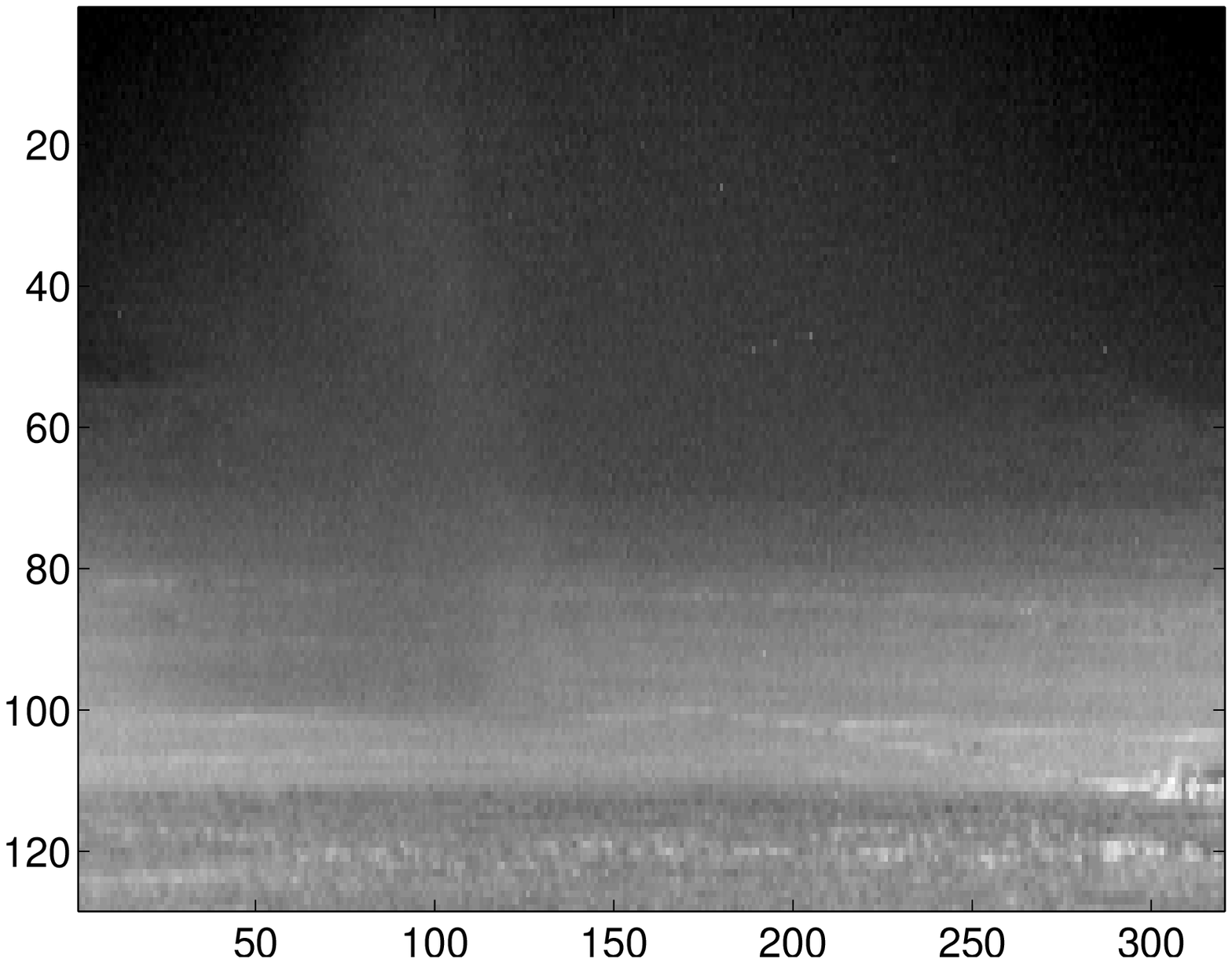}&
	\includegraphics[width=.23\textwidth]{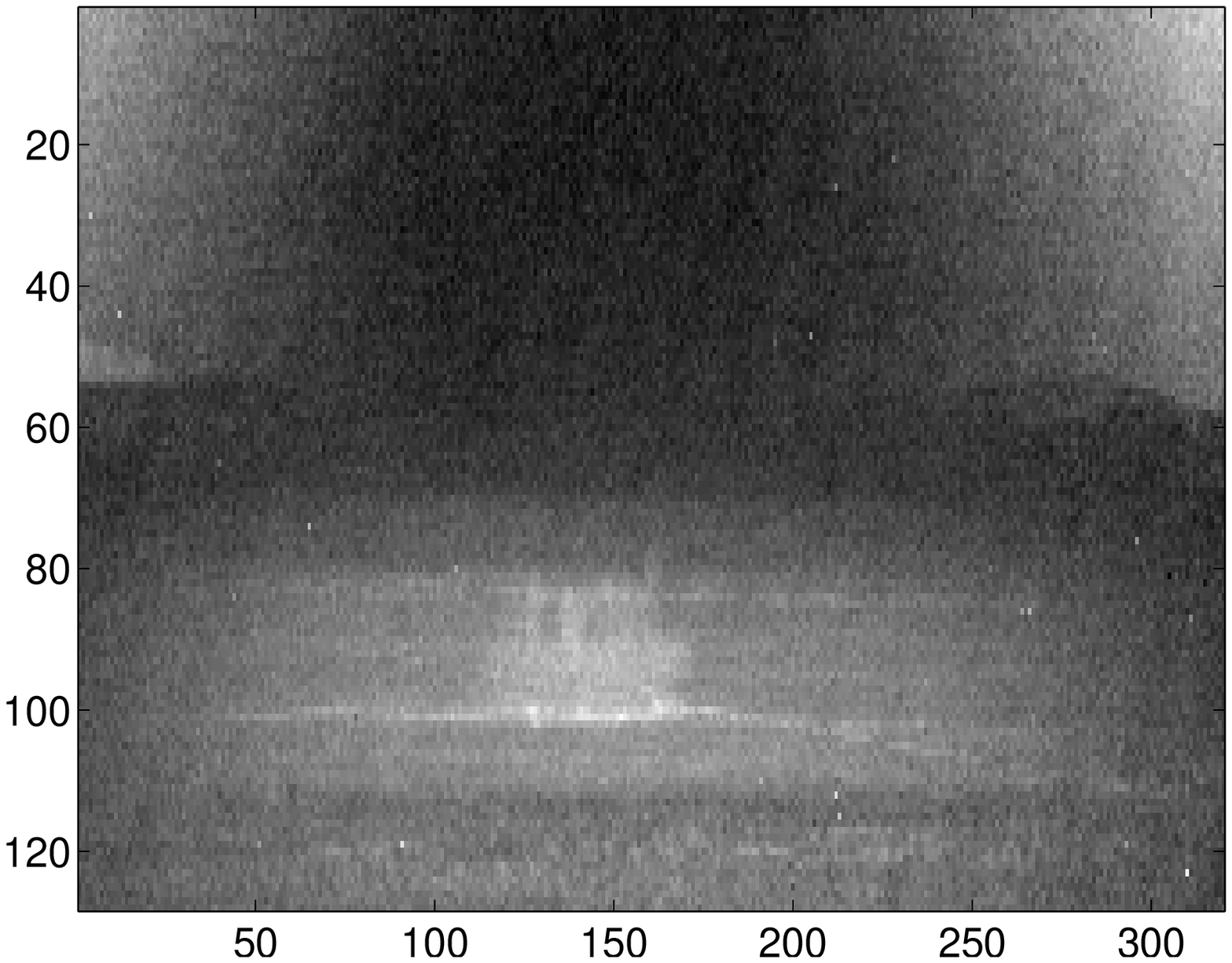}\\
		\multicolumn{4}{c}{Time frame 3}\\
	\includegraphics[width=.23\textwidth]{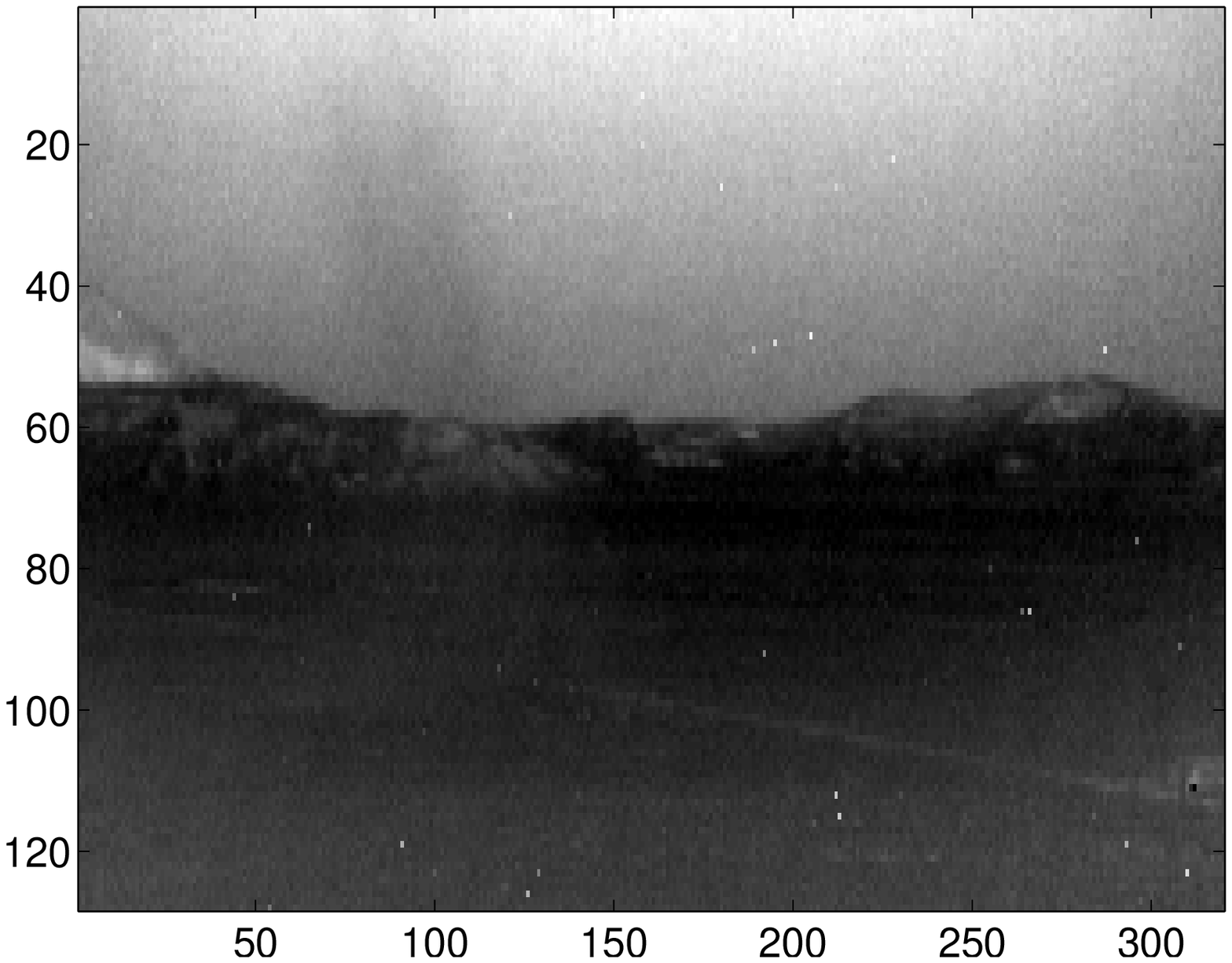}&
	\includegraphics[width=.23\textwidth]{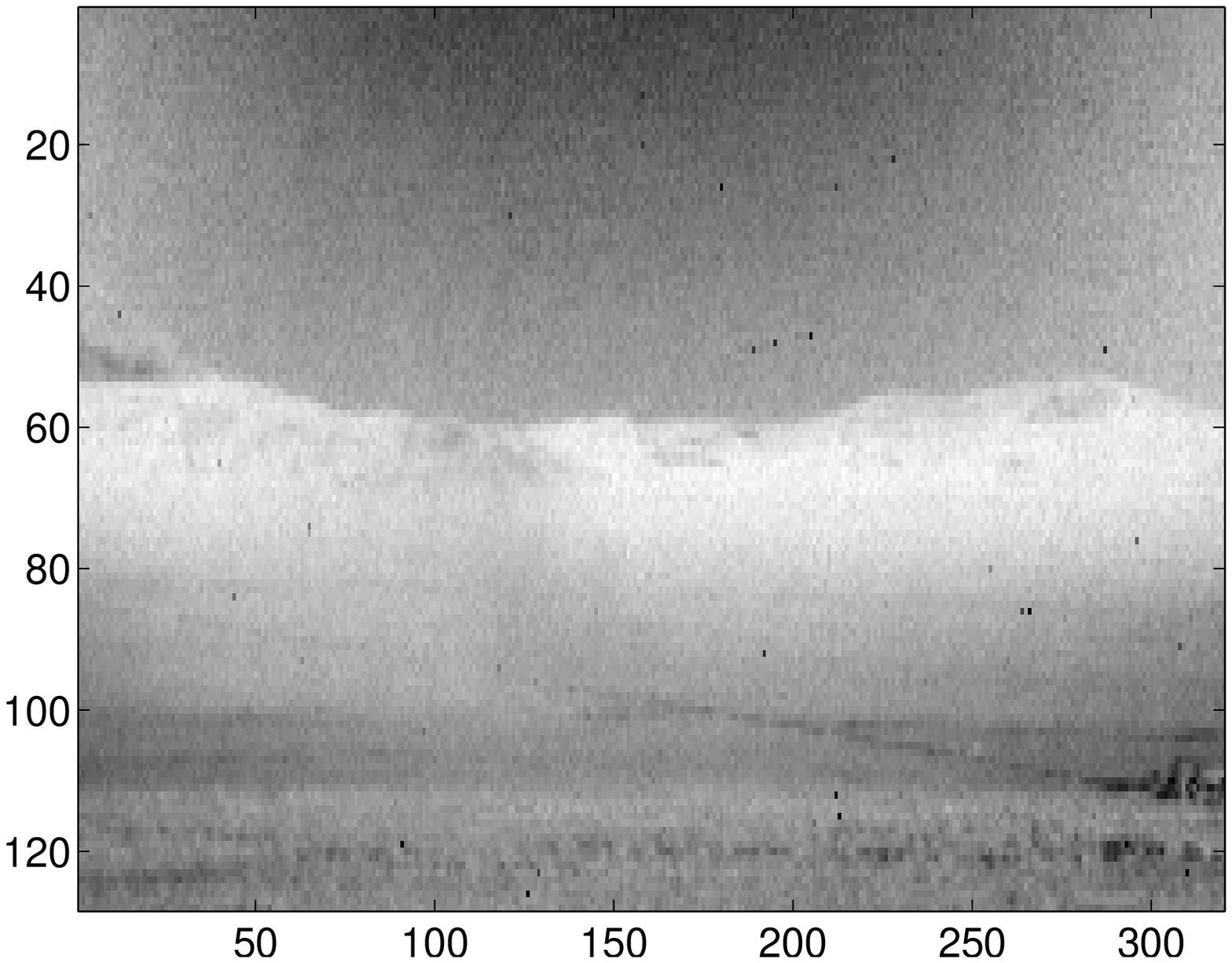}&
	\includegraphics[width=.23\textwidth]{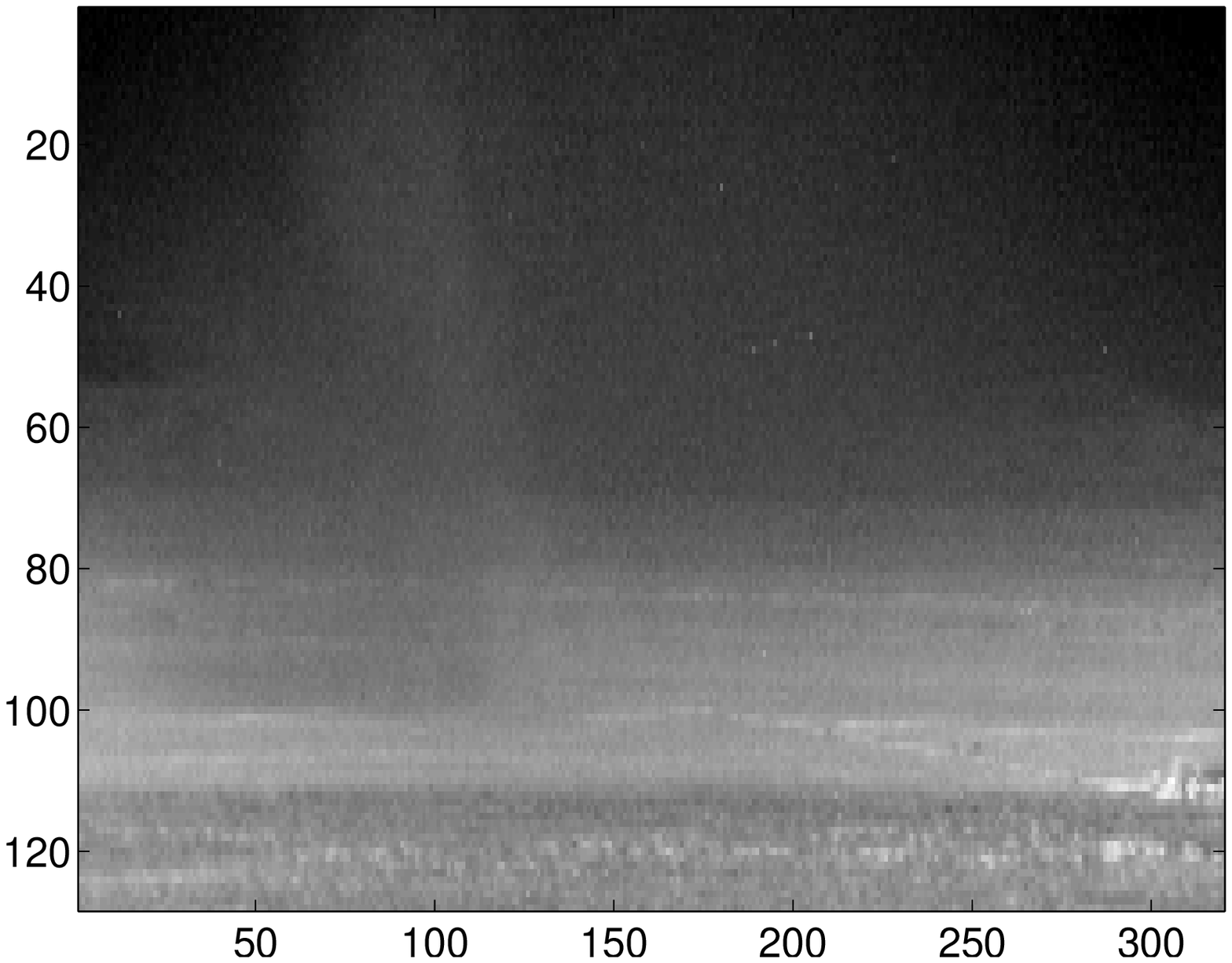}&
	\includegraphics[width=.23\textwidth]{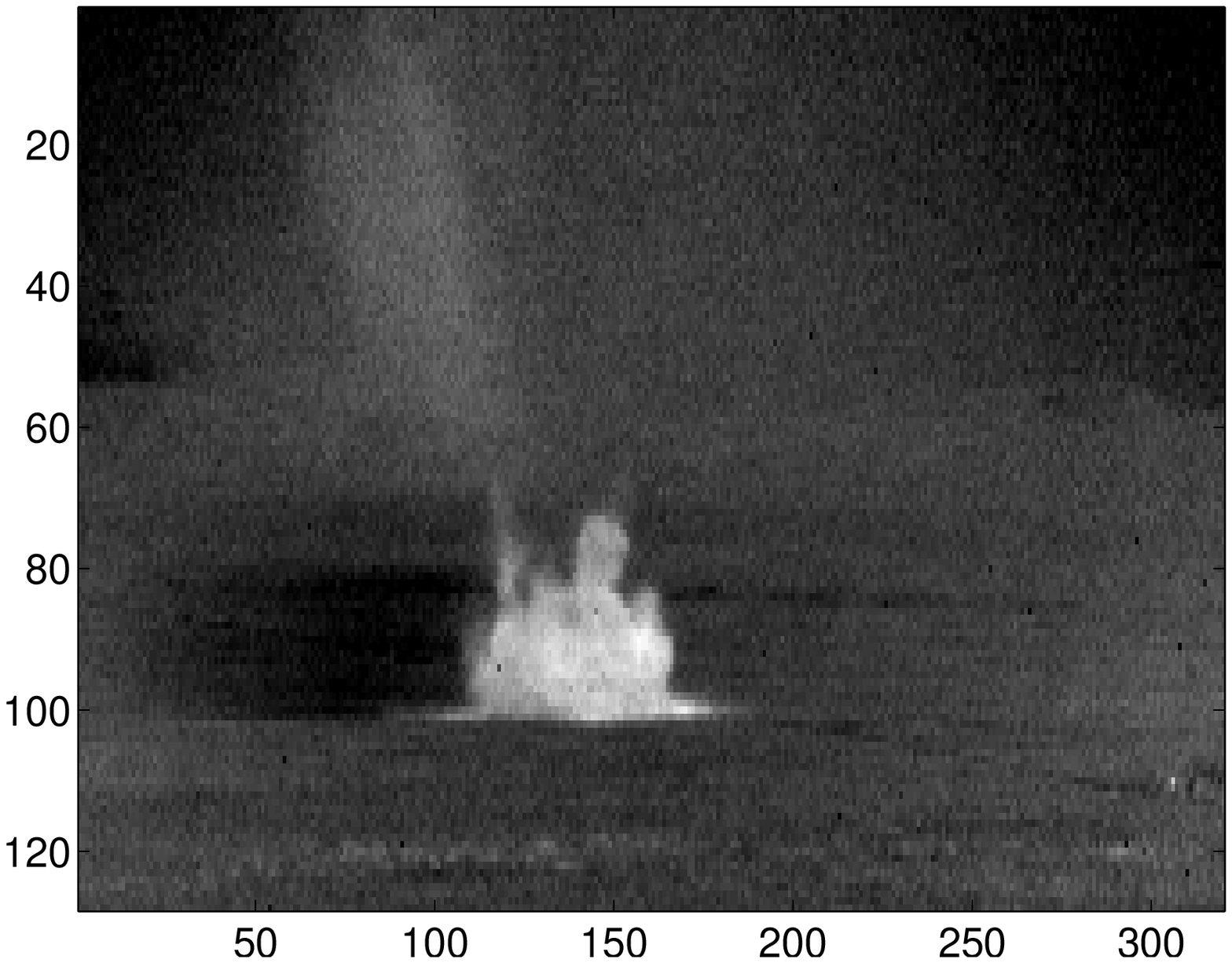}\\
		\multicolumn{4}{c}{Time frame 7}\\
	\includegraphics[width=.23\textwidth]{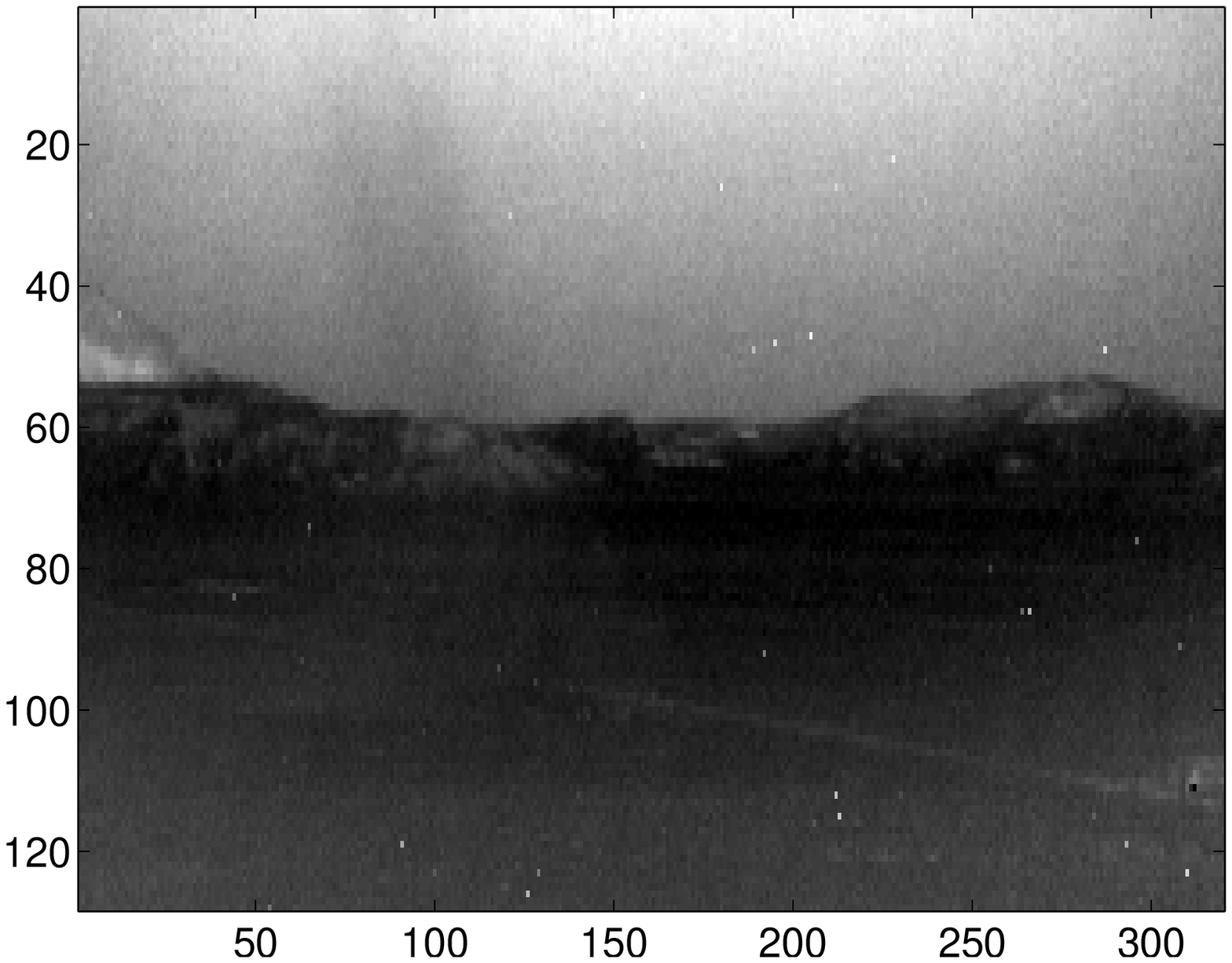}&
	\includegraphics[width=.23\textwidth]{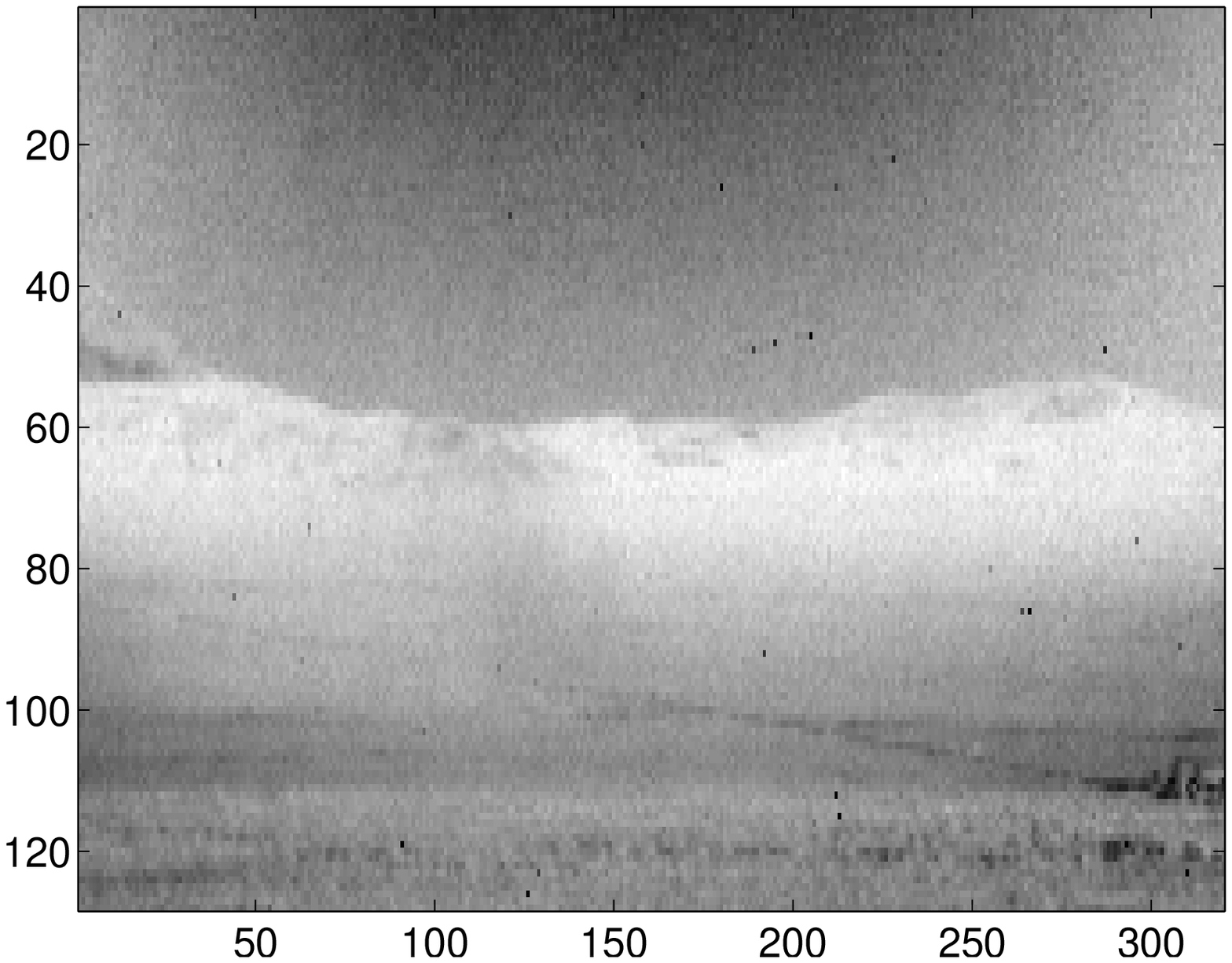}&
	\includegraphics[width=.23\textwidth]{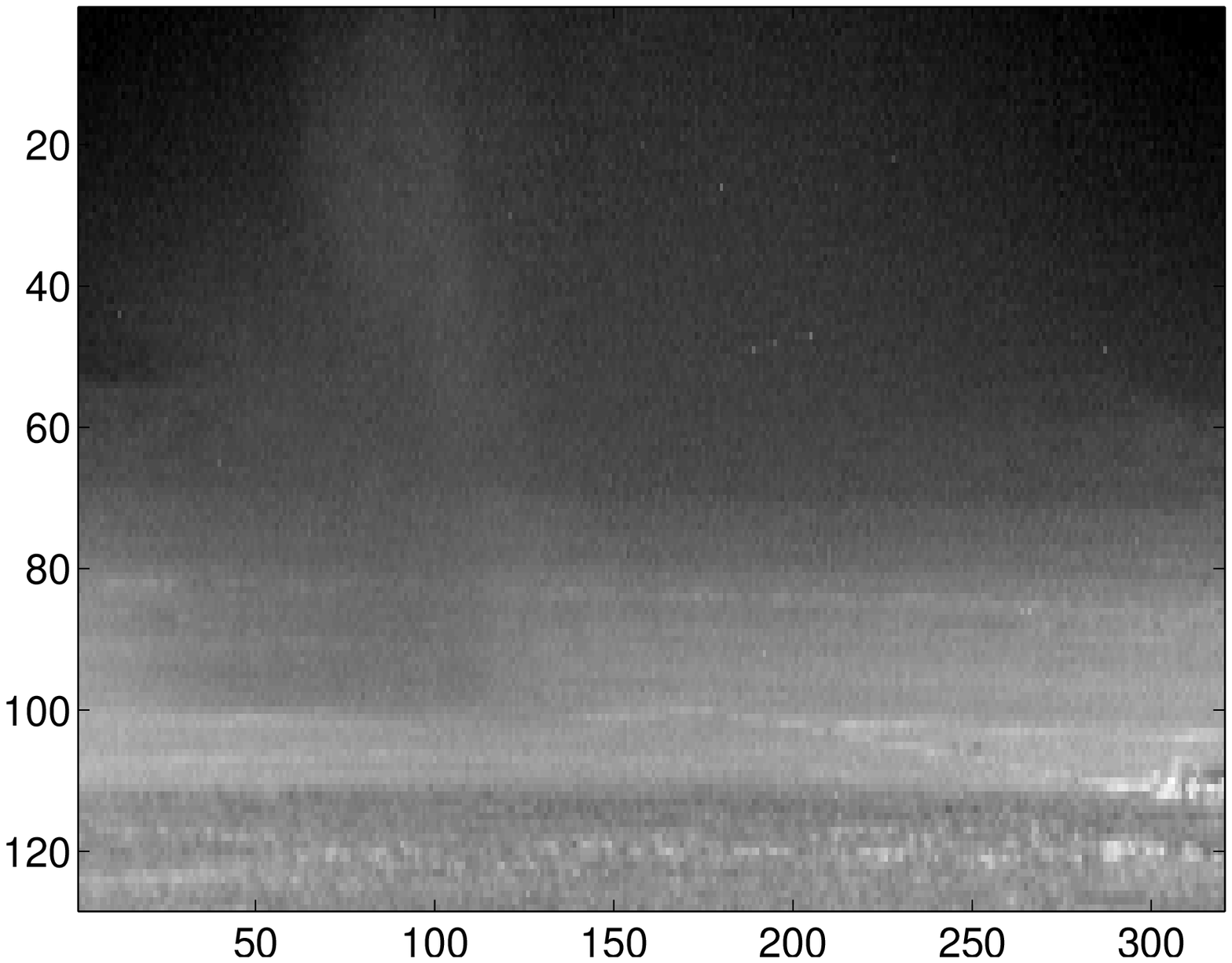}&
	\includegraphics[width=.23\textwidth]{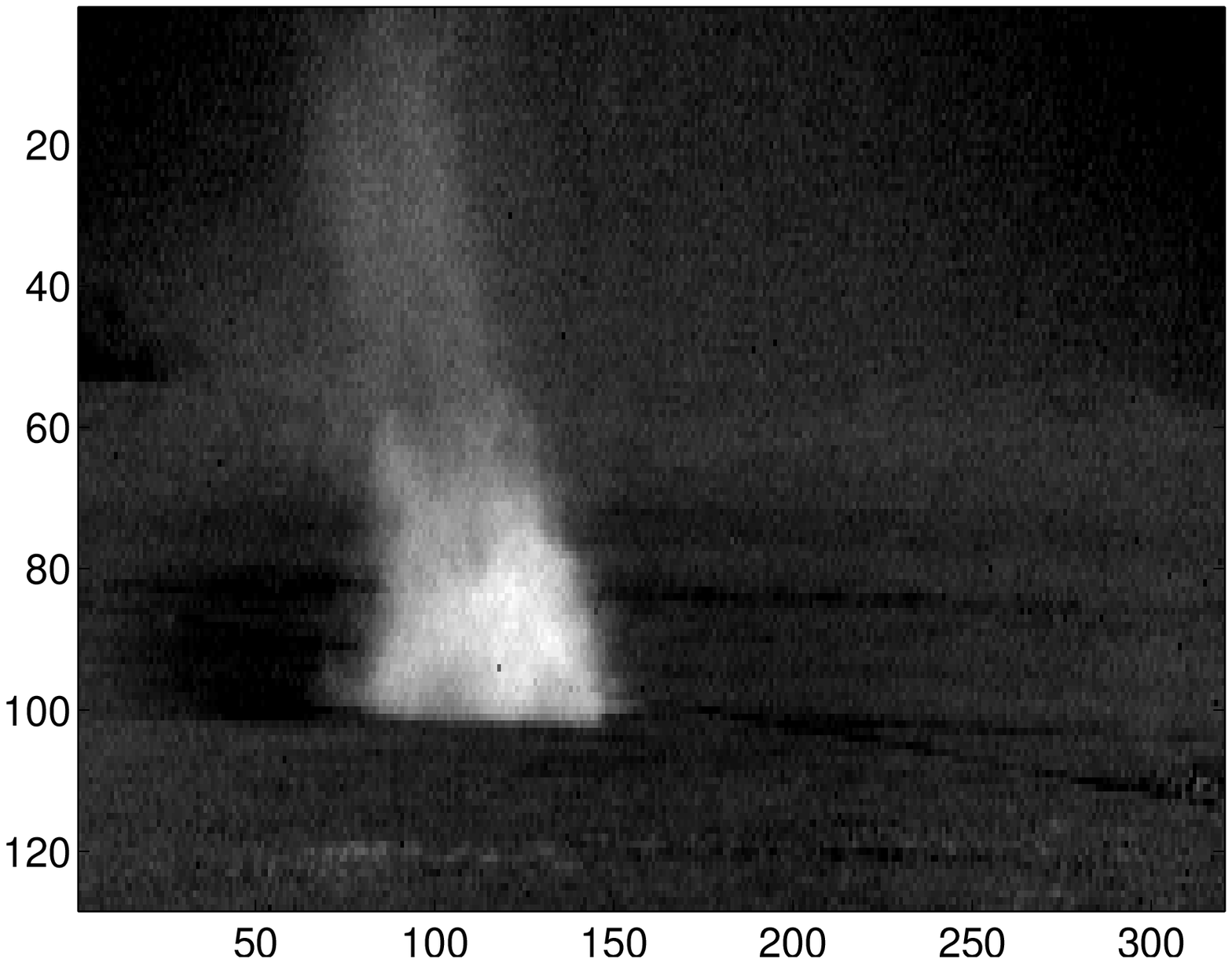}\\
		\multicolumn{4}{c}{Time frame 10}\\
	\includegraphics[width=.23\textwidth]{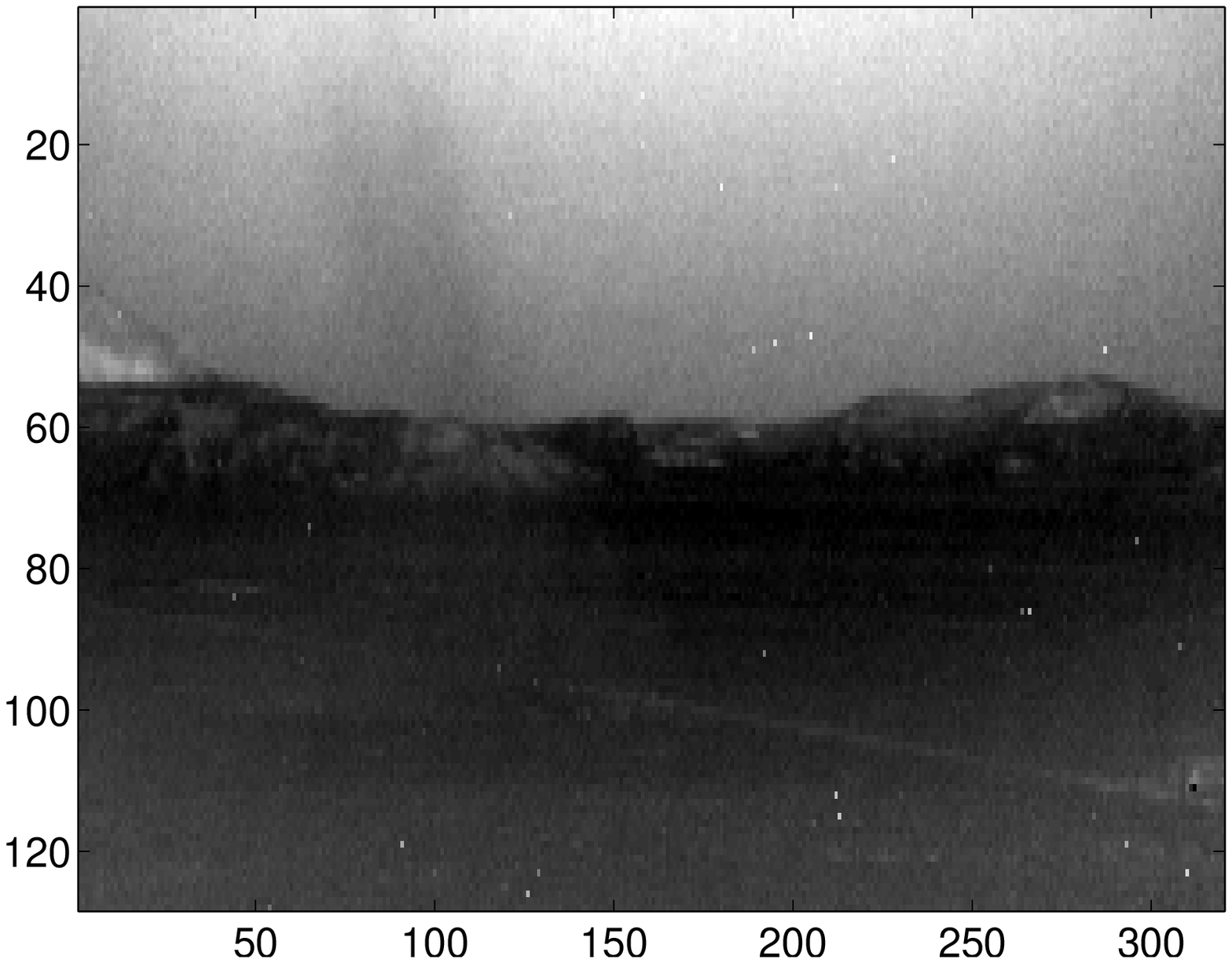}&
	\includegraphics[width=.23\textwidth]{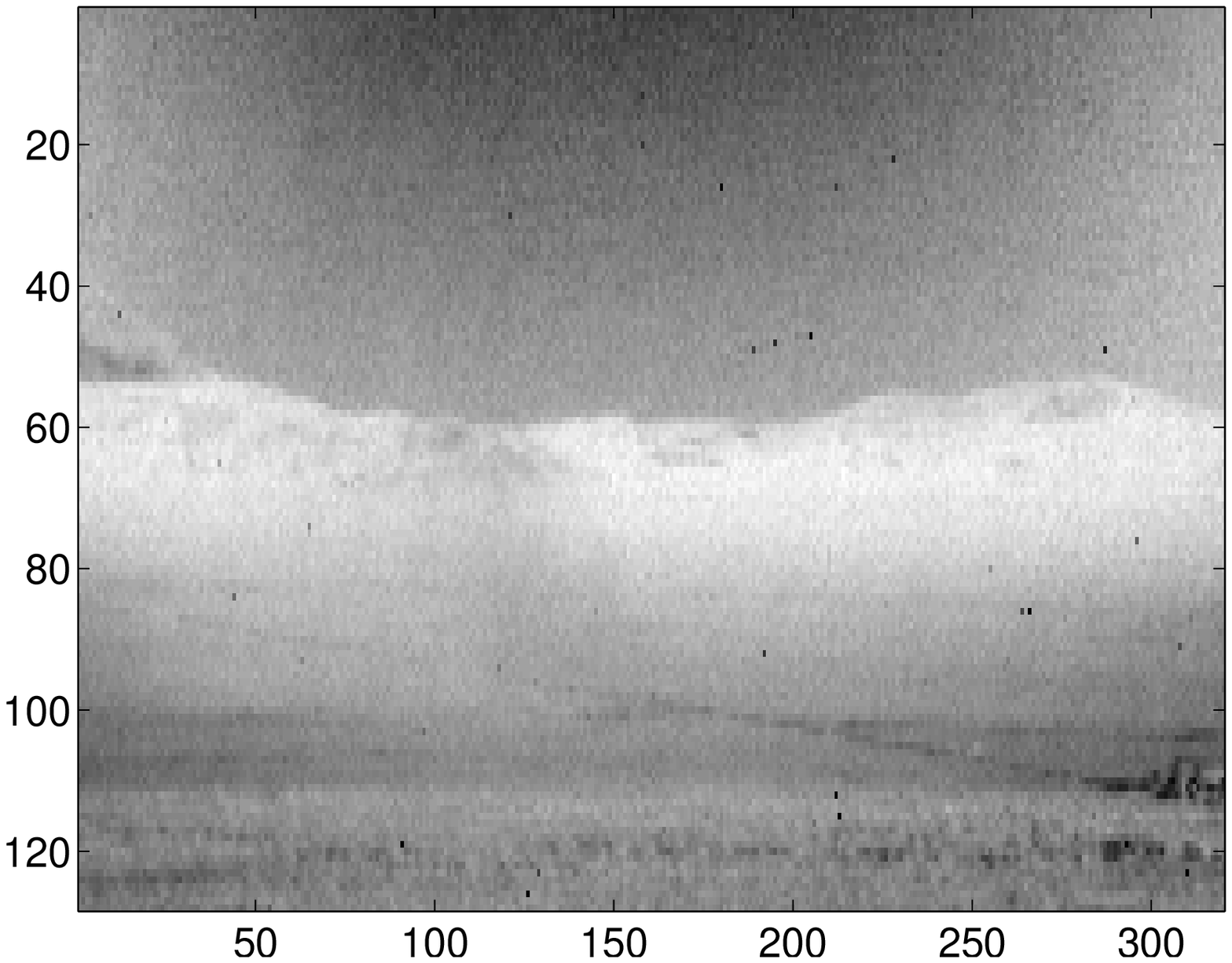}&
	\includegraphics[width=.23\textwidth]{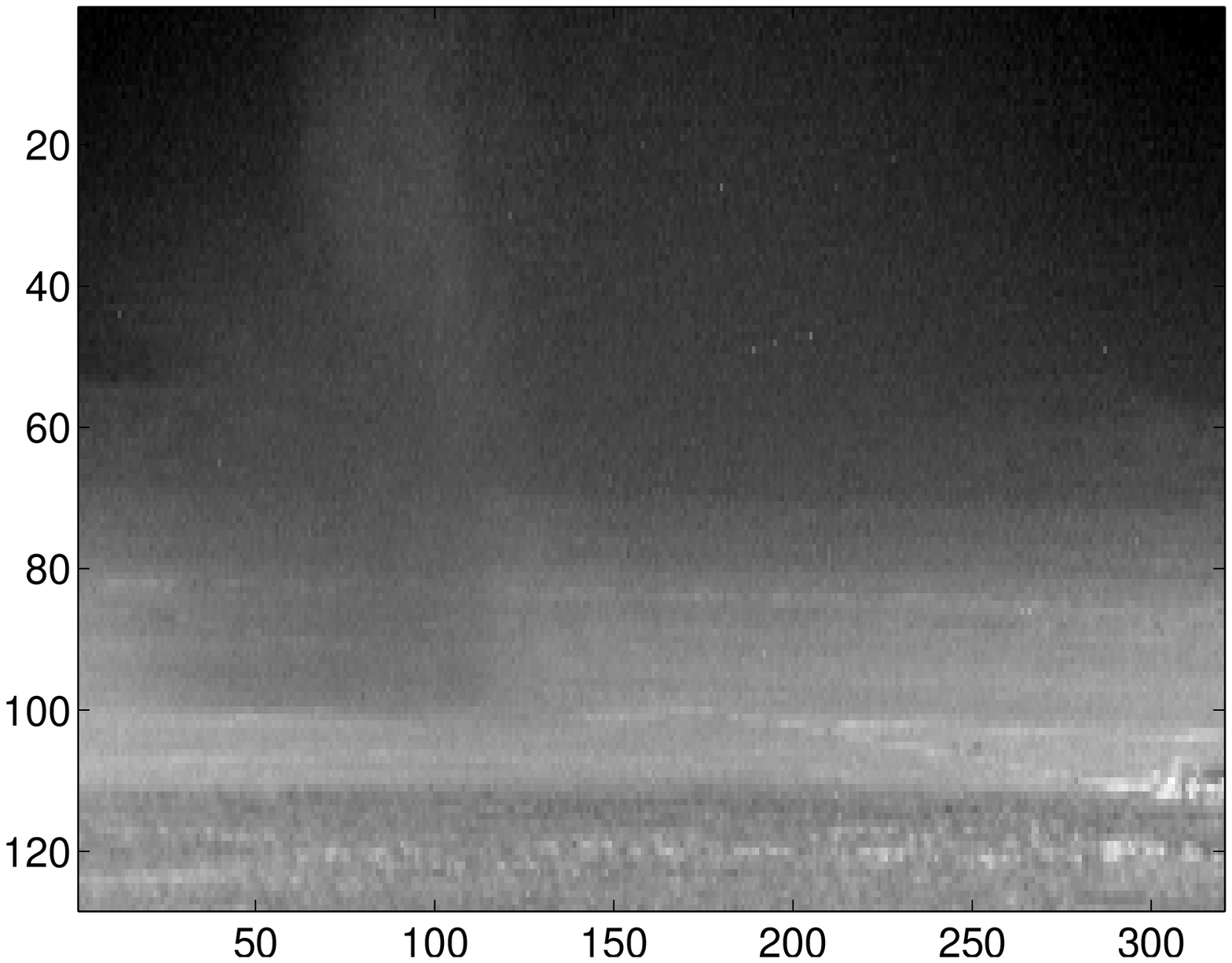}&
	\includegraphics[width=.23\textwidth]{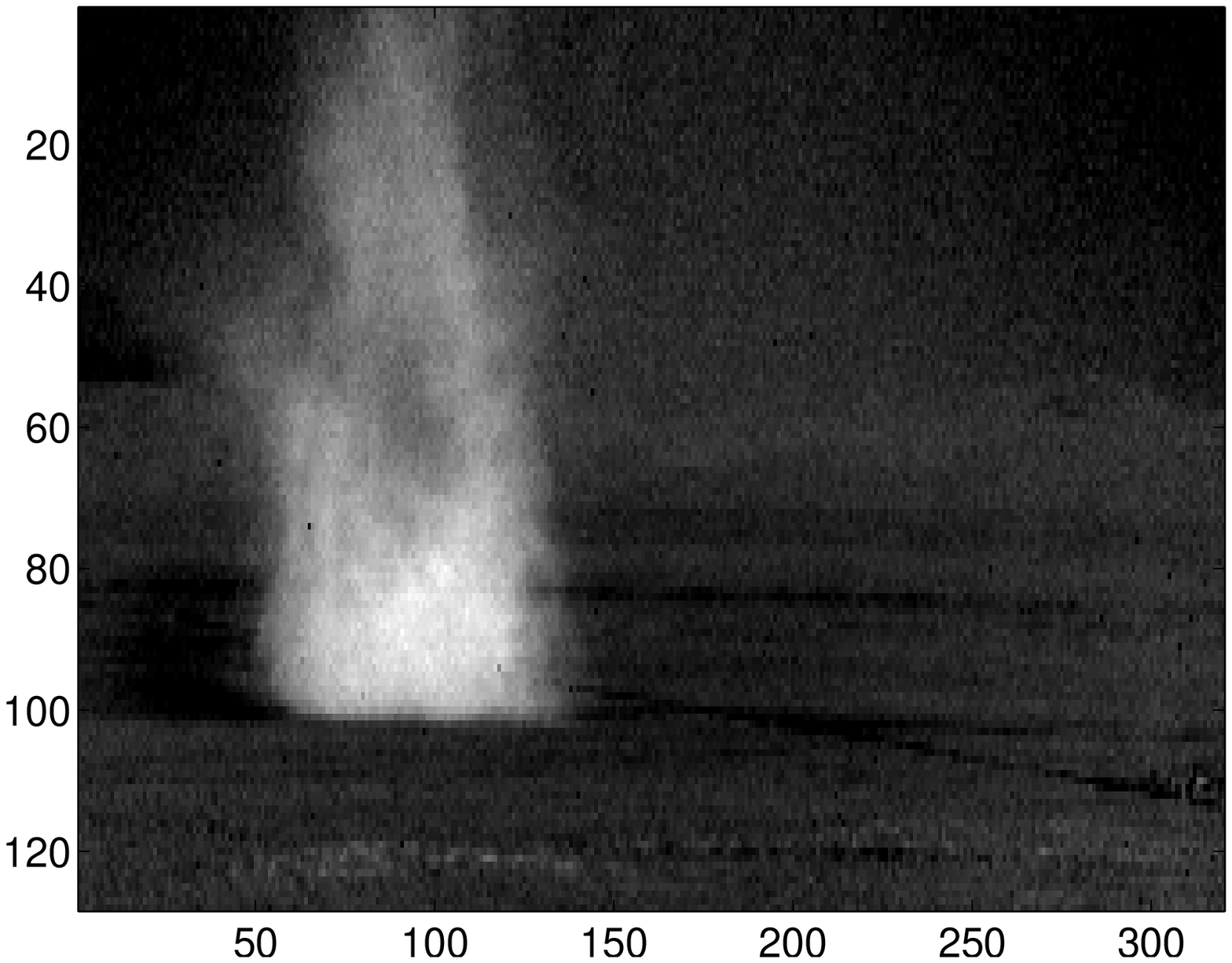}\\
	\end{tabular}
	\caption{\textcolor{black}{Abundance maps obtained by the joint unmixing method : estimated abundance maps of the first, third, seventh and tenth time frame. Each row corresponds to a time frame and each column to a particular source. Notice how 'ghosts' are significantly attenuated by accounting for the dynamical model of the data: abundances of sources 1 to 3 are barely corrupted by gas-plume related information (source 4).}}
		\label{fig:abund_joint}
\end{figure*}

\begin{figure*}[ht]
	\begin{tabular}{cccc}
	Source 1 & Source 2 & Source 3 & Source 4\\
		\multicolumn{4}{c}{Time frame 1}\\
	\includegraphics[width=.23\textwidth,height=.1\textheight]{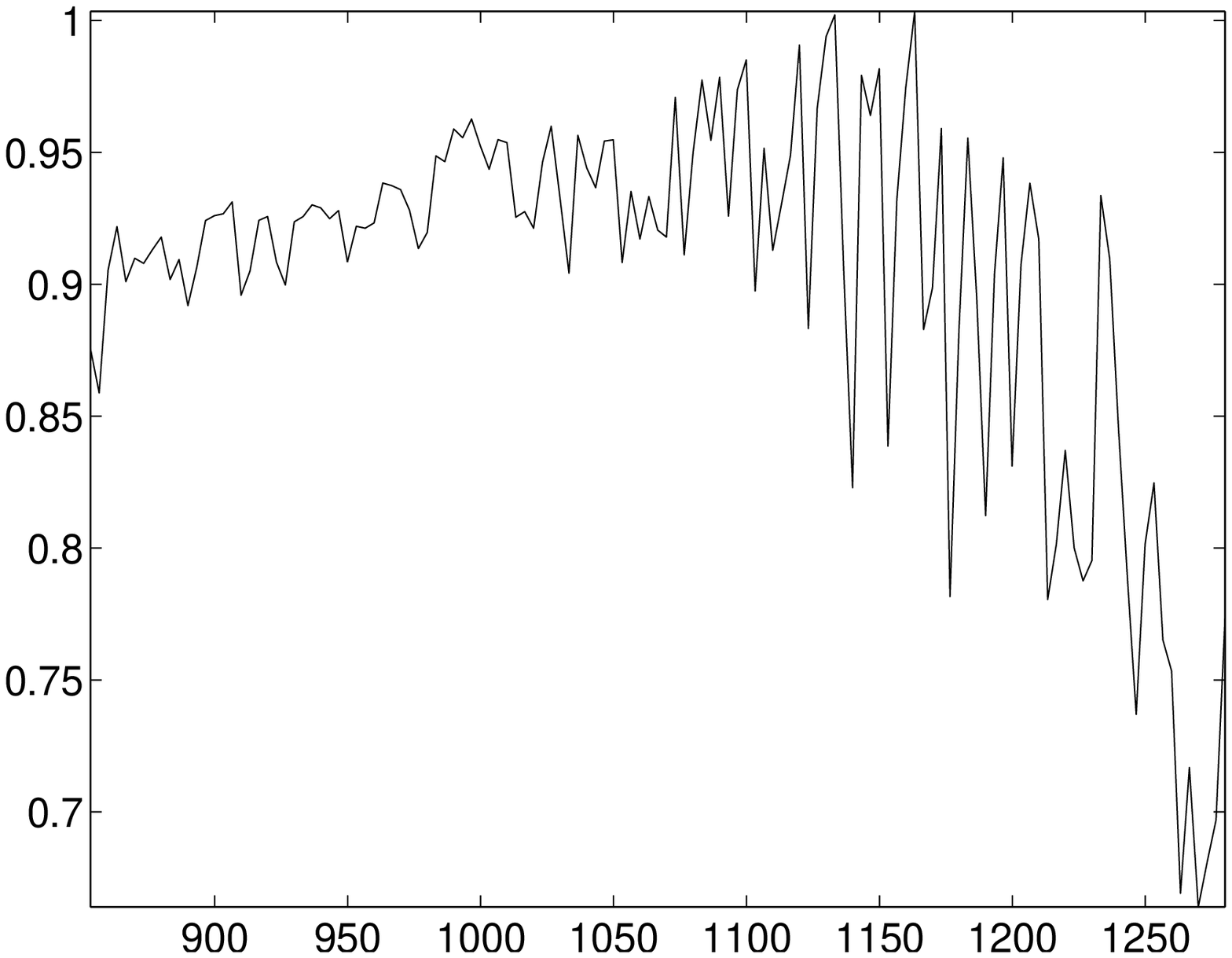}&
	\includegraphics[width=.23\textwidth,height=.1\textheight]{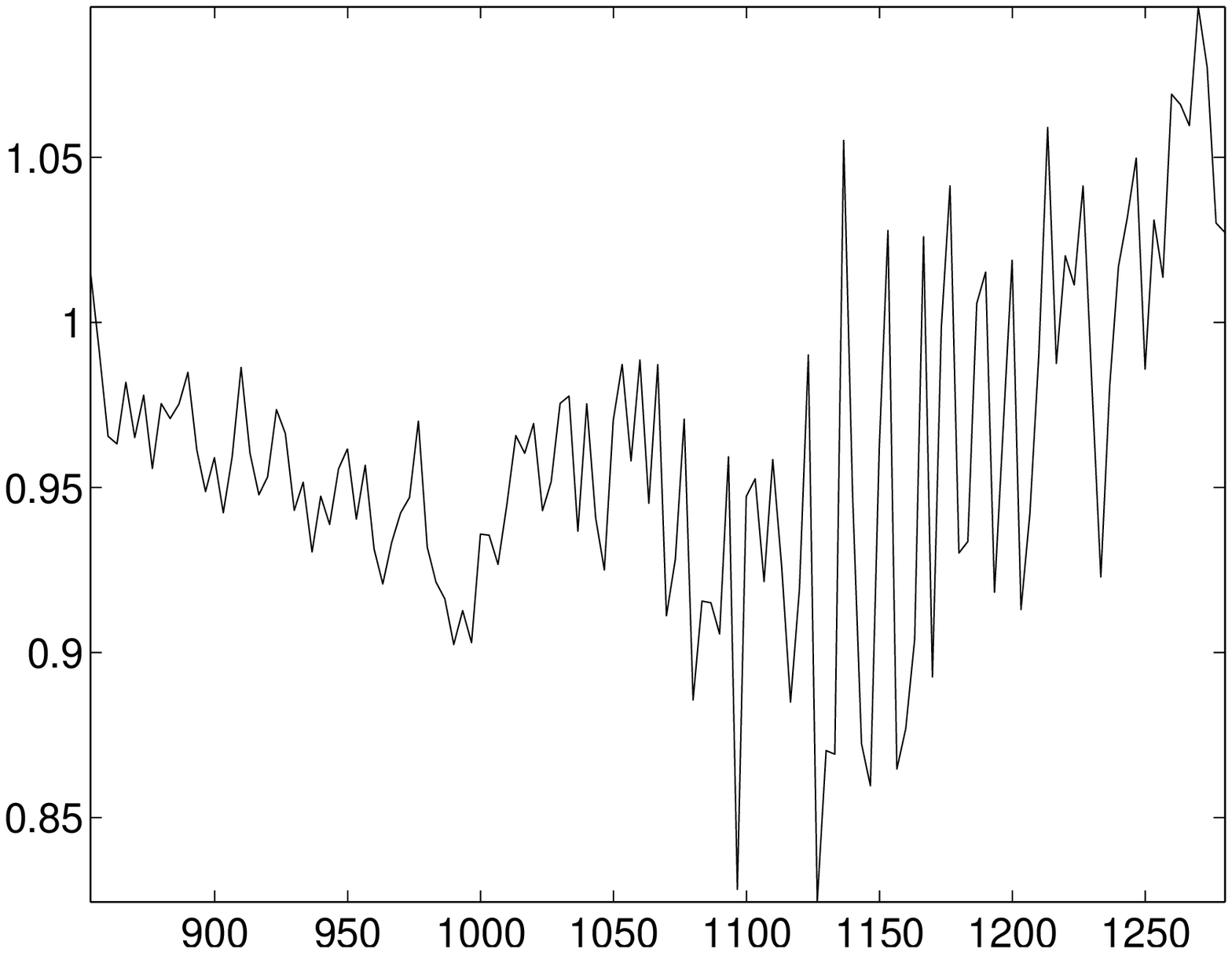}&
	\includegraphics[width=.23\textwidth,height=.1\textheight]{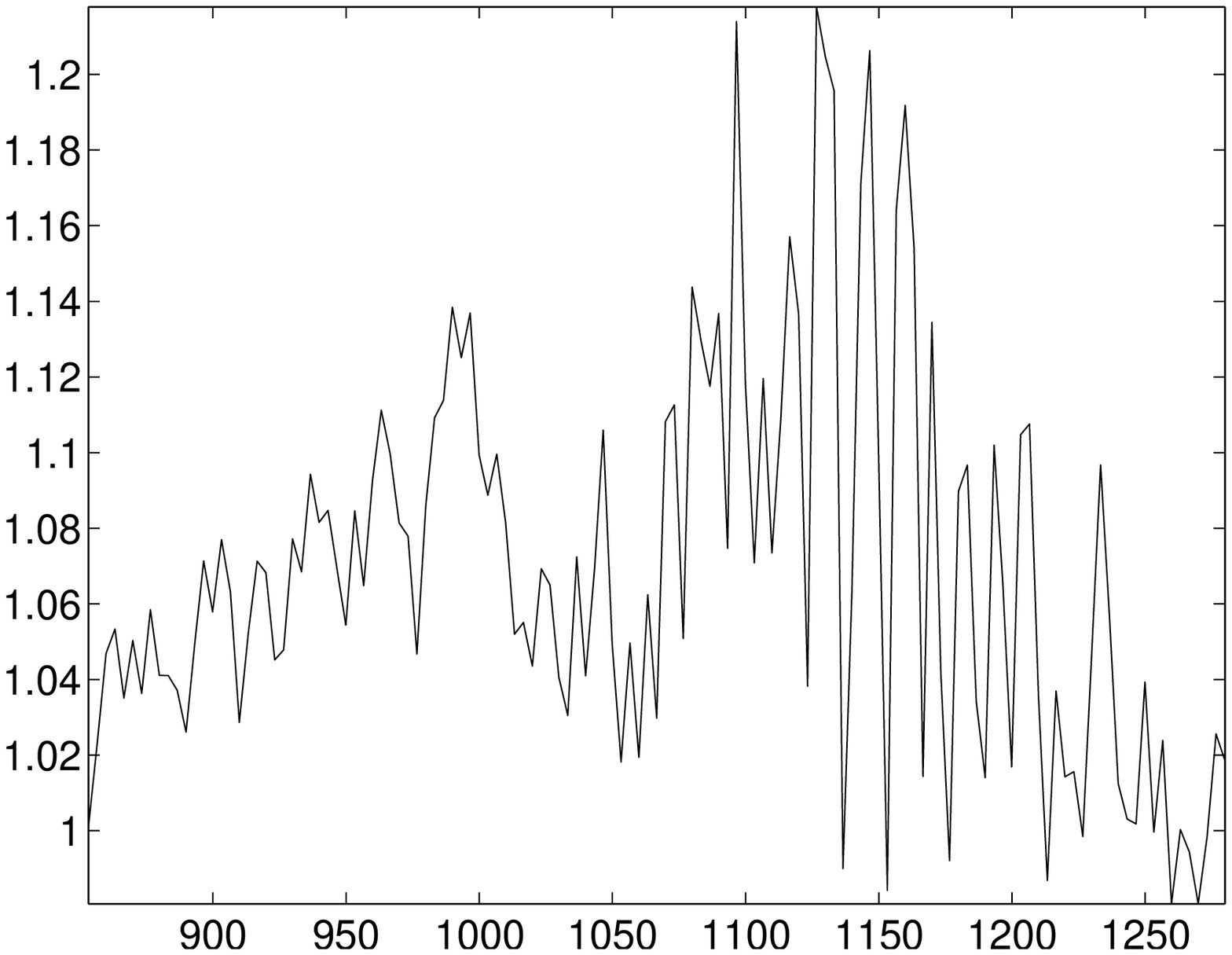}&
	\includegraphics[width=.23\textwidth,height=.1\textheight]{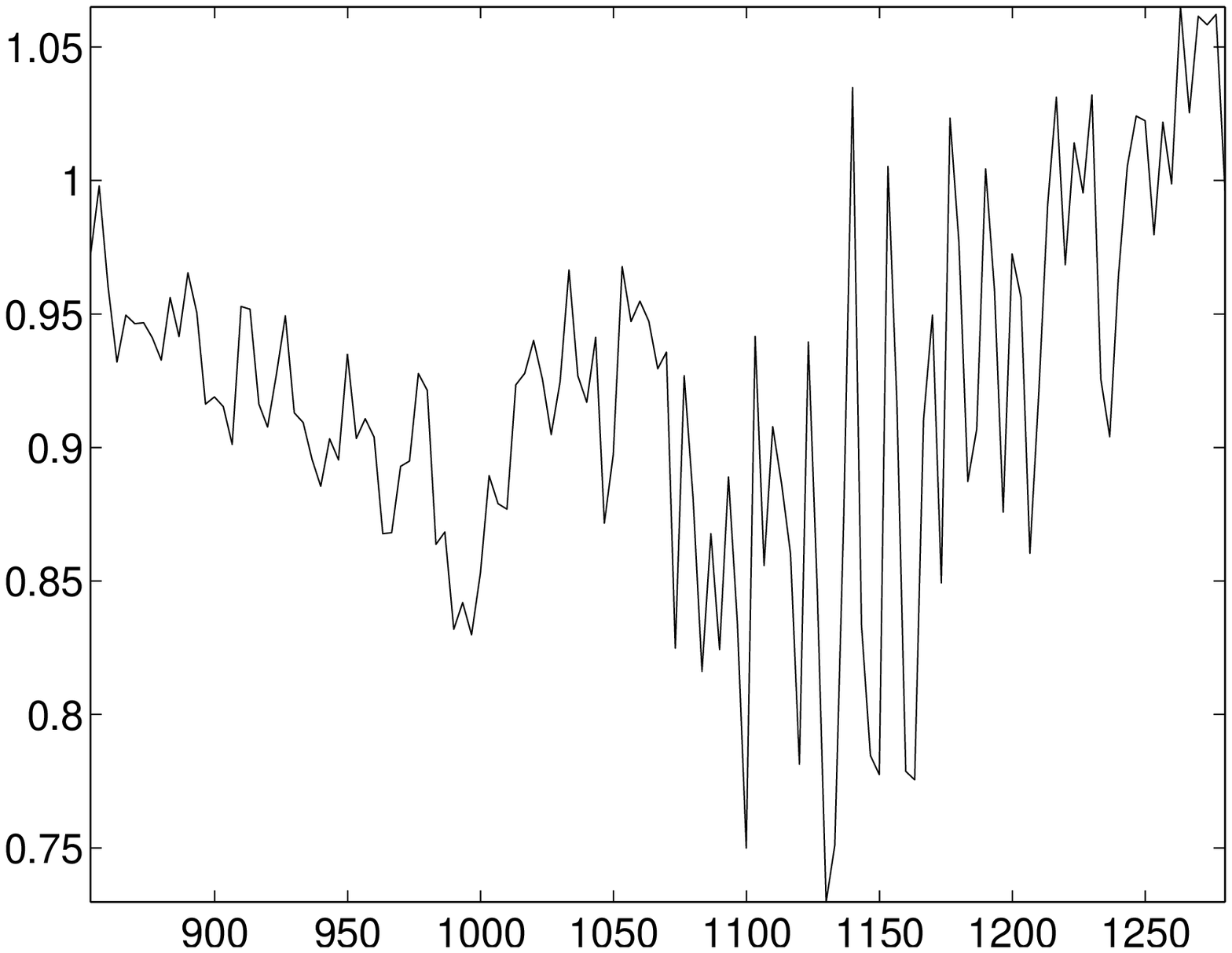}\\
		\multicolumn{4}{c}{Time frame 3}\\
	\includegraphics[width=.23\textwidth,height=.1\textheight]{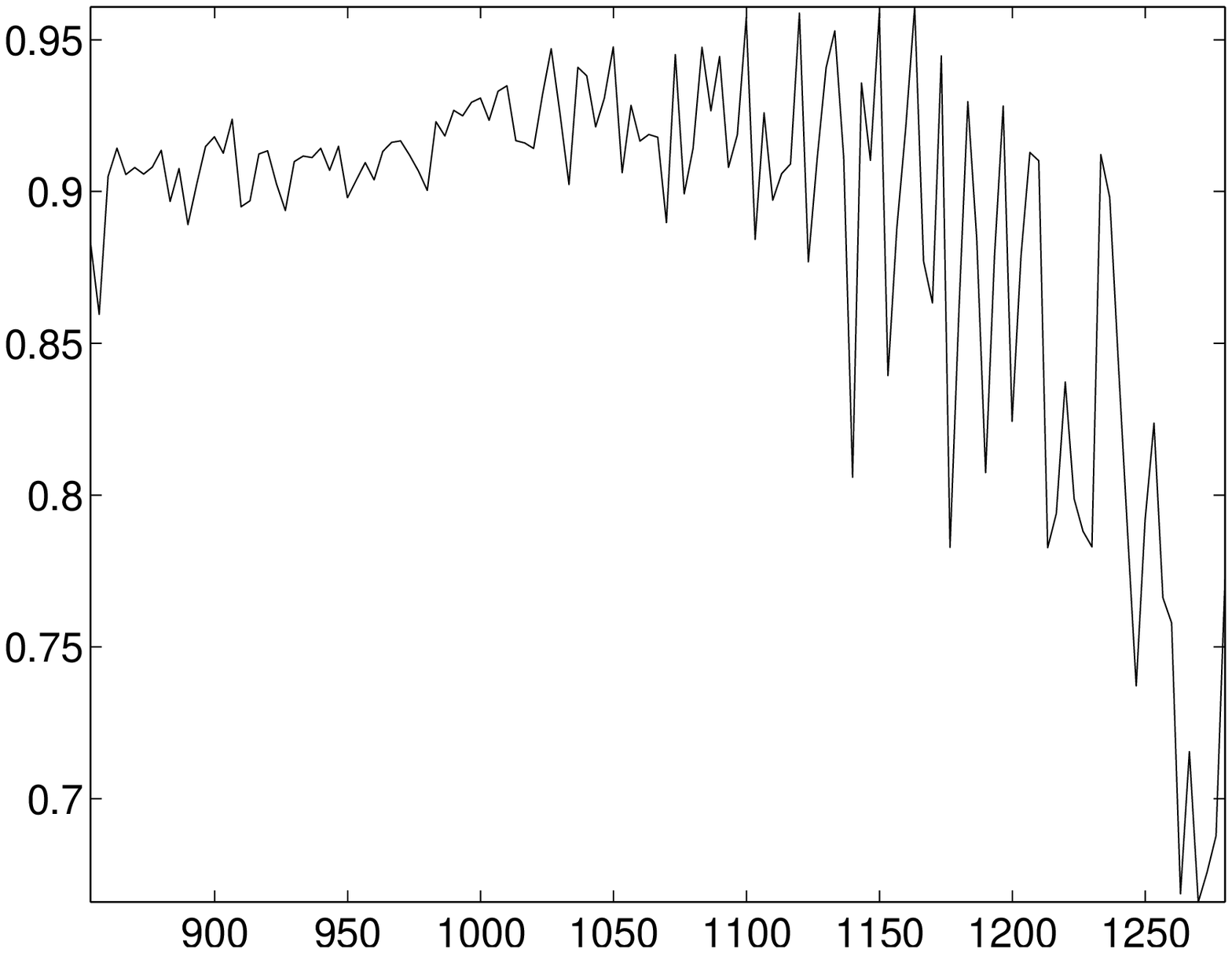}&
	\includegraphics[width=.23\textwidth,height=.1\textheight]{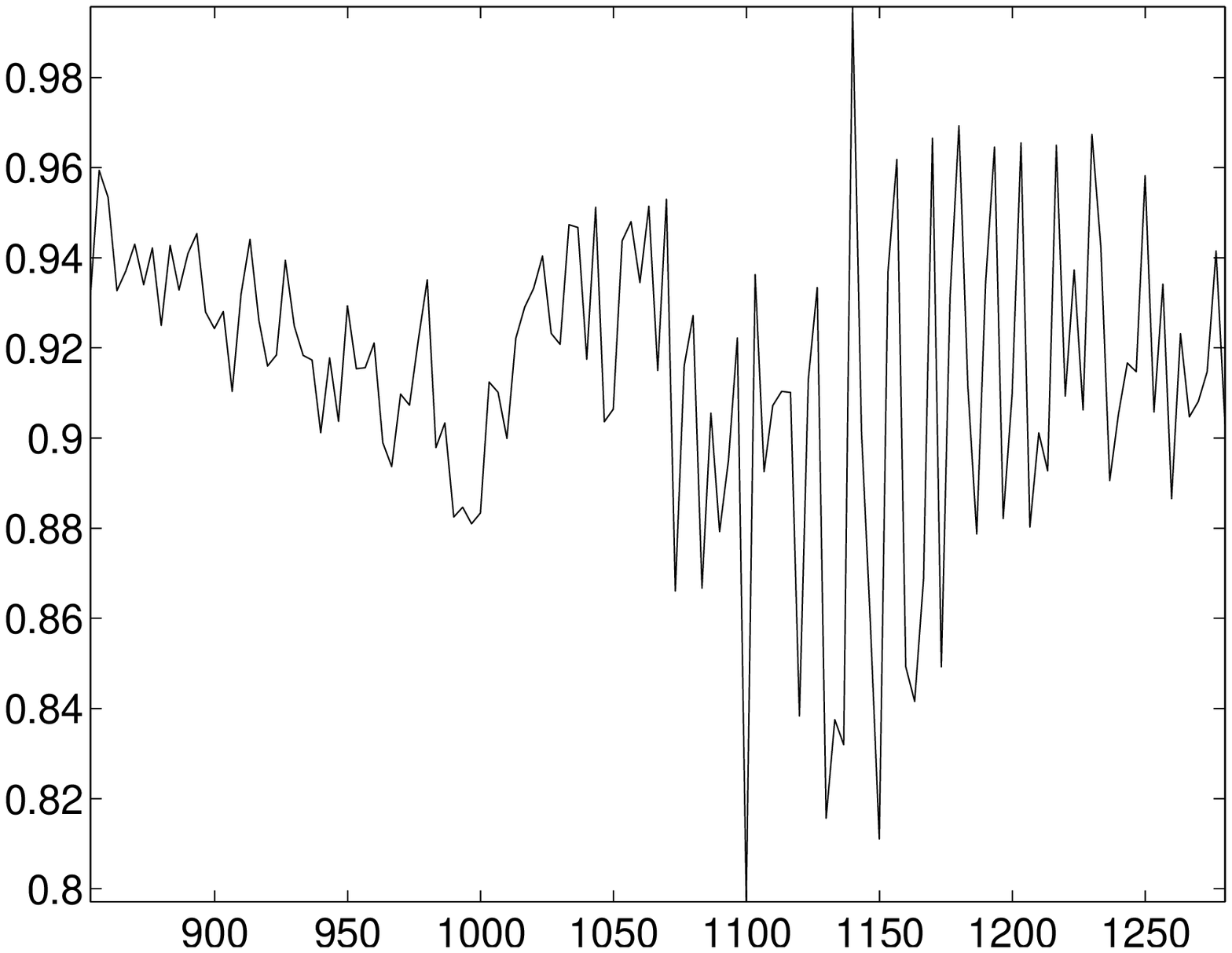}&
	\includegraphics[width=.23\textwidth,height=.1\textheight]{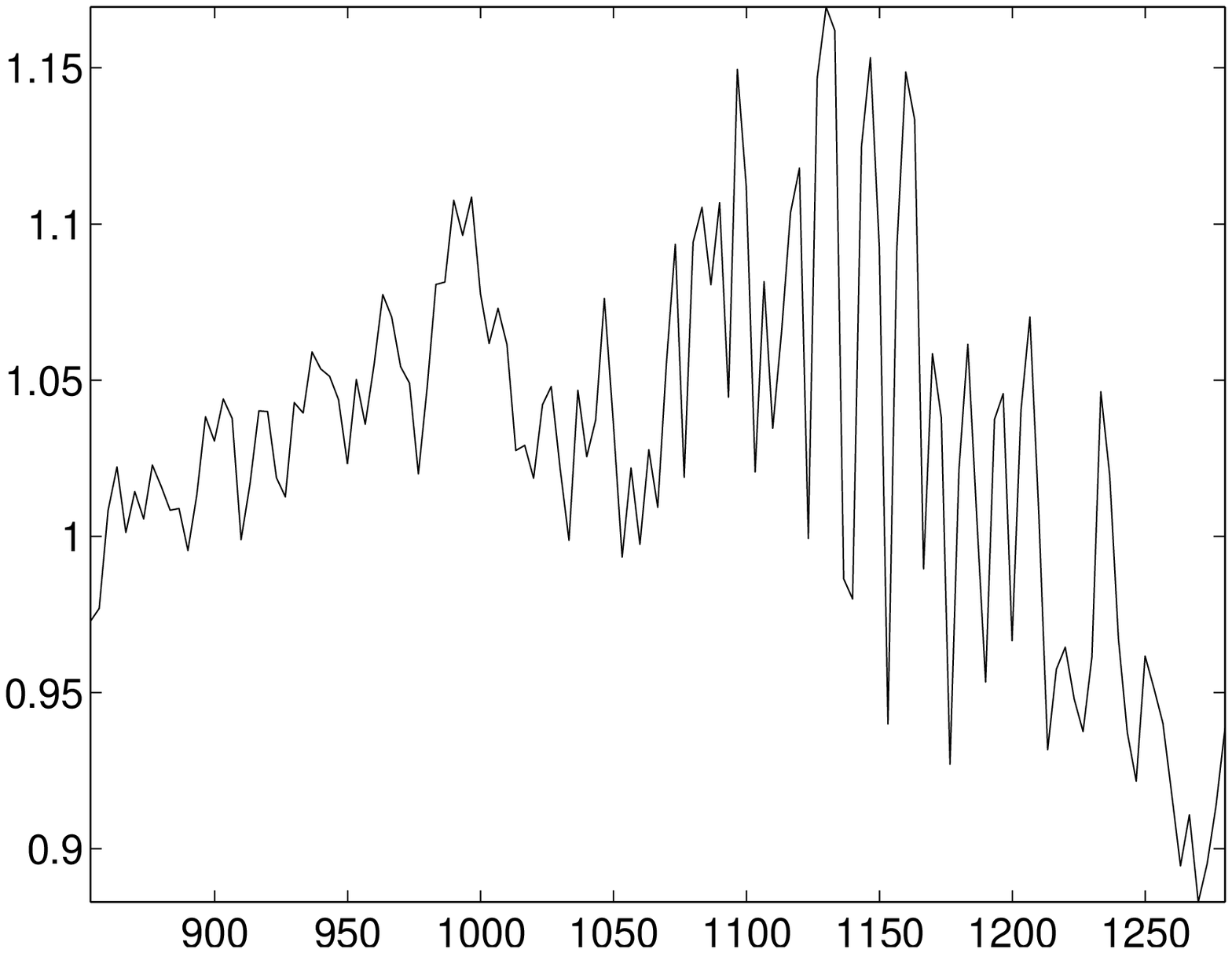}&
	\includegraphics[width=.23\textwidth,height=.1\textheight]{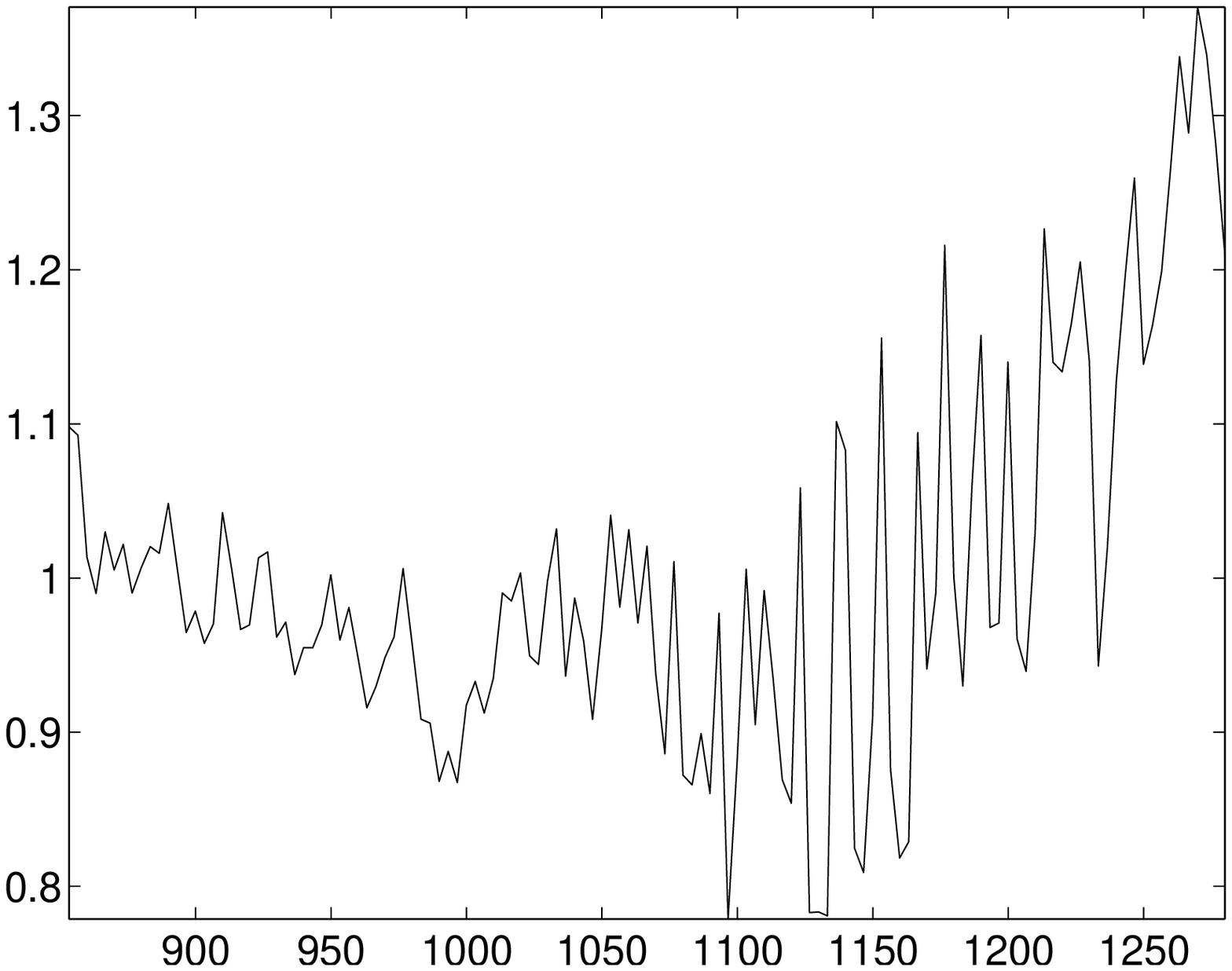}\\
		\multicolumn{4}{c}{Time frame 7}\\
	\includegraphics[width=.23\textwidth,height=.1\textheight]{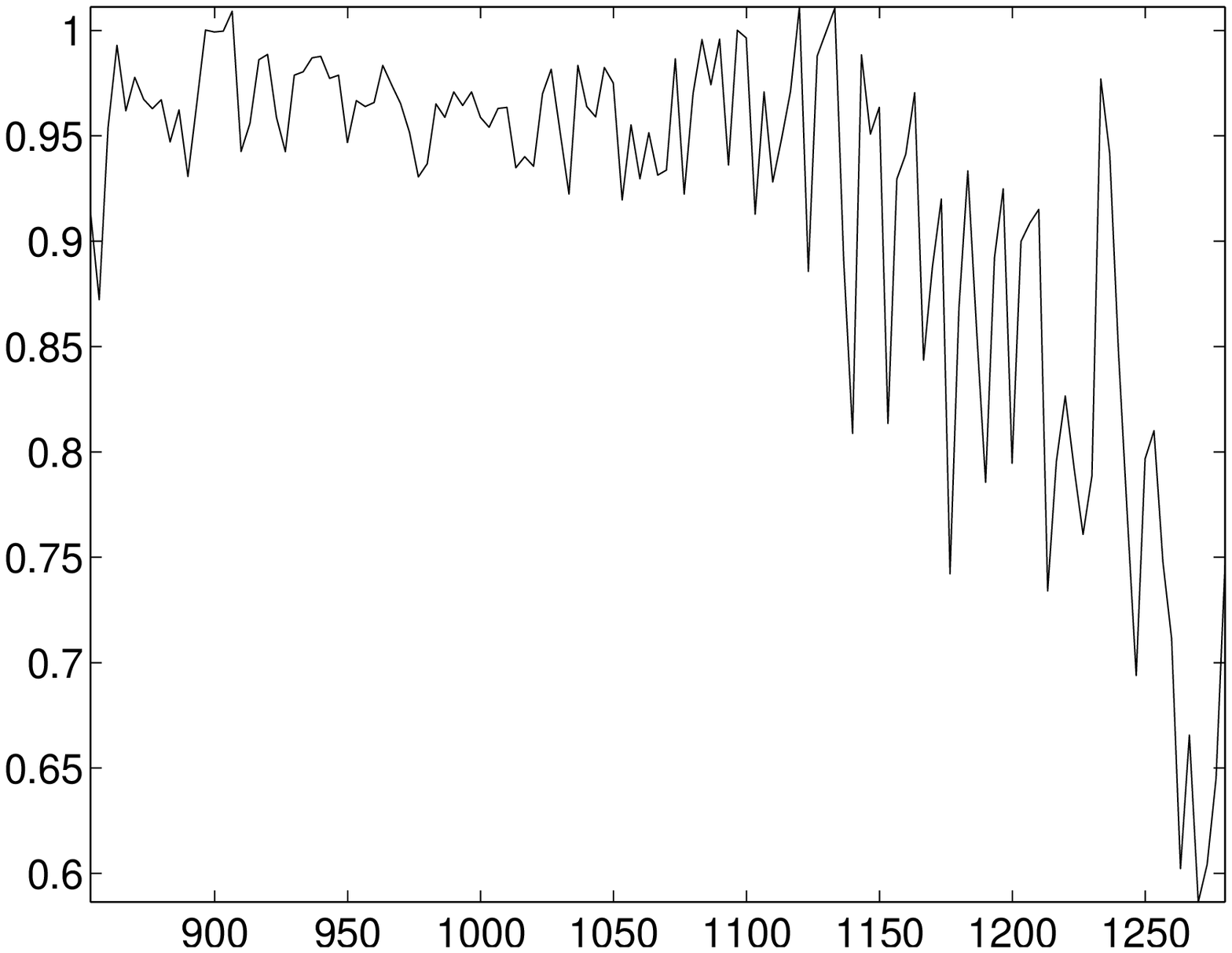}&
	\includegraphics[width=.23\textwidth,height=.1\textheight]{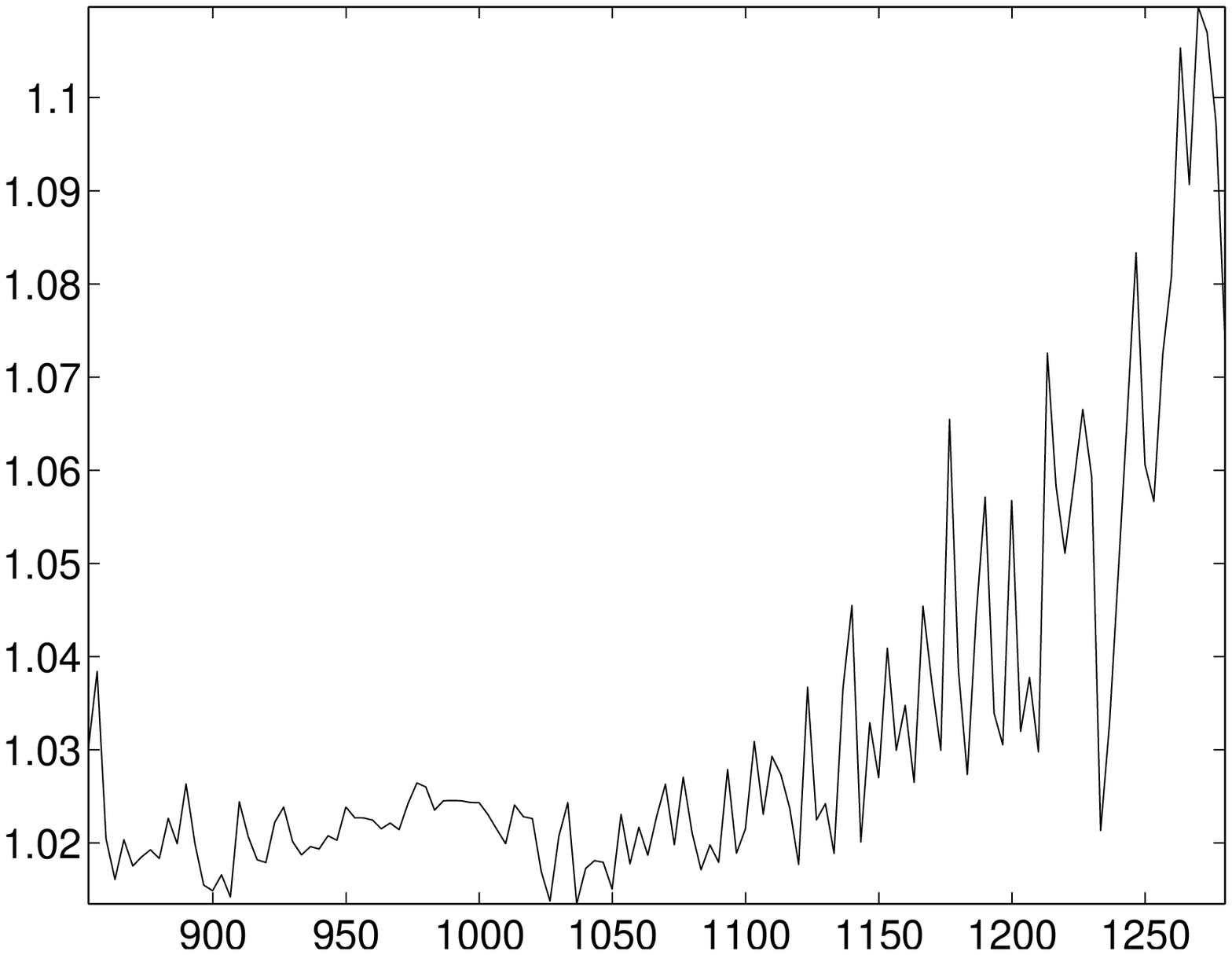}&
	\includegraphics[width=.23\textwidth,height=.1\textheight]{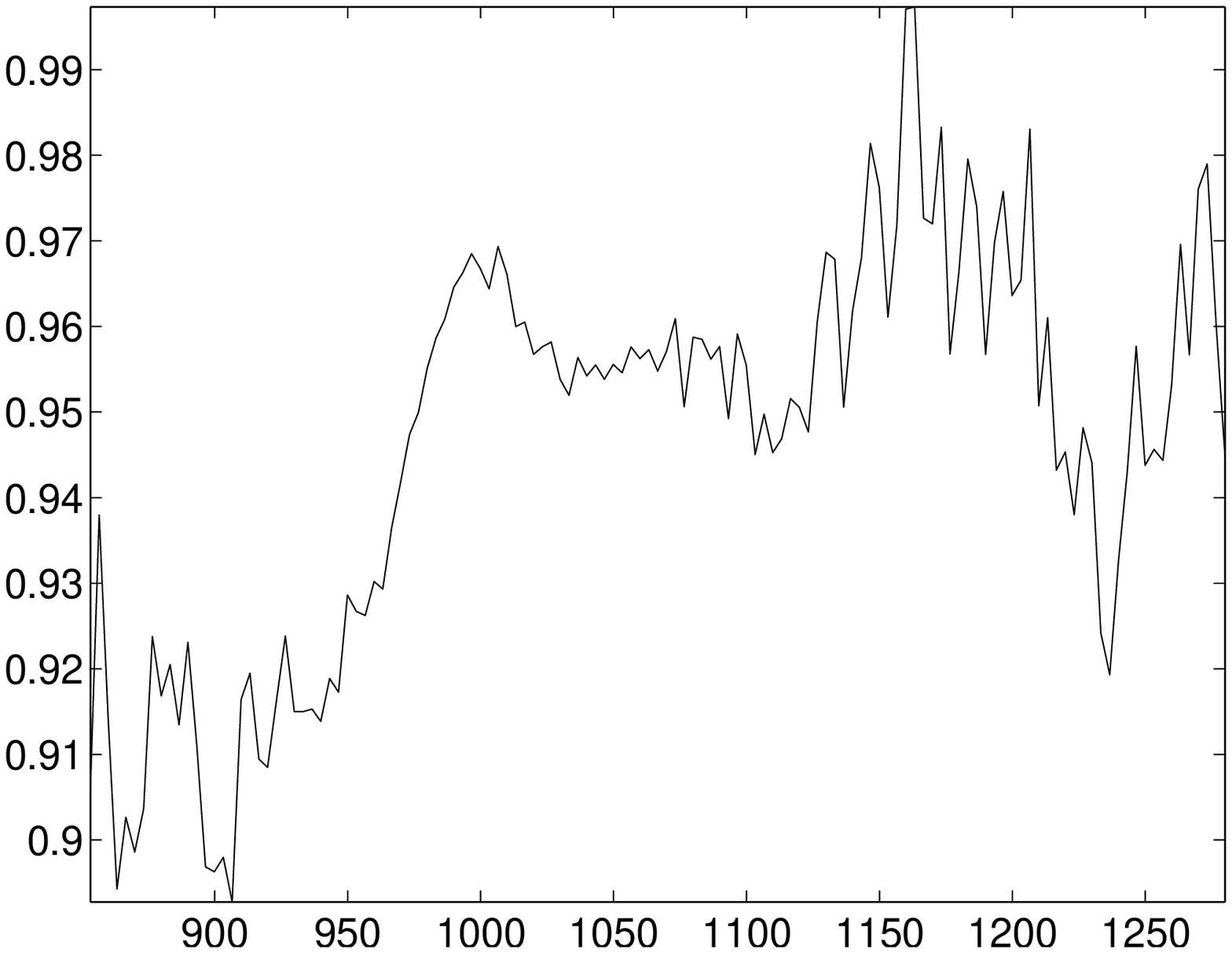}&
	\includegraphics[width=.23\textwidth,height=.1\textheight]{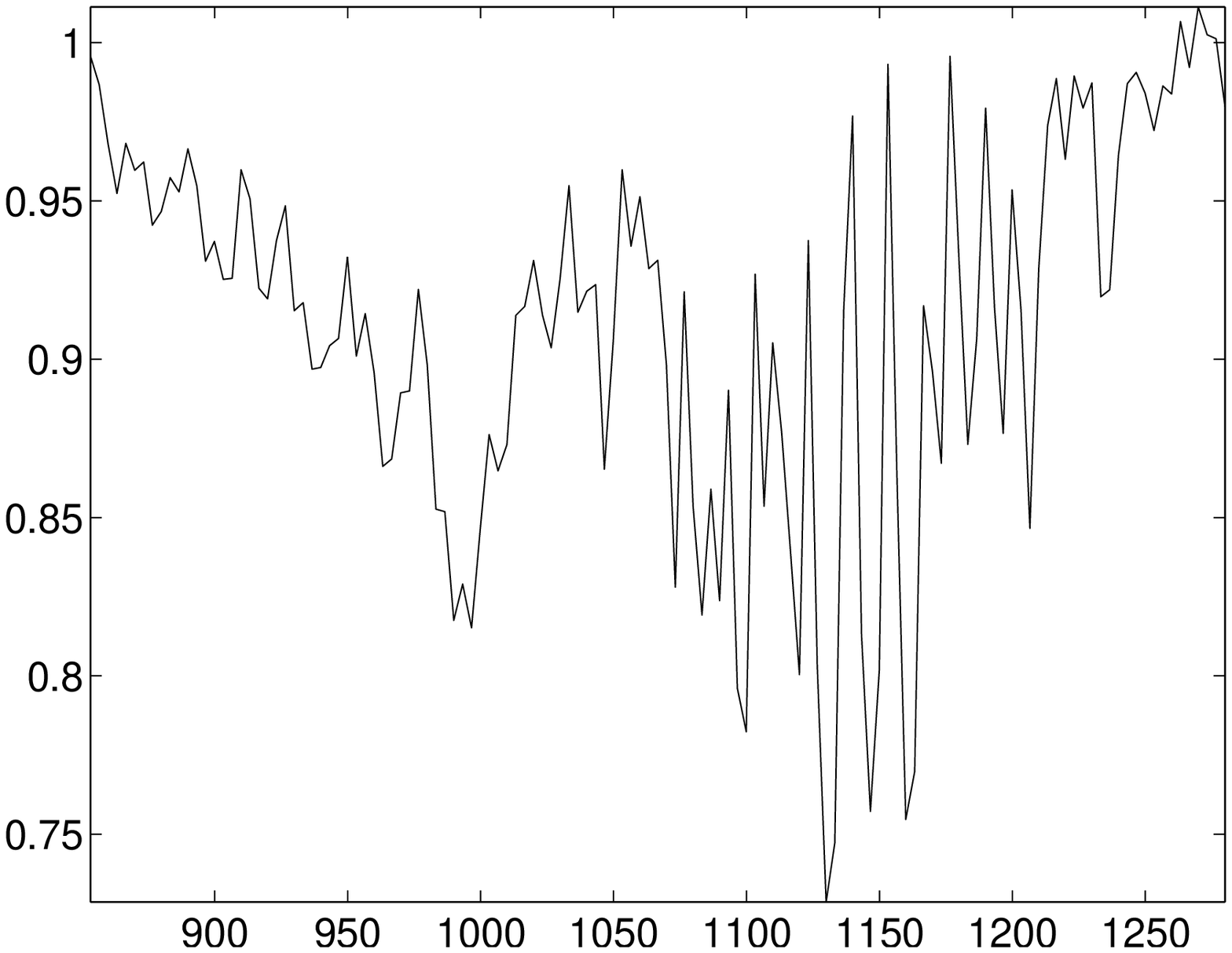}\\
		\multicolumn{4}{c}{Time frame 10}\\
	\includegraphics[width=.23\textwidth,height=.1\textheight]{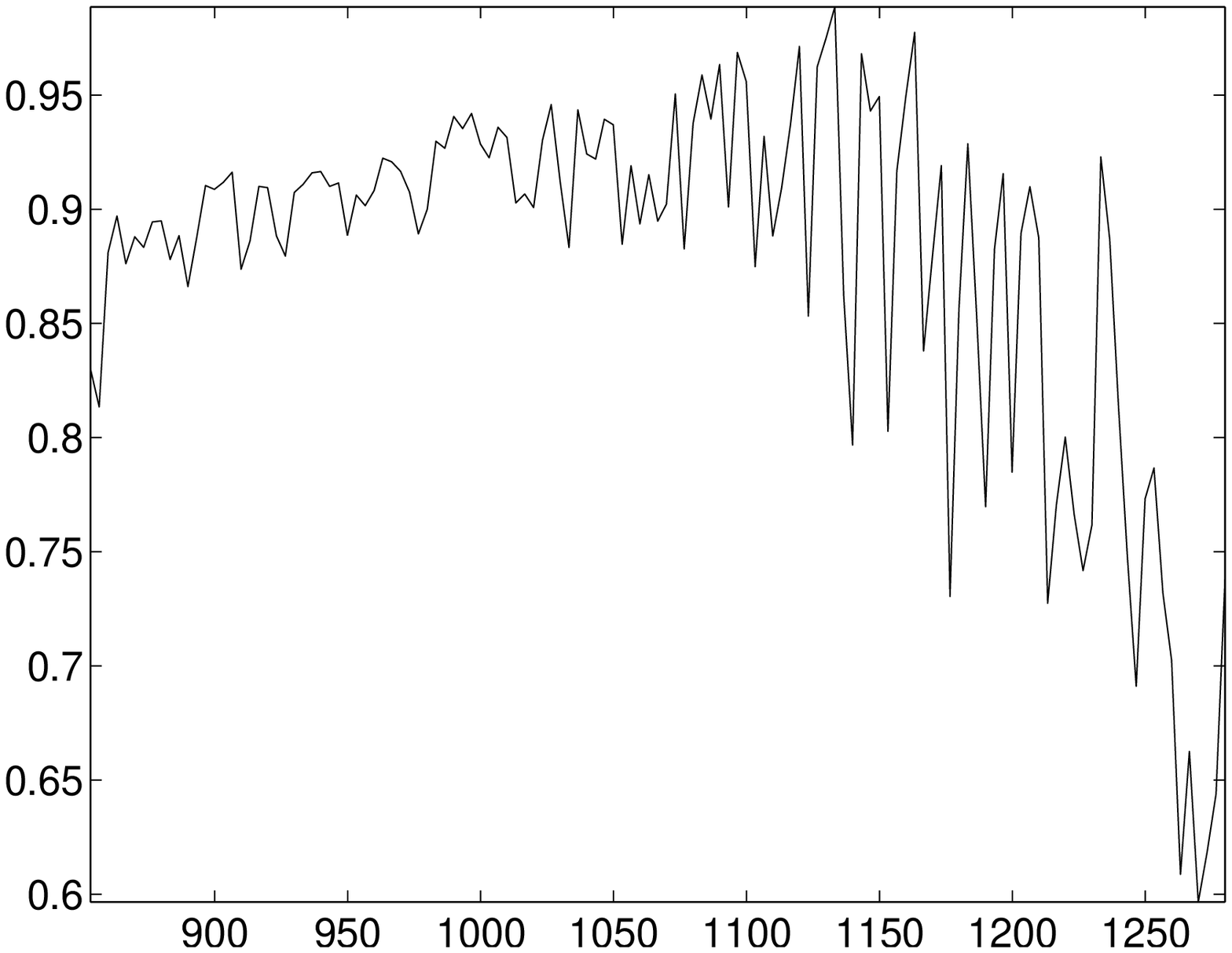}&
	\includegraphics[width=.23\textwidth,height=.1\textheight]{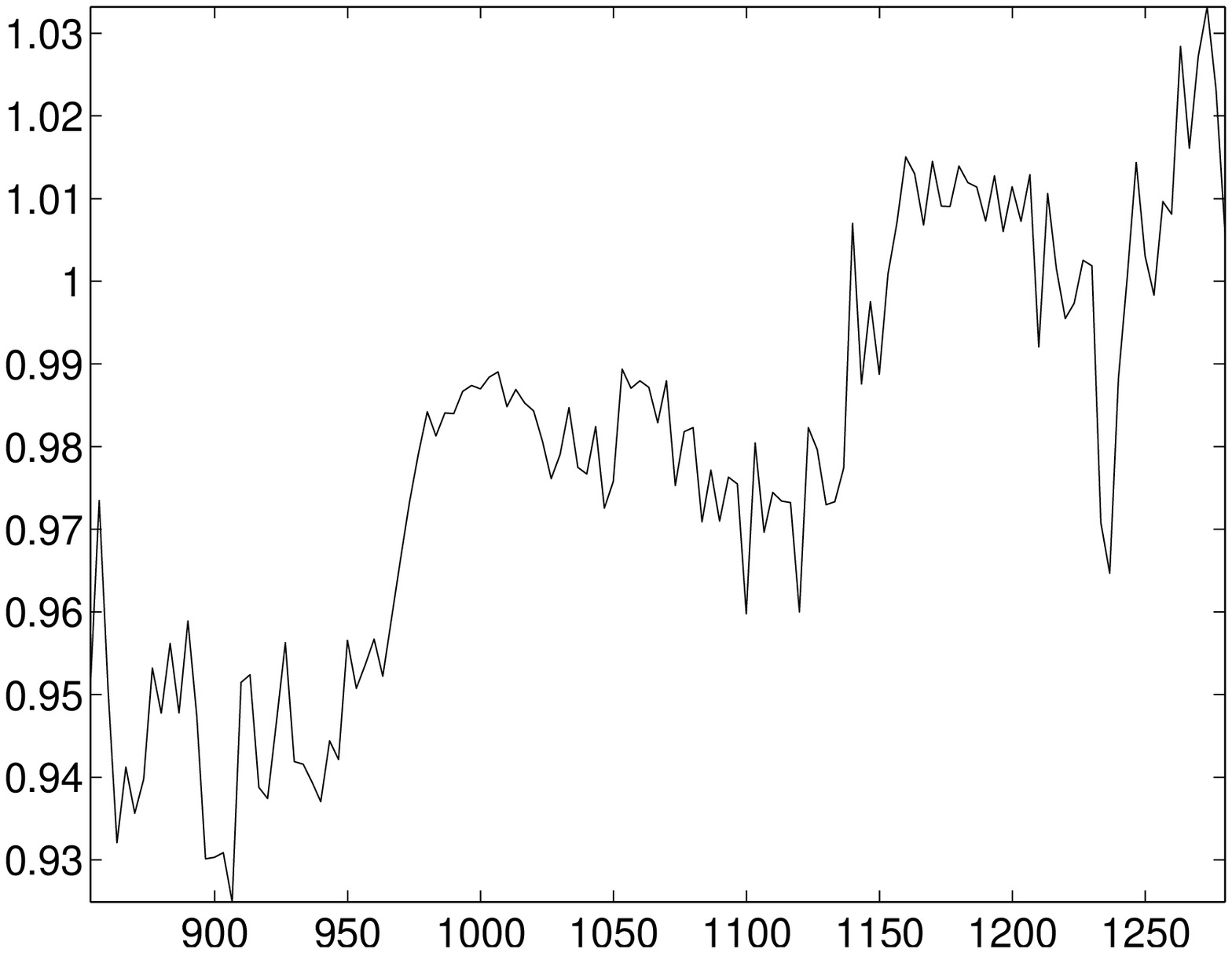}&
	\includegraphics[width=.23\textwidth,height=.1\textheight]{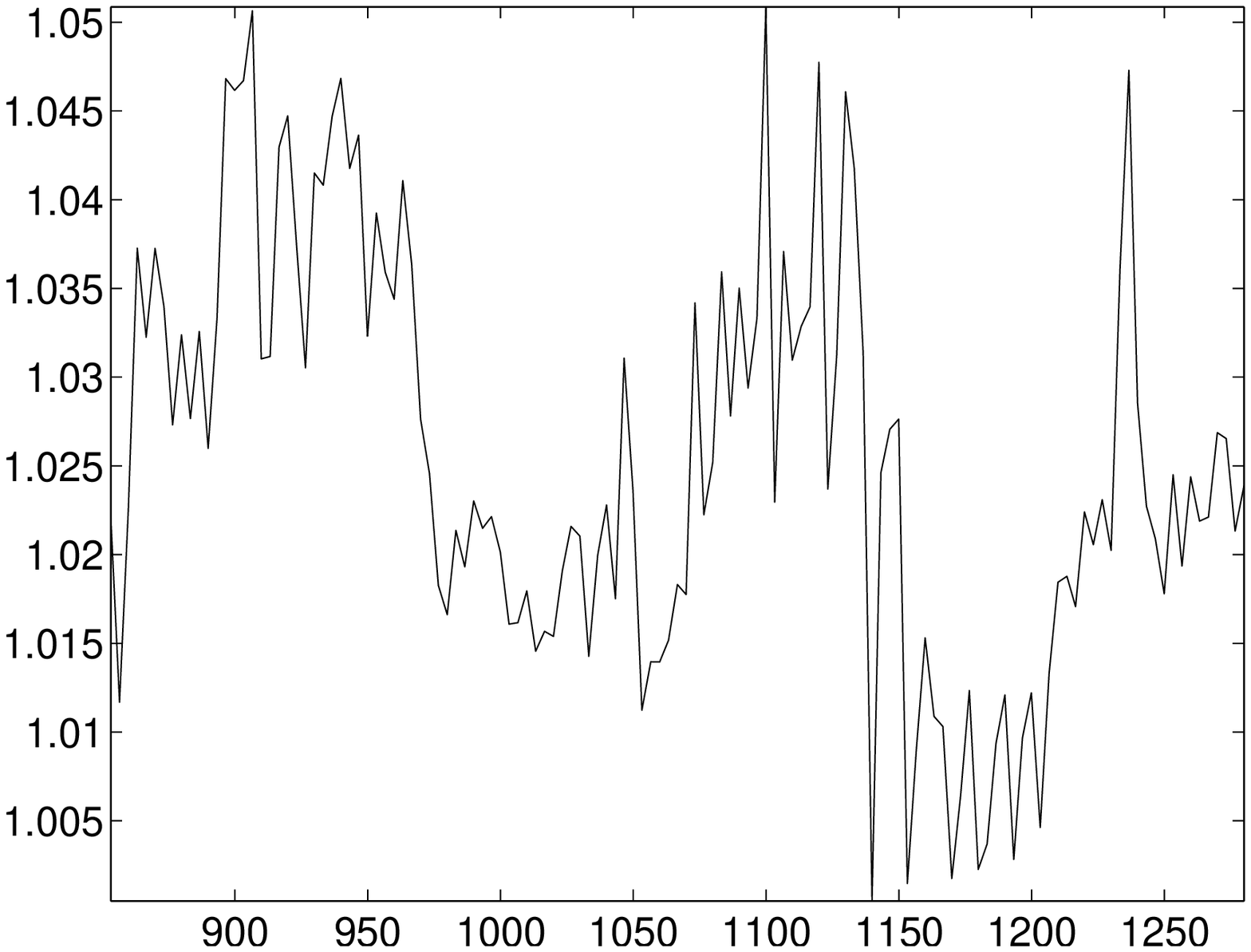}&
	\includegraphics[width=.23\textwidth,height=.1\textheight]{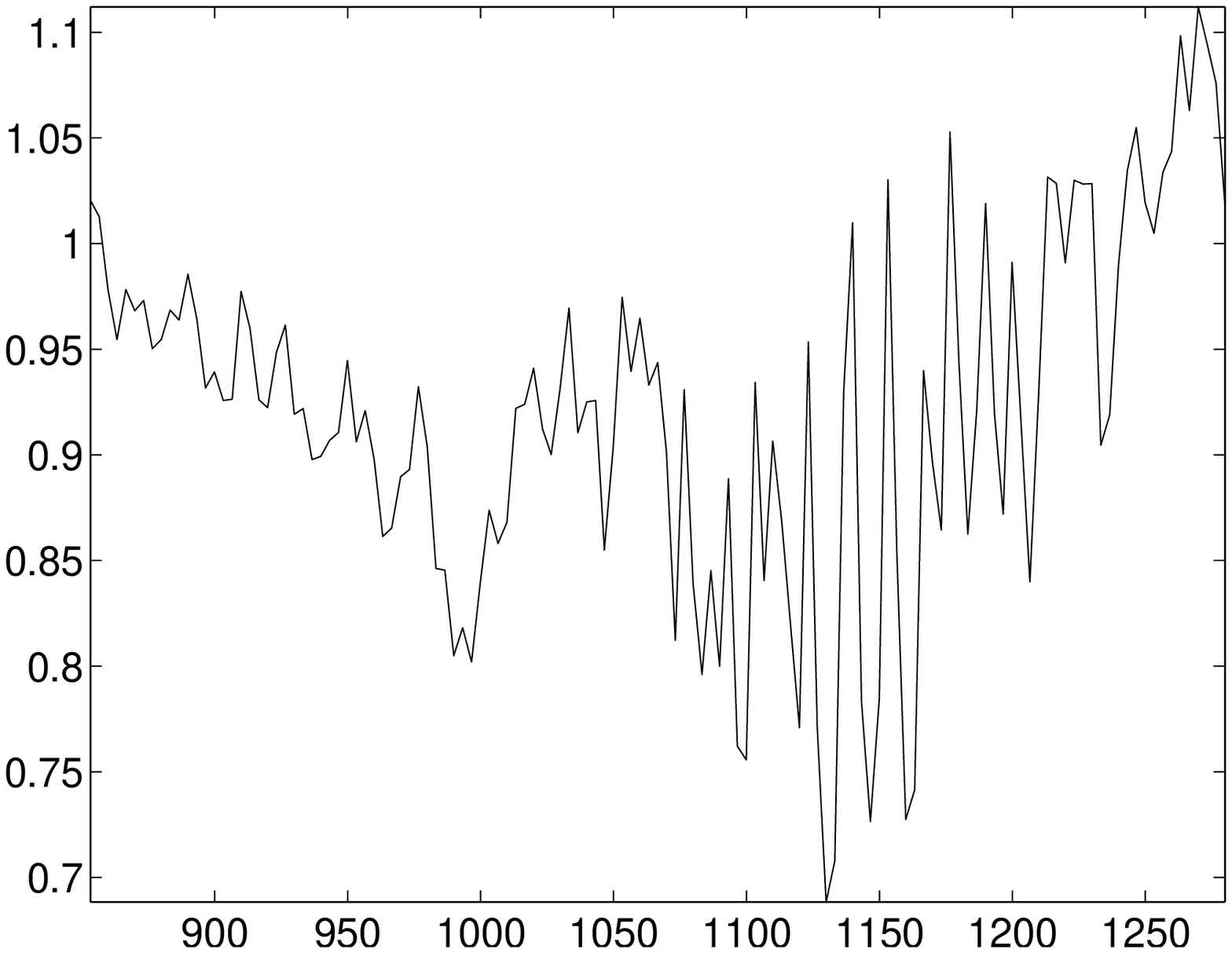}\\
	\end{tabular}
	\caption{\textcolor{black}{Endmembers obtained by the separate unmixing method : estimated spectra from the first, third, seventh and tenth time frame. Each row corresponds to a time frame and each column to a particular source. The larger the contribution of the 'ghost' to a specific abundance map, the larger the distortion of the corresponding extracted endmember: see \emph{e.g.} the seventh and tenth time frames of source 2 and 3.}}
		\label{fig:end_sepa}
\end{figure*}

\begin{figure*}[ht]
	\begin{tabular}{cccc}
	Source 1 & Source 2 & Source 3 & Source 4\\
		\multicolumn{4}{c}{Time frame 1}\\
	\includegraphics[width=.23\textwidth,height=.1\textheight]{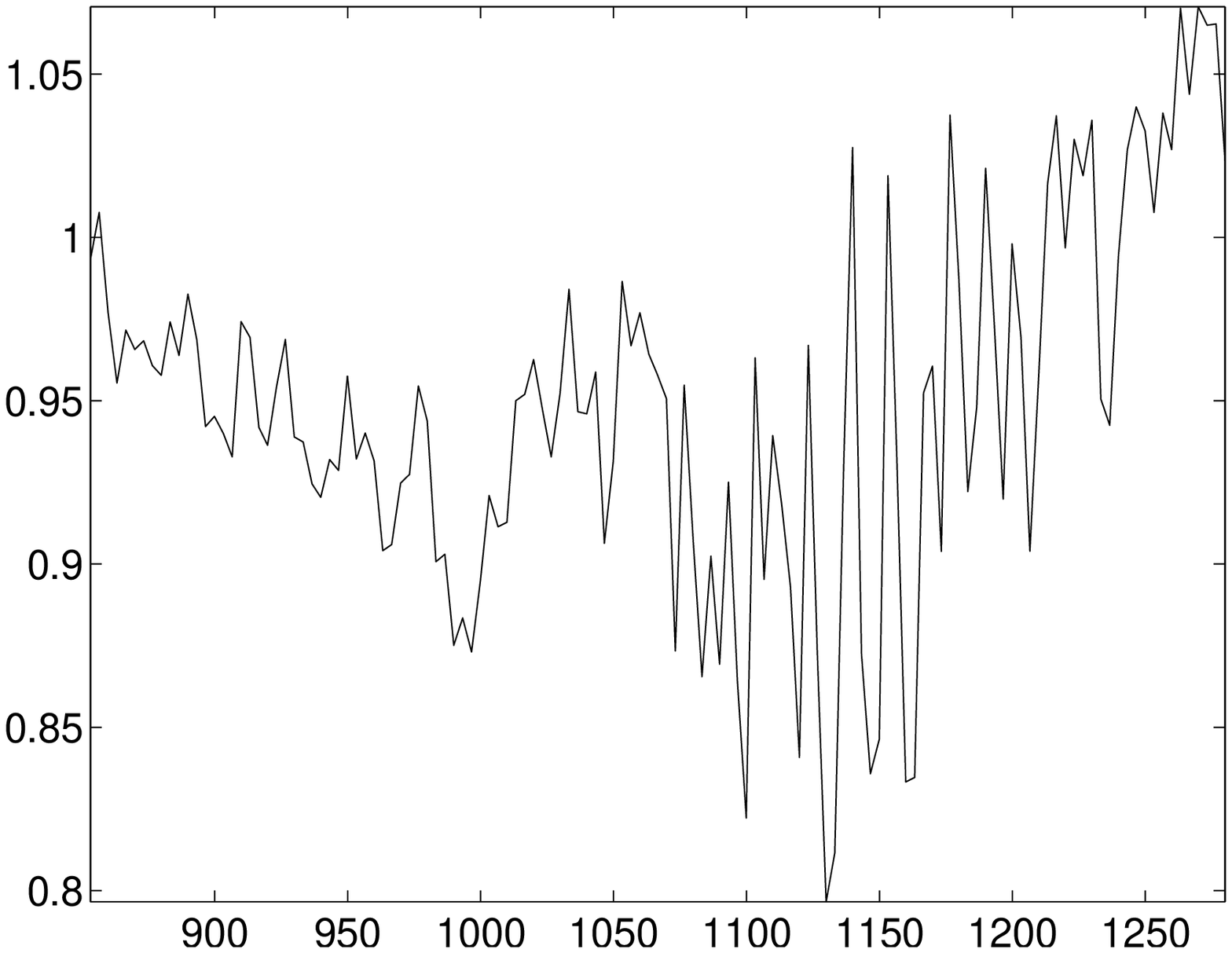}&
	\includegraphics[width=.23\textwidth,height=.1\textheight]{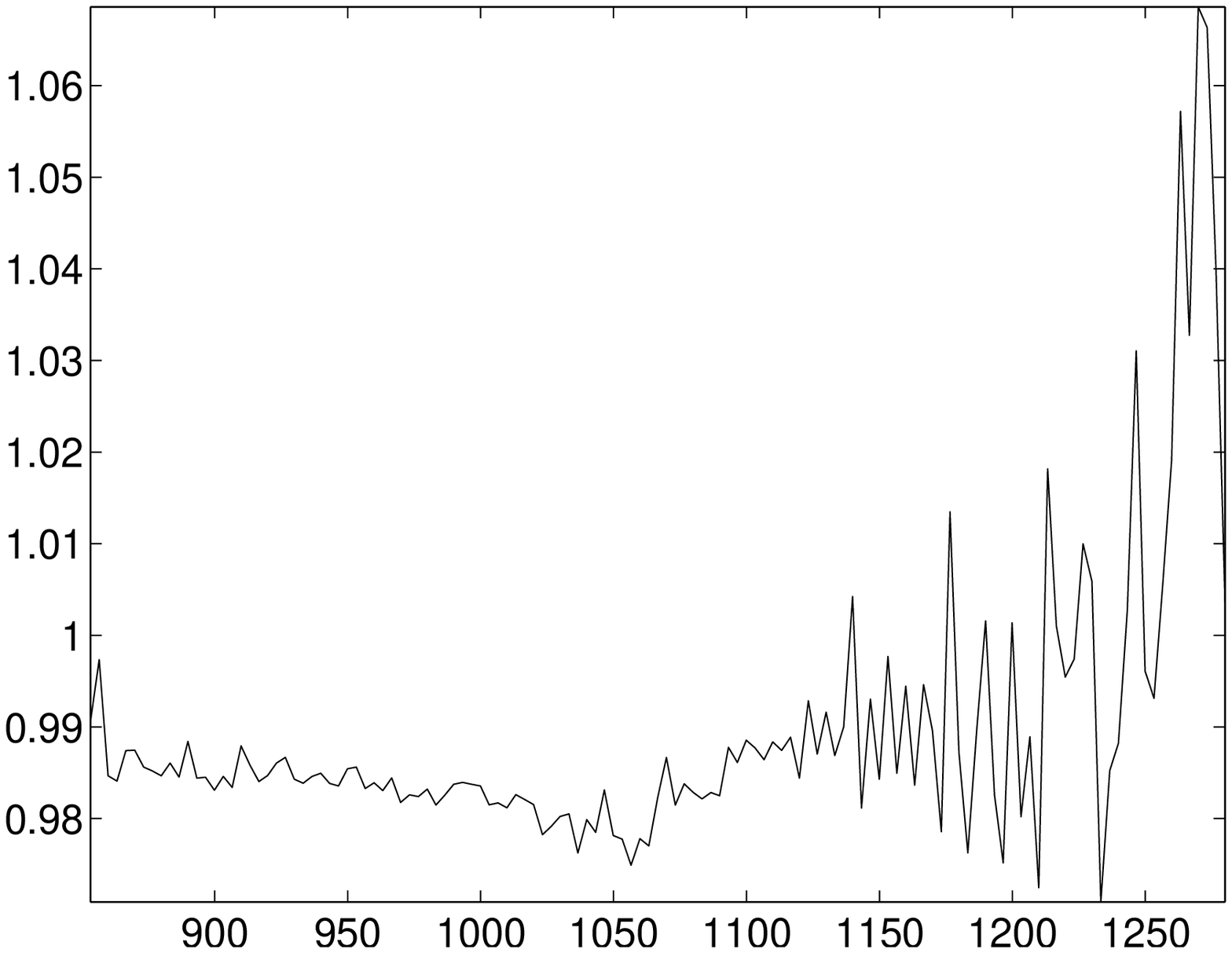}&
	\includegraphics[width=.23\textwidth,height=.1\textheight]{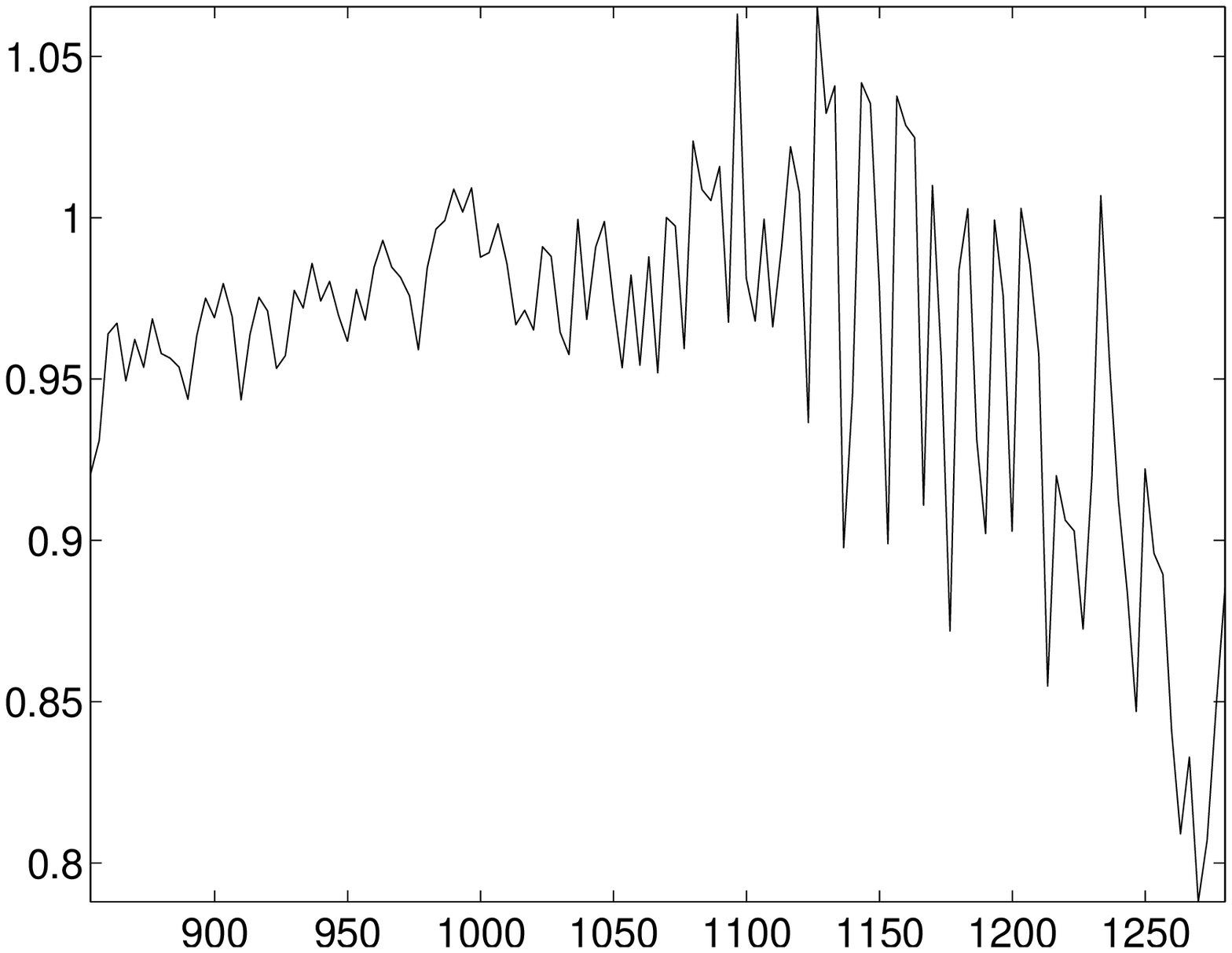}&
	\includegraphics[width=.23\textwidth,height=.1\textheight]{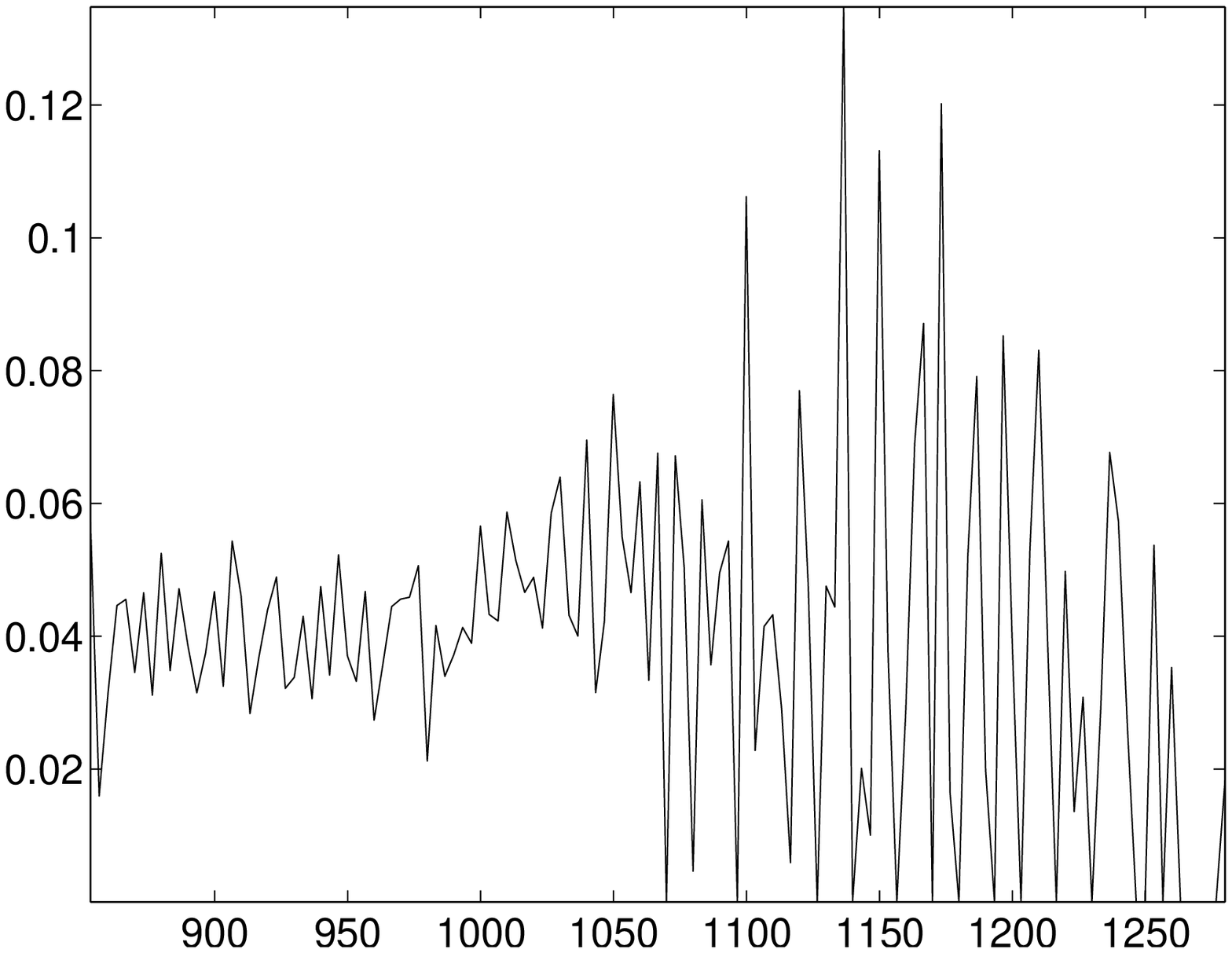}\\
		\multicolumn{4}{c}{Time frame 3}\\
	\includegraphics[width=.23\textwidth,height=.1\textheight]{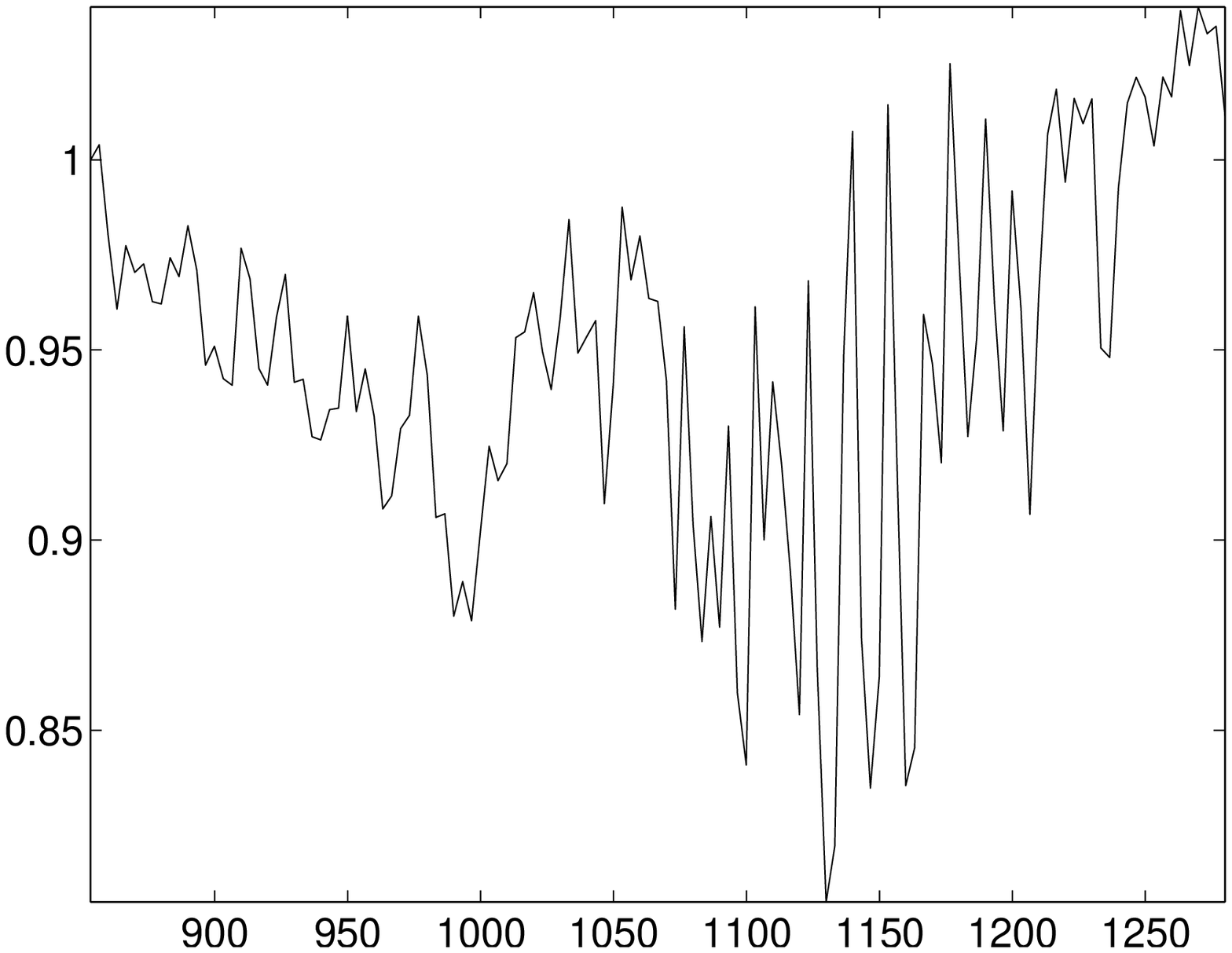}&
	\includegraphics[width=.23\textwidth,height=.1\textheight]{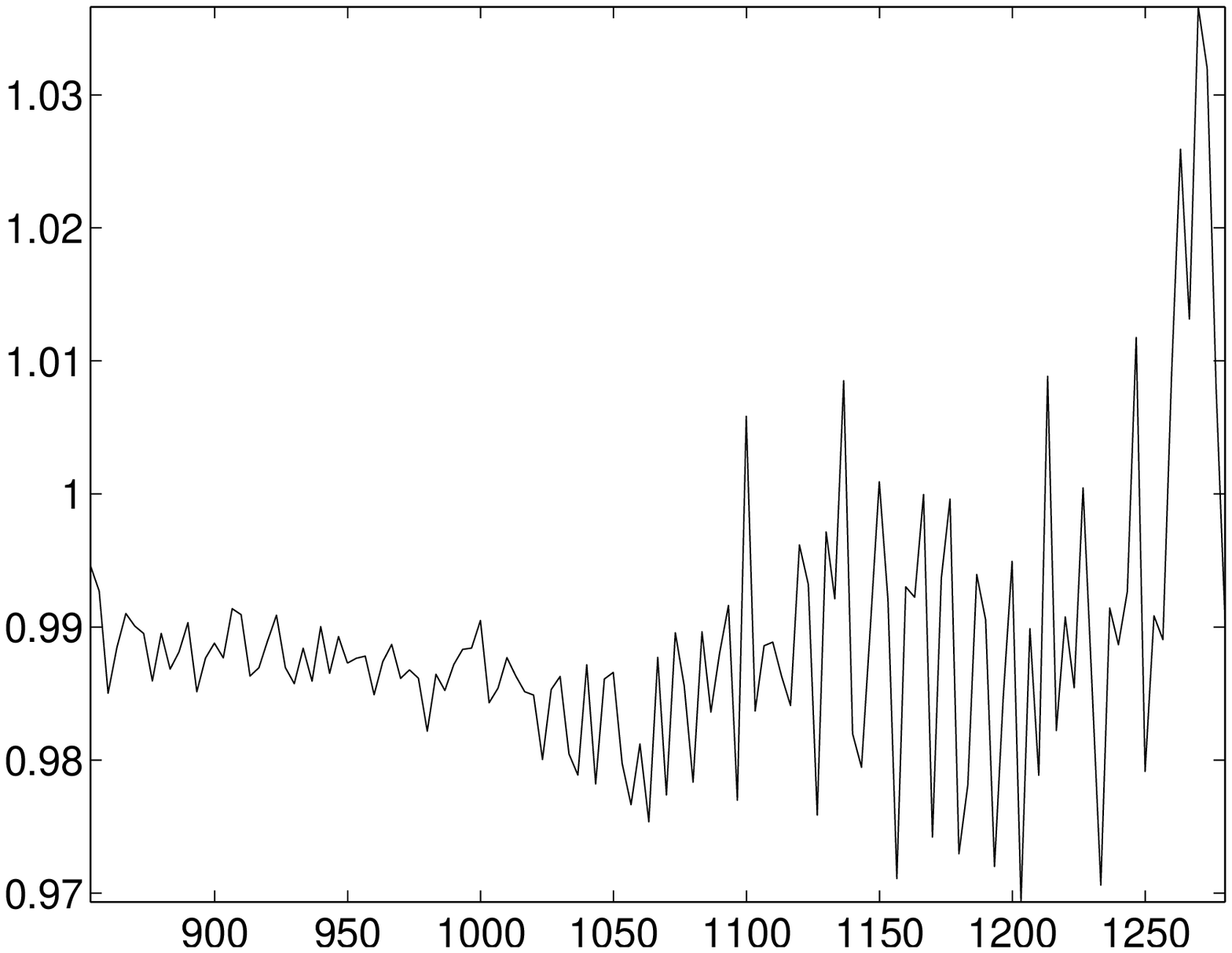}&
	\includegraphics[width=.23\textwidth,height=.1\textheight]{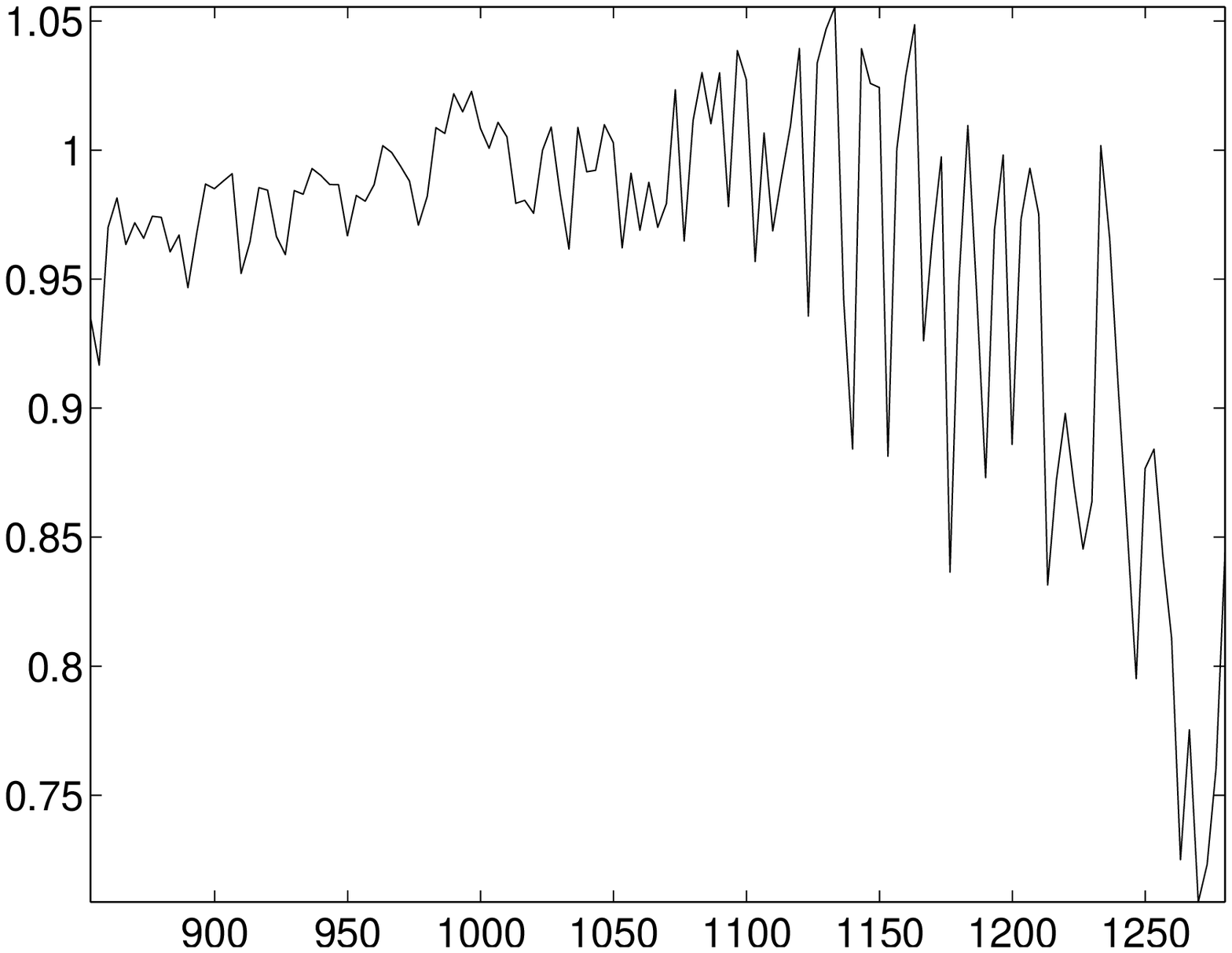}&
	\includegraphics[width=.23\textwidth,height=.1\textheight]{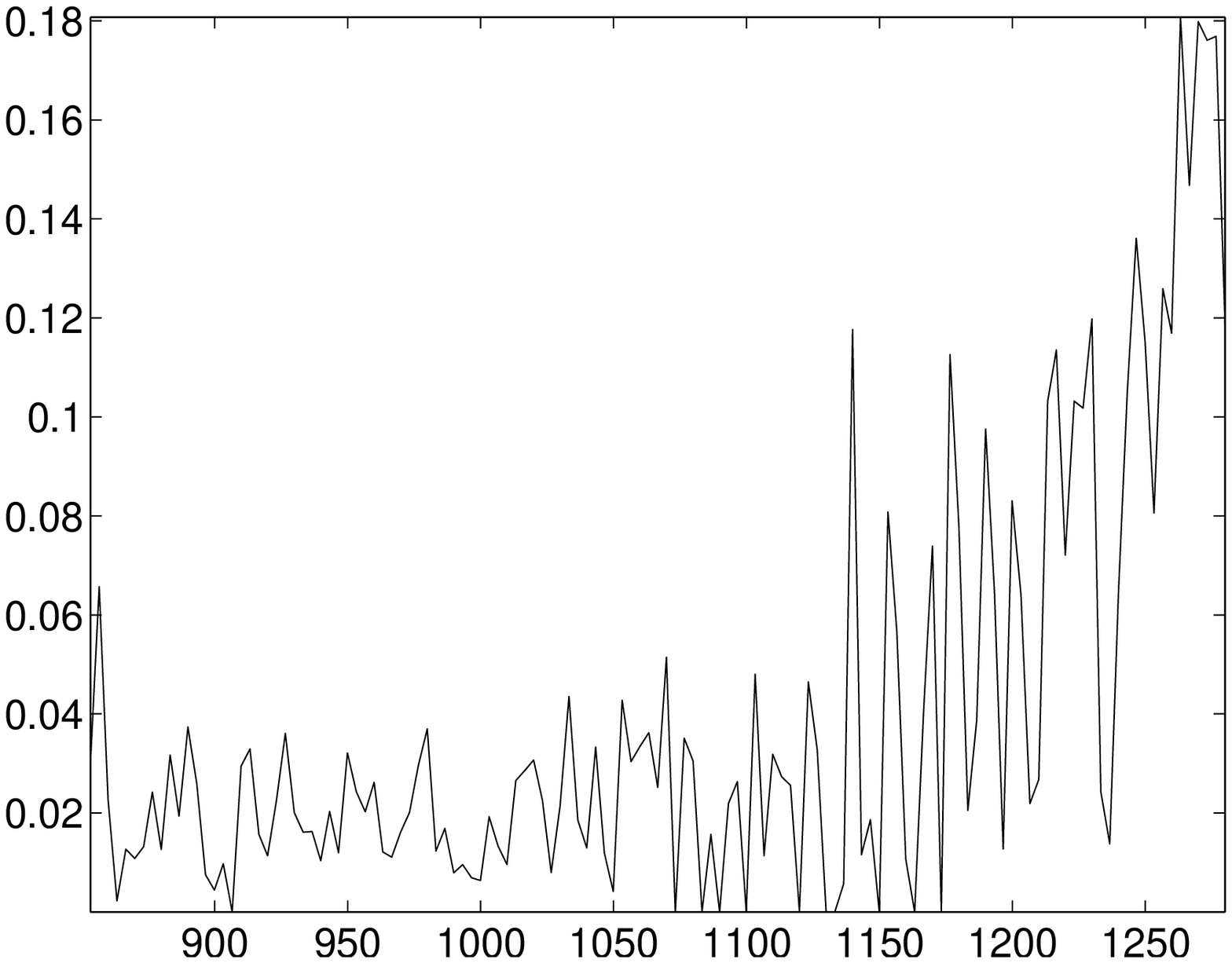}\\
		\multicolumn{4}{c}{Time frame 7}\\
	\includegraphics[width=.23\textwidth,height=.1\textheight]{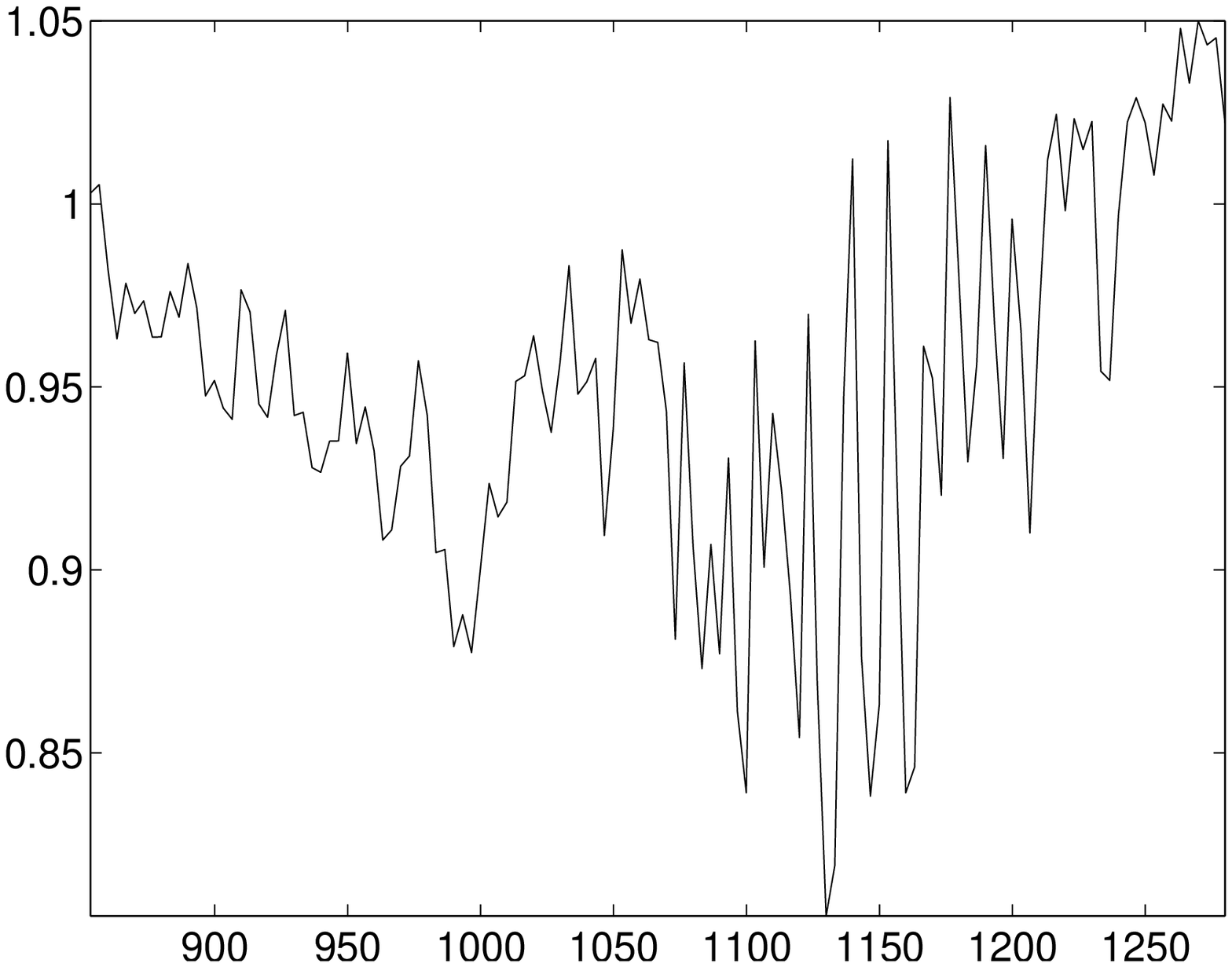}&
	\includegraphics[width=.23\textwidth,height=.1\textheight]{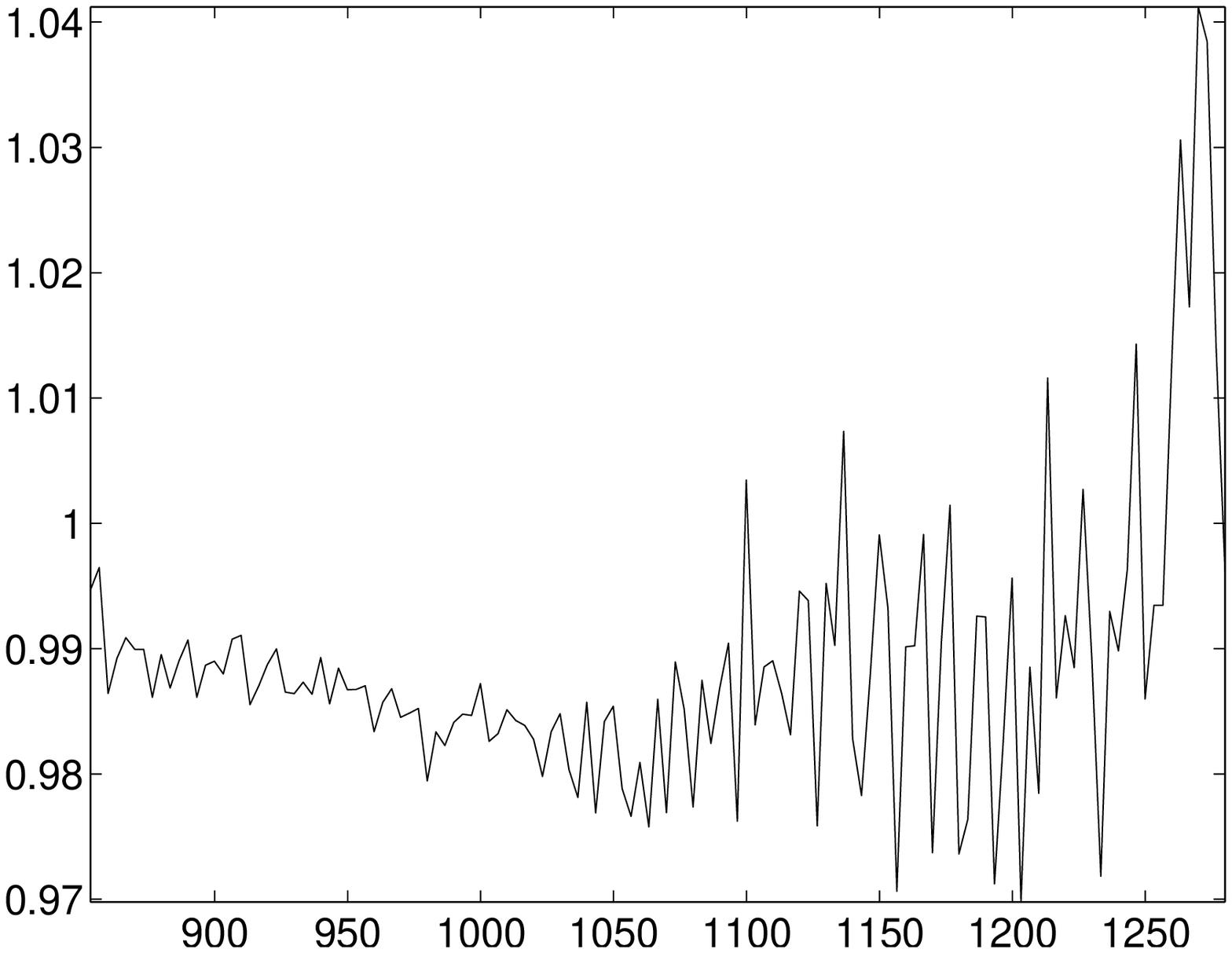}&
	\includegraphics[width=.23\textwidth,height=.1\textheight]{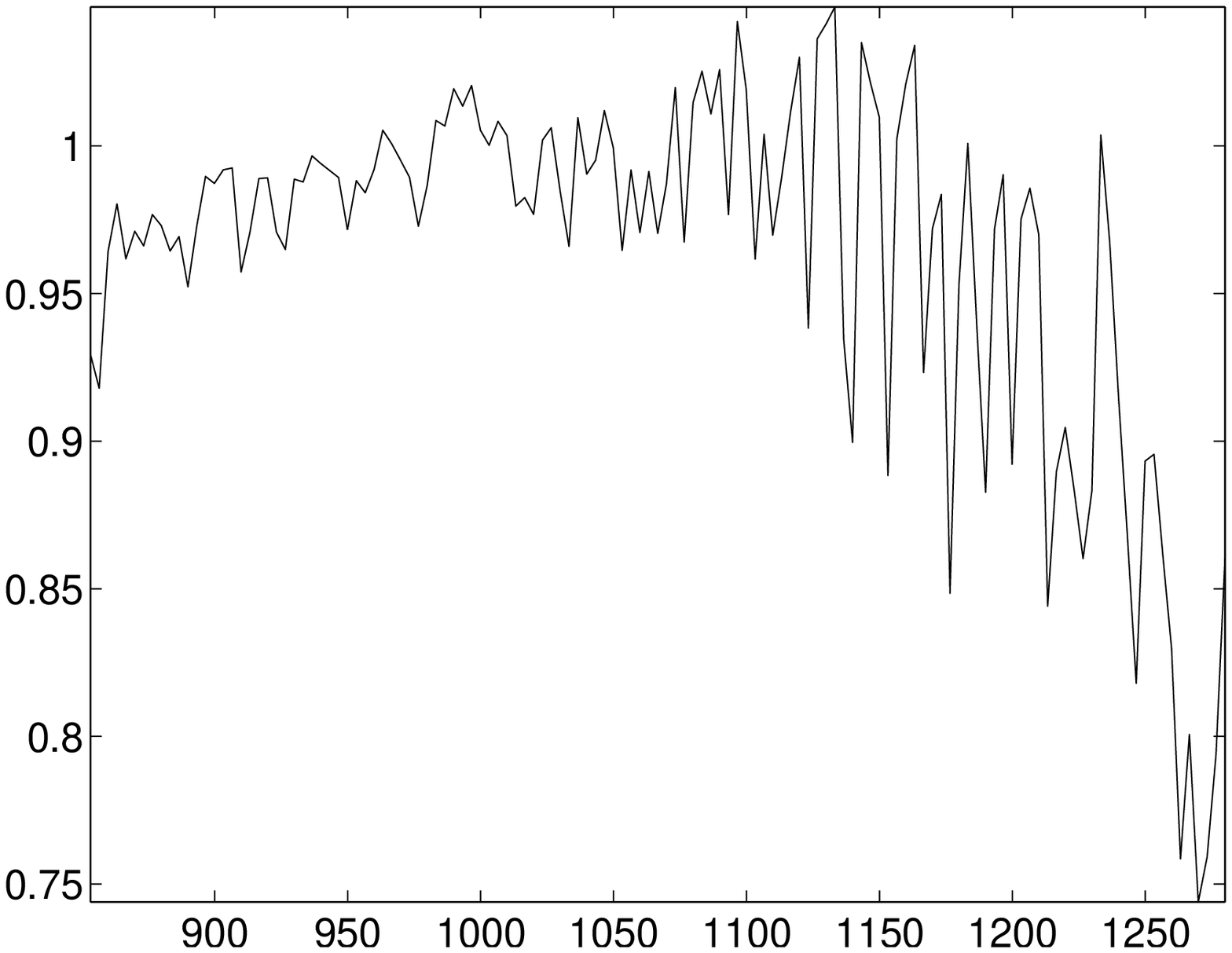}&
	\includegraphics[width=.23\textwidth,height=.1\textheight]{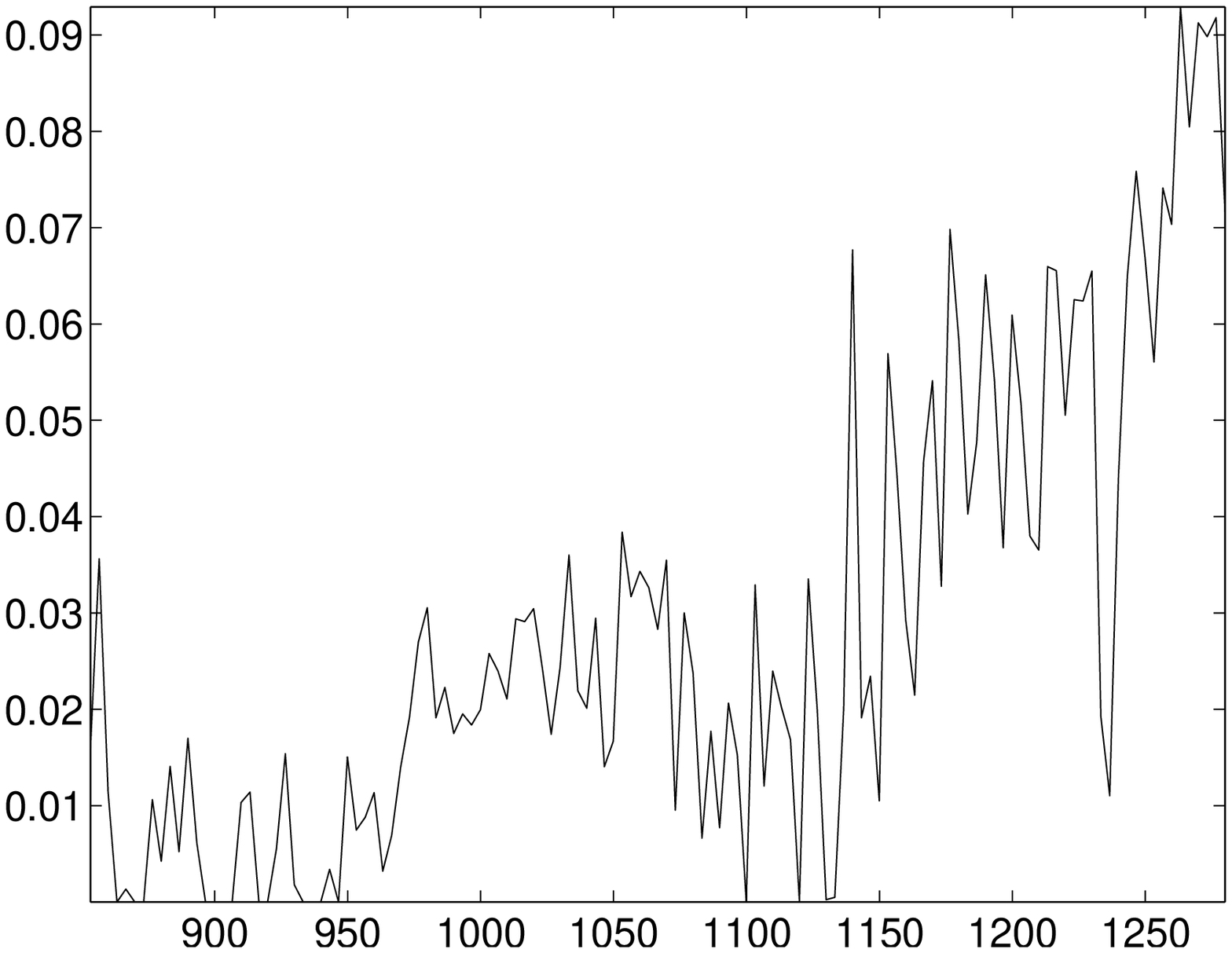}\\
		\multicolumn{4}{c}{Time frame 10}\\
	\includegraphics[width=.23\textwidth,height=.1\textheight]{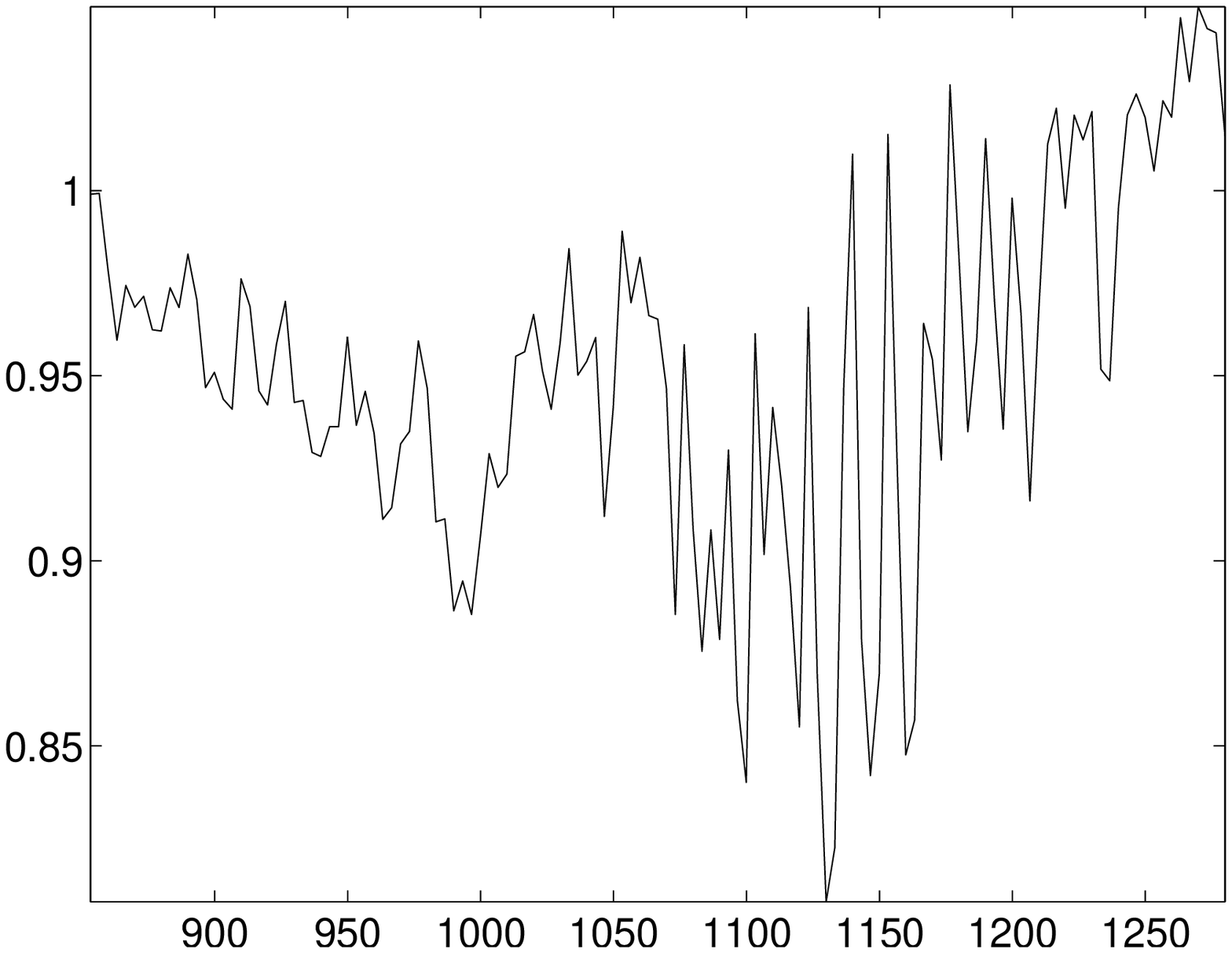}&
	\includegraphics[width=.23\textwidth,height=.1\textheight]{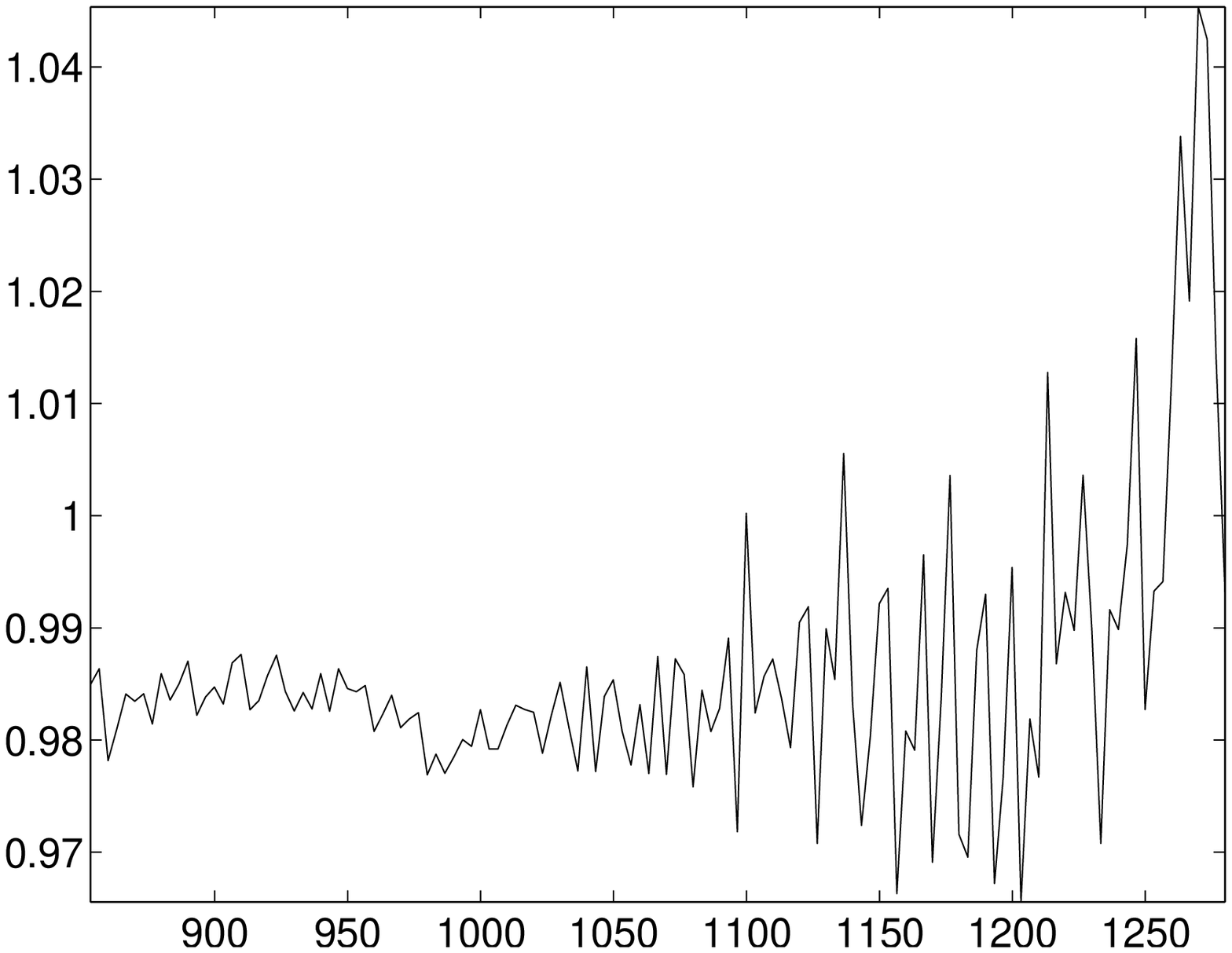}&
	\includegraphics[width=.23\textwidth,height=.1\textheight]{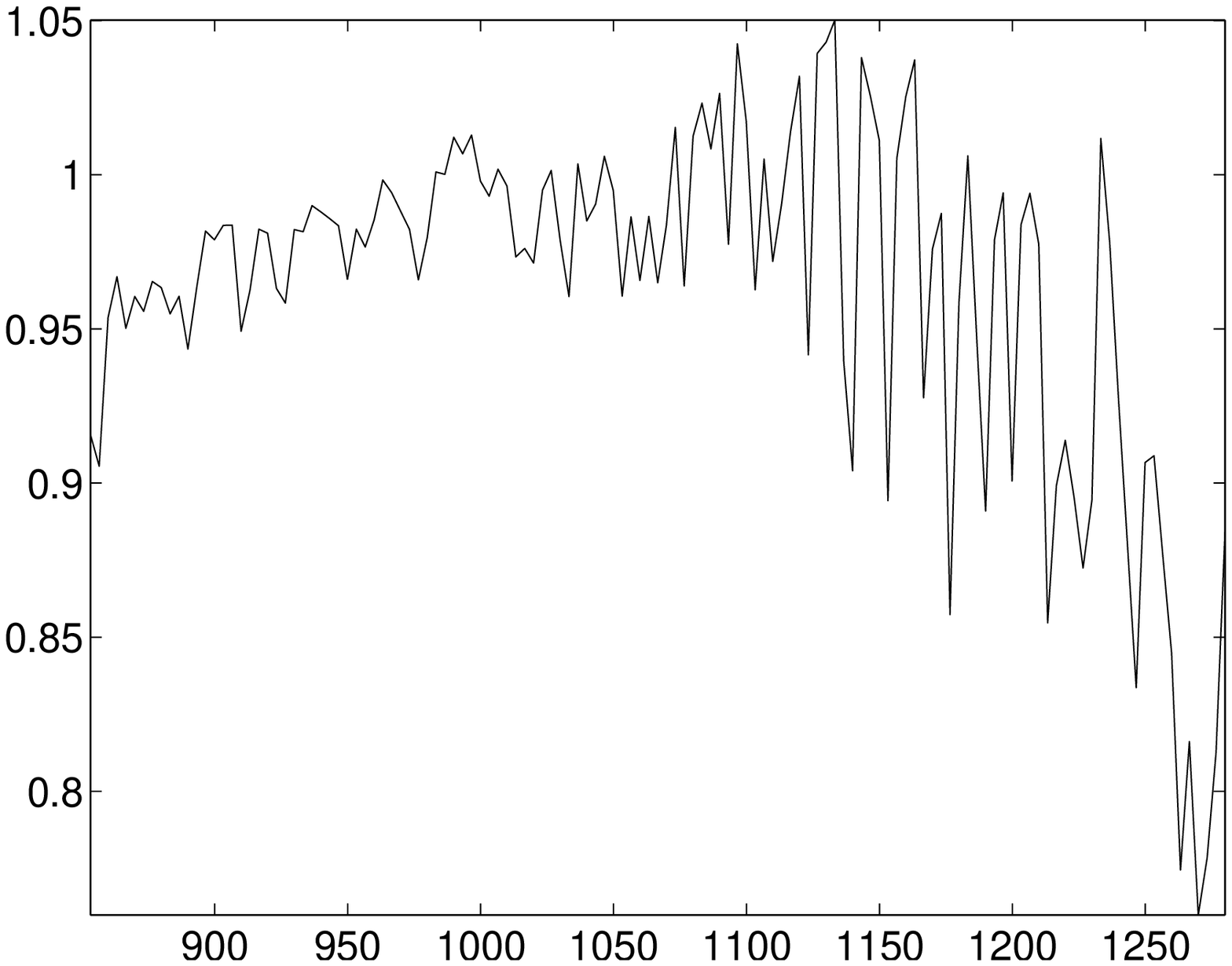}&
	\includegraphics[width=.23\textwidth,height=.1\textheight]{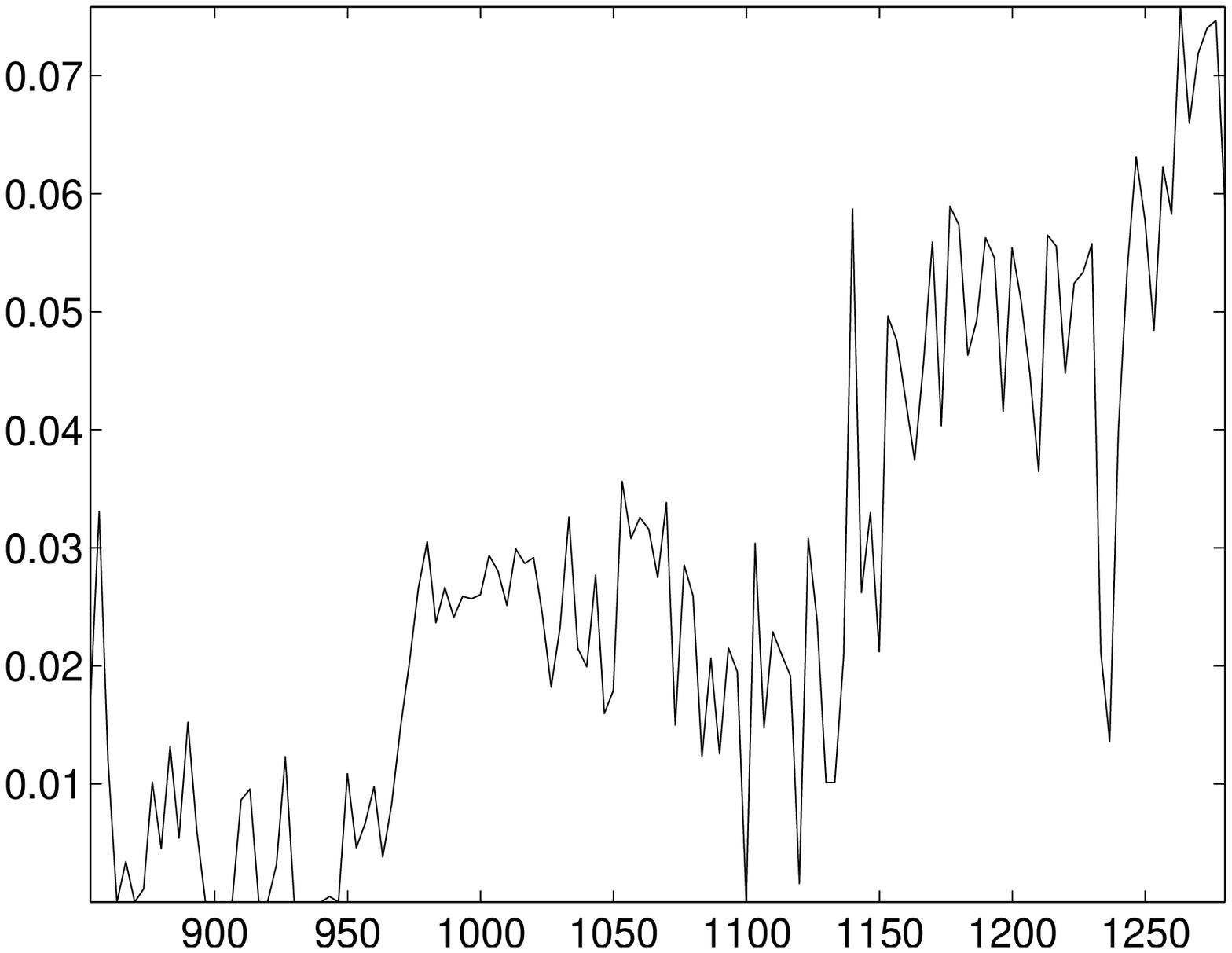}\\
	\end{tabular}
	\caption{\textcolor{black}{Endmembers obtained by the joint unmixing method : estimated spectra from the first, third, seventh and tenth time frame. Each row corresponds to a time frame and each column to a particular source. Notice how all spectra fit the proposed model of the data.}}
	\label{fig:end_joint}
\end{figure*}

We proceed to run the separate and joint unmixing strategies on the data set. Separate unmixing is again carried out using the default parameters. The parameters of the proposed joint unmixing algorithm are initialized in the following way:
\begin{itemize}
\item the initial abundance maps for the background endmembers, resp. denoted by 'mountain', 'sand' and 'sky', are set to the three maps extracted separately from the image using MVSA and SUNSAL. The initial abundance maps for the gas plume are matched to a segmentation of the plume, known beforehand~\cite{TOC14};
\item we obtain the reference spectra matrix $\mathbf{S}_0$ by performing nonnegative least squares inversion on matrix $\mathbf{A}_5^{\text{init}}$ (the time index being selected arbitrarily among all frames containing all four endmembers);
\item higher regularization parameters are set for the first three sources to enforce slower dynamics.
\end{itemize}
\textcolor{black}{The running times of both separate and joint unmixing methods are resp. 7 min and 44 min for the whole data set. Comparatively, nonlinear unmixing using the PPNMM takes approximately ten hours. The results are displayed in Figures \ref{fig:abund_sepa}, \ref{fig:abund_joint}, \ref{fig:end_sepa} and \ref{fig:end_joint}}.

The SU results on the first time frame show a discrepancy between the 'separate' and 'joint' method. The 'sand' endmember seems correctly extracted in both cases, but the abundance maps of the 'moutain' and 'sky' endmembers indicate that the 'joint' method unmixes them better than the 'separate' method. Conversely, a 'ghost' of the spatial support of the gas plume within all time frames can be observed on the three abundance maps produced by the 'joint' unmixing method on the current time frame. Removing this artefact caused by the proposed ALS implementation is one of the perspective of this work. The next time frames display a common extraction pattern: as expected, standard linear unmixing fails within the spatial support of the plume. Spectrally, it does seem that the gas is correctly attributed to a single endmember whose index varies along the time frames because of the permutation problem. However, the bad fit of the LMM in this spatial region causes the gas plume to appear on all four estimated abundance maps. Conversely, accounting for the temporal information in the joint unmixing approach allows to assign the gas plume to a single endmember (the fourth one).

\begin{figure*}[ht]
	\center
	\begin{tabular}{cccc}
	Source 1 & Source 2 & Source 3 & Source 4\\
	\includegraphics[width=.23\textwidth,height=.15\textheight]{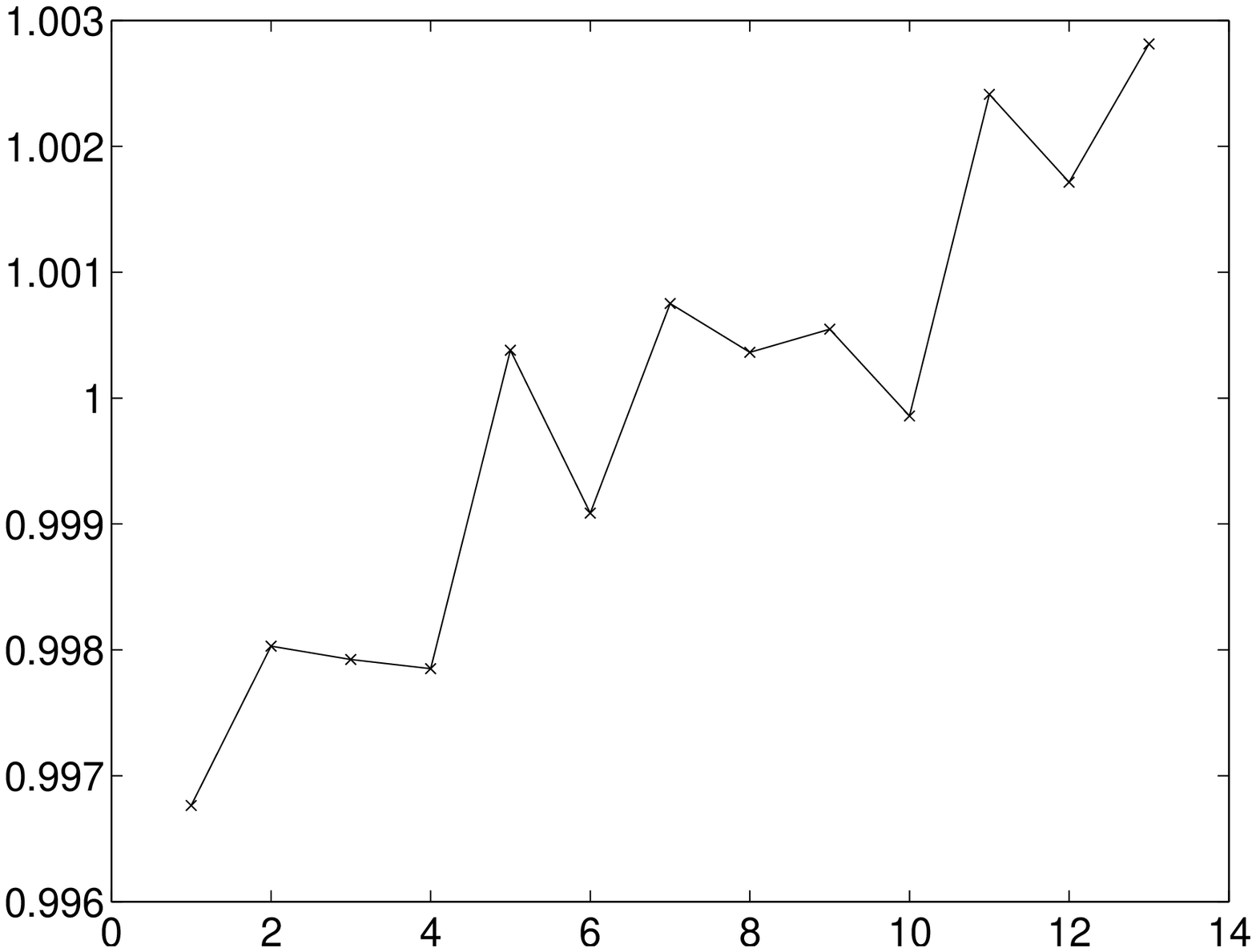}&
		\includegraphics[width=.23\textwidth,height=.15\textheight]{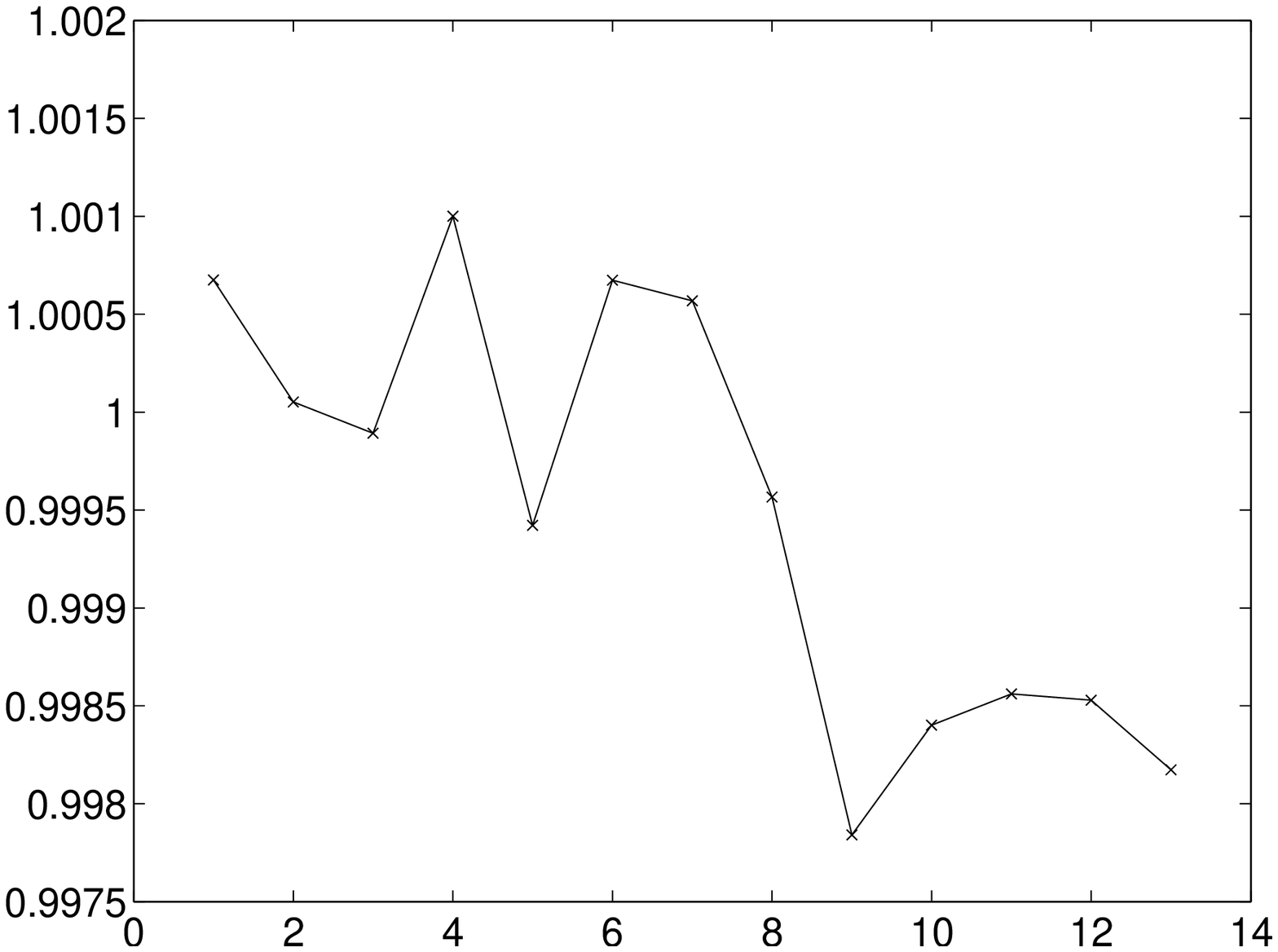}&
			\includegraphics[width=.23\textwidth,height=.15\textheight]{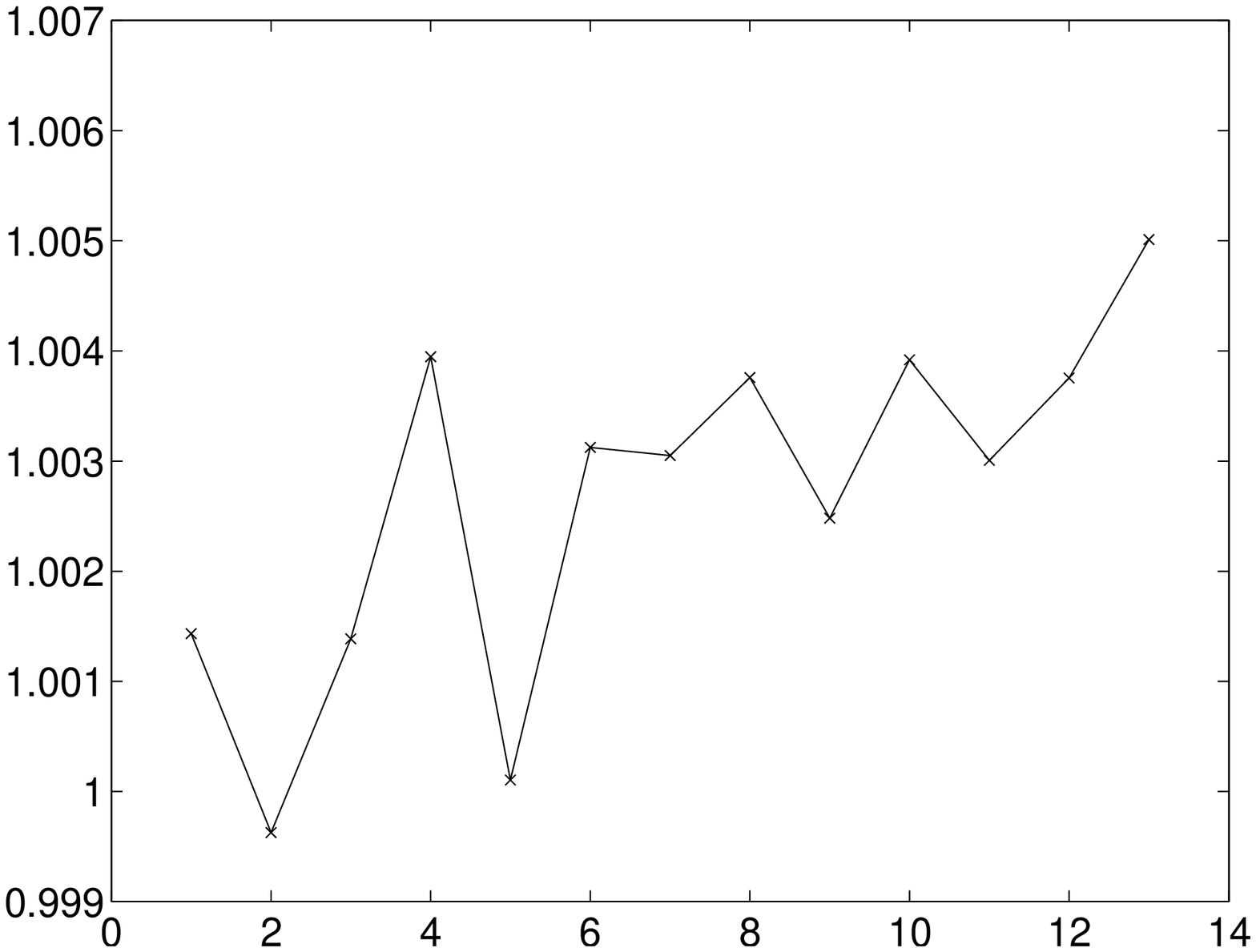}&
				\includegraphics[width=.23\textwidth,height=.15\textheight]{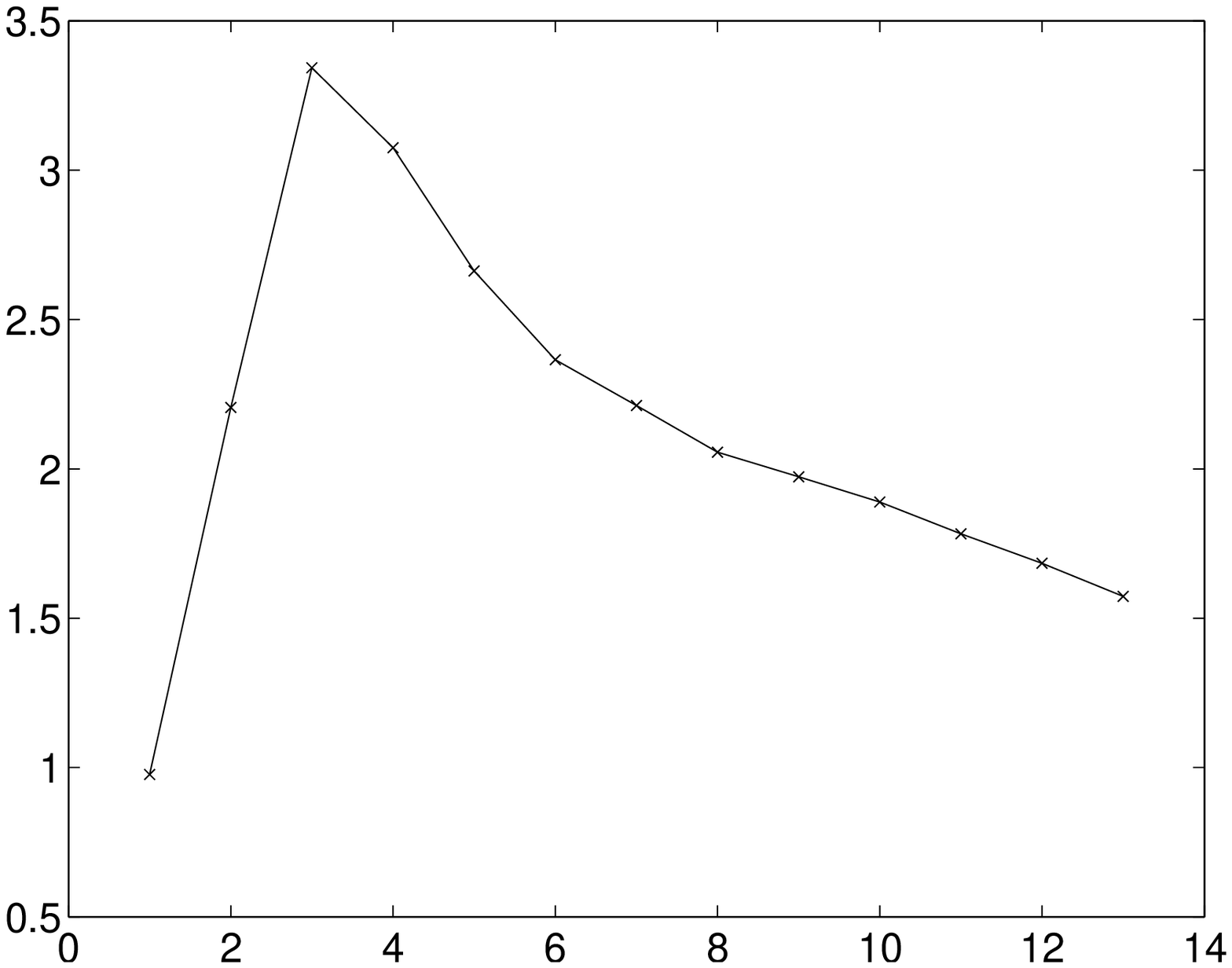}\\
	\end{tabular}
	\caption{\textcolor{black}{Evolution of scale factors over time : each plot corresponds to a particular endmember. Notice how the first three plots - 'sky', 'mountain', 'sand' - display tiny variations around unity, of the order of $10^{-3}$, whereas the fourth plot - gas plume- can be explained by the dynamical variation of the density of the plume.}}
	\label{fig:scales}
\end{figure*}

Spectral scale factors estimated by the joint unmixing method are displayed as a function of time in figure \ref{fig:scales}. The first three plots correspond to the natural endmembers in the scene - 'sky', 'mountain', 'sand' - and show very small perturbations around the unity value. Hence, the dynamics of these three endmembers can almost be considered as constant. The scale factor plot for the gas plume endmember displays a different pattern: the curve peaks at the third time frame, then slowly decays with time. Since the scale factor in endmember variability is linked to illumination properties of the material, the plot may be interpreted as the density of the plume, peaking just after the release then decreasing when the gas expands.

\begin{figure*}[ht]
	\begin{tabular}{cccc}
	\multicolumn{4}{c}{Time 4}\\
	Source 1 & Source 2 & Source 3 & Source 4\\
	\multicolumn{4}{c}{Joint unmixing}\\
	\includegraphics[width=.23\textwidth,height=.1\textheight]{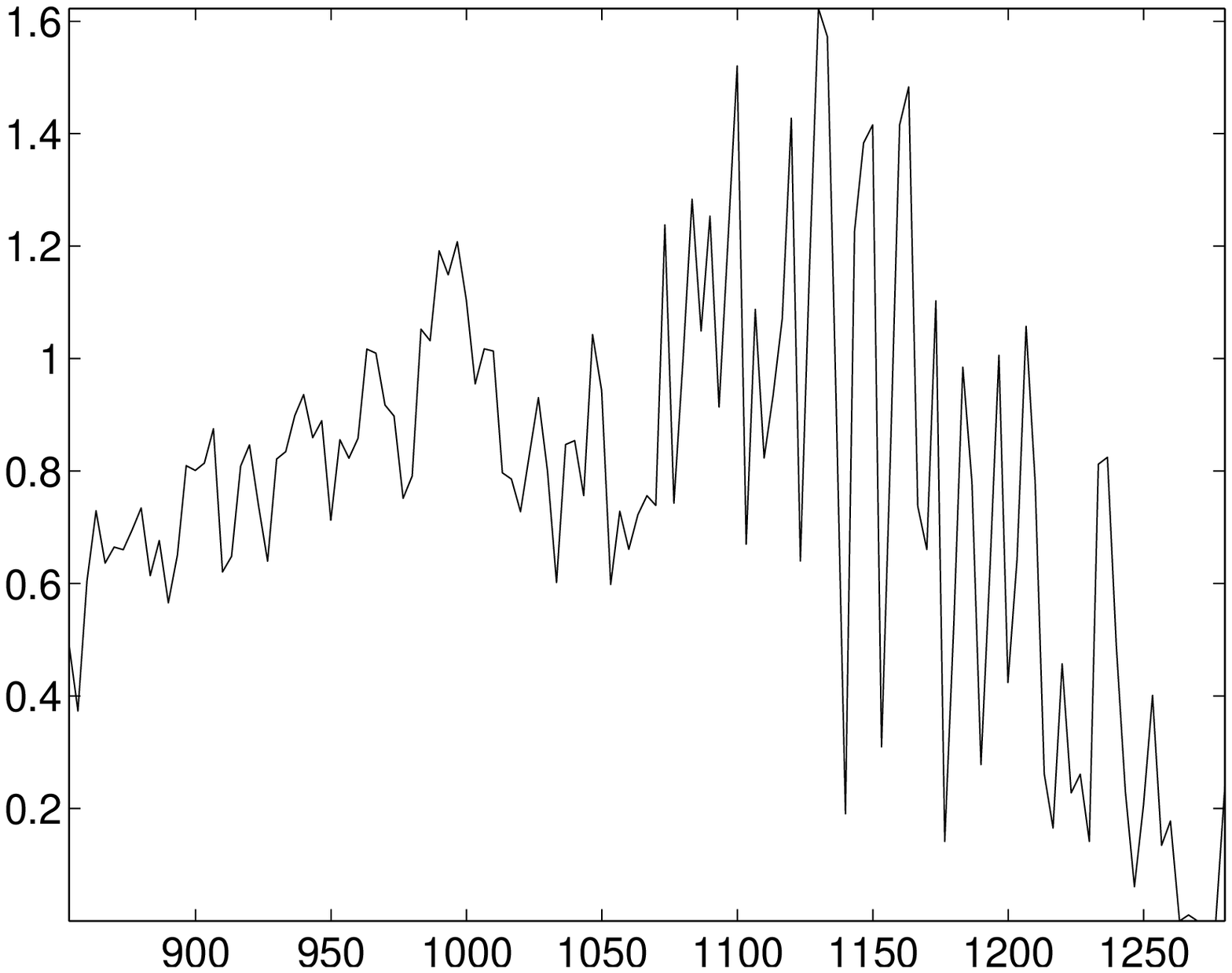}&
	\includegraphics[width=.23\textwidth,height=.1\textheight]{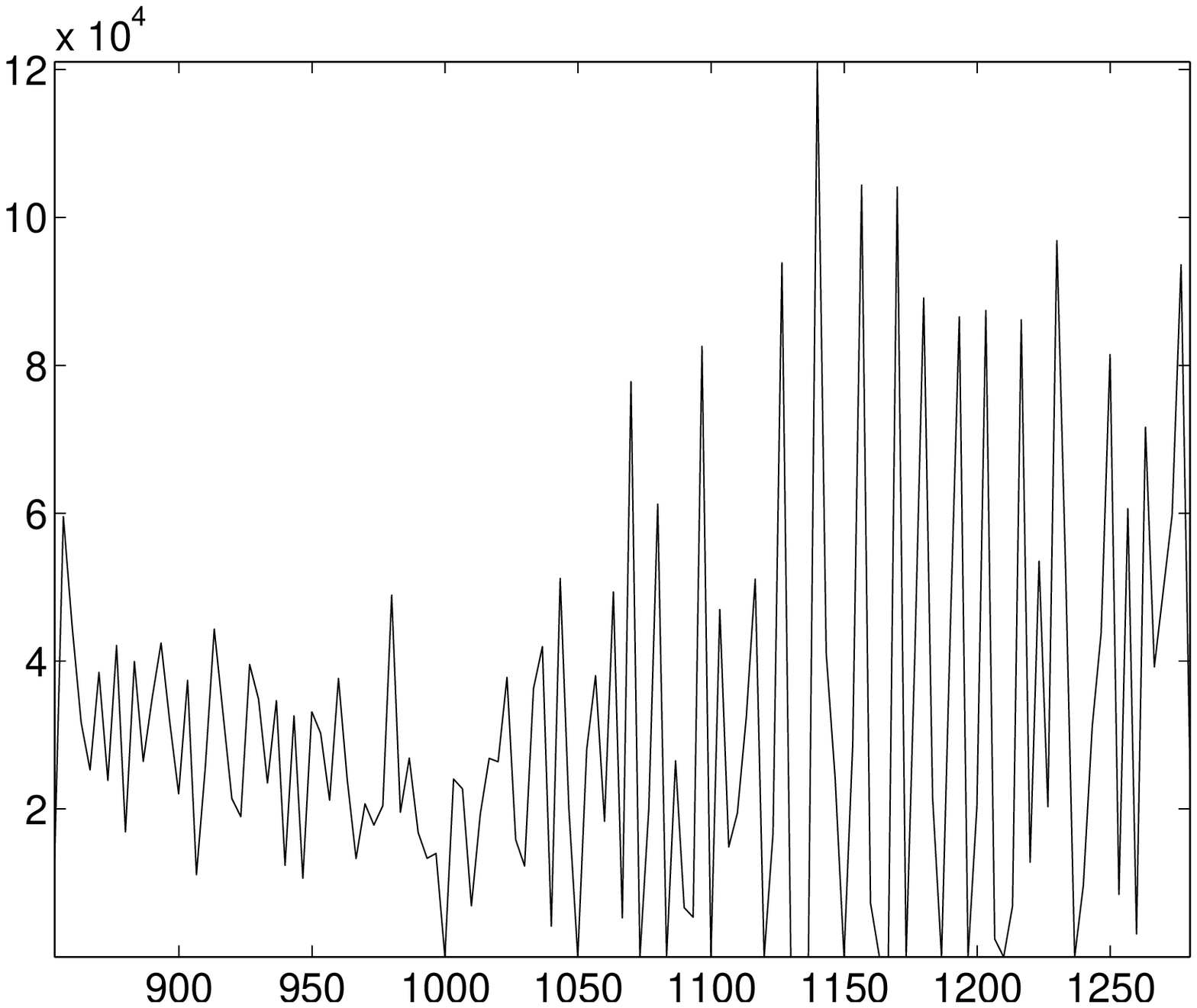}&
	\includegraphics[width=.23\textwidth,height=.1\textheight]{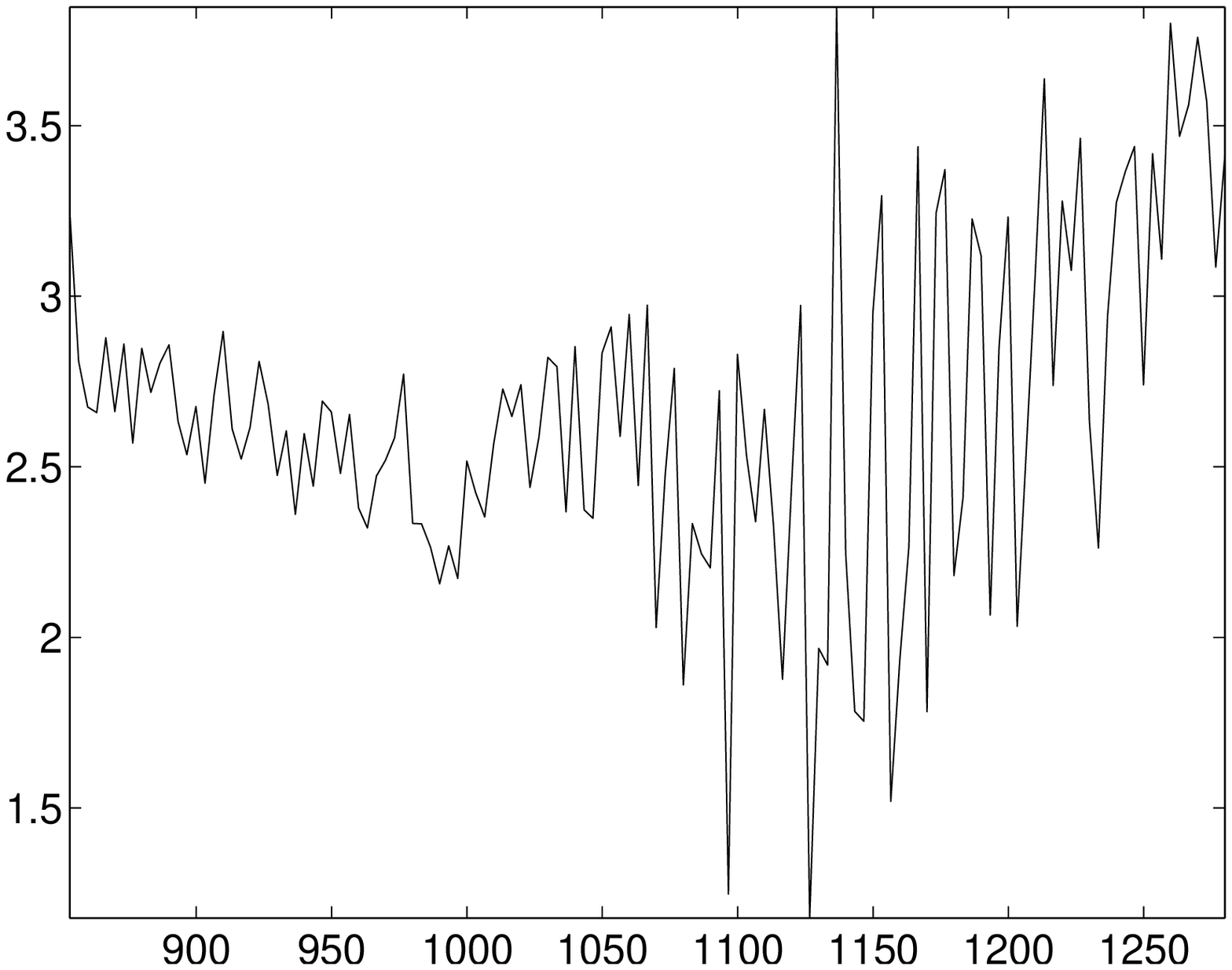}&
	\includegraphics[width=.23\textwidth,height=.1\textheight]{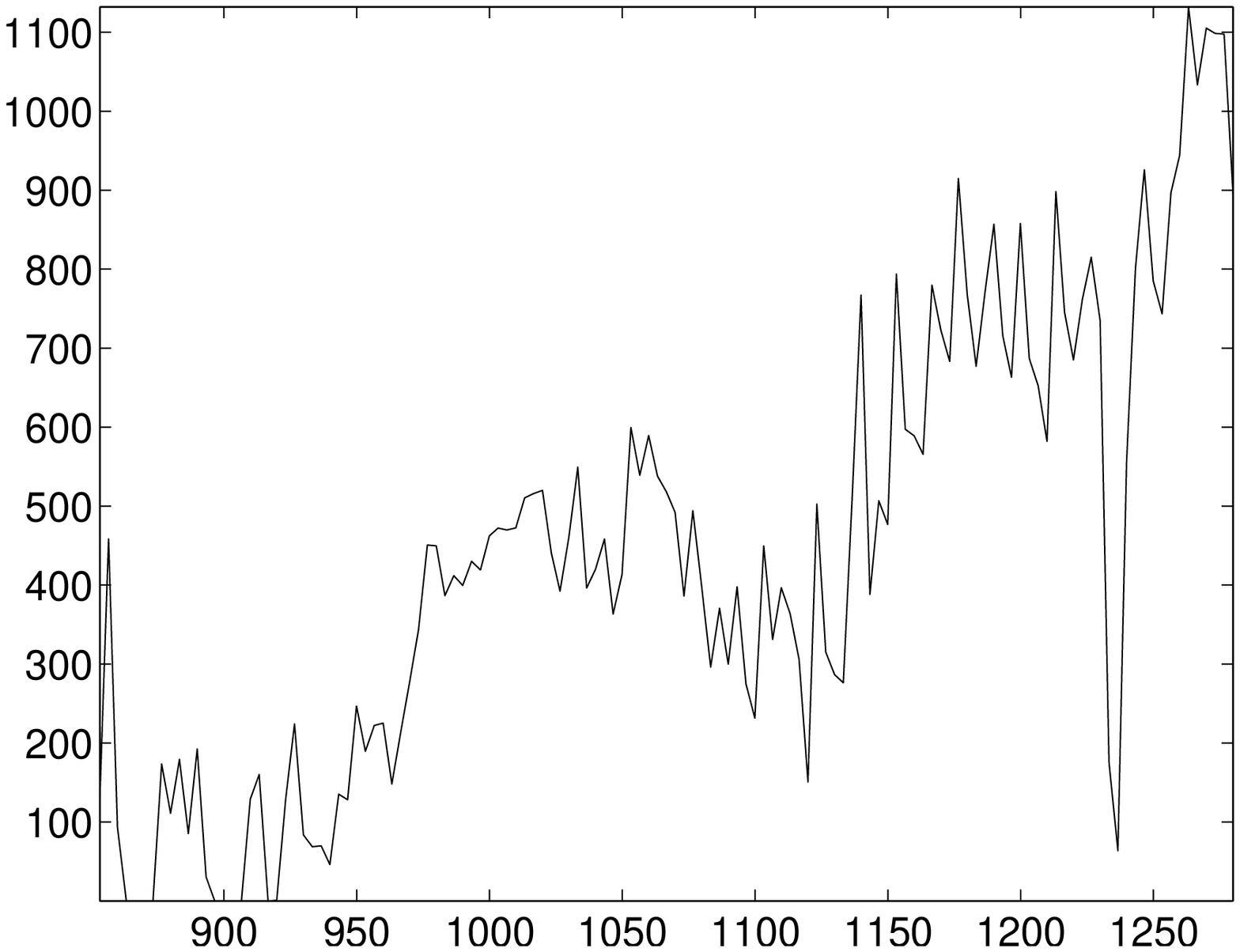}\\
	\includegraphics[width=.23\textwidth]{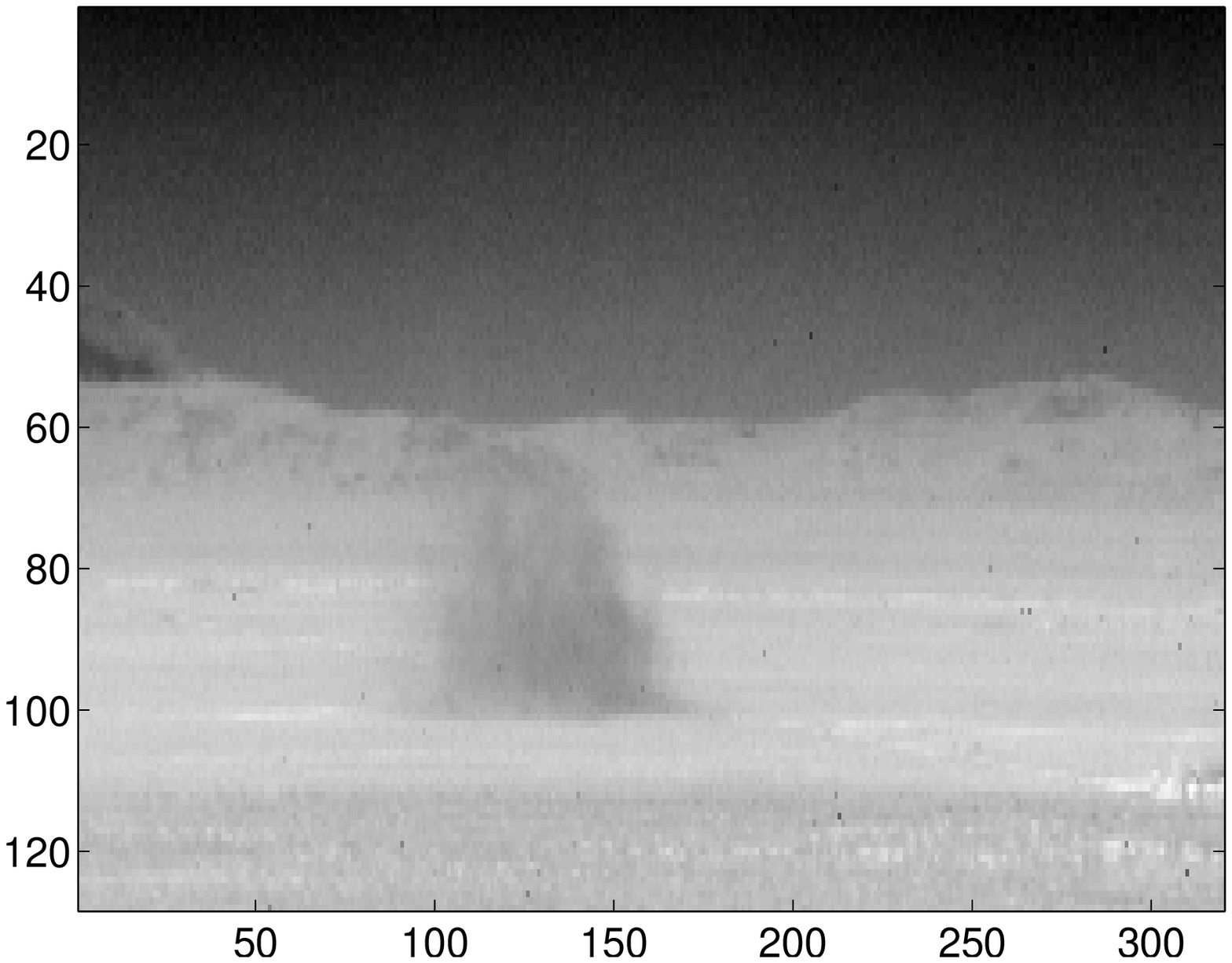}&
	\includegraphics[width=.23\textwidth]{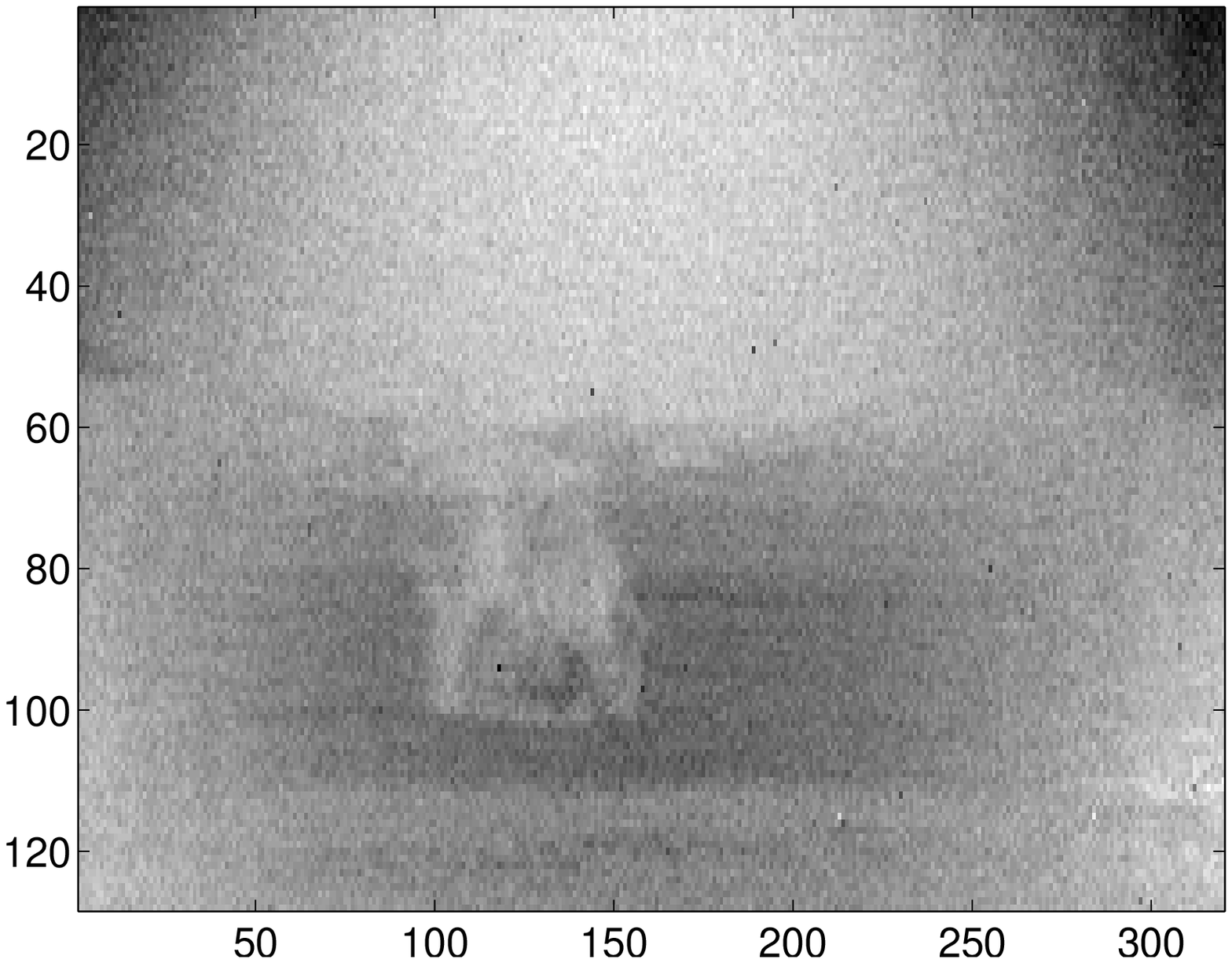}&
	\includegraphics[width=.23\textwidth]{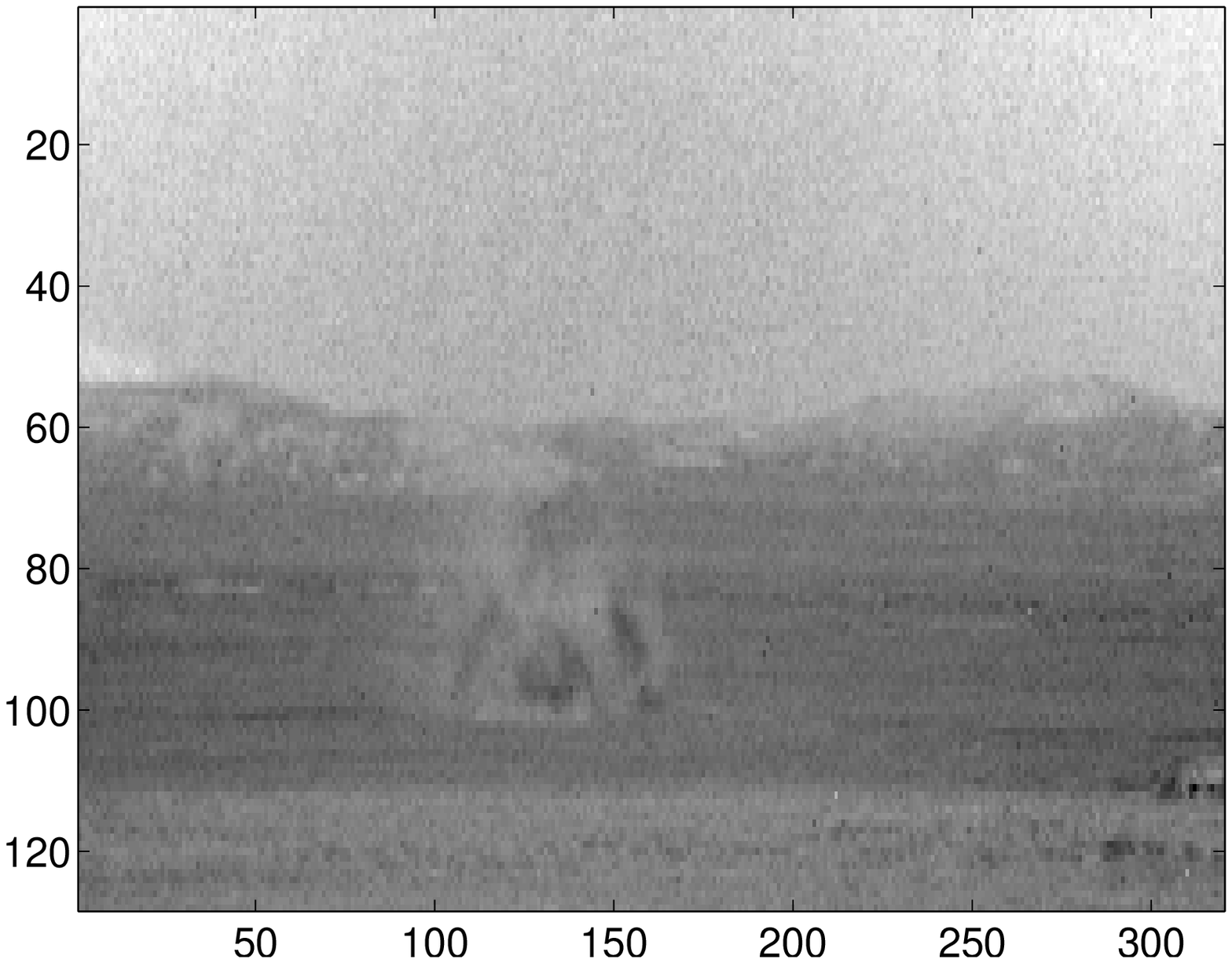}&
	\includegraphics[width=.23\textwidth]{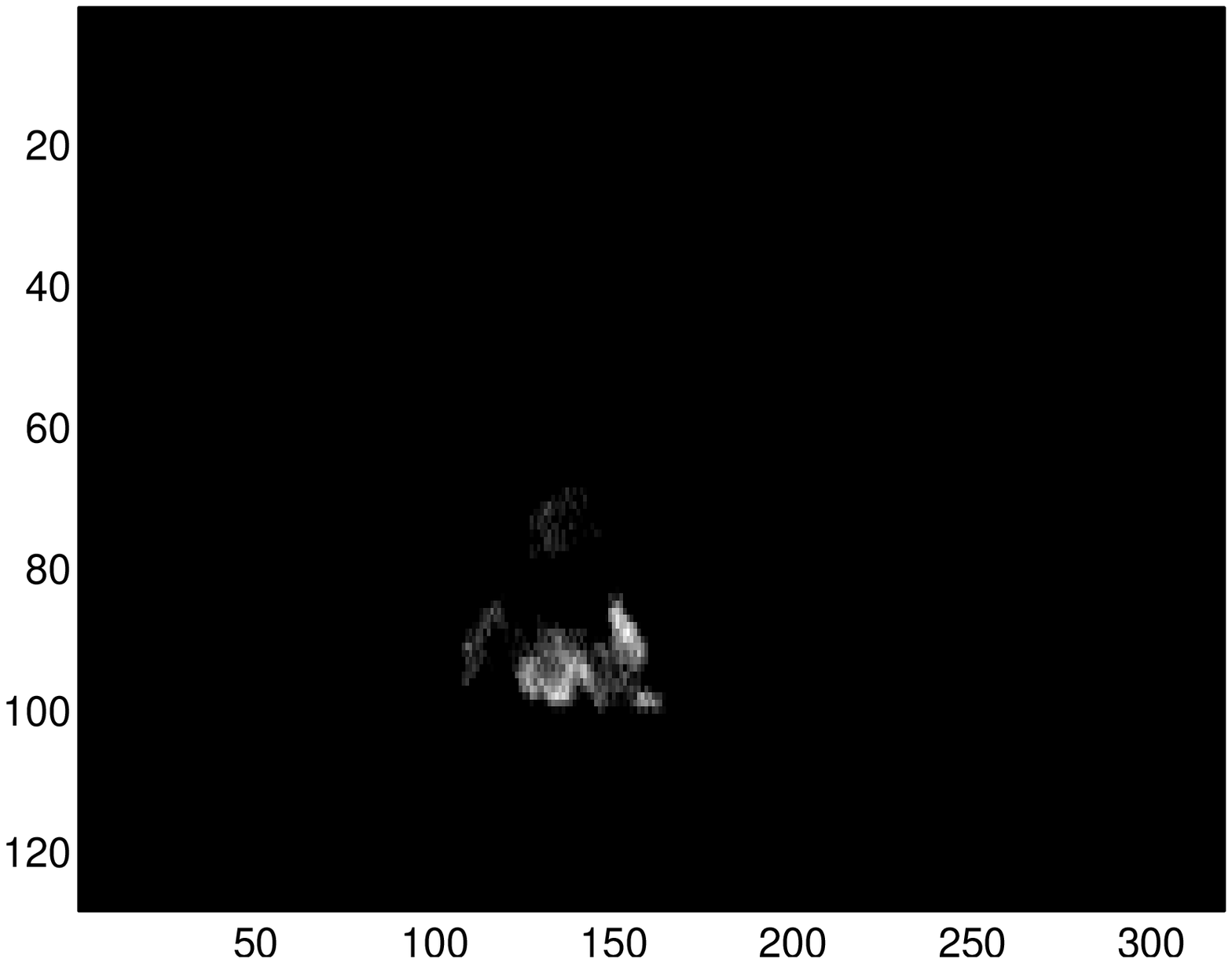}\\
	\end{tabular}
	\caption{\textcolor{black}{Unmixing of the fourth time frame with abundance maps initialized as constant images, using the 'joint unmixing' approach. Each column corresponds to a source, the first row displays the extracted spectra and the second row the estimated abundance maps. The joint method correctly attributes the gas plume to a unique endmember (the fourth one) even in the absence of any spatial prior knowledge of the scene.}}
	\label{fig:bad_init}
\end{figure*}

As a final note, we point out that the unmixing results are not obtained by relying solely on the proposed model. Indeed, we exploit a segmentation of this data set to perform the initialization step, which allows to avoid 'bad' local minima of the objective function~\cite{TOC14}. When less information about the data is available beforehand, a possible approach is to resort to multiple initializations in order to tackle the non convex nature of the problem, at the cost of a greater computational complexity. In the following experiment, the proposed algorithm is run with all abundance maps initialized as constant images. Figure \ref{fig:bad_init} displays the results obtained by the algorithm on the fourth time frame. As shown in the fourth panel of the figure, the gas plume is correctly extracted without using any segmentation technique. Other time frames are less physically interpretable, perhaps indicating that the method is stuck in a local minimum of the objective function. In the proposed strategy, the fourth time frame can then be used as a better initialization. We conclude from figure \ref{fig:bad_init} that the dynamical monitoring of the sources and abundances lead to promising results in multitemporal spectral unmixing.

\section{Conclusion}
	\label{sec:ccl}
	
	In this paper, we have proposed a general dynamical framework for spectral unmixing:
\textcolor{black}{
\begin{align*}
	\label{general_dynamic_model}
	& \left\{ \begin{array}{ll} \mathbf{X}_{k} &= \mathbf{S}_{k} \mathbf{A}_{k} + \mathbf{E}_{k}\\
	\mathbf{S}_{k} &= f_S (\mathbf{S}_{k-1}) + \mathbf{V}_{k}\\
	\mathbf{A}_{k} &= f_A (\mathbf{A}_{k-1}) + \mathbf{D}_{k}. \end{array} \right.
\end{align*}}
\textcolor{black}{The linear mixing model is assumed to hold at each time frame to benefit from the low computational complexity of linear unmixing methods. We make assumptions on the dynamics of the spectral signatures and abundance coefficients of sources, rather than on the data themselves, and hence fall within a data-driven framework. We propose a simplified version of this model tailored to the case of multitemporal hyperspectral images. We derive an efficient unmixing algorithm based on this model, which jointly processes all time frames in the image. The proposed method relies on an efficient nonnegative alternating least squares scheme. The performance of our approach is demonstrated on synthetic and real time series of hyperspectral images. We are currently investigating the extension of the method to account for more complex spectral variability schemes, in the temporal dimension as well as the spatial dimension of the image. Perspectives also include the extension of the model to nonlinear mixing processes. Finally, it is worth noting that the general dynamical framework is not restricted to the case of hyperspectral imaging. Because of its flexibility, it allows to analyze the dynamics of many types of multidimensional signals, \emph{e.g.} biomedicals signals such as electroencephalography (EEG) or magnetoencephalography (MEG).}

\section*{Acknowledgment}

We thank Andrea Bertozzi for her collaboration on the dataset provided by the US Defense Threat Reduction Agency and the National Science Foundation through NSF grant DMS-1118971. 

\newpage

\bibliographystyle{IEEEbib}
\bibliography{biblio_postdoc}

\end{document}